\documentstyle[twoside,a4wide,mcleg,11pt]{report}

\begin{document}

\pagenumbering{roman}




\newcommand{\newterm}[1]{{\sc #1}}
\newcommand{\acronym}[1]{{\sc #1}}
\newcommand{\scare}[1]{`#1'}
\newcommand{\shortquote}[1]{`#1'}
\newcommand{\latin}[1]{{\it #1\/}}
\newcommand{\foreign}[1]{{\it #1\/}}


\newcommand{\nlp}{{\sc nlp}}
\newcommand{\etal}{\latin{et~al}}
\newcommand{\sll}{{\sc sll}}

\newcommand{\argmax}[1]{\mbox{\it argmax\/}_{\,{#1}}}
\newcommand{\opt}{\mbox{\it opt\/}}
\newcommand{\alphaopt}{\alpha_{\opt}}
\newcommand{\countfn}{\mbox{\it count\/}}
\newcommand{\nullsym}{\mbox{\sc null}}
\newcommand{\pfromc}{p_c}
\newcommand{\vopt}{v_{\opt}}
\newcommand{\even}[1]{\mbox{\sc even}(#1)}
\newcommand{\odd}[1]{\mbox{\sc odd}(#1)}
\newcommand{\halfbin}{B_{\frac{1}{2}}}
\newcommand{\halfn}{\mbox{$\frac{n}{2}$}}
\newcommand{\halfnup}{\lceil \halfn \rceil}
\newcommand{\halfninc}{\mbox{$\frac{n+1}{2}$}}
\newcommand{\halfnincup}{\lceil \halfninc \rceil}
\newcommand{\halfndec}{\mbox{$\frac{n-1}{2}$}}
\newcommand{\halfndecup}{\lceil \halfndec \rceil}
\newcommand{\yaftern}{{}_n Y}
\newcommand{\ambig}{\mbox{\it ambig\/}}
\newcommand{\tla}{t_{\mbox{\sc la}}}
\newcommand{\tjo}{t_{\mbox{\sc jo}}}
\newcommand{\tsu}{t_{\mbox{\sc su}}}
\newcommand{\tsh}{t_{\mbox{\sc sh}}}


\newcommand{\lingform}[1]{{\it #1\/}}
\newcommand{\progname}[1]{{\sc #1}}
\newcommand{\publicationname}[1]{{\it #1\/}}
\newcommand{\articlename}[1]{\mbox{\sc #1}}
\newcommand{\predicate}[1]{\mbox{\bf #1}}
\newcommand{\concept}[1]{\mbox{\sc #1}}
\newcommand{\semrel}[1]{\mbox{\sc #1}}
\newcommand{\marker}[1]{\mbox{\bf #1}}
\newcommand{\tc}[2]{{\sc #1}\##2}
\newcommand{\tag}[1]{\mbox{\sc #1}}
\newcommand{\token}[1]{{\it $\|$#1\/$\|$}}
\newcommand{\stress}[1]{\mbox{\bf #1}}
\newcommand{\nonterm}[1]{\mbox{\sc #1}}

%
%
\newcommand{\np}{$_{\mbox{\footnotesize\sc np}}$}
\newcommand{\nbar}{$_{\bar{\mbox{\footnotesize\sc n}}}$}
\newcommand{\nn}{$_{\mbox{\footnotesize\sc n}}$}


\newtheorem{definition}{Definition}
\newtheorem{hypothesis}{Hypothesis}
\newtheorem{assumption}{Assumption}
\newtheorem{corollary}{Corollary}


\newsavebox{\citeforquote}
\newenvironment{citedquote}[1]{\begin{quote} \sbox{\citeforquote}{(#1)}}{\par \raggedleft \usebox{\citeforquote} \end{quote}}


\newenvironment{dialogue}%
{\begin{list}{}{\setlength{\itemsep}{0pt}\setlength{\parsep}{0pt}}}%
{\end{list}}
\newcommand{\turn}[1]{\item[\it {#1}:] }


%
%
\def\mytitlepage{ \newpage \thispagestyle{empty} }
\def\myendtitlepage{ \newpage }

\begin{mytitlepage}

\null\vspace*{1.5cm}

\includegraphics{logo-closed.eps}

\begin{center}
\Large \bf

\addvspace{-0.8cm}

{\huge \bf 
	Designing Statistical Language Learners:

\addvspace{0.25cm}

	Experiments on Noun Compounds
}

\addvspace{5.5cm}

	{\huge \bf Mark Lauer}

\addvspace{0.65cm}

	Department of Computing \\
	Macquarie University NSW 2109 \\
	Australia

\addvspace{1.7cm}

	Submitted in Partial Fulfillment of the Requirements \\
	of the Degree of Doctor of Philosophy

\addvspace{3cm}

	December, 1995

\end{center}
\end{mytitlepage}

\clearpage
\begin{center}
\begin{tabular}{l}
\vspace{6in} \hspace{4.5in} \\
{\small  Copyright \copyright\ Mark Lauer, 1995}
\end{tabular}
\end{center}
\thispagestyle{empty}

\vspace{4in}
\begin{center}
\begin{tabular}{r}
\vspace{3in} \\
{\large \it To Lesley Johnston,} \\
\large \\
{\large \it without whom nothing good can ever come.}
\end{tabular}
\end{center}
\thispagestyle{empty}



\begin{abstract}

Statistical language learning research takes the view that 
many traditional natural language processing tasks can be solved
by training probabilistic models of language on a sufficient 
volume of training data.  The design of statistical language learners
therefore involves answering two questions: (i)~Which of the
multitude of possible language models will most accurately
reflect the properties necessary to a given task?  
(ii)~What will constitute a sufficient volume of training data?  
Regarding the first question, though a variety of successful
models have been discovered, the space of possible designs
remains largely unexplored.  Regarding the second, exploration
of the design space has so far proceeded without an adequate answer.

The goal of this thesis is to advance the exploration 
of the statistical language learning design space.  
In pursuit of that goal,
the thesis makes two main theoretical contributions:  it identifies
a new class of designs by providing a novel theory 
of statistical natural language processing, and it presents the
foundations for a predictive theory of data requirements
to assist in future design explorations.

The first of these contributions is called 
the meaning distributions theory.  This theory specifies 
an architecture for natural language analysis in which
probabilities are given to semantic forms rather than to
more superficial linguistic elements.  Thus, rather than 
assigning probabilities to grammatical structures directly, 
grammatical forms inherit likelihoods from the semantic 
forms that they correspond to.  The class of designs 
suggested by this theory represents a promising new area
of the design space.

The second theoretical contribution concerns development
of a mathematical theory whose aim is to predict the
expected accuracy of a statistical language learning system
in terms of the volume of data used to train it.  
Since availability of appropriate training data is a key
design issue, such a theory constitutes an invaluable
navigational aid.  The work completed includes the development
of a framework for viewing data requirements and a number
of results allowing the prediction of 
necessary training data volumes under certain conditions.

The experimental contributions of this thesis
illustrate the theoretical work by applying
statistical language learning designs to
the analysis of noun compounds.  Both syntactic
and semantic analysis of noun compounds have been approached
using probabilistic models based on the meaning distributions
theory.

In the experiments on syntax, a novel model, 
based on dependency relations between concepts, was
developed and implemented.  Empirical comparisons
demonstrated that this model is significantly better than
those previously proposed and approaches the performance
of human judges on the same task.  This model also
correctly predicts the observed distribution of syntactic
structures.

In the experiments on semantic analysis, a novel model,
the first statistical model of this problem, was developed
and implemented.  The system uses statistics computed 
from prepositional phrases to predict a paraphrase with
significantly better accuracy than the baseline strategy.
The training data used is both sparse and noisy, and the
experimental results support the need for a theory of data
requirements.  Without a predictive data requirements
theory, statistical language learning remains an artform.

\end{abstract}


\chapter*{Acknowledgements}

To do justice to the people who have, in one way 
or another, made this work what it is, feels like it
would take more than the rest of the thesis.  
I can only say to everyone who has contributed
that I apologise for the entirely inadequate 
acknowledgements to follow.  I realise that 
the sheer number of names below will suggest
to the reader that each has contributed only a little.
What can I say?  All have made a significant
contribution; I have no real choice.

\subsection*{}

I could not ask for a pair of finer minds to share
this journey with than my supervisors, 
Robert Dale and Mike Johnson, whose brilliance
still amazes me on a regular basis.  Together, 
they are my Plato and my Aristotle, and I doubt 
I will ever shed their influences on my thinking.
I thank Mike for his infinite understanding
and Robert for his boundless energy.

Without the vision and determination of 
Vance Gledhill, the unique environment 
at the Microsoft Institute, and all the work 
that has emerged from it, would never have existed.
He deserves the success it currently enjoys,
and my heartfelt thanks.

I want particularly to thank Mark Johnson for
his assistance both in nurturing the development
of my own ideas, and in generously contributing 
his own.  Thanks also to Wayne Wobcke for
discussions and input at various times.

Special thanks are deserved by Ted Briscoe, 
Gregory Grefenstette, Karen Jensen and Richard Sproat, 
all of whom don't realise how much I have valued 
their encouragements and ideas.
I owe a great debt of gratitude to Philip Resnik, 
not only for his technical contributions, but also
for his faith, passion and friendship; I only hope 
that one day I can do them justice.

More than anyone else, the other Microsoft Institute
fellows have become part of the fabric of this thesis.
I am grateful to everyone here.  Especial thanks
must go to Richard Buckland and Mark Dras; both
are blessed with genius, as well as just being really 
friendly guys.  Particular contributions have also 
been made by Sarah Boyd, Maria Milosavljevic, Steven Sommer, 
Wilco ter~Stal, Jonathon Tidswell and Adrian Tulloch.

I wish also to thank the following people
for their friendship, which has all helped:
Ken Barker, Alan Blair, Tanya Bowden, Christophe Chastang,
Phil Harrison, Rosie Jones, Patrick Juola,
Elisabeth Maier, Michael Mitchell, Nick Nicholas,
Peter Wallis, Susan Williams and Danny Yee.

Without the financial support generously given
by the Microsoft Institute Fellowship Program and the
Australian Government Postgraduate Award Scheme,
this research would not have happened.

\subsection*{}

{\samepage
Every single one of the following has personally made a 
significant difference to the work presented here.  
I am sorry I cannot specifically thank you all.

\begin{tabular}{llll}
& & & \\
{\small John Bateman} &{\small George Heidorn} &{\small Malti Patel} &{\small Andrew Taylor} \\
{\small Ezra Black} &{\small Andrew Hunt} &{\small Pavlos Peppas} &{\small Lucy Vanderwende} \\
{\small Rebecca Bruce} &{\small Christian Jacquemin} &{\small Pam Peters} &{\small Wolfgang Wahlster} \\
{\small Ted Dunning} &{\small Geof Jones} &{\small David Powers} &{\small Bonnie Lyn Webber} \\
{\small Dominique Estival} &{\small Kevin Knight} &{\small James Pustejovsky} &{\small Yorick Wilks} \\
{\small Tim Finin} &{\small John Lafferty} &{\small Ross Quinlan} &{\small Dekai Wu} \\
{\small Norman Foo} &{\small Alon Lavie} &{\small Carolyn Penstein Ros\'{e}} &{\small Collin Yallop} \\
{\small Louise Guthrie} &{\small Chris Manning} &{\small Jeff Siskind} &{\small David Yarowsky} \\
{\small Marti Hearst} &{\small Jenny Norris} &{\small Mark Steedman} &{\small Kobayasi Yoshiyuki} \\
\end{tabular}
}

\subsection*{}

And now some very special people:
There is nothing I can ever do to repay the 
unerring support and care provided by my father,
my stepmother and my grandmother, without
which I would be lost.  

Finally, the love and friendship I have shared 
over the past few years with Andrew Campbell,
Christine Cherry and Lesley Johnston goes beyond
all words.  Each has saved me from despair more times 
than I can count.  They are the earth on which I stand,
the air which I breathe and the sunlight that banishes
my darkness.  

\subsection*{}

\subsection*{}

\subsection*{}
\vspace{2in}

\subsection*{Addendum to acknowledgements for first reprint}
This thesis has been accepted without modification by Macquarie University
in fulfillment of the requirements for the Degree of Doctor of Philosophy.
Since submission I have received insightful comments from my examiners
which have prompted me to make some small changes.
I would therefore also like to thank them: Eugene Charniak, Mitch Marcus
and Chris Wallace.

\chapter*{Preface}

The research represented in this thesis was carried out
at the Microsoft Institute.  All work reported here
is the original work of the author, with the following
two exceptions.
\begin{enumerate}
\item The reasoning given in section~\ref{sec:dr_beginning}
regarding empty and non-empty bins 
(pages~\pageref{pg:dr_beginning_MJstart}--\pageref{pg:dr_beginning_MJfinish})
was developed by Mark Johnson of Brown University.  The author's
contribution was to extend the results to even values
of $n$ (the initial work only considered odd $n$) and 
complete the proof for equation~\ref{eq:dr_beginning_emptybound}.
This work has been published as Lauer~(1995a) with Mark Johnson's
permission.
\item An original version of the probabilistic model given in 
section~\ref{sec:cy_model} was jointly developed by the author
and Mark Dras, and has been published in Lauer and Dras~(1994).
\end{enumerate}

Some parts of this thesis include revised versions
of published papers.  I would like to thank 
the Association for Computational Linguistics
for (automatically) granting permission to reuse
material from Lauer~(1995b) (this material is primarily contained
in sections~\ref{sec:cy_model}, \ref{sec:cy_results} 
and~\ref{sec:cy_comparisons}).  Similarly, the
Pacific Association for Computational Linguistics
has (automatically) granted permission to reuse
material from Lauer~(1995a) (this material appears
in sections~\ref{sec:dr_need} through~\ref{sec:dr_beginning}).
Finally, kind permission has been given to reuse material from
Lauer~(1995c) (``Conserving Fuel in Statistical Language Learning:
Predicting Data Requirements'' in Proceedings of the Eighth Australian Joint
Conference on Artificial Intelligence, pp.~443--450.
Copyright by World Scientific Publishing Co. Pte, Singapore, 1995)
which appears in sections~\ref{sec:dr_beginning} 
through~\ref{sec:dr_simulations}.

\tableofcontents
\listoftables
\listoffigures

\cleardoublepage
\chapter{Introduction}
\label{ch:intro}
\pagenumbering{arabic}


This thesis is about computers learning about language.  It is about bringing 
machines into communication with people, not by a laborious process of 
linguistic instruction, but by creating programs that learn about language for 
themselves.  Just as for other work in natural language processing (\nlp), the 
ultimate goal is to have computers understand language in the sense that 
humans do; in response to human language, computers should behave as we 
do, or at least analogously.  And, just as with other work in \nlp, it 
recognises that we must be content with meeker achievements in the 
short term.  But the vision that inspires this work is freedom from the myriad 
of minutiae that make up language.  Perhaps we, as teachers, will better 
serve our goal by showing computers how to learn and then leaving them to 
slave over the details of language.  After all, remembering thousands of tiny 
facts is just what they are designed for.

This vision is gaining currency amongst \nlp\ researchers.  This is in part due 
to the staggering volume of text that is now available to computers as a 
learning resource.  A single daily newspaper produces tens of millions of 
words of text every year, all stored electronically during typesetting.  
Electronic mail, business reports, novels and manuals are all being generated 
with every passing moment.  An immense stream of text is being produced.
Also, at the same time, modern computers are growing enormously powerful.  
Statistics about 
linguistic phenomena can be computed from millions of words of text in just 
a few seconds.  The question becomes how can we put such statistics to use 
as a learning mechanism in order to exploit the ocean of text at our disposal?

The area of research concerning this question has been called 
\newterm{statistical language learning} (\sll).  Already quite a bit of 
research has been devoted to its investigation, and it is by no means a recent 
idea.  Many designs have been proposed for a variety of tasks, and 
a few have shown promising success.  But the space of possible designs
is enormous, and the part so far explored is relatively small.
The goal of this thesis is to significantly further the
exploration of the \sll\ design space.
In order to advance, we must not only find previously uncharted
areas of the design space, but also build tools that help us to
navigate through it.  This thesis contributes in both of these ways.

\section{Theoretical Contributions}

The theoretical component of this thesis comprises two main elements.
First, it proposes an architectural theory of statistical natural
language processing that identifies a new class of \sll\ designs.
Second, it describes work on a mathematical theory of training data
requirements for \sll\ systems that constitutes a powerful tool
for selecting such designs.
The following are brief outlines of the two theories.
\begin{description}

\item[Meaning distributions:]  Existing statistical language learning models 
are defined in terms of lexical and syntactic representations of language.  
Probability distributions generally capture only grammatical knowledge.  
The architectural theory proposed in this thesis 
advocates statistical models defined in 
terms of semantic representations of language.  
Rather than representing grammatical knowledge probabilistically,
it views grammatical knowledge as a form of constraint.  
Syntactic structures inherit 
their probability distributions from semantic forms through these constraints.
The aim of this theory is to suggest new designs.  Thus, 
evaluation of the theory
must come from using it to build \sll\ systems and testing their
performance.  The value of the theory lies in pointing out a promising
direction for exploring the design space.

\item[Data requirements:]  The amount of text used to train a statistical 
language learning system is crucial to its performance.  Since there is no 
well-known theory that can predict the amount of training data necessary, 
the prevalent methodology in \sll\ research is to get as much text as you can 
and see if the your chosen model works.  
However, practical considerations
of data availability have a strong impact on model design.
Informed navigation of the design space rests on being able to
predict data requirements.
In this thesis, a framework for 
building a predictive theory is developed and several results are given that 
represent the first steps toward a general theory of data requirements.

\end{description}

Both of these theories have been investigated through experiments
on statistical noun compound analysis.

\section{Experiments on Noun Compounds}

The experimental component of this work concerns noun compounds in 
English.  Noun compounds are common constructions
exemplified by the last three words in example~\ref{eg:intro_cn}.
\begin{examples}
    \item This time, let's avoid buying those {\em styrofoam dinner plates}.    
	\label{eg:intro_cn}
\end{examples}
Because noun compounds are frequent, highly ambiguous and
require a great deal of knowledge to analyse,
understanding them represents an ideal problem though which
\sll\ designs can be explored.
Understanding noun compounds requires performance of at least two tasks:  

\begin{description}

\item[Parsing:] First, the syntactic structure can be ambiguous.  
Is \lingform{styrofoam dinner} a constituent (in the grammatical sense) 
or is \lingform{dinner plates} one?  
To choose between these two analyses, a parser must 
incorporate knowledge about styrofoam, about dinner and about plates.  This 
might be encoded in the corresponding lexical entries or might be 
represented as independent semantic knowledge.  Regardless of how it is 
stored and used, this knowledge must come from somewhere and statistical 
language learning might be able to provide it.

\item[Semantic Analysis:] Second, the semantic content of noun compounds 
is ambiguous.  In what relation does \lingform{dinner} 
stand to \lingform{plates}?  In order to 
understand the meaning of \lingform{dinner plates} one must be capable of 
identifying the implicit relationship.  Are the plates used for dinner, or 
are they formed from it (as in \lingform{dinner scraps})?  Once again, the 
necessary knowledge about plates and about dinners must come from 
somewhere.

\end{description}

An approach to parsing noun compounds using the meaning distributions
theory is reported in this thesis.  
It uses the text of an encyclopedia to acquire conceptual associations 
statistically and uses these to parse noun 
compounds. The results show that the method exhibits a high degree of 
accuracy within the limitations imposed by the formulation of the task.  The 
probabilistic model upon which the method is based illustrates an 
application of the meaning distributions theory.
In a series of experiments designed to make empirical 
comparisons, this model outperforms alternative models
based on previously proposed algorithms.
This is evidence that the new design territory identified by the
meaning distributions theory is fertile.

Experiments on a statistical method for paraphrasing noun compounds 
are also reported in this thesis.  This method uses 
statistics about prepositional phrases to select a paraphrase 
for a given noun compound
--- a strategy suggested by the meaning distributions theory.
This is the first statistical model of this problem.
The resulting performance is significantly better than the baseline,
but is difficult to evaluate both because comparable evaluation
of other approaches is rarely performed and because no applicable
theory of data requirements is available.
This lends further support to the position that a predictive theory 
of data requirements is urgently needed.

\section{An Overview of the Thesis}

In chapter~\ref{ch:background}, I will review the relevant literature in two 
areas, statistical language learning and noun compounds.  Readers who are 
familiar with the former should find no difficulty in skipping that half.  
Chapters~\ref{ch:md} and~\ref{ch:dr} are 
devoted to the two theoretical contributions, the 
meaning distributions theory and the work on data requirements.  

The experimental work is contained in chapter~\ref{ch:experimental}, which 
is divided into two parts, one for each of noun compound parsing and 
noun compound paraphrasing.  Many of the resources used for the 
paraphrasing experiment are described in the part on noun compound 
parsing since the two programs share these.  Finally, 
chapter~\ref{ch:conclusion} summarises the conclusions of the thesis and 
outlines possible extensions for future work.


\cleardoublepage

\chapter{Background}
\label{ch:background}


Both statistical language learning and noun compounds
have attracted a great deal of research.  This chapter
will review the work that is relevant to this thesis 
on each of these two topics.
Accordingly, the chapter is divided into two halves, 
the first for statistical language learning 
and the second for noun compounds. 

\section{Introducing Statistical Language Learning} 
\label{part:sn}

In the first half of this chapter I will review prior relevant work in 
statistical \nlp.  My chief aim is to make it accessible, and it is therefore 
fairly introductory.  Readers familiar with the area can easily skip to the 
summary given in section~\ref{sec:sn_review} and continue from there to the 
second half of the chapter.
%
%
%
%
%

In section~\ref{sec:sn_motivations}, I will introduce the general 
approach of statistical \nlp, in which statistics are used to 
extract information from large amounts of text.  This information 
is intended to replace hand-coded rules, with the advantage
that it is automatically acquired.
The information generally takes the
form of parameter values within a probabilistic model.  Such a model
assigns probabilities to the choices facing it in terms of
these parameters.  In addition to the advantage of automatic 
knowledge acquisition, a probabilistic representation also
enhances robustness.

Section~\ref{sec:sn_taggers} illustrates the approach 
with Markov model taggers. These have been shown to perform 
the tagging task with a high degree of accuracy.  A model for parsing, 
probabilistic context free grammars, is discussed in 
section~\ref{sec:sn_grammars}. However the performance of
such parsers is disappointing.  

An important distinction in statistical \nlp\ is that between
supervised and unsupervised learning.
In section~\ref{sec:sn_supervised}, I will discuss both of these
and cover several important techniques that have been developed for
the latter.  Without unsupervised techniques, the labour involved 
in marking up training data threatens to undermine the 
advantage of automatic acquisition.

The final three sections discuss approaches to enhancing probabilistic
parsers through sensitivity to words (section~\ref{sec:sn_lexical}),
specialising the model to handle particular grammatical constructs
(section~\ref{sec:sn_specialised}) and grouping words together
to reduce data requirements (section~\ref{sec:sn_conceptual}).
Each of these approaches bears strong relevance to the work
reported in this thesis.

\subsection{Motivations and Advantages of SLL} 
\label{sec:sn_motivations} 

It is now more than twenty years since Winograd~(1972) showed how to 
build a sophisticated natural language understanding system for a limited 
(and admittedly artificial) domain.  Yet the field is still in pursuit of 
technology that will perform similar tasks on unconstrained text.  Scaling up 
has proven to be extremely challenging, the requirement for ever more 
knowledge being seemingly unbounded.  Even if we accept for the moment 
that broad coverage \nlp\ is out of reach and focus instead on restricted 
domains, there is substantial effort involved in manually encoding the 
requisite knowledge.  Knowledge acquisition is an important process in \nlp, 
one that is critical to the success of any non-trivial system.

One currently popular, but still only promising, answer is to automatically 
learn from existing texts.  For quite some time, linguists have been building 
\newterm{corpora} (large text collections in appropriate electronic format) 
often annotated with linguistic analyses, and some of these are 
freely available.  In addition, 
there are vast volumes of electronically stored text.  These resources 
represent a knowledge source that, if tapped appropriately, promises barely 
imaginable detail and coverage and is growing continuously.  The challenge 
is to collect that knowledge without having to first solve all the problems 
of natural language understanding in order to get it.  

The science of statistics is concerned with deriving general conclusions 
about the behaviour of a system from a large amount of data, each
element of which is uncertain.  Viewing language as a system and 
making a few simplistic 
assumptions, we can convert text into exactly this kind of data, uncertainty 
arising from the simplifications.  Thus, statistical methods become powerful 
learning tools; applied correctly, they can extract general conclusions about 
the behaviour of language.  If the conclusions drawn can be arranged to 
coincide with the knowledge required for \nlp\ tasks, then knowledge 
acquisition can be entirely automated.  The vision of feeding text in one side 
and collecting knowledge at the other is a highly attractive conception of 
knowledge acquisition in \nlp.

Statistical language learning techniques have rapidly risen 
to occupy a substantial fraction of the research in \nlp\ over the past five 
years.  Starting out as a method for automatically assigning 
part of speech tags to 
corpora, \sll\ has spread to address a wide variety of tasks, including general 
parsing (Jelinek~\etal,~1992),  learning verb subcategorisations (Brent,~1993) 
and thesaurus categories (Grefenstette,~1994), sense disambiguation 
(Yarowsky,~1992), relative clause (Fisher and Riloff,~1992) and 
prepositional phrase attachment (Hindle and Rooth,~1993) and even 
anaphora resolution (Dagan and Itai,~1990).  The rise of statistical methods 
can also be seen as contributing to a shift in emphasis toward measuring 
development costs and evaluating performance quantitatively, as embodied 
by competitive meetings such as the message understanding conferences 
(Chinchor~\etal,~1993).

\subsubsection*{Some foundations of statistics}

Statistics is founded on probability theory.  It is impossible to give an 
adequate introduction to statistics and probability theory here.  An 
introduction to general statistical theory can be found in Walpole and 
Myers~(1993) and to categorical statistics in Agresti~(1990).  For an 
introduction to probabilistic representation and inference within 
artificial intelligence see Pearl~(1988).  
It will, however, be useful to briefly 
describe some basic elements here, so that the reader unfamiliar with the 
topic will be equipped with some understanding of the terminology.  In 
doing so I will skate over many important distinctions.  Set theory will be 
assumed.
 
A \newterm{random sample} is a set of observations, where the outcome of 
each observation is unaffected by earlier or later observations in the sample.  
The outcome of any given observation is called a \newterm{random 
variable}, usually denoted by a capital letter such as $X$.  For example, a 
geologist might observe the concentration of a mineral in the soil of several 
fields.  Since it is reasonable to assume that the outcome in one field is not 
changed by the outcome of the others, this constitutes a random sample.  The 
\newterm{population} from which this sample is drawn is the set of all fields 
which the geologist might have picked to measure.   

A \newterm{probabilistic model} is a mathematical construction that models 
the behaviour of the system (in this case mineral deposits).  Probabilistic 
models involve one or more random variables, each of which has an event 
space, the set of all possible distinct outcomes of an observation (in this case 
the event space is real-valued, ranging from zero to 1 million parts per 
million).  In most probabilistic models of language, the event space is 
discrete (say the set of all words), but can be infinite (the set of all 
sentences).  

For discrete event spaces, the \newterm{probability} of an outcome in the 
event space is a real number between 0 and 1 (inclusive) that represents 
how often we can expect that outcome to 
occur.\footnote{For continuous event 
spaces, we need to define probability density and probability then follows 
for subsets of the event space.}  It is formally written as $\Pr(X=a)$
where $a$ is the outcome.  If we 
imagine taking larger and larger numbers of random samples, the proportion 
of observations with outcome $a$ converges in the limit to the probability of 
$a$.  It is easy to see that probability is defined for any subset of the 
event space, simply by summing the probabilities of each event in the subset: 
$\Pr(X \in \{a, b\}) = \Pr(X=a) + \Pr(X=b)$. 

A \newterm{distribution} is an assignment of probabilities to each possible 
outcome in the event space, where the sum of all the probabilities is 1.  The 
constraint that probabilities must sum to 1 is necessary because, considering 
the entire event space as a subset of itself, the proportion of outcomes that 
are in the event space converges to 1 (in fact, it is always 1).

In general, probabilistic models give a parameterised class of distributions 
such that supplying values for the parameters fixes a particular distribution.  
For example, a binomial model uses an event space with two elements (say 
$a$ and $b$) and contains one parameter, $p$.  Given $p$, it assigns 
probability $p$ to outcome $a$ and $1-p$ to outcome $b$: that is, 
$\Pr(X=a) = p = 1-\Pr(X=b)$.  

A statistical learning algorithm uses a probabilistic model in order to make 
predictions.  To do this it first has to learn what the correct parameter values 
should be.  Thus, knowledge acquisition is performed by estimating values 
for the parameters of a model.  The process of acquiring these values from a 
random sample of training data is called \newterm{parameter estimation}.

A common and simple estimation method is called the \newterm{maximum 
likelihood estimator} (\acronym{mle}).  This method chooses the parameter 
values that give the observed sample the highest probability.  That is, out of 
the class of distributions given by the model, find the one that assigns the 
highest probability to the observations.  For binomial models, this sets $p$ 
to be the number of observations with outcome $a$ divided by the total 
number of observations (the sample size).

More complex event spaces can be composed from simpler ones.  A 
\newterm{joint distribution} is a distribution over an event space formed by 
the cartesian cross-product of other event spaces.  For example, consider the 
set of all words and the set of all parts of speech.  Selecting a word at 
random from a large corpus is an observation that defines a random variable 
over the words, $W$.  If we also observe the part of speech of that word, 
this observation defines a random variable, $T$ over the parts of speech.  
The cross-product of these two event spaces is a set of pairs $(w, t)$, 
representing the possible outcomes.  The distribution over this event space, 
written $\Pr(W, T)$, is referred to as a joint distribution of words and 
parts of speech.  

It is useful to define the notion of \newterm{conditional probability}, written 
$\Pr(X=a | Y=c)$.  Intuitively, this represents how often we can expect $b$ 
to occur if we know $c$ occurs at the same time.  For example, 
\lingform{green} is usually an adjective, so the probability 
$\Pr(T=\tag{adj} | W=\mbox{\lingform{green}})$ will be high.  Without 
the information that the word is \lingform{green} the probability of an 
adjective $\Pr(T=\tag{adj})$ is much lower.

Finally, we say that two random variables $X$ and $Y$ are 
\newterm{independent} when the probability of the joint event $(x, y)$, 
that is $\Pr(X=x, Y=y)$ is equal to the product of the 
two individual probabilities $\Pr(X=x)$ and $\Pr(Y=y)$.  
This is mathematically equivalent to the conditional 
and unconditional probabilities always being equal, that is, 
$\Pr(Y=y | X=x) = \Pr(Y=y)$ for all $x$ and $y$.  
Intuitively, the value of measurement represented by $X$
has no affect on the value of that represented by $Y$.

\subsubsection*{Applying statistics to learning language}

A thorough and accessible review of probability theory in computational 
linguistics can be found in Magerman~(1992) and is recommended for 
anyone interested in statistical \nlp\ methods.

The general architecture for statistical language learning includes the 
following essential elements.
\begin{enumerate}
\item A means of construing grammatical (or other analytical) relations as 
events.
\item A parameterised model of language production in terms of such events.
\item A statistical technique for estimating the model parameters from 
corpora.
\item A method for evaluating the likelihood of an event given the model 
and its parameters.
\end{enumerate}

In sections~\ref{sec:sn_taggers} through~\ref{sec:sn_conceptual} below a 
number of examples of probabilistic models will be discussed.  Commonly 
used examples of corpora include the Penn Treebank, Grolier's 
encyclopedia, collections from the \publicationname{Associated 
Press} newswire and the 
\acronym{atis} speech corpus.  The earliest corpora were built by linguists, 
the most famous being the Brown corpus constructed in 1963 at Brown 
University by Francis and Ku\v{c}era~(1982).  This consists of slightly over 
1 million words drawn from works written in 1961 and carefully selected to 
represent a range of genres.  As common sense suggests, the characteristics 
of the corpus used require serious consideration and in 
section~\ref{sec:md_register} I will return to this topic.

The size of a corpus is usually measured in words, or more properly 
occurrences of words.  The wordform \lingform{the} occurs many times in 
any English corpus, and increases the size of the corpus each time it occurs.  
The distinction between word occurrences and wordforms
is sufficiently important in \sll\ to warrant special 
terminology.  A \newterm{type} is a distinct value of a random variable; for 
example the word \lingform{the} is a type when the random variable ranges 
over words.  A \newterm{token} is an instance of a type.  On any particular 
occasion, the momentary value of a random variable is called a token; for 
example, each occurrence of the word \lingform{the} is a token.

This distinction becomes especially important when measuring 
the accuracy of an \nlp\ 
system.  Often test sets are designed to contain only distinct test cases.  That 
is, test cases do not occur multiple times in the test set.  If a proposed 
algorithm correctly handles 90\% of such a test set, then we expect it to 
handle roughly nine out of every ten {\em types}.  However, if some types 
occur more often than others (that is, tokens of one type are more frequent 
than tokens of another type), then the expected accuracy in practice could 
differ greatly from 90\%.  If the 10\% of types incorrectly handled represent 
80\% of the tokens, then the practical expected accuracy is a mere 20\%.  To 
properly evaluate the expected performance of an algorithm, the test set 
should contain multiple occurrences of test cases with distribution matching 
that of {\em tokens}.

\subsubsection*{Benefits of statistical language learning}

While I have positioned statistical learning methods as primarily motivated 
by the need to automatically acquire knowledge, there are other 
advantages.
\begin{description}
\item[Adaptability:] Language varies across time, place, situation and even 
from individual to individual.  For example, Lehman~(1992) shows that the 
usefulness of a limited domain parser is significantly 
increased if it adapts its 
grammar to the usage of individual users.  In principle, statistical \nlp\ 
systems automatically adapt to whatever type of language they are trained 
on.  If in one domain prepositional phrases attach to verbs highly 
frequently, then the system can learn and exploit this fact.  When retrained 
on text from a different domain where they usually attach nominally, the 
system can change in response to the new environment.  In contrast, any 
hand-coded knowledge base must be explicitly modified to allow for the 
new environment.
\item[Robustness:] Because of the explicit representation of uncertainty in 
probabilistic representations, knowledge is rarely absolute.  When 
unforeseen circumstances arise, probabilistic information can be used to 
make informed guesses, so that performance degrades gracefully.  The idea 
that linguistic restrictions should be regarded as 
preferences rather than strict 
rules was suggested by Wilks~(1975) and shown to provide measurable 
performance improvements in message understanding by Grishman and 
Sterling~(1989).  Probabilistic models are a mathematically principled 
representation for such preferences.
\item[Combination:] Weischedel~\etal~(1990) identify the 
\newterm{combination problem}, that is the difficulty of integrating knowledge 
from diverse sources into one decision procedure, as a key issue for \nlp.  
Many different kinds of cues affect language interpretation.  Finding 
appropriate methods for making use of these cues is important, but a means 
of combining the results from multiple methods is also necessary.  Where 
this problem has arisen in the past, many researchers have adopted 
\foreign{ad hoc} weighting or scoring schemes 
(see for example, McRoy,~1992).  In 
principle, probabilistic models provide for combination of information in a 
mathematically well-founded manner.  In fact, this is one of the central 
arguments put forward for the use of Bayesian network representations 
(Pearl,~1988:14--23).
\item[Efficiency:] The kinds of analysis pursued in \nlp\ are often formulated 
in terms of searching a space that is constrained by relatively simple rules, 
for example a grammar.  With any moderately sized grammar, there is a 
great deal of ambiguity and this results in rather unwieldy search 
complexity.  To build practical broad coverage natural language processors, 
a means for controlling the search is needed.  The classical artificial 
intelligence solution is to employ a heuristic during the search to order 
choices.  The heuristic should rank highly those options which are most 
likely to result in a successful analysis if they are chosen.  This is precisely 
what a probabilistic model provides: a set of empirically derived 
preferences.  Not only do statistical methods allow the correct answer to be 
derived (because appropriate knowledge is acquired automatically), but they 
also allow the analyser to arrive at the answer more quickly.
\end{description}

In summary, statistical methods are based on the mathematical foundation of 
probability theory and promise to allow us to make use of the continually 
growing resources of on-line texts to overcome the knowledge acquisition 
problem.

There are two other approaches to knowledge acquisition which might be 
viewed as alternatives to \sll.  First, the \acronym{cyc} project was 
established as a decade long project to build a large knowledge base of 
common sense rules (Lenat~\etal,~1986).  The project appears to have failed 
to meet its goals at this stage, although a large amount of knowledge has 
been encoded.  

Second, a large number of research groups have built special purpose 
systems for extracting knowledge from machine-readable dictionaries such 
as Longmans Dictionary of Contemporary English (\acronym{ldoce}; 
Proctor,~1978).  A survey is given in Lauer~(1992).  Early work (such as 
Chodorow~\etal,~1985) used simple patterns to extract hypernymic 
relations, and was extended to extract other kinds of 
lexical-semantic relationships (Markowitz~\etal,~1986; Janssen,~1990).  
Sense ambiguity of words in dictionary definitions often caused errors, 
leading to semi-automatic versions (Calzolari and Picci,~1989; 
Copestake,~1990).  Other research has applied these techniques directly to 
\nlp\ tasks, such as prepositional phrase attachment (Jensen and 
Binot,~1987) and sense-disambiguation (Braden-Harder,~1993).  The latter 
task has also been approached through both connectionist networks 
constructed from dictionary definitions (Veronis~\etal,~1991) and statistical 
word co-occurrence measures derived from \acronym{ldoce} 
(Wilks~\etal,~1990; Guthrie~\etal,~1991; Bruce and Guthrie,~1992; 
Cowie~\etal,~1992).  It is interesting to note that all of the latter work 
uses essentially statistical language learning techniques, using 
an on-line dictionary as a corpus.  
Viewed in this way, the dictionary is a tiny resource 
relative to the vast amount of text typically used for \sll.
More recent work combines the information provided by dictionary
definitions with co-occurrence statistics from a corpus (Luk,~1995).


\subsection{An Example: Probabilistic Tagging}  
\label{sec:sn_taggers} 

In this section I will use Markov model taggers to illustrate the use of 
probabilistic models.  The part of speech tagging problem is to take a text 
and assign to each word one of a fixed set of parts of speech such as singular 
noun, adjective, determiner and so on.  
In all taggers I am aware of, text is tagged 
sentence by sentence on the assumption that parts of speech are never 
affected by context beyond the sentence in which they are contained.

Markov model taggers were the earliest successful application of statistical 
learning to \nlp, a working one being reported by De Rose~(1988).  This 
tagger trains on part of speech bigrams (pairs of adjacent part of speech 
tags) and is therefore called a \newterm{bigram tagger}.  The obvious 
extension, trigram taggers (trained on part of speech triples) have found 
widespread use within the research community and are generally agreed to 
have effectively solved the part of speech tagging 
problem.\footnote{Although significant problems remain with unknown 
words.  See for example, Weischedel~\etal~(1993).}

Markov model tagging involves the four stage architecture outlined in 
section~\ref{sec:sn_motivations}.
In what follows, I will consider each of 
these for the bigram case.  My purpose is to give an intuitive understanding 
of how these models work and I will avoid mathematical detail as much as 
possible.  The reader desiring a deeper understanding should consult either 
Charniak~(1993:Ch.~3) or Allen~(1995:Ch.~7).

\subsubsection*{Event space}

First, parts of speech must be formally construed as events.  To do this we 
imagine drawing a random sentence from a corpus.  Suppose it contains $n$ 
words, $w_1 w_2\ldots w_n$, each with a part of speech, $t_1 
t_2\ldots t_n$.  Formally, this observation yields a joint event in the event 
space $W^n \times T^n$, thus defining $2n$ random variables which we can 
call $W_i$ and $T_i$ for $1 \leq i \leq n$.  Given a joint distribution over 
this composite event space, we can compute the conditional probability 
$\Pr(t_1 t_2\ldots t_n | w_1 w_2\ldots w_n)$ for all possible tag sequences 
and choose the most probable.\footnote{As is usual,
I have omitted the random variables from the probability expression here
because they can be inferred.  Technically the expression should
be $\Pr(T_1=t_1, T_2=t_2\ldots | W_1=w_1,\ldots)$.  Similar abuses of
notation will be made throughout this thesis.}

\subsubsection*{Probabilistic model}

Second, a probabilistic model is needed that provides the joint distribution 
once certain parameters have been estimated.  The bigram Markov model 
does this by making certain assumptions about the way in which sentences 
are generated.  What the model supposes is equivalent to the following 
imaginary generation process.  First make $n$ choices from $T$ to create a 
sequence of part of speech tags where the choice for the $i$th tag depends 
only on the $(i-1)$th tag.  Second, choose for each tag a word that can have 
that tag, where the choice of word depends only on the tag.  In this way
a sequence of tagged words is created.  Each of these $2n$ choices is 
probabilistic and the overall result is that an element of $W^n \times T^n$ is 
probabilistically selected.  

Since each element of $W^n \times T^n$ has 
some probability of being selected in this way, this process defines the 
required joint distribution.  The theory of probability allows us to compute 
this distribution if we know the distributions for the $2n$ choices.  
Since the model assumes that the choices are independent of the
position within the sequence, we only require two distributions, 
one for the tag choices and one for the word choices.

Formally, the distributions needed are written $\Pr(T_i | T_{i-1})$ and 
$\Pr(W_i | T_i)$.  The assumptions of the model state that words are only 
dependent on their respective tags, and that tags are only dependent on the 
previous tag.
\begin{eqnarray}
\lefteqn{\hspace{-1.5cm} 
\Pr(W_i=w | T_1=t_1,\ldots T_n=t_n, W_1=w_1,\ldots W_{i-1}=w_{i-1}, 
W_{i+1}=w_{i+1},\ldots W_{n}=w_{n}) = } \hspace{3in} & & \nonumber \\
 &  & \Pr(W_i=w_i | T_i=t_i) 
\label{eq:sn_taggers_lexgen}
\end{eqnarray}
\begin{eqnarray}
\lefteqn{\hspace{-1.3cm} \Pr(T_i=t_i | T_1=t_1,\ldots T_{i-1}=t_{i-1},
T_{i+1}=t_{i+1},\ldots T_n=t_n, W_1=w_1,\ldots W_{n}=w_{n}) = } 
\hspace{3in} & & \nonumber \\
 &  & \Pr(T_i=t_i | T_{i-1}=t_{i-1}) 
\label{eq:sn_taggers_taggen}
\end{eqnarray}

Importantly, it is easy to find examples where these assumptions are clearly 
violated.  In example~\ref{eg:sn_taggers_bad}\ref{eg:sn_taggers_bad_lex1}, 
the probability that the first word is \lingform{young} (given that all
tags and following word are known) is substantially
 
higher than it is in example~\ref{eg:sn_taggers_bad}\ref{eg:sn_taggers_bad_lex2},
contradicting equation~\ref{eq:sn_taggers_lexgen}.  In 
example~\ref{eg:sn_taggers_bad}\ref{eg:sn_taggers_bad_tag}, the 
probability that the third tag (in this case the tag of \lingform{eat}) is plural 
is substantially increased by the information that the first tag (of the word 
\lingform{Bears}) is plural too, because the verb is required to agree in 
number with its subject.  This contradicts 
equation~\ref{eq:sn_taggers_taggen}.
\begin{examples}
  \item \label{eg:sn_taggers_bad}
    \begin{subexamples}
      \item young/\tag{adj} lad/\tag{noun-sing}  
	\label{eg:sn_taggers_bad_lex1}
      \item young/\tag{adj}  fact/\tag{noun-sing}  
	\label{eg:sn_taggers_bad_lex2}
      \item Bears/\tag{noun-plur} often/\tag{adv} eat/\tag{verb-plur}
	\label{eg:sn_taggers_bad_tag}
    \end{subexamples}
\end{examples}
The hope, and one that works out in practice, is that these violations occur 
sufficiently infrequently and have sufficiently little affect on the tagging 
decisions to undermine the accuracy of the model.

The purpose of the model is to specify the joint distribution over $W^n 
\times T^n$ in terms of a small number of parameters.  In this case, the 
parameters are the probabilities comprising the two distributions $\Pr(T_i | 
T_{i-1})$ and $\Pr(W_i | T_i)$. This brings us to the next stage, estimating 
these parameters from a corpus.

\subsubsection*{Parameter estimation}

Third, applying a statistical method to data derived from a large text corpus 
allows the parameters of the model to be estimated.  For example, maximum 
likelihood estimates can be derived for the probabilities of the distribution 
$\Pr(T_i | T_{i-1})$ from a tagged corpus.  This distribution contains one 
probability for each pair of tags $(t_{i-1}, t_i)$, which we estimate by 
counting the number of occurrences of tag bigrams, that is sequences 
$t_{i-1} t_i$, in the corpus.  
The maximum likelihood estimate under the Markov 
model is given by equation~\ref{eq:sn_taggers_mle} in which bigram counts 
are denoted $\countfn(t_{i-1}, t_i)$.
\begin{equation}
\Pr(T_i=t_i | T_{i-1}=t_{i-1}) \stackrel{\rm MLE}{=}
\frac{\countfn(t_{i-1}, t_i)}{\sum_{t \in T} \countfn(t_{i-1}, t)}
\label{eq:sn_taggers_mle}
\end{equation}
The other distribution, $\Pr(W_i | T_i)$, is estimated similarly.  The process 
of computing these estimates from a corpus is called training.  Once trained, 
the model contains detailed information about the way in which words can 
be tagged and the typical sequencing of tags.  As De Rose~(1988) points 
out, this automatically acquired information can replace the costly 
hand-coded knowledge used in earlier taggers such as \acronym{claws}
(Marshall,~1983).

\subsubsection*{Analysis}

Finally, we are ready to apply the model to tagging fresh text.  
As noted above, when presented with a sentence, say $w_1 w_2\ldots w_n$, 
the tagger proceeds by computing the conditional probability 
$\Pr(t_1 t_2\ldots t_n | w_1 w_2\ldots w_n)$ for 
all possible tag sequences and choosing the most probable.  Fortunately, it is 
not necessary to explicitly construct every possible tag sequence, which 
would be computationally intractable.  The most probable sequence can be 
found quite efficiently using the Viterbi algorithm (Allen,~1995:201--203).  
This algorithm uses the probabilities of tag subsequences to prune the search 
for the most probable tag sequence, and is a good example of how a 
probabilistic representation can make processing more efficient.

To give an idea of how successful this approach to tagging is, 
Weischedel~\etal~(1993) report that their trigram tagger, trained on the 4 
million word Penn Treebank, achieves 97\% correctly tagged words.  Since 
human annotators disagree on about 3\% of tags, this rivals human tagging 
accuracy.  This degree of success should be qualified though by noting that a 
system which gives each word its most common tag regardless of context 
yields 91\% accuracy (Charniak,~1993:49).

Because training data can be unreliable, it is often beneficial to combine the 
predictions of more than one probabilistic model, using a process called 
\newterm{smoothing}.  This is especially important when data is sparse, 
leading to statistical uncertainty in parameter estimates.  For example, even 
though the trigram tagging model is more sensitive to context than the 
bigram model described above, it has many more parameters, so it requires 
more training data to acquire accurate parameter settings.  When data is 
limited, smoothing a bigram and a trigram model together can
result in greater performance than using either on their own. 

A frequently used smoothing method is called \newterm{deleted 
interpolation}, where the probabilities predicted by two or more
models are combined together by a weighted summation (Jelinek,~1990:456).  
The weightings given to each model, called the \newterm{interpolation 
coefficients}, are optimised empirically.  
There are many variations on smoothing but a 
survey of them would would take us too far afield. For now it
suffices to introduce the basic idea.

The promising start made by probabilistic taggers has inspired a great deal 
of work in applying statistical language learning to problems other
than part of speech tagging.  
A more challenging application of the \sll\ approach is probabilistic parsing,
which I will now introduce.  

\subsection{The Challenge of Probabilistic Grammars}
\label{sec:sn_grammars} 

Large scale grammars often generate many possible readings for a sentence 
and submitting all of these for semantic processing is computationally 
infeasible.  Instead, given a part of speech tagged sentence and a grammar, 
we would like to find the single most likely parse for the sentence according 
to the grammar.  To do this, we need a model of the distribution of parse 
trees, called a \newterm{probabilistic grammar}.  It is then a matter of 
computing the probability of each possible parse for the sentence and 
selecting the parse with highest probability.  Parsers for probabilistic 
grammars also benefit from the guidance of the probabilistic model and can 
be expected to perform better as a result.  

The simplest approach is to associate probabilities with each application of a 
rewrite rule in a context free grammar.  This approach, 
\newterm{probabilistic context free grammar}, has been studied in some 
detail (Jelinek~\etal,~1992) and is a good example of how statistical models 
can be structured in more sophisticated ways than simple sequences.

A context free grammar gives various possible rewrite rules for each 
non-terminal symbol, representing the choice available to a generator.  A 
{\em probabilistic} context free grammar construes 
each application of a rewrite rule as an event.  
It incorporates one distribution for each non-terminal, 
representing the probability of each possible rewrite rule with that 
non-terminal on the left-hand side.  For example, suppose that the possible 
rewrite rules with \nonterm{np} on the left-hand side are those in 
example~\ref{eg:sn_grammars}.
\begin{examples}
    \item    \label{eg:sn_grammars}
        \begin{subexamples}
            \item (0.4) \nonterm{np} $\rightarrow$ \nonterm{det} \nonterm{n}
            \item (0.2) \nonterm{np} $\rightarrow$ 
			\nonterm{det} \nonterm{ap} \nonterm{n}
            \item (0.1) \nonterm{np} $\rightarrow$ \nonterm{n} \nonterm{pp}
            \item (0.3) \nonterm{np} $\rightarrow$ \nonterm{n}
        \end{subexamples}
\end{examples}
The probabilities given at the left of each rule represent the chance of 
observing that an \nonterm{np} randomly selected from all parses is 
rewritten by that rule.  Notice that the probabilities add to 1 since these 
rules are all of the allowed rewrite rules for an \nonterm{np}.

Now each interior node of a parse tree represents the application of one rule, 
whose left-hand side is the node's label.  The probabilistic context free 
grammar model assumes that the probability of applying a rule depends only 
on the non-terminal being rewritten.  The choice at each node is made 
independently of the choices at the other nodes.  Probability theory then 
states that the probability of the parse tree as a whole is the product of the 
probability of each choice.
\begin{equation}
\Pr(T) = \prod_{n \in \mbox{\it interior}(T)} Pr(\mbox{\it rule}(n))
\end{equation}
Thus, we have a distribution over the set of all possible parse trees.  

The parameters of the model are the rule probabilities, so there is one 
parameter for each grammar rule.  A simple way to estimate these 
parameters is to take a parsed corpus and use maximum likelihood to 
estimate the probability of each rule.  For example, to find the probability of 
\nonterm{np} $\rightarrow$ \nonterm{det} \nonterm{n}, count the number 
of times that this rule is used in parses in the corpus and divide by the 
number of times an \nonterm{np} occurs.

Once these probabilities have been estimated, they can be used not only to 
rank the possible parses, but also to guide parsers in their search for the best 
analysis.  The Inside algorithm (Jelinek~\etal,~1992) provides an efficient 
way to compute the probability that a given non-terminal spans a given word 
string.  This can be used to guide a chart parser (see~Allen,~1995:213--215,
for an example).  This approach forces the parser to use a bottom-up 
method; however Corazza~\etal~(1994) show how to generalise the 
technique to guide an island-driven parser.

There are other probabilistic models of parsing, each varying in the way in 
which grammatical structure is construed as events.  For example, 
Seneff~(1992) builds augmented transition networks (a grammatical 
representation described in Woods,~1986) from a context free grammar and 
takes the transitions within these networks to be events.  Whenever more 
than one transition is available in the network, a probability distribution is 
estimated and these used to define the probability of different possible parse 
trees.

Briscoe and Carroll~(1993) explore a similar idea with an 
\acronym{lr}-parser.  Here \acronym{lr}-parse actions are the events 
represented probabilistically by the model.  The \acronym{lr}-parse table is 
first constructed from a generalised phrase structure grammar (Pollard and 
Sag,~1987) and then the probabilities of different parse actions are 
estimated.  Both this and Seneff's~(1992) model have the advantage of being 
integrated with efficient parsing systems making guidance of parsing direct 
and simple.  Training is performed by running the parser over an already 
parsed corpus and counting the appropriate parse events.

A somewhat more radical approach is to associate probabilities with 
arbitrary parse subtrees.  In the data-oriented parsing (\acronym{dop}) 
approach, the grammatical events are compositions where one subtree is 
inserted into another (Bod,~1993).  The probability of a parse is derived 
from the probabilities of subtrees.  However, unlike probabilistic
context free grammars, the decomposition is not required to 
continue all the way down to the level of individual parts of speech.
Large subtrees that appear in the training corpus are 
associated with additional probability that is not derived from their parts.
Every subtree occurring in the corpus, including entire parsed sentences, is 
associated with a probability.  The probabilistic model assigns probabilities 
to parses according to the likelihood of the subtrees that they contain, but it 
favours shorter sequences of compositions.  This encourages larger subtrees 
to be composed, so that common phrases can be treated as atomic
events, rather than being decomposed.

One problem with this approach is that there is no efficient 
algorithm for computing the probability of a parse.  To overcome this, 
Bod~(1993) uses Monte Carlo simulations to choose the most likely parse;
however, this strategy could not be used to guide a parser because it is too 
computationally expensive.

In all these approaches, probability is associated with constituents, that is, 
contiguous word sequences that form phrases and clauses.  This is because 
they are based upon grammatical formalisms which emphasise constituency.  
An alternative grammatical framework, dependency grammar 
(see for example, Mel'cuk,~1988), 
emphasises grammatical relationships that can span 
non-contiguous word groupings.  Milward~(1992) argues that this 
representation facilitates incremental construction of semantic forms.  The 
distinction between constituent-based and dependency-based grammar will 
become important later in this thesis.  Therefore, it is worth mentioning two 
proposals for probabilistic grammars based on dependency relations rather 
than constituents.\footnote{Carroll and Charniak~(1992) call their
model a probabilistic dependency grammar, but the model is actually
based on constituency.  I will discuss this work in 
section~\ref{sec:sn_supervised}.}
\begin{itemize}
\item Link grammar is based on typed dependency relations.  A probabilistic 
model for this grammar formalism has been developed by 
Lafferty~\etal~(1992).
\item Charniak~(1993:130--134) proposes a variant of the probabilistic 
context free model that conditions on the head of constituents, thus capturing 
dependency relations within a constituent-based formalism.
\end{itemize}
Both of these proposals involve the probabilities of grammatical structures 
being conditioned on words (as opposed to merely parts of speech).  
Unfortunately, no implementation of either has been reported
at this time.  

\subsection{Supervised and Unsupervised Training} 
\label{sec:sn_supervised}

All the training methods considered so far use a corpus that is annotated with 
answers.  Taggers use tagged training data and parsers use parsed training 
data.  This is called \newterm{supervised} learning.  As one might imagine, 
annotating a corpus with answers involves substantial effort.  Since one key 
motivation for statistical language learning is to avoid the effort of manually 
coding knowledge, reliance on supervised learning runs the risk of costing as 
much as it saves.

While some medium-sized parsed corpora are available, the effort involved 
in constructing them is enormous and relies on semi-automatic methods to 
be feasible (see for example, Marcus~\etal,~1993).  If \sll\ is to tap into the 
enormous volumes of on-line text, unsupervised techniques must be 
developed.  In this section I will consider some of the methods used to train 
probabilistic models using unannotated data.

\subsubsection*{Expectation maximisation}

The most common unsupervised algorithm for Markov model taggers is 
called the \newterm{forward-backward} method, a member of the class 
called \newterm{expectation maximisation} algorithms.  These algorithms 
iterate through a sequence of trained models, using earlier models in the 
sequence to modulate the training of later models.  To initiate the process, a 
simple probabilistic model is chosen that assigns some non-zero probability 
to each possible tag sequence for a sentence.  

Now each sentence of the untagged training data can be associated with a set 
of possible tag sequences, each with a probability according to the initial 
model.  One can view this as a kind of tagged data where each word has 
several tags, each with a probability weighting, a kind of weighted 
superposition of tags.  From this data, a new model can be built using counts 
of these superimposed tags.  The counts are weighted by the probabilities of 
the tags.  For example, suppose that the initial model gives all allowed tags  
for a word equal probability.  The sentence in 
example~\ref{eg:sn_supervised_init} might then have the probabilistic 
superposition of tags shown.
\begin{examples}
\item  
   All/\{$\tag{noun}_{0.33}$, $\tag{det}_{0.33}$, $\tag{adv}_{0.33}$\}
   bears/\{$\tag{verb}_{0.5}$, $\tag{noun}_{0.5}$\} 
   eat/\{$\tag{verb}_{1.0}$\}     \label{eg:sn_supervised_init}
\end{examples}
The data used to constructed the new model will then include the tag triple 
(\tag{det}, \tag{noun}, \tag{verb}) with weighted count 0.17, as well as the 
five other triples.  These counts are combined with those from all the other 
sentences in the training corpus, and used to estimate the parameters of the 
tagging model.  For example, this might lead to a model that assigns the 
probabilities given in example~\ref{eg:sn_supervised_next}.
\begin{examples}
\item  All/\{$\tag{noun}_{0.1}$, $\tag{det}_{0.6}$, $\tag{adv}_{0.3}$\}
	bears/\{$\tag{verb}_{0.3}$, $\tag{noun}_{0.7}$\} 
	eat/\{$\tag{verb}_{1.0}$\}     \label{eg:sn_supervised_next}
\end{examples}
Now this new model can be used to provide weighted counts for another 
iteration, leading to a third model and so on.  The process stops when the 
probabilities assigned by the model converge to stable values.

The choice of initial model does have a significant effect on the performance 
of the final model.  Merialdo~(1994) uses the uniform initial model (as in 
example~\ref{eg:sn_supervised_init}) and reports that tagging accuracy 
converges to 86.6\% after 10 iterations.  When the initial model is provided 
by maximum likelihood estimates from 100 sentences of supervised data, the 
accuracy converged to 92.6\%, so there is an undesirable sensitivity to the 
initial model chosen.

Also, unsupervised training with the forward-backward algorithm falls short 
of the performance achieved by supervised training, at least with the amount 
of training data Merialdo~(1994) uses.  He reports that
a smoothed trigram model using 
maximum likelihood estimates achieved 97.0\% accuracy on the same test 
set and training data.

For probabilistic parsing, there is an analogous algorithm called the 
\newterm{inside-outside} algorithm (Jelinek~\etal,~1992).  The method 
begins with an initial probabilistic grammar which is used to assign each 
sentence of the unparsed training corpus a weighted set of possible parses.  
These are used to generate weighted counts of rule applications and then a 
new model is estimated from these counts.  Again, the process is iterated 
until the probabilities converge.  
Fujisaki~\etal~(1991) report promising results with 72 out of 84 sentences 
parsed correctly (apparently using the inside-outside algorithm, though this is 
not directly stated).  However, the test sentences also appeared in the 
training corpus.

Furthermore, other researchers have not been able to achieve similar 
results even with supervised methods.
\begin{citedquote}{Allen,~1995:212}
Unfortunately, a parser built using [a supervised probabilistic context free 
grammar] turns out not to work as well as you might expect, although it does 
help.  Some researchers have found that these techniques identify the correct 
parse about 50 percent of the time.  It doesn't do better because the 
independence assumptions that need to be made are too radical.
\end{citedquote}
Therefore, there is still a wide margin for improvement in
statistical parsing models.

\subsubsection*{Grammar acquisition}

The inside-outside algorithm assumes that a context free grammar is 
available.  Several researchers have suggested that a grammar could be 
automatically acquired by, for example, beginning with a grammar 
containing all possible rewrite rules and eliminating those that the 
inside-outside algorithm assigns very low probability to.  An obvious 
problem is the huge number of possible rules.  A principled way of limiting 
the possible rules that is inexpensive in terms of labour is required.

Carroll and Charniak~(1992) explore the possibility of limiting the allowed 
rules to those which could be expressed by a dependency grammar.  That is, 
there is exactly one non-terminal symbol for each terminal symbol and all 
rules rewriting a non-terminal symbol must include the corresponding 
terminal symbol on the right-hand side.  In their experiments, a probabilistic 
grammar of this kind is used to generate an artificial corpus.  The goal is to 
use the inside-outside algorithm to automatically reconstruct the original 
grammar.  They conclude that further constraints (they give a set based on 
grammatical dominance relations) are necessary for learning to converge to 
the original grammar.  Though they refer to their grammar as a probabilistic 
dependency grammar, this is slightly misleading because the grammatical 
events are constituent rewrites, not dependency relations.  The probabilistic 
model is identical to that for a probabilistic context free grammar.

Briscoe and Waegner~(1992) take a similar tack, but allow two 
non-terminals for each terminal using the framework of
Jackendoff's~(1977) \={X} theory.  They 
also incorporate number and gender features, and enforce some general 
agreement constraints using a unification scheme.  Again, an artificially 
generated corpus is used; the results are similarly mixed.  It appears that 
something more is needed.

\subsubsection*{Noise elimination}

Apart from the iterative approach offered by expectation maximisation 
algorithms, there are other unsupervised learning techniques.  For example, 
Brent~(1993) uses a combination of simple patterns and statistical 
hypothesis testing to acquire verb subcategorisation frames.  Patterns based 
only on closed class words and morphemes are used to identify words that 
appear to be verbs in each of 6 syntactic configurations (simple transitive, 
infinitive, etc.) with high certainty.  Because the patterns are simple, they 
generate false positives, but most of the time they are correct.  By matching 
a binomial distribution to the resulting data, an estimate of the error rate of 
each pattern is derived.  This is then used to test the hypothesis that the 
observed data might have been caused purely by errors.  When this 
hypothesis can be rejected, the appropriate verb frame is added to the 
lexicon for that verb.

Finally, there is an altogether different technique that is of significance to 
this thesis.  Yarowsky~(1992) uses an on-line thesaurus to exploit 
known word equivalences.  His automatic sense disambiguation 
mechanism aims to use word co-occurrences to select an appropriate sense.  
For each target sense, a profile is collected of all the words that occur within 
50 words (on either side) of that sense in the training corpus.  When 
presented with an ambiguous word, it matches the profiles of the possible 
senses and chooses the sense with the greatest information value.  For 
example, the animal sense of \lingform{crane} has strong association (a 
large information score) with \lingform{species}, \lingform{nest} and so on, 
while the machinery sense of \lingform{crane} has strong association with 
\lingform{machine}, \lingform{engine} and so on.  Upon encountering the 
word \lingform{crane}, Yarowsky's algorithm computes the total 
information score of all surrounding words for the animal sense and for the 
machinery sense, and chooses whichever sense is greater.

The problem is that without a sense tagged corpus, it is not possible to 
collect the profiles for each sense of an ambiguous word.  The training 
algorithm will not be able to distinguish between occurrences of the animal 
sense and those of the machinery sense in the training corpus.  Yarowsky's 
solution is to treat every occurrence of \lingform{crane} as being both the 
machinery and the animal sense, but then add together the profiles of 
synonymous words.  All the profiles of words in the Roget's thesaurus 
category \concept{tools/machinery} are combined to form a profile for the 
machinery sense of \lingform{crane} (and for the appropriate sense of other 
words in that category).  Now all of the word co-occurrences with the animal 
sense of \lingform{crane} will be wrongly added into the profile for the 
machinery sense.  However, this noise will be relatively small since there are 
many words in the \concept{tools/machinery} category.  While the other 
words in that category may also be sense ambiguous and therefore contribute 
further noise, these different sources of noise will be scattered throughout 
the profile.  On the other hand, the correct co-occurrences (the signal) 
should be similar for all words in the same category, and therefore the signal 
should become concentrated and outweigh the noise.

Yarowsky applies this method to some 5000 occurrences of 12 polysemous 
words using Grolier's encyclopedia as a corpus.\footnote{This corpus 
contains 10 million words.}  The algorithm predicts the correct sense for 
92\% of these, which is very successful considering that the only linguistic 
knowledge used is the synonym groupings provided by Roget's thesaurus.  If 
similar performance could be squeezed from unsupervised learning methods 
on other natural language tasks then the need for 
manually coded knowledge would 
be enormously reduced.\footnote{Unsupervised training for sense 
disambiguation has attracted quite a bit of attention.  Hearst~(1991) uses a 
small amount of manually sense tagged data to train an initial model and 
uses this to provide more training data, in a similar vein to 
Merialdo's~(1994) tagger.  Sch\"{u}tze~(1992) uses singular value 
decomposition to discriminate senses on the basis of word 
co-occurrences.  Biber~(1993b) uses statistical factor analysis to discover 
word senses from collocations.}

It is now time to take stock.  This chapter began by introducing
statistical language learning as an approach to \nlp
based on probabilistic models.  These models
assign probabilities to linguistic analyses in terms of their parameters,
which in turn can be estimated from a corpus.  In this way, \sll\ systems
automatically acquire the knowledge necessary to their objectives.

Probabilistic taggers have been used to illustrate this approach.
These taggers use models that assign probabilities to 
part of speech sequences in terms of tag $n$-grams, and achieve
high accuracy rates.  However, extension of these ideas to
probabilistic context free grammars has been less successful.
We have also seen a range of techniques designed to 
allow unsupervised learning so that training corpora need not
be manually annotated.
These are all core topics in statistical language learning.

In the next three sections,
I will describe research that is more specifically relevant to the 
work reported in this thesis.
Each of these sections covers an approach to enhancing probabilistic
parsing: lexical sensitivity, specialised parsers and class-based
models.

\subsection{Using Lexical Statistics} 
\label{sec:sn_lexical}

In the taggers and probabilistic grammars discussed so far, the only 
sensitivity to words is through parts of speech.\footnote{As noted above, the 
proposals of Charniak~(1993) and Lafferty~\etal~(1992) do involve 
additional lexical sensitivity.}  Once the part of speech for a word has been 
identified, the probabilistic models do not distinguish between different 
words with the same part of speech.  Yet different words with the same part 
of speech can behave in radically different ways.  For example, in the 
tagging task, the plural noun \lingform{people} is more likely to be followed 
by a modal auxiliary than the plural noun \lingform{rocks}, as suggested
by example~\ref{eg:sn_lexical_modals}.
\begin{examples}
    \item  \label{eg:sn_lexical_modals}
        \begin{subexamples}
            \item People can be more efficient if they want to be.
            \item People might buy more cars.
            \item People should reuse plastic bags.
            \item People will have to adapt.
            \item Rocks can be difficult to move.
        \end{subexamples}
\end{examples}


Grammatical structure is also lexically sensitive.  For example, probabilistic 
context free grammars are incapable of distinguishing the difference in 
structure between the two noun compounds in 
example~\ref{eg:sn_lexical_cns}, because both involve exactly two 
applications of the phrase structure rule 
\mbox{\={N} $\rightarrow$ \={N} \={N}}.
\begin{examples}
    \item   \label{eg:sn_lexical_cns}
        \begin{subexamples}
            \item{[}breakfast [currant cake ] {]}
            \item{[}[cereal bowl ] manufacturer {]}
        \end{subexamples}
\end{examples}
Other probabilistic grammars (for example, Briscoe and Carroll,~1993:30) 
can distinguish the two structures, but always prefer one structure over the 
other regardless of the words involved.  This is because they ultimately deal 
with parts of speech rather than with words.  Their language models 
treat the occurrence of all words used in the same part of speech 
identically.  More sensitive language models are necessary to analyse noun 
compounds, since both structures illustrated by 
example~\ref{eg:sn_lexical_cns} are frequently used.

Furthermore, lexical sensitivity is a widespread property of language, not 
limited to esoteric parts of the grammar.  In fact, it lies at the root of the 
entire lexicalist movement in formal linguistics.  Therefore, I shall now turn 
to lexicalised probabilistic language models, those that account for the 
different effects of different words used in the same part of speech.

One class of such models is based on the notion of word association.  
Here, the language model contains one parameter for every pair of 
words.  Each of these parameters is referred to as the degree of 
association between the two words.  The underlying assumption is 
that most of the time only highly associated words will occur in the 
grammatical relations of interest, and that therefore a parser should 
pursue an analysis that relates associated words before one that does 
not.  Naturally the question of determining which words are 
associated (that is, estimating the parameters) is crucial.  Different 
methods for measuring association will correspond to different 
underlying assumptions about the likelihood of certain grammatical 
relations.

An example of this is the \newterm{distituents} model described in 
Marcus~(1991).  In this account, parsing is driven by the search for 
constituent boundaries.  It is assumed that words within a single 
constituent are generally more likely to be adjacent throughout the 
corpus than those on either side of a major boundary.  Association is 
therefore measured by word adjacency within the training corpus, and 
parsing proceeds by marking boundaries within a sentence wherever 
the association between adjacent words is lowest.  The association 
measure used is mutual information which, intuitively speaking, 
captures the degree to which one word can be predicted given the 
other.
\begin{equation}
\mbox{\cal M\cal I}(X, Y) \stackrel{\rm def}{=} 
	\log{
		\frac{\countfn(X, Y)}
	        {\countfn(X) \times \countfn(Y)}
	}	
\end{equation}

Church and Hanks~(1989) also use mutual information.  But rather than
simply counting adjacent words in the corpus, they preprocess
the corpus with a parser and count words in certain grammatical 
relations.  For example, given the phrase \lingform{read a book}, 
Marcus~(1991) would associate \lingform{read} with \lingform{a} 
and \lingform{a} with \lingform{book}.  Church and Hanks~(1989) 
would first parse to reveal the verb-object relation, and then associate 
\lingform{read} with \lingform{book}.  Because their model is sensitive 
to the grammatical structure of text, it is capable of making more 
sophisticated predictions.  The underlying assumptions (for example, that 
the subject of a verb is most likely to be a word which is used frequently as 
the subject of that verb in the corpus) are more realistic than those of
Marcus~(1991).  However, they require a parsed training corpus.  
Also, Church and Hanks do not aim to improve parsing 
performance, instead proposing the system as a tool for lexicographers.

What is required is a lexically sensitive probabilistic model that incorporates 
some syntactic structure.  Church~\etal~(1991a) take a similar position.
\begin{citedquote}{Church~\etal,~1991a:110}
Syntactic constraints, by themselves, though are probably not very 
important. \dots On the other hand, collocational factors (word 
associations) dominate syntactic ones so much that you can easily measure 
the influence of word frequency and word association norms on lexical 
retrieval without careful controls on syntax. \dots We believe that syntax 
will ultimately be a very important source of constraint, but in a more 
indirect way.  As we have been suggesting, the real constraints will come 
from word frequencies and collocational constraints, but these questions will 
probably need to be broken out by syntactic context.
\end{citedquote}
The theory proposed in chapter~\ref{ch:md} of this thesis concurs
with this suggestion.  According to this theory, 
processing is driven primarily by lexical-semantics,
mediated by syntax.

\subsection{Specialising Probabilistic Parsers} 
\label{sec:sn_specialised}


The probabilistic parsing models covered so far are intended to represent 
knowledge about the entire grammar.  More recently, some work has looked 
at modelling specific grammatical constructs, focusing on those parts of the 
language that pose the most difficulty for traditional \nlp.

A lexically sensitive approach of this kind, and one to which this 
research owes much, is that of Hindle and Rooth~(1993).  Their 
system focuses on prepositional phrase attachment ambiguity, a 
problem that has long been regarded as one of the more difficult in \nlp.  
In particular, they considered the choice presented to a parser when a 
transitive verb construction is followed by a prepositional phrase.
Three examples based on theirs appear in  
example~\ref{eg:sn_specialised_pps}.
\begin{examples}
   \item	\label{eg:sn_specialised_pps}
     \begin{subexamples}
        \item Germany sent tanks into Poland. 
		\label{eg:sn_specialised_pps_verbal}
        \item They ignored doors into the kitchen. 
		\label{eg:sn_specialised_pps_nominal}
        \item They mined the roads along the coast. 
		\label{eg:sn_specialised_pps_indeterm}
     \end{subexamples}
\end{examples}
The parser must choose whether \lingform{into Poland} qualifies the 
verb \lingform{send} or the noun \lingform{tanks}.  These two possibilities 
are illustrated by 
examples~\ref{eg:sn_specialised_pps}\ref{eg:sn_specialised_pps_verbal} 
and~\ref{eg:sn_specialised_pps}\ref{eg:sn_specialised_pps_nominal} 
respectively.

Hindle and Rooth's~(1993) approach defines an association between 
prepositions and their attachment points (in the example, \lingform{into} is 
associated with \lingform{sent} and \lingform{doors}, but not 
\lingform{tanks} or \lingform{ignored}).  
Given an ambiguous prepositional phrase, their system compares two 
association values: the preposition-verb association and the 
preposition-object association.  To measure the association values, they use 
a corpus of 13 million words of newswire text.  

For such a large training corpus, an unsupervised learning method is 
required.  Their system exploits a small amount of grammatical knowledge 
to find unambiguous prepositional phrase attachments and uses these to 
acquire initial estimates, which are then used to choose an attachment for 
ambiguous examples.  These latter examples form further training data so 
that the association values can be cyclically refined.  The end result is a set 
of associations which select the correct prepositional phrase attachment in 
almost 80\% of the test cases.

In performing this study, Hindle and Rooth encounter two interesting 
phenomena which are also relevant to the work described here.  First, 
they observe that some examples exhibit what they call \newterm{semantic 
indeterminacy}.  This arises when the meanings corresponding to each 
attachment are the same.  For instance, in 
example~\ref{eg:sn_specialised_pps}\ref{eg:sn_specialised_pps_indeterm} 
above, it makes no difference in most contexts whether we speak 
of the roads along the coast being mined or of the roads being mined 
along the coast.  It would seem that the syntactic representation 
makes distinctions which are not necessary to understanding.  A 
similar effect arises in compound nouns, as in \lingform{city sewerage 
systems}.  Hindle and Rooth discard such examples from their test set, a 
policy also followed in the experimental work in 
chapter~\ref{ch:experimental}.

Second, Hindle and Rooth point out that many prepositional phrase 
attachments are ambiguous even for human readers.  The 
oft-cited example of seeing a man with a telescope illustrates a 
case where both analyses are easily conceivable.\footnote{If I see a man 
with a telescope, am I looking through the telescope or is he carrying it?}  In 
such cases, more context is necessary to select the correct attachment.
Since Hindle and Rooth's system only pays attention to three words (the 
verb, the noun and the preposition), it cannot always make the right
choice.  If it chooses to attach \lingform{with} verbally (to 
\lingform{seeing}) then there will be cases when the same three words 
appear, but \lingform{with} attaches nominally.  Because of the limited 
context used, it is not possible for the system to achieve 100\% accuracy.

Hindle and Rooth argue that because it is not possible for humans to 
achieve 100\% accuracy based only on the verb, object and preposition,
it is difficult to judge how well the system mimics human performance.
To provide a performance goal, they conducted an experiment in which 
human subjects were asked to select attachments using only the verb, object 
and preposition.  The humans achieved roughly 85\% accuracy (compared to
the 80\% accuracy of the system), suggesting that the algorithm 
might still be improved a little without introducing further context.

Also relevant to the current work is the nature of the training process.  
Their unsupervised learning method allows the technique to be easily and 
simply applied to new training data, and even to new domains, without 
annotation.  However, such a powerful advantage is not free.  As mentioned, 
the learning proceeds on the basis of unambiguous examples of attachment.  
In particular, verbal attachment is unambiguous when the object noun is a 
pronoun, as in 
example~\ref{eg:sn_specialised_sure}\ref{eg:sn_specialised_sure_verb}.
Nominal attachment is unambiguous when the prepositional phrase 
precedes the verb, as in 
example~\ref{eg:sn_specialised_sure}\ref{eg:sn_specialised_sure_noun}.  
\begin{examples}
   \item	\label{eg:sn_specialised_sure}
     \begin{subexamples}
        \item He hooked it with his four-iron. 
		\label{eg:sn_specialised_sure_verb}
        \item Every man with a donkey was dead. 		
		\label{eg:sn_specialised_sure_noun}
     \end{subexamples}
\end{examples}
The price paid by utilising unambiguous examples is that one further 
assumption must be made; attachment properties of prepositional 
phrases are assumed to be independent of the existence of ambiguity.  
Thus, Hindle and Rooth assume that prepositional phrases qualifying 
transitive verbs occur in the same distributions when the verb's object 
is a pronoun as when not.  This is an important assumption which 
is not made explicit.  The noun compound models in 
chapter~\ref{ch:experimental} are closely related to the
prepositional phrase attachment one of Hindle and Rooth, as are the
training methods.  Therefore, analogous assumptions will be made in 
the those models.

\subsection{Using Class-based Models and Conceptual 
Association} \label{sec:sn_conceptual} 


The approaches presented so far either consider the properties of individual 
words or those of entire grammatical classes of words, the parts of speech.  
In the latter case, as argued above, the models are not sufficiently sensitive 
to variation in the usages of words.  
However, in the former case we would expect 
the number of parameters of the probabilistic model to be many more 
than necessary.  For example, the words \lingform{automobile} and 
\lingform{car} have very similar properties, so that learning these properties 
independently is wasteful of resources and fails to capture 
generalisations that humans would readily use.  

This increase in the number of parameters
takes on particular importance because most systems that are based on 
corpus statistics are plagued by data sparseness.  A large proportion of the 
word tokens in text are relatively infrequent words.  Since statistical 
confidence rests on sample size, rare words require huge amounts of text to 
be processed before an \sll\ system can learn about them.  In fact, one could 
fairly say that progress in the field of statistical \nlp\ is shaped by the 
availability of data and estimation techniques that conserve it.  While a 
range of techniques are used to combat data sparseness, including deleted 
interpolation and other forms of smoothing, any principled means of 
reducing the number of parameters in a model is highly valuable.

This has led a number of researchers to advocate class-based modelling 
where words are allocated to classes but where the classes are finer than the 
part of speech divisions and motivated by different distinctions.  Words 
within a class are assumed to behave in a similar manner and statistical 
inference proceeds in terms of word classes instead of individual words in 
much the same way as parts of speech are treated by Markov model taggers 
and probabilistic context free grammars.\footnote{An interesting variation is 
similarity-based modelling where, rather than having rigid class boundaries, 
the probabilistic model uses a similarity metric to incorporate information 
derived from the behaviour of similar words (Dagan~\etal,~1993).}  For 
instance, Brown~\etal~(1992b) derive word classes by a clustering
algorithm designed to optimise their probabilistic language model.  
While purely statistical approaches have been shown to derive part of 
speech groupings (for example, Finch and Chater,~1992), many of the 
classes derived by Brown~\etal~(1992b) are unintuitive and do not appear to 
be useful for much more than tuning their language 
model.\footnote{Statistical word clustering is one of the oldest goals of 
statistical \nlp\ (starting at least as early as Hirschman~\etal,~1975) and has 
many other applications, even automatic thesaurus generation 
(Grefenstette,~1992).}

Another approach is to manually assign each word in the lexicon a set of 
semantic markers relevant to the text content and take these to delineate the 
word classes.  Velardi~\etal~(1991) use this in a limited domain to address 
various \nlp\ tasks including prepositional phrase attachment (more recent 
work by the same group is reported in Basili~\etal,~1993).  By assigning one
or more of 12 semantic tags to each open class word in their lexicon, useful 
parameter estimates are acquired from a relatively small corpus of just half a 
million words.  
However, the restriction to a limited domain means that the technique is not 
directly applicable to unconstrained text.  Also, one must be careful that the 
development effort involved remains small relative to the cost of building a 
traditional \nlp\ system for the domain.

An approach that is not restricted to a limited domain is that developed by 
Resnik and Hearst~(1993), who use the semantic classes provided by an 
on-line thesaurus.  The intent of their prepositional phrase attachment system 
is to extend Hindle and Rooth~(1993) to allow sensitivity to the object 
of the preposition.  Thus an attachment choice was made on the basis of 
the verb, its object, the preposition and the object of the preposition.  
This permits a distinction between the two attachments shown in 
example~\ref{eg:sn_conceptual_ppobj}, where the attachment decision 
critically depends on the final word.  
\begin{examples}
   \item	\label{eg:sn_conceptual_ppobj}
     \begin{subexamples}
        \item eating fish with bones 
        \item eating fish with chopsticks 
     \end{subexamples}
\end{examples}
Since the new model provides sensitivity to more factors than Hindle and 
Rooth's model does, it requires significantly more training data.  

To reduce this requirement, Resnik and Hearst define \newterm{conceptual 
association}, by analogy with Hindle and Rooth's lexical association.  
Associations are recorded between classes of words (as defined in the 
thesaurus) rather than individual words, on the assumption that word 
properties are uniform within a thesaurus category.  

One interesting viewpoint is to consider such categories as representing 
mental concepts.  Under this view, Resnik and Hearst have moved from a 
string-based representation to one based on psychological constructs.  
Associations are no longer distributions of surface textual elements; they are 
distributions of mental concepts.  An important consequence is that different 
senses of a word are separately represented by the probabilistic model.  
When a word is polysemous, it appears in several thesaurus categories and 
the behaviour of each sense can be represented differently.  
Yarowsky's~(1992) sense disambiguation model works on this principle.

The thesaurus used by Resnik and Hearst is WordNet (described by
Miller,~1990), an on-line structured 
network containing more than 100,000 words, which provides a hierarchical 
taxonomy of small groups of synonyms, called \newterm{synsets}.  
Resnik and Hearst's system creates one class for each synset.  This class
contains not only the words within the synset, but also all words in synsets
below it in the hierarchy.\footnote{This choice of classes 
leads to widely varying class 
sizes.  A different method for creating word classes from WordNet which 
results in roughly uniform classes is described by Hearst and 
Sch\"{u}tze~(1993).}
The system then collects associations from the corpus for each such 
class.  Under this grouping scheme each word appears in every 
class which is a hypernym of its synonym set.  For example, 
\lingform{penny} is a member of many classes some of which we might call 
\concept{coins}, \concept{cash}, \concept{money}, \concept{possessions}, 
\concept{objects} and \concept{entities}.  Therefore, the properties of 
\lingform{penny} are represented by properties of a sequence of classes 
ranging from the most specific to the most general.  

This creates a problem of deciding which class to use for making the 
attachment decision when faced with an ambiguous prepositional phrase 
involving the word.  Resnik and Hearst's solution was to perform a paired 
$t$-test between the two alternatives across the entire sequence of possible 
classes.  They gave no intuitive motivation for this choice; however the 
approach seemed to work, producing a small improvement over the 
technique of Hindle and Rooth.


%
 
\subsection{Brief Review}
\label{sec:sn_review}

The first half of this chapter has served to introduce statistical 
language learning.  We have looked at probabilistic taggers
and grammars, to illustrate the general approach, and considered
three research directions that have been pursued to improve
probabilistic parsing.  These systems have demonstrated 
that statistics computed from a corpus can provide useful knowledge 
for natural language processing tasks.  
However, each of these systems represents just one possibility
out of an enormous range of potential designs.  As yet, we 
have only just begun to explore this potential.  If we could map 
out the world of possible statistical models of language,
all the designs yet proposed would comprise, at most, a small 
corner of that map.  

While models for tagging have yielded good results,
probabilistic context free grammars fall short of solving
the parsing problem.  Better designs still need to be explored.
Sections~\ref{sec:sn_lexical} through~\ref{sec:sn_conceptual}
reviewed research on three design aspects of parsing models: 
lexical sensitivity, specialisation of the model to 
particular syntactic forms and use of conceptual representations.  
Each of these three research directions represents a new area
on the map.

In chapter~\ref{ch:md}, I will propose a theory of statistical
natural language understanding that extends these
research directions by identifying a new class of designs.
In doing so, the theory points the way to an entirely new part
of the map, an unexplored area that might yield the designs we seek.
Importantly, it is an empirical question whether the new designs
will prove useful.  To test the theory, we need to build systems 
and measure their performance.  Only then can we fill in 
the details of the map.

To begin evaluation of the theory, I have applied it
to the task of analysing noun compounds.  This work
will be described in chapter~\ref{ch:experimental}.
The second half of this chapter will therefore review 
existing work on noun compounds.


\section{Introducing Noun Compounds} 
\label{part:cn}

If parsing is taken to be the first step in taming the 
natural language understanding task, then broad coverage 
\nlp\ remains a jungle inhabited by wild beasts.
For instance, parsing noun compounds appears to 
require detailed world knowledge that is unavailable
outside a limited domain (see Sparck Jones,~1983,
for a detailed argument).  
Yet, far from being an obscure, endangered species, 
the noun compound is flourishing in modern language.  
It has already made five appearances in this paragraph 
and at least one diachronic study shows a veritable population
explosion (Leonard,~1984).  

Both the challenges posed by noun compounds and their abundance
have attracted a substantial body of work in linguistics 
and computational linguistics.  The second half of this chapter 
reviews this work, with an eye to providing the necessary background 
for the experimental work described in this thesis.  

%
%
%
%

In section~\ref{sec:cn_motivations}, I will argue that noun compounds
are an important subject of study because they both occur frequently
and present significant difficulties for current \nlp\ techniques.
Section~\ref{sec:cn_nature} settles on a precise definition
of noun compounds and describes some important aspects, including
their appearance in many languages, their productivity and their
functions.
 
Noun compounds are syntactically ambiguous when they contain more
than two words.  In section~\ref{sec:cn_grammar}, I will discuss
various theories of noun compound syntax and explain how this
ambiguity arises.  While there is agreement about the syntax
of noun compounds, no consensus has been reached regarding
their semantics except at the broadest level of
classification.  A variety of semantic theories will be described
in section~\ref{sec:cn_meaning}.  In section~\ref{sec:cn_accommodation},
I will cover the notion of semantic granularity, which is vital
to some of the theories of noun compound semantics.

One important property of noun compounds is their context dependence;
this will be discussed in section~\ref{sec:cn_context}.  
Most computer systems designed to handle noun compounds
ignore this aspect and aim to give the best out of context analysis.

Algorithms have been proposed or implemented  
for many tasks, including phonological stress assignment and 
direct machine translation.  A review of the computational tasks
that have been addressed forms section~\ref{sec:cn_computational}.

Finally, sections~\ref{sec:cn_knowledge} and~\ref{sec:cn_statistical}
cover the existing work on parsing and semantic analysis of noun
compounds using knowledge-based methods and statistical methods,
respectively.

\subsection{Motivations for Noun Compound Research} 
\label{sec:cn_motivations}

Why should we be interested in noun compounds?  In this section 
I will suggest that not only is noun compound understanding a 
vital component of any sophisticated natural language understanding 
system, but that noun compounds also constitute a promising area 
for research that exposes some of the weaknesses of current 
natural language technology.

There has been much emphasis in recent \nlp\ research on measuring
performance, especially within statistical circles.  In such an environment
the first question to ask in regard to any proposed technique is: how
often does the need for the technique arise?  
The perceived value of a method is in
direct proportion to its breadth of immediate applicability.  From this 
(somewhat mercenary) point of view, noun compounds represent a 
significant haul.  They are indisputably common in most texts, and in certain 
genres (for example, technical descriptions) they dot the landscape 
like Australian sheep.  

Leonard~(1984) reports the results of a diachronic study of noun compound  
use in fictional prose.  It shows a steady increase in their frequency over the 
last two centuries.  Table~\ref{tb:cn_motivations_leonard} is reproduced 
from her work.  
Assuming 15 words per sentence, \publicationname{Down 
There on a Visit} has an average of one compound in every five sentences.  
Yet fiction genres appear to have relatively few noun 
compounds.  Warren's~(1978:235) study of the Brown corpus places 
fictional text as the least compound rich text type, apart from \scare{belles 
lettres}.  

\begin{table*}[h] 
\centering 
\begin{tabular}{|l|l|l|r|r|}  \hline
Date & Author & Work & Tokens & Types \\
\hline
1759 & Johnson, Samuel & 
	\publicationname{Rasselas} & 10 & 8 \\
c.1813 & Austen, Jane & 
	\publicationname{Pride and Prejudice} & 39 & 34 \\
1840 & Dickens, Charles & 
	\publicationname{The Old Curiosity Shop} & 89 & 75 \\
1875 & Meredith, George & 
	\publicationname{The Ordeal of Richard Feverel} & 147 & 112 \\
1962 & Isherwood, Christopher & 
	\publicationname{Down There on a Visit} & 229 & 198 \\
\hline
\end{tabular}
\caption{Frequencies of noun compounds per 17,500 words of fictional 
prose compiled by Leonard~(1984)} 
\label{tb:cn_motivations_leonard}
\end{table*}

The ratio of types to tokens here is also noteworthy.  Noun compounds, at 
least in these texts, are productive.  Each one encountered is typically a fresh 
construction, which must be interpreted anew.  Whatever the mechanism for 
dealing with noun compounds, it must be dynamic if it is to be broadly 
useful.  It will not be practical to merely add noun compounds statically to 
the lexicon.

Noun compounds are even more evident in press and technical materials.  In 
regard to the former text type, McDonald~(1982:125) identifies 252 noun 
compounds (types) in around 500 sentences of 
\publicationname{Newsweek} and \publicationname{New York Times} 
articles.  A previously unseen noun compound must be dealt with on 
average in every second sentence.

For the latter text type, ter~Stal~(1994a:9) has made a study of compounds in 
a corpus consisting of 293 technical abstracts from 
\publicationname{Engineered Materials Abstracts}.  His list contains some 
3514 noun compounds (types), yielding an average of 12 per abstract.

Taking a broader definition of noun compound, an even higher density has 
been reported by Beardon and Turner~(1993:2--3) in 
\publicationname{Computer Graphics} abstracts.  They found that their 
sample of six abstracts had \shortquote{an average of 27\% of all words 
participating in potential complex nominals}.\footnote{The word potential 
reflects the fact that one of the goals of the study is to define complex 
nominals.  Many kinds of noun pre-modifiers are included; however, they do 
not count predicating adjectives.}  While this is a very small sample, and 
clearly text of extraordinary density, it does vividly depict the quandary 
faced by any \nlp\ system ill-equipped to deal with noun compounds.

These figures demonstrate that the likelihood of a natural language 
processing system encountering a noun compound is high.  But do noun 
compounds play an important role in comprehension?  If not, perhaps they 
could just be skipped.  As we might expect, the content words that are added 
to form a compound are instrumental to correct interpretation; they have no 
other reason to exist (unlike closed class words, which on occasion function 
merely to make a sentence grammatical).  Compounding serves a number of 
important linguistic functions, chiefly in refining otherwise under-specified 
references.  For example, both words in 
example~\ref{eg:cn_motivations_lb} below are polysemous --- compounding 
serves to select an appropriate sense for each.  Virtually all practical 
applications of \nlp\ will be degraded by inability to make use of these 
functions.
\begin{examples}
\item lift button   \label{eg:cn_motivations_lb}
\end{examples}

In section~\ref{sec:cn_grammar} below, I will discuss the syntactic 
character of noun compounds.  One of their striking aspects is that on 
one hand, their grammar is extremely simple, while on the 
other hand they exhibit explosive syntactic ambiguity.  The latter property 
means that syntax-first \nlp\ techniques are not well-suited to handling noun 
compounds; the former makes noun compounds a good area for gaining 
insight into the weaknesses of this traditional approach.  Their grammatical 
simplicity prevents complex syntactic issues from clouding the conclusions.

Furthermore, noun compounds pose such large difficulties for \nlp\ that they 
have been the canonical example used to point out the great distance 
remaining between state-of-the-art \nlp\ technology and \shortquote{more 
powerful and comprehensive [language-processing] programs}~(Sparck 
Jones,~1983:363).  The influence of pragmatic issues and seemingly 
arbitrary amounts of world knowledge mean that any serious approach to 
noun compounds must address the key issues of knowledge representation, 
knowledge acquisition and robustness.  Since these have wider significance 
to language processing in general, progress on the problems of noun 
compounds necessarily has broad impact on other aspects of \nlp.

Independent of the wider implications for \nlp, there are important direct 
applications of noun compound processing techniques.  I will mention two 
promising areas.
\begin{description}
\item[Machine translation:] Use of compounds varies from one language to 
another, so that automatic translation systems cannot rely on lexical 
substitution to translate compounds.  For example, French makes limited use 
of compounding, while in German it is heavily productive.  It is therefore 
commonly necessary to render a German compound into an alternative 
French syntactic construction, such as a prepositional phrase.  In 
section~\ref{sec:cn_nature} I will cite some work specifically on this task.
\item[Caption/abstract retrieval:] Compounding is commonly used to 
compress information into a smaller amount of text and therefore pervades 
captions, abstracts and signs.\footnote{A small exercise for the reader: on 
your way home today take note of any signs you see.  Count the proportion 
of words used that are part of a noun compound.  On Australian roads the 
rate is close to 100\%, chiefly because of the high frequency of one sign: 
\lingform{road work}.}   Recent advances in multi-media technology have 
resulted in large image collections appearing with no means of indexing 
retrieval except via captions (see for example, Rowe,~1994).  Sophisticated 
retrieval must therefore process caption text and inevitably the noun 
compounds found there.
\end{description}

So, to summarise, the following properties of noun compounds make them 
an important subject for research:
\begin{itemize}
\item They are frequent.
\item They have linguistic functions that need to be represented.
\item They have a simple grammar, yet exhibit high parsing complexity.
\item They demand the application of large amounts of knowledge.
\item Apart from more general implications for \nlp, noun compound 
processing technology also has worthwhile direct applications.
\end{itemize}

For these reasons, chapter~\ref{ch:experimental} will be concerned
with experiments on applying statistical \nlp\ techniques
to the problem of noun compound processing.

\subsection{The Nature of Noun Compounds} 
\label{sec:cn_nature}  

The species of linguistic construct that is referred to as the noun compound 
has received regular attention from linguists.  There are many and varied 
subspecies, a broad range of habitats and a multitude of nomenclatures.  In 
this section, I will review a number of aspects of noun compounds and work 
towards a definition for the purposes of this work.  The definition I will 
adopt follows that of Downing~(1977).  Finally,  I will briefly mention some 
of the functions performed by compounding. 

\subsubsection*{Varieties and habitat}

There are almost as many names for noun compounds and their relatives as 
there are linguistic studies of them.  I will survey a selection of related 
constructions when considering definitions below.  However, it is worth 
listing the names here for the record:
\begin{itemize}
\item noun compounds, compound nouns, compounds;
\item nominal compounds, compound nominals, complex nominals;
\item noun premodifiers;
\item nominalisations;
\item noun sequences, noun-noun compounds, noun + noun compounds.
\end{itemize}
While the definitions vary, all these terms describe very similar classes of 
constructions.  All appear where nouns do, all involve multiple open class 
morphemes and, in each case, members of the class exist that contain no 
closed class morphemes.

Noun compounds are common in English, but are by no means limited to 
one language.  Relatives can be found in virtually all widely spoken 
languages.  Work specifically addressing noun compounds investigates their 
use in many languages, including:
\begin{description}
\item[Chinese:] Pun and Lum~(1989) describe an algorithm for analysing 
Chinese noun compounds and related constructions using case grammar.
\item[Japanese:] Fujisaki~\etal~(1991) report some experiments on applying 
a probabilistic grammar to parsing Japanese noun compounds.
\item[French:] Bourigault~(1993) uses a corpus-based technique to parse 
various constructions, including noun compounds, in French.
\item[German:] Rackow~\etal~(1992) use a monolingual corpus to find 
appropriate English translations of German noun compounds.
\item[Italian and Modern Greek:] Di Sciullo and Ralli~(1994) contrast the 
ordering constraints of noun compounds in English, Italian and Modern 
Greek.
\end{description}
In some languages (for example, Dutch and German) noun compounds are 
orthographically single words and a joining morpheme may appear between 
two lexemes.  In these languages, and those without orthographic word 
markings like Chinese, there is an additional ambiguity introduced by the 
need to segment the noun compound into individual lexemes.  
Sproat~\etal~(1994) addresses this problem for Chinese.  

The present work addresses only English noun compounds.  While each 
language presents its own special difficulties, the qualitative differences in 
noun compound behaviour between languages are relatively few.  There are 
no apparent reasons to suggest that the techniques developed here could not 
be adapted to other languages.

\subsubsection*{Lexicalisation}

One of the more confusing aspects of noun compounds is the fact that they 
exhibit both morphological and syntactic behaviour, cutting across the usual 
linguistic divisions.  Though it is more common in certain other languages, 
English noun compounds can be rendered orthographically as single words, 
either by hyphenation, or unmarked concatenation.  There are often several 
forms of the same noun compound in use.
\begin{examples}
   \item	
     \begin{subexamples}
        \item bed spring
        \item bed-spring
        \item bedspring
     \end{subexamples}
\end{examples}
Noun compounds seem to live in the border zone between word formation 
and syntax, exhibiting adaptations to both.  Some researchers (for example, 
Sproat,~1992:37) distinguish 
between syntactic and morphological compounds, but this leads to difficulty 
in distinguishing the two because the boundary is not clear cut.

The morphological character of noun compounding is related to another 
well-known aspect called \newterm{lexicalisation}.  As is argued in many 
accounts (for example, Levi,~1978:10) noun compounds can acquire
idiomatic meanings which are not derivable from the component nouns.  This 
appears to result from the fact that compounds can acquire the status of 
independent words.  \publicationname{The Oxford Shorter English 
Dictionary} (Onions,~1973) has not only subentries under \lingform{foot} 
for \lingform{-bridge}, \lingform{-hill}, and \lingform{-work}, but also full 
entries for the words \lingform{football}, \lingform{footman} and 
\lingform{footpath}.  The non-compositional meanings expressed by these 
words are not dissimilar in character to the noun compound in
example~\ref{eg:cn_nature_lexicalised}.
\begin{examples}  
  \item  \label{eg:cn_nature_lexicalised}
	foot fault
\end{examples}  
Not only is this example commonly used as a fixed word combination, but 
its meaning would change if semantically similar words were substituted: 
compare \lingform{toe fault} and \lingform{leg fault}.  Therefore, such 
idiosyncratic meanings must be encoded on an individual basis.  

By contrast, other examples exhibit highly productive combinations.  The 
noun compounds in example~\ref{eg:cn_nature_novel}
have corresponding siblings \lingform{leg brace} and \lingform{toe size}.  
\begin{examples}  
  \item  \label{eg:cn_nature_novel}
    \begin{subexamples}
      \item foot brace
      \item foot size
    \end{subexamples}
\end{examples}  
In such cases, the multitude of possible combinations would make any 
attempt to exhaustively encode meanings for each compound hopeless.  
Consider the following examples.
\begin{examples}  
  \item  
    \begin{subexamples}
      \item nose ring
      \item neck ring
      \item ankle ring
      \item belly button ring
    \end{subexamples}
  \item  
    \begin{subexamples}
      \item mongoose attack
      \item mongoose mating
      \item mongoose habitats
      \item mongoose escape
    \end{subexamples}
\end{examples}  
I will adopt the term \newterm{novel compounds} to refer to the highly 
productive kind, following Downing~(1977).  Since these constitute a 
substantial portion of noun compounds, any sophisticated \nlp\ system must 
incorporate a dynamic mechanism to deal with these cases.

\subsubsection*{Definition}

We still lack a precise definition of noun compounds.  As mentioned, there 
are a multitude of possible definitions, each with its own arguments.  Four are 
sufficiently popular to be mentioned.
\begin{description}
\item[Noun premodifiers:] Quirk~\etal~(1985:1321--1346) adopt a very open 
definition.  Their grammar permits virtually any constituent to appear before 
a noun to form a pre-modified noun.  Their definition thus includes 
\lingform{out-in-the-wilds cottage} and similar constructions.  The difficulty 
with this definition
lies in distinguishing compounding from adjectival modification, which I 
will have more to say about shortly.  
\item[Compounds:] Chomsky and Halle~(1991:16,91--93) take a 
phonological definition in which words preceding a noun form a compound 
if they receive the primary stress.  Thus \lingform{blackboard} is a 
compound, while \lingform{black board} is not.  The problem with this tack 
is that pronunciations vary amongst speakers, so that what is a compound for 
one person may not be for another.
\item[Complex Nominals:] Levi~(1978:1) chooses to include certain 
adjectives along with nouns as possible compounding elements and so calls 
her noun compounds \newterm{complex nominals}.  The adjectives included are 
non-predicating adjectives (ones that supposedly cannot be ascribed via a 
copula construction).  An example then is \lingform{electrical engineer}, 
because \lingform{That engineer is electrical} is ungrammatical.  Since 
adjectives are difficult to divide into predicating and non-predicating, this 
ambiguity causes computational difficulties.
\item[Noun + Noun Compounds:] Downing~(1977) defines noun 
compounds as any sequence of nouns that itself functions as a noun.  While 
this is more restrictive than any of the previous three, it is relatively 
unambiguous.  Leonard~(1984) also takes this path, using the term 
\newterm{noun sequences}.
\end{description}
In this thesis, I will follow Downing~(1977) in restricting attention to noun 
sequences.  However, there are some special noun sequences to note.

While nouns marked for the genitive, as in 
example~\ref{eg:cn_nature_genitive}, are frequent premodifiers, they have 
explicitly marked semantics, and it has been argued (Chomsky,~1970) that
their syntactic behaviour differs from other noun compound modifiers.
\begin{examples}  
  \item  \label{eg:cn_nature_genitive}
    \begin{subexamples}
      \item dog's breakfast
      \item senate's directives
      \item Noam's objections
    \end{subexamples}
\end{examples}  
The last of these raises another special case, that of names.  While proper 
nouns form compounds with equal readiness to common nouns, as in 
example~\ref{eg:cn_nature_proper}, names have an arbitrariness that 
is clearly of a different nature; consider the names in
example~\ref{eg:cn_nature_names}.
\begin{examples}  
  \item  \label{eg:cn_nature_proper}
    \begin{subexamples}
      \item Boston music   
      \item August snowfalls
      \item Christian God
    \end{subexamples}
  \item  \label{eg:cn_nature_names}
    \begin{subexamples}
      \item Charles River  
      \item Pitt Street 
      \item Lake Tahoe
    \end{subexamples}
\end{examples}  
A similar behaviour occurs with artificial naming conventions such as those 
used to refer to chemical substances, as in 
example~\ref{eg:cn_nature_chemicals}.
\begin{examples}  
  \item  \label{eg:cn_nature_chemicals}
    \begin{subexamples}
      \item calcium chloride
      \item porphyrin quinone phenanthrene
    \end{subexamples}
\end{examples}  

I will exclude all these special cases (proper nouns are not a special case 
unless the noun compound is a name).
\begin{definition}[Noun Compound]
A {\em noun compound\/} is any consecutive sequence of nouns 
at least two words 
in length that functions as a noun, but which contains no genitive markers 
and is not a name.
\end{definition}
This definition does not include orthographically joined morphemes such as 
\lingform{firetruck} and \lingform{steamboat}.  I will assume that such 
words are joined because they are sufficiently common as to warrant 
inclusion in the lexicon, and thus do not require dynamic processing.
A few examples to illustrate the definition are shown in 
example~\ref{eg:cn_nature_definition}.
\begin{examples}
   \item	\label{eg:cn_nature_definition}
     \begin{subexamples}
        \item stone fish  
        \item party animal 
        \item ball park
        \item outback adventure tour
        \item emergency bus fuel 
        \item anchor nose ring 
        \item killer whale attack 
     \end{subexamples}
\end{examples}
Note also that the definition allows for gerunds, so that the following 
examples are all valid noun compounds.
\begin{examples}  
  \item  \label{eg:cn_nature_gerunds}
    \begin{subexamples}
      \item laughing children
      \item horse riding
      \item leading soprano
    \end{subexamples}
\end{examples}  

Since I have no special need to reserve other terms, I will refer to noun 
compounds throughout this thesis variously as noun compounds, compound 
nouns, or simply compounds, interchangeably.

The working definition given above is adequate to support the 
experimental work and corresponding argumentation, but should not 
be regarded as limiting the scope of the problem in any 
fundamentally important way.  At this stage, there is no evidence to 
suggest that the results are not applicable to, for example, adjectival 
modifiers of the kind that Levi~(1978) allows.  

There remains a small difficulty with this definition, caused by part of 
speech ambiguity, when a word is classed both as noun and adjective.  It is 
somewhat surprising that few researchers adopting definitions similar to the 
one used here have remarked upon this problem.  Apart from 
Warren~(1978:65), who devotes two sentences to the topic, I have not seen 
it mentioned.  There are many nouns and adjectives that are orthographically 
identical.  Consider the nominal forms in example~\ref{eg:cn_nature_adjs}.  
In each case, the modifier can be used both as a noun and as an adjective.  
\begin{examples}  
  \item  \label{eg:cn_nature_adjs}
    \begin{subexamples} 
      \item plastic ball \label{eg:cn_nature_adjs_pb}
      \item American law
      \item light pole
    \end{subexamples}
\end{examples}  
Fortunately, the different part of speech assignments usually lead to distinct 
interpretations.  In 
example~\ref{eg:cn_nature_adjs}\ref{eg:cn_nature_adjs_pb}, either the ball 
is made of the substance plastic (whether hard, soft, rigid or otherwise ---
compare \lingform{acrylic ball}) or the ball is particularly flexible 
(regardless of its material composition --- compare \lingform{malleable 
ball}).  In the former case, the substance plastic is clearly represented by a 
noun, while in the latter case, the characteristic of being plastic is 
represented by an adjective.\footnote{Although this is not the only way to 
draw the distinction; the choice of part of speech also has syntactic 
implications.  Compare the use of an adjectival modifier in 
\lingform{recycled plastic ball} to the adverbial modifier in 
\lingform{highly plastic ball}.}  
The other two examples behave similarly.  Laws
might be drafted in America (adjectival reading) or be
about Americans (compounded reading).  Poles might support lights (compounded
reading) or not be heavy (adjectival reading).
Without context, it is not possible to be sure of either reading.  
Therefore, it is difficult to automatically distinguish
between these two types of premodification.

The definition also does not distinguish between novel and lexicalised 
compounds.  The goal of developing noun compound processing techniques 
is primarily motivated by novel compounds.  The idiosyncrasy of 
lexicalised compounds requires them to be handled on an individual basis.   
However, since currently available lexical resources are insufficient to easily 
exclude lexicalised compounds, it is simpler to include both kinds.  This 
tactic removes the task of establishing a precise distinction between these 
two kinds of compounds, a task that involves some theoretical 
complications.  For example, many noun sequences are in 
common use and yet have not acquired non-compositional meanings.  
\begin{examples}  
  \item  
    \begin{subexamples} 
      \item kitchen table
      \item dog food
      \item wine glass
      \item field trip
    \end{subexamples}
\end{examples}  
The experiments I have conducted to date have not required this 
issue to be addressed, and it can in principle be solved by 
providing lexical entries for all lexicalised compounds.

\subsubsection*{Functions}

Before turning to the syntactic properties of compounds let's consider the 
linguistic functions performed by compounds.  Since they act as nouns their 
primary effect is to indicate  things.  Modifiers assist in this function by 
reducing the set of possibilities.  However, there seems to be an alternative 
function: providing additional information about the thing already identified.  
This is analogous to relative clauses, which have restrictive and 
non-restrictive versions.  Bolinger~(1967) makes a similar observation 
regarding adjectival modifiers, dividing them into reference-modifying and 
referent-modifying groups.  I call compounds sharing the former property 
\newterm{indicative}, and those sharing the latter \newterm{informative}.
Quirk~\etal~(1985:1322) also draw this distinction.  While most compound 
noun modifiers are indicative, example~\ref{eg:cn_nature_informative} 
shows some that are informative.
\begin{examples}  
  \item  \label{eg:cn_nature_informative}
    \begin{subexamples} 
      \item The {\em idiot doctor} gave him the wrong tablets.
      \item You might already have won a {\em dream holiday}.
    \end{subexamples}
\end{examples}  

Marsh~(1984) observes that compounding is a common text 
compression strategy in navy messages.  When space is limited, use of noun 
compounds serves to convey more content with fewer 
words.\footnote{Although see Dras~(1995) for work on the use of 
nominalisations (and support verbs) in making texts wordier.}  
Fewer words overall entails fewer 
closed class words, with the result that linguistic cues to sentence structure 
become sparser:  \shortquote{\dots the function words that so often direct 
a parsing procedure and reduce the choice of possible constructions are 
frequently absent \dots and \dots structural ambiguity becomes a serious 
problem}~(Marsh,~1984:505).  Her solution involves reliance on 
semantics rather than syntax.  
\shortquote{By utilizing the semantic patterns that are derived 
from a sublanguage analysis, it becomes possible to properly bracket 
complex noun phrases}~(Marsh,~1984:507).

The same theme also appears in Sparck~Jones~(1983), although she is more 
pessimistic about how useful these semantic patterns might be.
\begin{citedquote}{Sparck~Jones,~1983:377}
Summarising a series of experiments, Marslen-Wilson and Tyler argue that 
human beings show abilities to recognise words correctly, even before they 
have been completely uttered, which necessarily imply not only that 
extensive top-down work is being done on the speech signal as it arrives, but 
that this work involves inference, including pragmatic inference.

This argument would appear to imply that compound noun interpretation can 
be carried out in an essentially predictive, i.e. strictly expectation-driven, 
manner, relying only on long-term memory and local context.  This is very 
strong, perhaps too strong to accept.
\end{citedquote}

The extent of the local context referred to here is not clear.  
However, emphasis is certainly being placed on the role of semantic 
expectations in compound noun processing.  This is a topic to which I shall 
return in chapter~\ref{ch:md}.

\subsection{The Grammar of Noun Compounds} 
\label{sec:cn_grammar}  

Compound nouns form a small corner of the grammar of English.  
Syntactically speaking, they follow quite simple rules.  Traditional phrase 
structure grammar, dealing with sequences of categories, provides little to 
assist the analysis.  Quirk~\etal~(1985:1321--1346) specifies that any noun 
can be premodified by one or more other nouns.  A single application of this 
rule is sufficient to generate arbitrarily long noun sequences, all without 
structural ambiguity.  The ambiguity inherent in noun compounds arises 
from the fact that premodifying nouns can themselves be premodified by 
multiple applications of the rule.  

These properties can be economically captured by a grammar rule of the 
form shown in example~\ref{eg:cn_grammar_cnrule}, because the constituent 
formed by compounding is identical to the two compounded elements and 
can participate in further compounding, either as left or right member.  
\begin{examples}
\item \={N} $\rightarrow$ \={N} \={N}  \label{eg:cn_grammar_cnrule}
\end{examples}
Multiple premodifiers of one noun are then generated by right recursion.  
For the purposes of this work I shall assume that this rule (or some further 
restriction of it) is sufficient to cover all noun compounds.  
Applied repeatedly, this rule generates binary tree structures, where 
each noun in the sequence lies at a leaf of the tree.  In fact, it allows every 
possible binary tree with $n$ leaves to be a possible parse of a compound of 
length $n$. Examples exist which cannot be analysed this way.
For instance, some words appear to require more than one modifier, 
as in the case in example~\ref{eg:cn_grammar_2place}.
However, these appear to be infrequent.
\begin{examples}  
  \item US Soviet relations \label{eg:cn_grammar_2place}
\end{examples}  

According to the rule, all compounds longer than two words
are syntactically ambiguous.  Take for instance the two compounds in 
example~\ref{eg:cn_grammar_ambig}.
\begin{examples}
   \item	\label{eg:cn_grammar_ambig}
     \begin{subexamples}
       \item {[}china [tea cup \nn] \nn{]}                  
       \item {[}[hydrogen ion \nn] exchange \nn{]}
      \end{subexamples}
\end{examples}
Both consist of a sequence of three nouns, yet the structures differ.  
Importantly, different structures lead to different interpretations as in 
example~\ref{eg:cn_grammar_both}, in which the company may either 
institute the policy or be the tax payer.  
\begin{examples}
   \item company tax policy   \label{eg:cn_grammar_both}
\end{examples}

Yet, at the same time as being critical to interpretation, such structural 
ambiguities multiply rapidly.  It is easy to show that the number of possible 
parses for a compound grows exponentially with its length.  To see this, 
build up the noun sequence from right to left.  Each new noun after the 
second can modify (at least) the whole sequence so far, or the 
leftmost noun so far.  Therefore, each noun (at least) doubles the 
ambiguity of the sequence.  ter~Stal~(1994a) reports extracting compounds 
up to 8 words in length.  The burgeoning of possibilities makes
it vital to find efficient methods of disambiguating compounds.

\subsubsection*{Nominalisation}

Beyond the rewrite rule given in example~\ref{eg:cn_grammar_cnrule}
above, syntactic theories of noun compounds identify 
two general types of compounding.  
The first type is formed by transformations of clausal 
structures of one kind or another.  The idea behind this is that the syntactic 
behaviour of these compounds can be derived from the syntactic behaviour 
of verbal structures.  The most usual manifestations of this have a rightmost 
noun related morphologically to a verb.  Thus, the verb \lingform{collect} is 
nominalised to form the noun \lingform{collection} and complements of the 
verb become modifiers of the noun.  
Example~\ref{eg:cn_grammar_nominalisation} shows such a compound and 
the verb phrase from which it is derived.
\begin{examples}
   \item	\label{eg:cn_grammar_nominalisation}
     \begin{subexamples}
       \item stamp collection
       \item to collect stamps
      \end{subexamples}
\end{examples}

Once again, there are numerous nomenclatures.  Levi~(1978) calls these 
compounds \newterm{nominalisations}, in line with Fraser~(1970) and 
Chomsky~(1970). Warren~(1978) calls them \newterm{verbal-nexus 
compounds}, while Leonard~(1984) uses the term \newterm{sentence compounds} 
and Meyer~(1993) calls them \newterm{sortal nouns}.  
Finin~(1980) extends this class of 
compounds to include \newterm{role 
nominals}, where nouns can be associated with verbs to which they are not 
morphologically related.  For instance, the noun \lingform{food} is 
considered to derive from the verb \lingform{eat} and therefore inherits a 
subject (the eater) so that the compound \lingform{dog food} can be 
explained by the same syntactic rules as those applying to 
explicitly marked nominalisations.

While all these different names are justified by various theoretical 
differences, at a coarse level they all capture essentially the same process, 
whereby clausal syntax is inherited by nouns derived from verbs.  
Furthermore, the syntactic constraints resulting from these theories do not do 
anything substantial to reduce the structural ambiguity.  There is, for 
instance, no requirement for syntactic agreement between the two nouns, 
even when the verb phrase from which they are derived does require
agreement.  Also, Cumming~(1991) argues that nominalisations behave like 
clauses in some ways but not others, and that therefore two separate 
syntactic representations are required.

There is one interesting property arising from all these transformational 
accounts though.  They are often based around distinctions that have a 
semantic rather than syntactic flavour.  For example, Lees~(1970) bases his 
treatment of compounds on what he calls \scare{very deep grammatical 
structure}, a representation close to semantics.  Likewise, Fraser~(1970) 
defines three types of nominalisation, factive, action and substantive, based 
upon semantic distinctions.  This suggests that semantic elements play an 
important role in controlling the structure of noun compounds.

\subsubsection*{Non-verbal-nexus compounds}

The second general type of noun compound comprises those compounds 
which are not nominalisations.  Levi~(1978) calls these 
\newterm{deleted predicate nominals}, Warren~(1978) calls them 
\newterm{non-verbal-nexus compounds} and 
Meyer~(1993) calls them \newterm{relational nouns}.  
Finin~(1980) claims that the 
extended definition of role nominal allows this class to be eliminated.  
However this seems optimistic in the light of Warren's~(1978) study, which 
is restricted just to non-verbal-nexus compounds.  

As far as the syntax of 
non-nominalisations goes, there is little more than the structural rule given 
in example~\ref{eg:cn_grammar_cnrule} above.  
Probably the only concrete constraint is the morphological rule 
described by Quirk~\etal~(1985:1333--5), which dictates that modifying 
nouns are singular in number.  While this rule even applies to nouns that 
normally do not have singular forms 
(example~\ref{eg:cn_grammar_morph_strong}), it still has many 
exceptions.  Example~\ref{eg:cn_grammar_morph_weak} shows two 
common ones.
\begin{examples}
   \item	\label{eg:cn_grammar_morph_strong}
     \begin{subexamples}
       \item trouser leg
       \item scissor factory
      \end{subexamples}
   \item	\label{eg:cn_grammar_morph_weak}
     \begin{subexamples}
       \item arms race
       \item tropical gardens expert
      \end{subexamples}
\end{examples}

All these theories of compound noun syntax assume that 
compounding results in a form that is a suitable element for further 
compounding.  The general rule, \mbox{\={N} $\rightarrow$ \={N} 
\={N}}, thus captures the gross syntactic behaviour of compounds and is 
recursively applicable without limit.  However, there is one theory that is an 
exception.  While other theories are fully recursive, Marcus~(1980) holds 
that processing limits restrict the possible structures, 
so that the general rule 
given above is not quite idempotent.  In particular, Marcus~(1980) prohibits 
structures in which three or more nouns all premodify another noun.  Put 
another way, the rewrite rule cannot be reapplied to its right member more 
than twice.  Examples of the structures disallowed are shown in 
example~\ref{eg:cn_grammar_marcus}, taken from Finin~(1980:47).
\begin{examples}
   \item	\label{eg:cn_grammar_marcus}
     \begin{subexamples}
       \item {[}Aluminum [automobile [water pumps \nn] \nn] \nn{]}  
       \item {[}plastic [toy [fire truck \nn] \nn] \nn{]}             
       \item {[}[back yard \nn] [brick [dog house \nn] \nn] \nn{]}  
     \end{subexamples}
\end{examples}
For the purposes of this thesis, I will assume that Marcus's~(1980) theory is 
false, as is evidenced by these examples.
However I do not consider the implications of his theory to have 
significant impact on the results presented here in any case.

Regardless of the grammatical stance taken, the syntax of compounds that 
are longer than two nouns is underconstrained by the grammar, resulting in 
syntactic ambiguity.  In fact, even with Marcus's constraint, the degree of 
ambiguity is exponential in the length of the compound.  Furthermore, 
analysis of these ambiguities appears to depend primarily on semantic 
expectations.  The lack of syntactic and morphological constraints 
immediately forces us to confront the use of semantics during the syntactic 
analysis, suggesting an interleaved approach.  The question becomes 
how do we efficiently bring the appropriate semantic constraints to bear,
and how do we acquire the constraints in the first place?  

\subsection{The Meaning of Noun Compounds} 
\label{sec:cn_meaning}  

Since the purpose of compounding is to identify things by way of relating 
them to others, a semantic analysis of a compound involves identifying the 
particular relationship involved.  For instance, in 
example~\ref{eg:cn_meaning_implicit}\ref{eg:cn_meaning_implicit_time} 
the time of the activity is being specified.  In 
example~\ref{eg:cn_meaning_implicit}\ref{eg:cn_meaning_implicit_cause} 
a causal relationship is denoted, and we take 
example~\ref{eg:cn_meaning_implicit}\ref{eg:cn_meaning_implicit_made} 
to mean a pillar made of cement.  
\begin{examples}
   \item	\label{eg:cn_meaning_implicit}
     \begin{subexamples}
       \item morning prayers   \label{eg:cn_meaning_implicit_time}  
       \item drug deaths     \label{eg:cn_meaning_implicit_cause}  
       \item cement pillar   \label{eg:cn_meaning_implicit_made}  
     \end{subexamples}
\end{examples}
But the relationship in each case is left implicit, so that if automatic 
interpretation is desired an additional problem of identifying this 
relationship arises.  In this section, I will review linguistic 
theories that have been proposed regarding the semantics of noun 
compounds.

First though I will describe one aspect that is universally recognised.  
There are two roles to be played in 
compounding: one noun denotes the thing to be identified, exactly as if it 
were not part of a compound, while the other noun denotes a related 
thing.\footnote{Warren~(1978:19) observes that this is arguable in the case 
of so called \newterm{dvandva} compounds, for example 
\lingform{poet-painter}.}  The former is called the \newterm{head}, the 
latter the \newterm{modifier}.  The class of things denoted by the compound 
as a whole is generally a subset of the class denoted by the head.  The 
modifier determines which subset.  For instance, 
example~\ref{eg:cn_meaning_implicit}\ref{eg:cn_meaning_implicit_time},
the things denoted are prayers.  The word \lingform{morning} denotes a related
thing (the early part of the day) which identifies a subset of the possible
prayers.

The semantic head is also the syntactic 
head: the agreement features of the whole compound are inherited from the 
head.  In English the head of a compound is almost always the rightmost 
noun.  This is true of all the examples seen so far.  Exceptions, such as those 
in example~\ref{eg:cn_meaning_leftheads}, are usually lexicalised and often 
borrowed from other languages.
\begin{examples}
   \item	\label{eg:cn_meaning_leftheads}
     \begin{subexamples}
       \item attorney general
       \item fetucinne arabiata
     \end{subexamples}
\end{examples}

\subsubsection*{Range of semantic theories}

Apart from right-headedness, the semantic properties of compounds have 
been hotly debated in linguistics, with numerous contradictory views being 
proposed.  At the optimistic end of the scale, it has been suggested that there 
exists a small set of semantic relationships that compounds may imply.  
Levi~(1978) propounds a theory along these lines, which I will describe 
below.  If these theories are correct, they provide a firm basis for 
computational interpretation, since they specify a concrete list of 
possibilities.  

In contrast, the most pessimistic views claim that the implicit relationship 
between head and modifier is entirely unconstrained.  For example, 
Downing~(1977) performed a series of psychological experiments to 
support her argument that the semantics of compounds cannot be exhausted 
by any finite listing of relationships.  
Her results suggest that the number 
of relationships is very large.  However, she does observe that certain types 
of relationship are more common than others (including purpose, part-whole 
and place).\footnote{Another advocate of the view that implicit semantic 
relationships are unconstrained is Nishikawa~(1988), although the 
arguments he puts forth also apply to the whole of language: all utterances 
can be used to mean anything because of shared socio-cultural knowledge.
If his argument is accurate, the entire \nlp\ enterprise is impossible.}

If Downing's~(1977) hypothesis is true in a strong sense, then noun 
compound interpretation requires complete pragmatic context-dependent 
inference and therefore any attempt to handle noun compounds will involve 
detailed manual knowledge coding.  However, noun compound semantics do 
form certain patterns.  There is certainly hope that adopting a 
more restrictive theory that relies on these patterns will prove useful, 
even if there are cases that it is 
unable to handle.  This is exactly the situation in which statistical language 
learning earns its living.  

Between the two ends of this scale, a range of noun compound
semantics theories exist which I will survey in a moment.  There are 
two methodologies used to develop such theories.  The first, which I will call 
\newterm{example-based}, is to collect a set of example compounds, most 
usually without regard to the source, 
with the aim of arriving at as large a list 
of examples as possible.  The list that results contains compounds, but no 
contexts, so that the theory is based on the \scare{out of context 
interpretation} of the examples.  Levi~(1978) supplies her list of around 200 
examples in an appendix.  Vanderwende~(1993) also follows this 
methodology, using 59 compounds (but does not supply the list).  Others 
base their theory on examples, but do not appear to have a fixed list (for 
example, Finin's,~1980, thesis contains on the order of 100 scattered 
examples).

As I will argue in section~\ref{sec:cn_context}, the interpretation of 
compounds can readily change in different contexts.  We have already seen an 
instance where this is possible because of syntactic ambiguity in 
example~\ref{eg:cn_grammar_both} of section~\ref{sec:cn_grammar}.  
Therefore, example-based theories of compound noun semantics base their 
conclusions (at least ostensibly) on only context free interpretations.

The second methodology, which I will call \newterm{corpus-based}, 
acquires noun compounds from a corpus.  In these studies, every compound 
from a corpus is used, thus requiring the theory to be comprehensive, at least 
for the text types that the corpus represents.  This approach also permits the 
research to measure the relative frequencies of different semantic relations, 
which ensures that emphasis is given to semantic behaviours that are more 
likely to be encountered.  Furthermore, it results in each occurrence of a 
compound having a known context, within which it can be 
interpreted.  There is no longer any need to assume that compound meaning 
is well-defined outside of any context.  The corpus-based methodology 
focuses on tokens, the example-based one on types.  Warren~(1978) and 
Leonard~(1984), both described below, use a corpus-based methodology.

Three linguistic theories of semantics and one \nlp\ theory are sufficiently 
prominent to be worth describing.  In order of the degree to which they 
constrain the possible semantics, they are:
\begin{itemize}
\item Levi's~(1978) recoverably deleteable predicates and nominalisations;
\item Leonard's~(1984) typology of compounds in fictional prose; 
\item Warren's~(1978) analysis of Brown corpus compounds; and
\item Finin's~(1980) role-filling theory, incorporating role nominals.
\end{itemize}
I will describe each of these in turn.

\subsubsection*{Levi's RDPs}

Levi~(1978) pursues the idea that compound nouns express only a small set 
of semantic relations within the generative semantics theory.  She proposes 
nine \newterm{recoverably deleteable predicates}, which are expressed 
either as prepositional phrases or as relative clauses involving a small set of 
verbs.  These are:
\begin{itemize}
\item \semrel{in}, \semrel{for}, \semrel{from} and \semrel{about}; and
\item \semrel{cause}, \semrel{have}, \semrel{make}, \semrel{use} and 
\semrel{be}.
\end{itemize}
Thus, \lingform{electricity station} implicitly involves
a predicate which can be expressed as \lingform{station that makes 
electricity}.  She claims that, apart from nominalisations, these predicates 
are the only relationships possible.  To the extent that this is true, it provides 
a practical basis for analysing compound noun meanings: we need only 
choose which predicate has been deleted in order to perform interpretations 
(or at least reduce the task of detailed interpretation to something simpler).

In addition, Levi identifies a separate class of compounds, the 
nominalisations.  These are produced through a somewhat different 
process, being derived from verbal constructions.  The head noun is 
derived from a verb by morphological modification and carries its 
arguments along with it.  The semantics are defined in terms of the following 
four verb roles:
\begin{itemize}
\item \semrel{act}, \semrel{product}, \semrel{agent} and \semrel{patient}.
\end{itemize}
For example, \lingform{truck driver} expresses an \semrel{agent} role 
because it is derived from the clause \lingform{agent drives a truck} and  
\lingform{student discontinuations} expresses an \semrel{act} role because 
it is derived from the clause \lingform{students act to discontinue}.
Thus Levi claims that all noun compounds express one of thirteen
predicates.

\subsubsection*{Leonard's typology}

While Leonard~(1984:6) gives a shorter typology of noun compounds,
it is derived only from fictional prose.  The types are 
defined chiefly with reference to possible paraphrases.  An example
of each type follows.
\begin{itemize}
\item Sentence: \lingform{hire car}
\item Locative Sentence: \lingform{breakfast-room}
\item Locative: \lingform{country free-school}
\item Annex: \lingform{mountain top}
\item Equative: \lingform{baby crocodile}
\item Material: \lingform{stone lion}
\item Additive: \lingform{Lampton-Lufford report}
\item Reduplicative: \lingform{plop-plop}
\end{itemize}
Each of the eight types is characterised by 
an associated paraphrase.  For example, the Annex type compound 
\lingform{blood pressure} can be paraphrased as \lingform{pressure of 
a blood/bloods}.  

The typology is based on a test set of 1944 example compounds 
taken from a 305 thousand word corpus of fictional prose.  Here, as in the 
other three theories reviewed in this section, number and definiteness are not 
captured by the semantic representation so that 
alternatives appear in the paraphrase.\footnote{Chambers~(1994) observes that 
both number and definiteness are sometimes determined
by the semantic relationship, which might be used to fill this gap.}  
Leonard~(1984:70) also describes an implemented algorithm 
for analysing compounds into these eight groups.  
It is based on a hand-crafted lexicon (specific to the test set).  
Each word is marked with one or more of 59 markers, which are then used 
by a sequence of rules to choose the correct type.  The type is then used to 
produce a paraphrase, which is evaluated for felicity by hand.  
\shortquote{At a generous count, 76\% of the interpretations are possible 
ones in English}~(Leonard,~1984:v).

\subsubsection*{Warren's taxonomy}

Warren's~(1978) theory is far less constraining.  
She has made a comprehensive study of compound nouns in 360 
thousand words of the Brown corpus.  She manually extracted every 
non-verbal-nexus noun compound from the sample, which included text 
from eight of the registers represented (recall that the Brown corpus contains 
a carefully selected range of genres).  This yielded 4557 different 
compounds (all her statistics are given as counts of types, rather than tokens; 
see Warren,~1978:54).  She then developed a taxonomy of implicit semantic 
relations, with four hierarchic levels of abstraction: major semantic classes, 
minor semantic classes, main groups and subgroups.  Each compound was 
assigned to a subgroup (or main group where subgrouping was not 
distinguished within the group).  Her major semantic classes were:
\begin{itemize}
\item \semrel{constitute}: divided into Source-Result, Result-Source and 
Copula classes;
\item \semrel{possession}: divided into Part-Whole, Whole-Part and 
Size-Whole classes;
\item \semrel{location}: divided into Goal-\semrel{obj}, 
Place-\semrel{obj}, Time-\semrel{obj} and Origin-\semrel{obj} classes;
\item \semrel{purpose}: divided directly into main groups;
\item \semrel{activity-actor}: divided directly into subgroups; and
\item \semrel{resemblance}: consisting only of the Comparant-Compared 
class.
\end{itemize}

The distribution of compounds across the taxonomy is quite informative 
(although it is given in terms of types rather than tokens, so it reflects the 
productivity of the different semantic relations as much as it does the 
frequency).  The counts of each group are tabulated in each section, with 
counts by semantic class tabulated in the summary (Warren,~1978:229).  
The distribution strongly favours some classes, with the most common being 
Whole-Part (23\%).  Only six of the classes cover more than 5\% of the 
distribution (Whole-Part, Source-Result, Purpose, Place-\semrel{obj}, 
Part-Whole and Origin-\semrel{obj}).  Copula compounds occupy 5\% of 
the distribution, and the major semantic class Resemblance, just 1.8\%.  If 
the major semantic classes can be considered coarse semantic relationships, 
then the distribution is evidence that a theory such as Levi's has 
some explanatory power, in spite of readily discoverable exceptions.

\subsubsection*{Finin's role nominals}

While these three linguistic theories of compound noun semantics treat  
nominalisation as one of two distinct processes (Warren,~1978, is careful to 
exclude verbal-nexus compounds; Leonard's,~1984, program uses a special set 
of secondary, lexical-valued features to analyse Sentence and Locative 
Sentence types), another view holds that most compounds are implicit 
nominalisations, even if the head noun is not morphologically derived from a 
verb.  

Finin~(1980) adopts this perspective, claiming, for example, that 
\lingform{recipe book} is an implicit nominalisation of a verbal construction 
involving \lingform{write}, rather than a deletion of the predicate expressed 
by \semrel{about}.  He calls such constructions role nominals.  A 
semantic analysis of a compound consists of first identifying the event 
denoted by the implicit verb and then assigning the objects denoted by other 
nouns to roles of the event.  In this example, \lingform{recipe} denotes an
object that fills the topic role of a writing event that is implied by 
\lingform{book}.  Under this view, the set of possible semantic relations is 
arbitrarily large, determined by the range of possible implicit
verbs and their roles.

One problem with this approach is that we are forced to posit large numbers 
of lexical relations without direct linguistic justification (although qualia 
theory might be adapted to this end, Pustejovsky~\etal,~1993).  For 
example, we must suppose that every physical object implies a composition 
event for which something has filled the material role.  Thus, 
\lingform{fuselage} must imply a verbal construction involving 
\lingform{made}, which has a composition role fillable by 
\lingform{aluminium}, in order to analyse \lingform{aluminium 
fuselage}.\footnote{Finin~(1980) and Isabelle~(1984) both explore 
sublanguages about aircraft maintenance.}  This difficulty is also observed 
by Isabelle~(1984), who adopts the role nominal theory.  \shortquote{For 
those role nominals where there is no morphological evidence of relatedness 
with the underlying verb, one is forced to rely mostly on 
intuitions.}~(Isabelle,~1984:511).

\subsubsection*{Common elements}

All of these theories have strengths and weaknesses, so rather than
commit to any one, I will simply mention four properties that these
and almost all others share.
\begin{description}
\item[Permanence:] Most theories observe that semantic relations between 
nouns in a compound involve inherent or typical properties of the objects 
denoted.  For example, a \lingform{sea mammal} never denotes a mammal 
that happens momentarily to have gone for a dip.  In theories that constrain 
the set of possible relations, this causes all the relations to be interpreted 
with this in mind.
\item[Negative relations:] Even the most pessimistic theories agree that 
modifiers never denote objects that are specifically {\em not} in a certain 
relationship to the head.  For example, even if all the paintings in a
room depict tubas except for one, \lingform{tuba painting} will 
never be used to denote that one painting.
\item[Nominalisations versus predicates:] Apart from Finin's~(1980) role 
nominals, two different types of semantic relation, corresponding to the two 
different syntactic types identified in section~\ref{sec:cn_grammar}, are 
always suggested.  Also, most theories include a copula predicate.
\item[Appeal to paraphrase:] In almost all theories the most explicit 
diagnostic for determining the semantics of a compound is through 
comparison of possible paraphrases.  For example, the most concrete 
method of checking for a Purpose relation involves considering a paraphrase 
employing the preposition \lingform{for}.  I will return to this point in more 
detail in section~\ref{sec:ce_problem}.
\end{description}

%

\subsection{Accommodation and Semantic 
Granularity} 
\label{sec:cn_accommodation}

The hypothesis that compounds can express only a small set of semantic 
relationships rests on the view that semantic representations are limited in 
their detail.  Not every detail of the relationship will be explicit in the 
representation.  For instance, to say that a causal relationship is implicit in 
the compound \lingform{earthquake casualties} is a summary of finer 
semantic details.  Nothing is said as to whether the causer is the primary 
influence, or merely a contributor.  Nor is it specified whether the effect was 
intentional or otherwise.  We accept that a causal relationship exists, even 
though there are many and varied shades of causation.  Regardless of which 
theory is adopted, meaning analyses are necessarily limited in resolution.

As Ryder~(1994:91) observes, listeners adapt the general meanings of the 
nouns involved in order to better fit one another and the context.  She calls 
this process \newterm{accommodation}.  In fact, accommodation is a 
fundamental part of all language.  There are arbitrarily fine shades of 
meaning to be captured by a discrete symbolic language, resulting in a loose 
fit between words and interpretations.  A person reading words must 
accommodate the out of context meanings of words to find an in context 
interpretation.

This is apparent in the divisions created by word senses in a dictionary.  
As Kilgarriff~(1992) argues, each word has a potentially infinite variety 
of usages, dependent on the context in which it is used.  
The lexicographer chooses to divide this space 
into discrete senses because it is impossible to give an exhaustive listing of 
the slight nuances conveyed by the word in each different context.

Importantly, in dividing the space, the lexicographer has some choice.  In 
large, detailed dictionaries, there will be many fine-grained senses, each 
carefully distinguished from one another.  In smaller dictionaries, the senses 
must be coarse-grained, making fewer distinctions.  Thus, descriptions of 
word meaning have \newterm{granularity}, which may range from coarse to 
fine, depending on the degree of detail distinguished.

Likewise, a semantic analysis of compound nouns is necessarily given at a 
certain level of granularity.  A highly sophisticated analysis might give great 
detail about the relationship between the two objects, whereas a simpler 
analysis might summarise finer points under one heading.  In fact, 
Levi~(1978:85) devotes some discussion to justifying her particular choice 
of grain size (the nine recoverably deleteable predicates) and concludes that 
her analysis \shortquote{incorporates neither the maximal nor the minimal 
degree of generalisation possible, but rather an optimal degree}.

The evaluation of a representation, including the appropriateness of the 
granularity, naturally depends on the goal.  It should be judged by its 
usefulness in performing the intended task.  Therefore, is it perfectly 
legitimate to claim that \lingform{hydrogen bomb}, \lingform{machine 
translation}, \lingform{smoke signal} and \lingform{steam iron} are 
expressing an Instrument relation, if this is a useful classification for the 
purpose at hand.  

This issue is fundamental to natural language interpretation and raises some 
questions about the distinction between syntax and semantics.  The syntactic 
analysis of compound nouns is isomorphic to a semantic analysis in which 
just one general relationship, that expressed by something like 
\semrel{is-a-modifier-of}, is modeled.  This homomorphism is one that I 
will exploit in the experimental work on parsing in 
chapter~\ref{ch:experimental}.  However, for now I turn to a more 
practical problem posed by compound nouns: dependence on context.

\subsection{Context Dependence} 
\label{sec:cn_context}  

The meaning of a compound noun, and even its syntactic structure, can vary 
under different conditions.  Information that contributes to the analysis can 
come from other words in the same sentence, surrounding sentences, 
non-linguistic knowledge or even pragmatic considerations.  It is not 
generally sufficient to analyse a particular noun compound (token) based 
only on the noun compound itself (type).  

For a start, word meanings vary depending on context.  If the most common 
meaning of a word is not the one being used, this can affect the preferred 
reading of a compound.  In example~\ref{eg:cn_context_senses} we expect 
a right-branching analysis (a fee to cover drinks exacted by a nightclub upon 
entrance).  
\begin{examples}
\item club cover charge   \label{eg:cn_context_senses}
\end{examples}
However, if the context reveals that the speaker is in a golfing shop 
purchasing accessories, then the senses of both \lingform{club} and 
\lingform{cover} will be different and in this case a left-branching 
structure is more likely (the price of a plastic jacket used to 
protect golf sticks).

Even if the word senses are known, some context sensitivity remains.  In 
example~\ref{eg:cn_context_syntax} context is needed to determine 
whether the system referred to is a system for defense which involves 
missiles (right-branching) or a system for defense against missiles 
(left-branching).  
\begin{examples}
\item missile defense system   \label{eg:cn_context_syntax}
\end{examples}
In both these examples, the semantics is dramatically affected by context of 
various kinds.  Information determining which reading is intended could be 
derived from a range of sources including pragmatic inference, discourse 
structure and topic.  Other examples are discussed in Sparck Jones~(1983).

While context dependence is an obvious and universal feature of 
language,
it is perhaps of particular 
significance in noun compounding.  Because one of the functions of noun 
compounds is to compress text, it is plausible that a greater reliance is 
placed on context than in other types of syntactic construction.  If there is 
less information on the page, there must be more information coming from 
elsewhere.  More detail must be left implicit, and thus context dependent.

Meyer~(1993) places great emphasis on context, dividing his thesis into two 
halves, dealing with compounds in isolation and compounds in context, 
respectively.  He gives a formulation of compound noun meanings within a 
formal semantics framework where compounds are treated as anaphors.  
When interpreted in context, the semantics of a compound are defined by 
reference to a discourse representation.  According to his theory, 
\shortquote{concepts denoted by novel \acronym{nn}-compounds must be 
linked to conceptual nets denoted by preceding 
discourse}~(Meyer,~1993:169).  Therefore, context must have a significant 
influence.  In fact, the limitation of sources of contextual effect to {\em 
preceding text} is too restrictive.  Example~\ref{eg:cn_context_senses} 
might appear before a price on either a small tag in a golf shop or a sign 
beside the doorway of a night spot.  In both cases there would be no 
preceding text and yet the concepts denoted would differ.  

Almost all computational approaches to noun compounds have made the 
practical compromise of assuming that context dependence is either 
negligible or of little impact.  However, at least two works incorporate a 
mechanism that, in principle, allows for analyses to depend on the 
denotation of preceding text, or even earlier texts.

McDonald~(1982:20) utilises a scoring function to control application of a 
range of heuristics (his work will be described in more detail in 
section~\ref{sec:cn_knowledge} below).  One of the these, called the 
\newterm{cognate heuristic}, gives preference to conceptual structures that 
have already been stored in memory.  
When there is more than one of the possible 
structures stored in memory, the \newterm{instances heuristic} then applies, 
favouring the one appearing most often.  This mechanism is used to 
manually supply the meanings of lexicalised compounds to the system, but, 
in principle, earlier processing could 
lead to cognates in memory and therefore to context dependent effects.

A similar strategy constitutes the entire noun compound interpretation 
algorithm described by Lehnert~(1988).  This algorithm is used by the 
\progname{researcher} system, built by Lebowitz~(1983), which 
incorporates an episodic representation of memory.  As text is processed, 
interpretations are added to an inheritance hierarchy, which is then searched 
during later interpretations and used in a form of case-based reasoning.  
Again, frequency is used as a criterion when multiple alternatives exist.  The 
result demonstrates powerful context modeling effects.  However,  complete 
reliance on episodic knowledge is probably weaker than complete reliance 
on context independent semantic knowledge.  A method for combining the 
two (such as McDonald's,~1982) seems more promising.

In any case, there are strong arguments for requiring context dependence in 
compound noun analysis.  In section~\ref{sec:cy_human} I will report a 
study on the extent of context dependence in noun compound syntax.  
However, except for quantifying its effects, I shall henceforth follow most 
others in attempting to ignore it.  Such insensitivity is a weakness of the 
present work, although at this stage it appears to be a necessary one.

\subsection{Computational Tasks} 
\label{sec:cn_computational}   

Researchers in computational linguistics have addressed at least five 
computational tasks to do with noun compounds, 
all of which could benefit from statistical \nlp\ 
techniques.  In this section I will review existing work on:
\begin{itemize}
\item identification of compounds from amongst other text;
\item syntactic analysis of structurally ambiguous compounds;
\item assignment of implicit semantic relations;
\item prediction of prosodic features of compounds; and
\item directly translating compounds in one language to phrases in another.
\end{itemize}
Of these, parsing and semantic analysis have received most attention
and therefore will be reviewed in more detail in 
sections~\ref{sec:cn_knowledge} 
and~\ref{sec:cn_statistical}.\footnote{Another 
interesting task is the generation of 
compounds that are synonymous with given query terms for information 
retrieval.  Since compound nouns are frequent in abstracts, especially in 
technical domains, Norris~(1993) predicts that accurately generating 
compounds from synonymous expressions will have an impact on 
retrieval performance.}

\subsubsection*{Identification}

Nouns are commonly ambiguous for part of speech. Thus parsers may easily 
confuse noun compounds for other constituents.  
For instance, example~\ref{eg:cn_computational_ident} could be 
taken as a verb phrase rather than a noun compound because
\lingform{ditch} may be a verb.  This analysis could lead to 
interpretation of the compound as an instruction to discard 
the digging machine.
\begin{examples}
\item ditch digging machine \label{eg:cn_computational_ident}
\end{examples}

Arens~\etal~(1987) describe a system for 
analysing digital system specifications in which
this problem arises frequently.  Their 
solution uses hand-coded rules called \newterm{pattern-concept pairs}, 
which encode semantic expectations in the domain.  These expectations are 
then applied by a heuristic that also checks number agreement constraints.  If 
the system encounters an ambiguity that it cannot resolve, it requests the 
user's assistance.  No coverage or accuracy figures are reported (the lexicon 
contains 25 verbs and 100 nouns).

Using hand-coded rules may be adequate for limited domains, but is expensive in 
development effort and is difficult to scale up to broad coverage 
language processing.  Statistical methods offer a cheaper means of acquiring 
domain specific expectations and may even be applicable to 
unconstrained text, if a sufficiently large corpus can be obtained for training.  
While exploration of this problem is not pursued in this thesis, it is possible 
that the same statistics used in the experimental work on parsing in 
chapter~\ref{ch:experimental} could be used in solving this problem.  

Note that the related problem of identifying terms or collocations has 
received a great deal of attention, but is tangential to the main line of the 
present work.  A good starting point is the work of Smadja~(1993).

\subsubsection*{Parsing}

As we have seen, compounds longer than two words are syntactically 
ambiguous and, as demonstrated in section~\ref{sec:cn_grammar}, the parse 
ambiguity of compounds grows exponentially with their length.  
Since ter~Stal~(1994a) reports finding compounds of up to eight words in 
length in a relatively small corpus, the level of ambiguity could be crippling 
to any uninformed parser.  

Existing work on parsing noun compounds falls into two classes:
knowledge-based and statistical methods.  Since 
sections~\ref{sec:cn_knowledge} and~\ref{sec:cn_statistical} below
give details of these, I will delay review of such work until then.
It is however noteworthy that 
most of the early knowledge-based approaches to noun compounds (for 
example, Finin,~1980) perform parsing in combination with semantic 
interpretation.  ter~Stal~(1994b) suggests that treating parsing as an 
independent aim may be inappropriate because subsequent semantic 
processing applies the same knowledge.  Nonetheless, syntactic analyses have 
independent value beyond being a stepping stone to semantics.  For 
example, they are useful for predicting prosody in text-to-speech systems,
even when no semantic analysis is carried out.
It is therefore useful to distinguish
parsing and semantic analysis.
The first half of chapter~\ref{ch:experimental} of this thesis
will be devoted to work on applying statistical methods to parsing 
compound nouns.  

\subsubsection*{Semantic analysis}

As outlined above in section~\ref{sec:cn_meaning}, compounds can express 
a variety of relationships, 
with selection of the appropriate relation being left 
up to the reader.  Computationally, the task of selecting the most likely 
implicit relationship is highly knowledge intensive.  As noted above, 
Leonard~(1984) has built a system for assigning a semantic analysis 
to compounds in fictional prose
based on lexical features and a set of rules.  
The system requires detailed entries in 
the lexicon to achieve about 76\% accuracy on the development set.
Other relevant work employs knowledge-based methods and will be
reviewed in section~\ref{sec:cn_knowledge}.

Knowledge acquisition is the central barrier to solving this task in 
unconstrained text.  The second half of chapter~\ref{ch:experimental}
of this thesis will 
be devoted to work on applying statistical methods to compound noun 
semantic analysis.  While Johnston~\etal~(1994) propose a method for 
acquiring appropriate lexical-semantic information from corpora, the results 
reported later in this thesis represent the first empirical study 
of which I am aware.  

\subsubsection*{Prosody}

Work in text-to-speech systems has investigated the 
prosodic features of compounds.  The phonological stress of noun 
compounds varies with their syntactic structure, among other things.  In 
order to synthesise realistic speech, a text-to-speech system must be capable 
of placing the correct emphasis on elements of a compound.  
Example~\ref{eg:cn_computational_prosody} shows compounds with their 
typical stress patterns; intonationally prominent words are in bold face.
\begin{examples}
    \item  \label{eg:cn_computational_prosody} 
        \begin{subexamples}
	\item \stress{panic} attack
	\item \stress{living} room \stress{table}
	\item \acronym{risc} \stress{instruction} set
        \end{subexamples}
\end{examples}

Sproat and Liberman~(1987) describe a rule-based method for assigning 
stress to compounds for text-to-speech purposes.  More recently, 
Sproat~(1994) has applied statistical methods to the problem and given a 
quantitative evaluation of the performance on two word compounds.  A test 
set of 940 compounds (types) were analysed with 84\% accuracy (assigning 
leftmost accent yields 70\%).  When combined with the previous rule-based 
method by employing statistics only when no rules apply, a different test set 
of 1138 compounds (types) was analysed and achieved an average 
agreement with human judges of 85\%.  The rules were used in only 15\% of 
cases, showing the limitations of the manual coding approach.  Inter-judge 
agreement rates were 91\%.  

There is a close relationship between syntactic structure and accent contours 
in compounds.  However, the relationship is not straightforward.  As 
Sproat's~(1994) work shows, even syntactically unambiguous compounds 
have varying stress.  The implied semantic relationship can be a factor, as 
shown by the two readings in 
example~\ref{eg:cn_computational_semprosody}.  The first is typical when 
a Purpose relationship is implied, the second in the case of a Made-Of 
relation.
\begin{examples}
    \item \label{eg:cn_computational_semprosody}
        \begin{subexamples}
	\item \stress{chocolate} money 
	\item chocolate \stress{money}
        \end{subexamples}
\end{examples}
In this work I will not address the problem of stress assignment, but 
strong parallels exist between work on stress assignment and 
the syntactic and semantic analyses pursued in chapter~\ref{ch:experimental}.

\subsubsection*{Direct translation}

While one of the important motivations for performing syntactic and 
semantic analyses of compound nouns is to allow sophisticated machine 
translation of compounds, some research simply sets the task to be direct 
translation from one language to another, bypassing the need for language 
independent syntactic and semantic representations.  

Rackow~\etal~(1992) describe an innovative noun compound translation 
algorithm that employs statistics from an English corpus to select the 
appropriate translation of a German compound.  First, segmentation rules 
are used to break down the German compound into morphemes, which are 
then translated into sets of possible English words.  Generation rules are 
then used to formulate a set of possible translations, including English 
adjectival premodifiers, prepositional phrases and noun compounds.  

The correct translation is selected by counting the occurrences of each of the 
possible translations in the English corpus and choosing the one appearing 
most frequently.  The selection of lexical items and syntactic construction is 
performed solely by the corpus statistics, allowing the lexical transfer and 
generation rules to overgenerate heavily.  A strong advantage of this 
approach is that only a monolingual (target language) corpus is required.  
Rackow~\etal~(1992) give no quantitative evaluation.

Along similar lines, Jones and Alexa~(1994) describe a method for aligning 
English word sequences with German compounds in a parallel corpus based 
on the alignment model of Brown~\etal~(1993).  Statistical associations are 
derived between German compounds and word sequences commonly found 
in the corresponding English sentences, with the aim of avoiding the need 
for manual development of symbolic translation rules.  Unfortunately, the 
corpus is very small (2543 English words) and only qualitative evaluation is 
reported.  Also, a sentence aligned parallel corpus is required.

Finally, Maas~(1994) reports on \progname{mpro}, a transfer-based 
approach to translating compounds from German to French.  Much of the 
work is done by phrasal transfer, with whole compounds appearing in the 
lexicon.  For other compounds, a simple rule is used that expands the 
German compound modifier into a French prepositional phrase using 
\lingform{de} (\scare{of}).  No quantitative evaluation is reported.

In this thesis, direct translation will not be addressed.  There is no reason, 
though, why the probabilistic models and training methods given in 
chapter~\ref{ch:experimental} for the syntax and semantics of noun 
compounds could not be applied to translation.  In particular, the 
probabilistic models could replace the frequency counting method used by 
Rackow~\etal~(1992), either to overcome data sparseness or to make use of 
a source language corpus rather than a target language one.

\subsection{Knowledge-based and Dictionary-based 
Approaches}   \label{sec:cn_knowledge}   

Most of the prior work proposes highly knowledge intensive algorithms for 
analysing compound nouns.  Early work was concerned with making use of 
the knowledge representation schemes developed in the late 1970s, and 
worked primarily by means of slot filling.  The key idea of these methods is 
that the concept denoted by one noun contains slots and that the concept 
denoted by the other noun should fill one of these slots.  Syntactic and 
semantic analysis is performed by deciding which concept should occupy 
which slot.  The algorithms are therefore concerned with evaluating the 
appropriateness of concepts as possible fillers of slots.

The first such system was \progname{jets} (Finin,~1980), 
a program for analysing 
compounds in queries given to an aircraft maintenance database (called 
\progname{planes}).  It employs a set of rules to interpret compounds by 
slot-filling.  

For example, the noun \lingform{flight} is associated with a frame called 
\scare{to-fly}, which has a slot called \scare{time}.  This slot has associated 
with it various facets, including specifications of what fillers are the 
preferred, the default and the typical filler, and also what properties are 
required of the filler.  Given example~\ref{eg:cn_knowledge_finin}, the 
relevant slot filling rule matches the frame for \lingform{January} against 
the various slots of \scare{to-fly}.  
\begin{examples}
    \item January flight \label{eg:cn_knowledge_finin}
\end{examples}

A scoring system evaluates each possibility.  For example, if the modifier 
matches the default for a slot, this contributes 8 points to the score.  Once 
scores are computed for each possible slot-filling rule and frame-slot 
combination, the system selects the highest scoring analysis.  

The system contains close to 200 concept frames and six interpretation rules.  
No quantitative evaluation is given, although it is clear that substantial effort 
must be expended to encode the necessary knowledge for even a very 
restricted domain.

A similar approach, but in this case aimed at unrestricted text, is described in 
McDonald~(1982).  Here, the knowledge used is expressed using semantic 
networks in which each concept has associated slots with selectional 
restrictions.  By a process of matching slots, a structural analysis is produced 
that also supplies the implicit semantic relationships.  

The system has a vocabulary of just under 200 words and works for around 
25 noun compounds.  Yet McDonald claims that most of the over 600 
examples given in the appendix could be handled if only the hand coded 
semantic knowledge were available.
\begin{citedquote}{McDonald,~1982:125, his bold}
Once this list of more than six hundred compounds was created, each 
compound was examined to see how well the model presented above would 
process it.  The program itself was {\bf not} run on all these examples.  
Instead the processing for this set of compounds was done by hand using the 
algorithms described above.  There are two reasons \dots [First, it would 
be] necessary to add a large amount of knowledge to the data base.  All this 
knowledge would have to be added to the data base by hand \dots [Second, 
the] amount of memory available to store the knowledge is very limited.
\end{citedquote}

In fact, even given the questions raised by this method of evaluation, the 
results are disappointing.
\begin{citedquote}{McDonald,~1982:158--9}
The program as it currently stands can [given appropriate knowledge] 
process about 60\% of the compounds encountered in real-world 
text \dots Another approximately 30\% of the compounds can be processed 
correctly if some reasonable assumptions are made and if a couple of new 
patterns are added to the system.
\end{citedquote}

It seems likely that encoding the knowledge McDonald requires for anything 
broader than a narrow domain is a highly difficult (if not impossible) task, 
especially since each new piece of knowledge would potentially introduce 
errors in the earlier correct analyses.
It is therefore far preferable to turn to 
an automatic method of acquiring and applying the required knowledge.  

This conclusion is supported by an information retrieval study by Gay and 
Croft~(1990).  They build a slot-filling noun compound interpretation 
program and evaluate its impact on retrieval performance in a limited 
domain.  Their conclusion is that the cost of building such a system cannot 
be justified, primarily because a simple statistical strategy, using a three 
word window achieves comparable performance.
\begin{citedquote}{Gay and Croft,~1990:36}
To obtain these performance improvements, which are almost certain to be 
extremely small, a prohibitive effort involving construction of the 
knowledge base and processing document text would be required.
\end{citedquote}

More recently, with the emergence of unification-based \nlp, there have been 
some proposals for noun compound interpretation algorithms based on 
unification.  In these proposals, feature structures replace the earlier 
slot-filling rules, 
with unification being used to constrain the possible fillers 
(see, for example, the proposal of ter~Stal and van~der~Vet,~1994).  While 
the underlying knowledge representation is more advance, these proposals 
can only work for a limited domain because of the requirement for detailed 
hand-coding.  

Wu~(1992) proposes a novel combination of statistical and 
unification-based methods, in which a probabilistic model of 
feature structures is developed.  
Probabilities are assigned to uninstantiated feature structures 
and the maximum entropy principle used to define the probabilities of fully 
instantiated ones.  An efficient algorithm for approximating the probabilities 
of different possible analyses is given.  However, analysis of noun 
compounds still requires manual construction of feature structure 
representations, and estimation of the probabilities of the uninstantiated 
feature structures poses an additional problem.  No evaluation of the 
algorithm's performance is given.

Another recent proposal, described in detail by Hobbs~\etal~(1993), 
holds that the meaning of a compound can be arrived at by abductive 
inference.  The referents of nouns within the compound are denoted by 
logical literals and an explanation for their relationship is sought through 
abduction.  While Hobbs~\etal\ claim that the computational problems 
involved in using general inference can be solved by means of weighted 
abduction (see Charniak,~1986 and Goldman and Charniak~1990), the 
problem of furnishing the enormous amount of world knowledge required 
remains.

All these proposals rely on hand crafted symbolic knowledge bases.  As a 
result, any system implementing these proposals must have a very limited 
vocabulary.  Even though such limits are acceptable within a narrow 
domain, the development effort required for each system makes these 
approaches very expensive.  None of the knowledge-based systems built by 
\nlp\ researchers have been subjected to quantitative evaluation.  This is to 
be contrasted with Leonard's~(1984) computer program, where a careful 
corpus-based performance test was conducted.  While adding knowledge to 
noun compound processors appears tempting, careful attention must be paid 
to how this knowledge can be provided.

One possible approach to knowledge acquisition is the use of dictionaries as 
discussed in section~\ref{sec:sn_motivations}.  Vanderwende~(1993) describes 
a design of this kind for the task of assigning a semantic relationship to 
two word compounds.  It uses a pattern-based method to acquire 
knowledge from dictionary definitions.\footnote{See 
section~\ref{sec:sn_motivations} for background on pattern-based methods
of dictionary processing.}  Using a set of 25 weighted analysis 
heuristics that use the knowledge from these definitions, a score is given to 
each of 11 possible relationships. For example, one of the definitions for 
\lingform{sanctuary} is \scare{an area for birds \dots }, which results in 
\lingform{sanctuary} having an \semrel{is-for} attribute 
with value \lingform{birds}.  
This is used by one of the heuristics to give a score of 0.9 to a Purpose 
relation in the compound \lingform{bird sanctuary}.  The scores are adjusted 
manually.
\begin{citedquote}{Vanderwende,~1993:172}
The specific values for the likelihood factors result from experimentation, 
but are not ad hoc; their effects converge, in a set of heuristics, to give 
consistently satisfactory results.
\end{citedquote}
The heuristics are based on a development set of 59 compounds, 
of which 49 are correctly analysed.

A corpus-based evaluation is reported in Vanderwende~(1994).  The 
algorithm tested is an extended version of that in Vanderwende~(1993), 
having 13 possible semantics relations and 34 heuristics.  Some search is 
involved in applying the heuristics, since the extended algorithm 
incorporates a general inheritance mechanism via hypernymic and 
meronymic relations extracted from the dictionary.  The extended 
development set contains 100 compounds, of which 79 are correctly 
analysed.  

Of a test set of 97 compounds taken from the Brown corpus, 51 were 
analysed correctly, giving a 52\% accuracy rate.  This approaches 
McDonald's theoretical accuracy of 60\%, but requires much less 
development effort.  Vanderwende's~(1994) accuracy figure is the only 
empirical performance result for this task to date, Leonard's~(1984) figure of 
76\% being for the development set.  In section~\ref{sec:ce_results} I will 
report the empirical accuracy of a statistical method for semantic analysis of 
two word compounds.

\subsection{Statistical Approaches} 
\label{sec:cn_statistical}

Given the importance of compound nouns and the recent interest in 
statistical methods it is not surprising that several proposals 
for applying statistical methods to noun compounds have been put 
forward in the last couple of years.  Statistics have been employed both in 
the assignment of accent to compounds for text-to-speech and in the parsing 
of compounds.  Work on the former task has already been described in 
section~\ref{sec:cn_computational}.  In this section I will review the work 
on statistical noun compound parsing, both because it has received more 
attention and because it is of more relevance to this thesis.

All of the algorithms I will review in this section are variants of one 
proposed by Marcus~(1980:253).  Therein, the procedure is stated in terms 
of calls to an oracle which can determine if a noun compound is acceptable.  
It is reproduced here for reference:

\begin{citedquote}{Marcus,~1980:253}
Given three nouns $n_1$, $n_2$ and $n_3$:
\begin{itemize}
\item If either [$n_1$ $n_2$] or [$n_2$ $n_3$] is not
semantically acceptable then build the alternative structure;
\item otherwise, if [$n_2$ $n_3$] is semantically
preferable to [$n_1$ $n_2$] then build [$n_2$ $n_3$];
\item otherwise, build [$n_1$ $n_2$].
\end{itemize}
\end{citedquote}

Only more recently has it been suggested that corpus statistics might provide 
the oracle, and this idea is the basis of the algorithms described below.  
Since the algorithm evaluates the acceptability of only adjacent pairs of 
nouns, I will call any analysis procedure which follows this general outline 
an \newterm{adjacency algorithm}.  Note that this strategy is designed to 
ensure that the deepest constituent is as acceptable as possible.

The simplest of these algorithms
is reported in Pustejovsky~\etal~(1993).  Given a three 
word compound, a search is conducted elsewhere in the corpus for each of 
the two possible subcomponents.  Whichever is found is then chosen as the 
more closely bracketed pair.  For example, when \lingform{backup compiler 
disk} is encountered, the analysis will be:
\begin{examples}
   \item	\label{eg:cn_statistical_adj}
     \begin{subexamples}
        \item{[}backup [compiler disk \nn] \nn{]}   
	when \lingform{compiler disk} appears elsewhere
        \item{[}[backup compiler \nn] disk \nn{]} 
	when \lingform{backup compiler} appears elsewhere
     \end{subexamples}
\end{examples}
Since this is proposed merely as a rough heuristic, it is not stated what the 
outcome is to be if neither or both subcomponents appear, nor is there any 
evaluation of the algorithm.

Bourigault~(1993) proposes the same algorithm for parsing all noun 
phrases, not just noun compounds, in French.  In this case, if neither or both 
competing substructures appear, the algorithm refuses to answer.  While the 
accuracy for noun compounds is not reported, the overall correctness is 
70\%, with no answer being given in a further 27\% of cases.

The proposal of Liberman and Sproat~(1992) is more sophisticated and 
allows for the frequency of the words in the compound.  Their proposal 
involves comparing the mutual information between the two pairs of 
adjacent words and bracketing together whichever pair exhibits the highest.  
There is no evaluation of the method other than a demonstration that four 
examples work correctly.  

The most complex proposal to be made appears in Resnik~(1993:126),
and once again is based on the adjacency algorithm. 
The \newterm{selectional association} between a 
predicate and a word is defined based on the contribution of the word to the 
conditional entropy of the predicate.  The association between each pair of 
words in the compound is then computed by taking the maximum selectional 
association from all possible ways of regarding the pair as predicate and 
argument.  Whilst this association metric is complicated, the decision 
procedure still follows the outline devised by Marcus~(1980) above.  

Resnik~(1993:128) used unambiguous noun compounds from the parsed 
\publicationname{Wall Street Journal} corpus to estimate the association 
values and analysed a test set of around 160 compounds.  The accuracy of 
the basic algorithm was 66\%, but by adjusting a threshold on the association 
value and selecting a default (bracket the first two nouns together)
when the value fell below this, the accuracy was raised to 73\%.  
This outperforms the baseline strategy of 
always choosing the default which achieves 64\% accuracy.

All four existing statistical methods for parsing noun compounds use
adjacency algorithms.  Even being generous, the best performance
achieved is 73\% on three word compounds.
In the experimental work on parsing noun compounds presented in the first 
half of chapter~\ref{ch:experimental}, I will propose a model that does
not follow the adjacency algorithm and compare this empirically
to an equivalent adjacency algorithm.
The new algorithm will emphasise dependency relations rather
than constituents.

It is noteworthy that Bod's~(1993) data oriented parsing scheme (see 
section~\ref{sec:sn_grammars}), if extended to include lexical items in 
subtrees, would enact a variant of the adjacency algorithm when presented 
with three word noun compounds that have not appeared in the training 
corpus.  If $n_1 \, n_2 \, n_3$ did not appear in the training corpus, 
and [$n_1 \, n_2$] or [$n_2 \, n_3$] did, 
then the algorithm would choose to bracket 
together whichever pair appeared more frequently.
Once again, the statistical parsing model is based on constituency,
just as the adjacency algorithm dictates.

Finally, I should mention one more recent study aimed at parsing
noun compounds by Abe~\etal~(1995).  This work does not give 
a specific algorithm for analysing noun compounds; instead, 
their program seeks to learn the acceptability of noun pairs as compounds,
leaving the use of this information up to the parser.  
Given a set of two word compounds, it learns a binary relation.  
That is, for each pair of nouns it learns whether or not that pair is a 
possible noun compound (this is in contrast to the measures given by 
Resnik,~1993, and Liberman and Sproat,~1992, which yield a graded 
score).  It is not possible to conclude how useful their results are in parsing 
noun compounds.

\subsection{Brief Review}

The second half of this chapter has reviewed the existing work on noun
compounds.  We have seen that they are both common and ambiguous, and
thus pose an important challenge to \nlp.  Many compounds are
productive, requiring a dynamic mechanism to process, and doing so
involves substantial knowledge, especially semantic expectations.
These characteristics make noun compounds an ideal medium through
which to explore statistical language learning designs.

The existing computational work on syntactic and semantic analysis 
of noun compounds has been either knowledge-based, 
and thus limited by the availability
of suitable hand-coded knowledge, or statistical.  In the latter case
only parsing has been attempted and then with limited success.
All existing statistical parsers have been based on one scheme,
the adjacency algorithm.

In chapter~\ref{ch:experimental}, I will report a series of experiments
on both parsing and semantic analysis of compound nouns.  The parsing
model differs from the adjacency algorithm in being based on
dependency relations and is motivated by the theory of statistical
\nlp\ to be proposed in chapter~\ref{ch:md}.  The semantic
analysis model is also motivated by this theory and represents
the first application of statistical language learning techniques
to this problem.

\cleardoublepage

\chapter{Meaning Distributions}
\label{ch:md}




\section{Introduction}

In this chapter I will give a theory of 
automatic natural language interpretation.  The goal of the theory is to 
provide an architecture for interpretation tasks within which several 
independent components sit.  Naturally the overall performance of 
systems designed according to this theory will depend heavily on these 
components.  Nonetheless, the theory does not specify these components 
in any more detail than is necessary to precisely describe the 
architecture.  
The components are
\begin{itemize}
\item a semantic representation,
\item a probabilistic model of this representation, and
\item a lexico-grammatical analyser.
\end{itemize}
While I will discuss certain requirements placed by the theory upon these 
components (in particular the functions they are to perform) the theory 
will not make any commitments to particular internal architectures for 
these components.  In applying the theory many additional decisions 
must be made regarding these components, an exercise which will be 
carried out in the experimental work in chapter~\ref{ch:experimental}.

In the discussion below I will consider the unit of interpretation to be the 
sentence (or I will use the term utterance).  However, the theory is 
intended to apply generally to many linguistic levels: clauses, phrases or 
even individual words.  If a suitable semantic representation of 
paragraphs were available, perhaps even units larger than sentences 
could be addressed.\footnote{Zadrozny and Jensen~(1991) give one 
semantic representation for paragraphs.}  The smaller the unit, the 
greater will be the influence of context on interpretation, but the basic 
architecture remains applicable.  
In chapter~\ref{ch:experimental}, the experimental work takes the
unit of interpretation to be the noun compound.

Finally, the rationale behind the theory is to suggest appropriate 
designs for real natural language processing systems.  I am not making 
the claim that systems built according to the theory {\em must\/} outperform 
alternative designs.  Rather the theory represents a synthesis of a wide 
range of ideas, including some derived from culturally determined 
viewpoints, and as such only serves to point out one promising way forward.  
The ultimate worth of such 
theories should only be judged empirically: by building various 
alternative systems and comparing their performance.  In 
chapter~\ref{ch:experimental} I will make a range of comparisons
between a compound noun parser based on the theory and other such parsers.

%
%
%

Keeping this qualification in mind, 
the inspiration for the theory derives from a conventional 
metaphor that views human communication as the transfer of pieces
of knowledge, as will be discussed 
in section~\ref{sec:md_meaning}.
According to this metaphor,
knowledge of meanings (semantic expectations) must play 
a central role in guiding interpretation.  

However the idea that such knowledge might be
responsible for control of the interpretation process has been
pursued at great length in the past and is commonly deemed
to have failed.  
Section~\ref{sec:md_revising} discusses
why the present theory differs from this earlier work and provides
a rough sketch of the theory.

In section~\ref{sec:md_priors}, a probabilistic representation
of semantic expectations will be proposed.  The section will also
discuss the functions required of such a representation by the
theory and argue that a representation of this kind is both
currently lacking and ultimately necessary for statistical \nlp.

Semantic expectations are derived from a wide variety of
sources.  While the theory treats all sources of expectations
uniformly, an implementation must choose which sources it will
model.  Section~\ref{sec:md_context} considers several possible
types of contextual information that provide semantic
expectations, ranging from local discourse effects to world
knowledge.

The core of the theory will be presented 
in section~\ref{sec:md_linking}.  The theory assumes that 
a lexico-syntactic module is available that supplies
a bidirectional many-to-many mapping between utterances
and semantic forms.  Thus, for any utterance, there is a set
of allowed meanings.  The theory claims that the most probable
allowed meaning is the best interpretation, regardless
of the particular syntactic configuration.  That is, 
syntactic forms inherit likelihoods directly from semantic
ones and do not have independent probabilities of their own.

Section~\ref{sec:md_linking} also discusses the possible
advantages of adopting this theory and some of the implications
of it.  Finally, section~\ref{sec:md_register} covers
some important issues regarding the selection
of training corpus register, in light of the theory.


\section{Meaning As Pilot}
\label{sec:md_meaning}

Automatic natural language processing systems necessarily 
rely on a control strategy that is ultimately responsible 
for deciding which knowledge resources 
get brought to bear on the task.  
One often reads that traditional \nlp\ consists of a pipeline, beginning 
with morphological analysis, passing through syntactic analysis and 
thence semantic interpretation, and finally undergoing discourse and 
pragmatic processing.  This is an architecture that comes from taking 
linguistic levels literally.  
Still, there is no question 
that it has proven useful (it can be found in many implemented systems, 
even as far back as Green~\etal,~1961).  It is probably not coincidental 
that the first two modules of the pipeline are concerned with morphology 
and grammar, certainly better understood aspects of language than 
semantics and pragmatics.  The placing of syntax prior to semantics can 
be seen as an attempt to gain leverage from the work on grammar in 
linguistics.  

Simplifying greatly, traditional \nlp\ 
holds grammar to be the guide of the language understanding process 
and grammatical knowledge is given precedence over semantic 
knowledge.  
%
However, grammar rules alone are not sufficient to determine a unique
syntactic analysis of a text, 
at least not with our current conception of grammar.  
In fact, syntax rules substantially underconstrain syntactic structure.
The most obvious symptom of 
this is the staggering number of parses admitted by broad coverage 
grammars.  Briscoe and Carroll~(1993:42) report that dictionary 
definitions~30 words in length \shortquote{often have many hundreds or 
even thousands of parses} under the \progname{alvey natural language 
tools} grammar.  While this might be attributed to a failure of the 
particular grammar, the field still lacks a broad-coverage parsing system 
which can fully disambiguate syntactic structure.  If there was any 
uncertainty in the first place, this fact gives strong evidence that 
substantial ambiguity remains after the application of syntactic 
knowledge, and that a control strategy that pursues syntax to its limits 
prior to any semantic processing is, at best, inefficient.

A number of researchers (for example, Mellish,~1985, Hirst,~1987, 
Haddock,~1987) have attempted to address this problem by
interleaving syntactic and semantic processing, with promising
results.  Interleaved systems give control alternately to 
one component and then the other.
While this affords a greater degree of integration, syntax
is still considered to be an autonomous component.

An alternative control strategy is inspired by the conception that the 
purpose of language is the communication of meaning and that therefore 
the structure of language should primarily follow the contours of 
meaning.  This idea has a certain intuitive elegance and has been the 
foundation of a substantial body of research which can be lumped 
together under the title \newterm{semantically driven nlp}.  Under this 
approach, primacy is given to semantic knowledge.  It is knowledge of 
the meaning of the language that shapes the interpretation of it, with 
grammar playing a minor part, if any at all.

The early champions of this approach were Schank~(1975) and his students.  
The theory of \newterm{conceptual dependency}, and the ensuing 
processing systems based on this theory, 
aimed to represent meaning independently of 
language by composing a small set of semantic primitives.  By means of 
this representation, expectations about the meanings of words could be 
recorded.  These expectations played the role of pilot, guiding the 
interpreter in arriving at a representation of the meaning of the sentence.  
The control strategy was:
\begin{citedquote}{Schank,~1975:13}
[The parser] attempts to recognise each word, and if it can, puts it into its 
semantic model.  Then if the meaning is known, a good idea of what 
other meanings are going to be around it can be established because 
predictions can be made on the basis of the meaning of that word.
\end{citedquote}

Another early semantically driven \nlp\ theory was Wilks's~(1975) 
\newterm{preference semantics}.  His system relied on semantic templates 
defined in terms of~70 semantic primitives.  Analysis in this system was 
driven by matching these templates to maximise \newterm{semantic 
density}, another representation of expectation about meaning.  

Despite the large volume of work in semantically driven \nlp, the notion 
of a primarily semantic control strategy is no longer popular.  For a 
variety of reasons, which I will come to in a moment, the entire paradigm 
is now regarded as an error of judgement.  Yet, the reason for discussing 
these approaches here is the fact that the strategy I will propose shares 
several key elements with them.  I claim that some aspects of these 
approaches are worthwhile, or even necessary, and it is only the 
historical inclusion of other faulty elements under the same banner that 
has led to their failure (both perceived and real).

In particular, this work shares two important bases with the older 
semantically driven \nlp\ approaches.
\begin{itemize}
\item Meaning is represented by simple combinations 
of a set of primitive elements that are relatively language independent, 
symbolic and finite.
\item Primary control over the analysis process is driven by expectations 
about the possible meanings of the text being processed.
\end{itemize}
The first of these is hardly controversial in \nlp; the crux of the 
matter lies with what we consider to be simple.  Meaning 
representations formed from the combination of symbolic and language 
independent elements appear in almost every \nlp\ system.  However, it 
might be claimed that a certain amount of complexity in semantic 
representation is necessary before that representation can prove 
worthwhile (or indeed even language independent).  The strategy that I 
will propose requires the meaning representation to be simple for certain 
mathematical reasons which will become clear in succeeding sections.  
Thus, this work shares with earlier semantically driven \nlp\ approaches, 
a reliance on the simplicity of the meaning representation, a property not 
required by other modern approaches.

The second point requires somewhat more justification than the first.  
For this reason, I will now briefly review the motivations for
salvaging the idea that semantic expectations should play a central role.
These stem from culturally determined intuitions about human language
processing and therefore constitute no more than plausible suggestions.
In what follows, I do not intend to 
suggest that the proposed control strategy bears any resemblance to 
psychological reality.  The ultimate test of the proposal lies in the 
performance of \nlp\ systems that employ it.  

\subsection*{Communication as transfer}

Most work in \nlp\ is based on a metaphor 
in which communication is regarded as the transfer of knowledge.  
Beliefs, propositions, ideas and dispositions are regarded as 
objects to be copied from the writer to the reader.  The metaphor is both 
entrenched and extensive.  Most natural language understanding systems 
make use of some form of representation of these objects, whether they are 
algebraic (like logical forms), graphical (like semantic networks), or 
topological (like nodes of a neural network).  None of this is surprising; 
the metaphor is a well-accepted convention of our culture (Lakoff and 
Johnson,~1980:10, use it as their example of an entrenched 
conventional metaphor, calling it the \newterm{conduit} metaphor).

One of the exceptions is constituted by the efforts in 
statistical machine translation found in Brown~\etal~(1993) in which
any representation of meaning is avoided.  The 
omission of a first class representation of such objects is exactly what I 
wish to argue against, and for this reason, I will explore the metaphor a 
little more closely in this section.  Again, I do not intend to make any 
claims about how humans communicate.  Rather, I want to accept certain 
assumptions about language (assumptions which are almost universally 
adopted by \nlp\ systems anyway) and from them deduce some simple 
principles for machine emulation of the interpretation process.  

According to the metaphor, communication assumes two sides, each 
regarded as a container in the sense that they may hold items of 
knowledge.  These items of knowledge are representations.  They may be 
{\em about\/} states, objects, actions and so on, that appear in the world, 
or they may be {\em about\/} other items of knowledge, but they are 
obviously not the same as elements of the world.  So there are two 
distinct sets: senses (the items of knowledge) and referents (the elements 
of the world which knowledge is about).\footnote{The metaphor seems 
to assume that senses can exist independently of language; that is, 
intelligent beings can reason without necessarily being able to 
communicate.}

In the simplest, declarative case, the source (writer/speaker) possesses some 
item of knowledge that the recipient (reader/listener) does not, and the 
objective is to create a copy of this for the recipient.  Since items of 
knowledge are not tangible, a somewhat indirect procedure is used to 
perform the transfer.  This is the point where language becomes 
necessary.  Language provides a third set of elements, a 
system of signs, allowing the source to indicate the item of 
knowledge being communicated.  Just as senses are not identical with their 
referents, neither are signs identical with their senses. 
Rather, each linguistic element indicates some aspect 
of the intended sense and the combination of these indications constrains 
the interpretation; that is, the surface linguistic forms narrow down the 
semantic {\em space}.

So it is part of the metaphor that individual meanings (senses) 
can be thought of 
as distinct points within a space.  Now to make this feasible, we need to 
distinguish that which is {\em meant\/} from that which is merely 
logically {\em entailed} or pragmatically {\em connoted}.  
Each communication conveys a single meaning, but from this 
meaning the recipient may make further implications.  These 
implications may or may not have been intended by the source.  But, 
regardless of whether the implications were intended, the communication 
is only perceived to have one distinct meaning in the truth-functional 
sense; its meaning is just the propositional content.  Despite the fact that 
this perspective ignores a host of pragmatic effects, I will adopt this 
simplification throughout, under the assumption that it will make little 
difference to the performance of \nlp\  systems, though this remains an 
empirical question.

Now, returning to the metaphor, by composing a series of linguistic 
elements, the source overlays more and more constraints on the possible 
meanings, narrowing the space with each.  If communication is 
successful, the accumulation of constraints eventually describes fully the 
item of knowledge being transferred.  If the recipient successfully 
interprets each linguistic element, and composes the corresponding 
constraints correctly, then the resulting composition will indicate the 
intended meaning.  A copy of the item of knowledge has therefore been 
successfully made.

Naturally, the space of possible meanings is extremely large and 
complex.  Yet a few short syllables is usually sufficient to indicate an 
arbitrary point in it.  This, according to the metaphor, can only be 
explained by the effects of context.  The source and recipient already 
share a great deal of knowledge about the likely intended meanings.  
The context has already selected a relatively tiny portion of the space, 
allowing relatively few constraints to fully indicate the information desired.  
While the metaphor has little to say about how context might be 
represented, it attributes enormous power to it as a communicative force.  
Out of all the infinite variety of possible human meanings, somehow the 
context eliminates all but a few salient meanings.  It is then only 
necessary for the source to distinguish which of these few is the desired 
one.

The Gricean maxim of Quantity (Grice,~1975:45) states that utterances
should be concise.  If we accept this principle as well as 
the metaphor of communication as transfer, then it is a principle of 
language that all utterances should only distinguish between the 
relatively few {\em salient\/} meanings.  Any additional linguistic signs 
which serve to rule out meanings already disallowed by context are 
redundant, and contradict the principle of concision.  Therefore, the 
language produced by the source must distinguish between all and only 
those possible meanings allowed by the context.  Summarising, the language 
production task begins with the intended meaning {\em plus\/} a small set 
of contextually relevant semantic possibilities.  The goal is then to use 
appropriate linguistic knowledge to indicate the intended meaning, 
distinguishing it only from the other relevant possibilities.\footnote{This
view is explicitly adopted by Dale~(1989) as the basis of
his generation algorithm for referring expressions.}

Correspondingly, the language comprehension task requires the use of 
the same linguistic knowledge to infer which of these possibilities is the 
intended one.  From this standpoint, knowledge of 
the semantic possibilities is crucial to language understanding.  It is vital 
that the recipient be able to identify the small set of contextually relevant 
semantic possibilities that the source is distinguishing amongst.  
Therefore, adopting the conventional metaphor of natural language 
understanding, and assuming the Gricean maxim of concision, leads us to 
the conclusion that knowledge of semantics must play a central role in 
comprehension.  It should be the resource that is applied first, with 
grammatical knowledge serving as the final step in refining meaning.  

Yet this was the principle claim of the early 
semantically driven \nlp\ work.  It 
is therefore natural to ask why, after so many years, it is still worth 
pursuing this idea.
In the next section, I will argue 
that the failure of these earlier works can be attributed to weaknesses of 
other ideas held concurrently by early semantically driven \nlp\ 
proponents and that therefore it is still worth taking up the semantically 
driven banner.

\section{Revising Semantically-driven NLP}
\label{sec:md_revising}


I claim that two design flaws were responsible for the failure of early 
semantically driven \nlp\ systems to scale up.  They are:
\begin{itemize}
\item dependence on a small set of semantic primitives; and
\item rejection of the need for syntactic knowledge.
\end{itemize}
In addition, I claim that two points of the philosophical stance taken by 
semantically driven \nlp\ researchers were responsible for warranted 
skepticism of their research.  They are:
\begin{itemize}
\item the claim that semantically driven systems are models of human 
language behaviour; and
\item the claim that the {\em referents\/} of utterances hold the key to 
guiding language understanding (as opposed to their senses).
\end{itemize}
To avoid criticism on the basis of these last two points, 
it is only necessary to avoid making these two claims.  
Regarding the first claim, I have already 
stated that the evaluation of the theory presented here is purely on the 
grounds of empirical performance, with no pretense to psychological 
validity.  As for the second, the theory below carefully distinguishes 
senses from referents, and relies on knowledge of the former for 
controlling analysis.  I will not discuss further why these claims warrant 
skepticism, focusing instead on the two design flaws.

The semantically driven parsers developed by Schank and his students 
were based on the conceptual dependency knowledge representation 
(Schank,~1975:Ch.~3).  At the heart of this representation were~11 
primitive event types which could be linked through case roles with symbols 
representing objects in the world, and with one another through causal 
and instrument links.  Similarly, Wilks's~(1975) semantic preferences 
parser was based on~70 semantic primitives.  It was assumed that all 
meanings could be captured by combinations of these primitives.  This is 
a highly questionable assumption.  

To start with, if there were such primitives, it seems reasonable
that our languages would have special words to indicate them, 
words which children would be likely to acquire very early.  
Concepts such as \concept{sign} (a communication 
event) and \concept{mbuild} (a mental construction) are certainly not 
like this.  Also, we would expect the construction of human language 
dictionaries to be much simpler than it is in reality.  Lexicographers 
would only need to define the words for the semantic primitives in a 
special section and then restrict themselves to using those words 
everywhere else.  In fact, the Longmans Dictionary of Contemporary 
English (Proctor,~1978) is constructed by attempting to use as few 
defining words as possible.  Even if only stems are counted, the defining 
vocabulary exceeds~2,000 words and many of these words have several 
distinct senses (each sense presumably corresponding to a separate 
semantic primitive).

So, even if the notion of semantic primitives is in principle workable, the 
versions of it adopted by the early semantically driven \nlp\ researchers 
were far too spartan.  They could not even approximate the richness of 
meaning expressed by natural language.  It is therefore not surprising that 
their parsing systems could not handle a broad-coverage input and came 
to be regarded as toy systems.

What I am arguing against here is the limitation to a {\em small\/} set of 
semantic primitives.  I do not believe that meaning cannot be represented 
by the combination of a finite set of semantic elements; in fact if there is 
to be any symbolic natural language understanding, then such 
representation must be possible.  Rather, it seems clear that the number 
of such elements needs to be large, of the order of thousands at least.  

It is even conceivable that the set of such primitives differs from person 
to person.  A method of representing this is provided by the notion of 
\newterm{episodic memory} developed in Lebowitz~(1983), where new 
concepts are built in response to new texts.  Lehnert~(1988) applies this 
idea to compound nouns, placing emphasis on the dependence of 
compound meaning on previously seen semantic structures.  The models 
of compound nouns presented later in this thesis also use previously seen 
semantic structures to infer interpretations for compound nouns.  While 
they are based on semantic primitives of a kind, the number of primitives 
is far larger than those used in early semantically driven \nlp; there are 
over a thousand of them for representing noun meanings alone.

The second design flaw of early semantically driven \nlp\ was the 
complete rejection of syntactic representation:  
\begin{citedquote}{Schank,~1975:12}
We have never been convinced of the need for grammars at all.
\end{citedquote}
\begin{citedquote}{Wilks,~1975:265}
Preference Semantics contains no conventional grammar for analysis or 
generation: its task is performed by strong semantics.
\end{citedquote}
As these quotes 
show, the enthusiasm for adopting semantic knowledge as the main 
resource for natural language comprehension led to the 
position that syntax could be dispensed with.  Yet Lytinen~(1987) 
showed that semantically driven systems contained syntactic 
information disguised as semantics.  At the same time as arguing that 
syntactic knowledge was unnecessary, the advocates of semantically
driven \nlp\ were forced to make many 
grammatical distinctions to get their parsers to work.  Because their
systems lacked a separate syntactic representation, 
syntactic distinctions became intertwined 
with the semantic representation.  Lytinen's solution was to have both a 
semantic and syntactic representation working in tandem, with neither 
having priority.  
 
Another well-known and vocal opponent of semantically driven \nlp\ was 
Ritchie~(1978).  He argued that 
\begin{citedquote}{Ritchie,~1978:273}
\dots notions of semantically-based grammar are not workable in the 
long term.  This is not an argument in favour of traditional syntax, but 
rather an attempt to prevent an inadequate system (autonomous syntax) 
being replaced with an equally unrealistic extreme (autocratic 
semantics).
\end{citedquote}
Again, the central criticism lay with the rejection of grammatical 
knowledge.\footnote{Ritchie~(1983) carried the argument further and also took 
issue with the two philosophical points mentioned above: cognitive 
plausibility and the use of {\em referential} semantics.}  

Since early semantically driven \nlp\ systems suffered from these 
flaws, it is reasonable to attribute their current 
poor status to just these sources.  It is not necessary to conclude that the 
notion of a semantic representation being the primary controlling influence is 
itself flawed; early systems could just as easily have failed for other 
reasons.  It is therefore reasonable to pursue further semantically driven 
systems, provided that
the other reasons for failure are avoided in some fashion.  
In the theory I will develop below, syntax plays a crucial role; 
this is a very different story from that 
adopted by early semantically driven \nlp\ researchers. 

\subsection*{A new semantically driven approach to NLP}

Following the ambitious claims of the early work, the~1980s saw a 
return to the syntactically driven paradigm for \nlp.  By this time, the 
need for semantic knowledge was widely recognised and many efforts 
were focussed on mechanisms for applying semantic knowledge (for 
example the interleaved approaches mentioned above, like Hirst,~1987). 
But the notion of a purely 
semantically driven natural language system had fallen into disrepute.

More recently there has been a resurgence of research that places 
emphasis on semantic expectations in just the same way as early work.  
Hahn~(1989) puts forward detailed arguments in support of semantically 
driven processing for text summarisation, even going so far as to 
dispense once more with syntax.
\begin{citedquote}{Hahn,~1989:357; his emphasis}
Parsing natural language is achieved [by] mapping natural language 
utterances {\em directly into semantic representation structures\/} without 
considering a logically separated, intermediate level of syntactic 
representations.
\end{citedquote}
However, his main points centre around the efficiency of only analysing 
those parts of a text that are necessary for the understanding task at hand.  
This doesn't preclude a separate syntactic representation; it only requires 
that primary control of the analysis is driven by semantics.  

Mauldin~(1991) uses a semantically driven parser based on an extended 
form of conceptual dependency to \shortquote{push [information] 
retrieval performance beyond the keyword barrier}~(Mauldin,~1991:6).  
The resulting system shows promise, but is limited by the availability of 
scripts and frames, which must be hand-coded.  Perhaps then these 
knowledge resources could be acquired by statistical methods.

Hobbs~\etal~(1993) propose to integrate syntactic and semantic 
processing by reformulating comprehension in terms of abduction, 
claiming that almost every remaining problem in natural language 
interpretation can be solved by use of weighted abductive inference
(including noun compound interpretation, see section~\ref{sec:cn_knowledge}).  
The abductions proposed are essentially the application of knowledge of 
semantic expectations, although the issue of how these might be acquired 
in the first place is not addressed.  

Even Wilks's preference semantics parser has been reincarnated in a 
new-look broad-coverage form called \progname{premo} (Slator and 
Wilks,~1991), which interleaves semantic expectations with syntactic 
knowledge.  

In all this work, the controlling influence of semantic expectations is 
recognised, but no longer is there a reliance on a small set of semantic 
primitives, nor is grammatical knowledge denied its proper place 
(excepting Hahn,~1989).  
In the remainder of 
this section, I will roughly sketch a revised version of semantically 
driven \nlp.  The details of this new version will be the topic of the rest 
of this chapter. 

The approach I am proposing I will call \newterm{meaning 
distributions}.  According to this approach, the process of language 
interpretation begins with knowledge of the possible intended meanings 
in the form of semantic expectations.  We have at our disposal complete 
information about which meanings are more or less expected across the 
entire meaning space.  
We know, for example, that a sentence describing the diet
of animals is to be expected in an encyclopedia and that one
describing the diet of shoes isn't.
Given two arbitrary propositions, we can choose which of the two is 
more {\em expected}.

It isn't difficult to see that this knowledge is highly context dependent.  
Intended meanings will vary from situation to situation, and so too 
should the semantic expectations.  For example, we expect statements to 
be relevant to the topic at hand, whatever it might be, so as topics shift so 
must expectations.  As it stands the meaning distributions approach 
admits that such shifts exist and that they need to be modelled, but does 
not suggest how this might be achieved.  However, I will have more to 
say about these issues when I consider context dependence in 
section~\ref{sec:md_context}.

At this point, I have already assumed a great deal: a complete ordering of 
all possible meanings on the basis of how expected they are.  In fact, 
without having examined the text we wish to interpret, we are already 
quite some way towards an interpretation.  The semantic expectations 
allow us to delimit the likely interpretations to a small set of contextually 
relevant meanings.  
%
It is here that lexical and syntactic knowledge become important.  The 
meaning distributions approach 
assumes that sufficient lexical and syntactic machinery is in 
place to allow input text to be mapped onto its possible meanings.  That 
is, there is a mapping available between, on the one hand, grammatical 
constructions of words and, on the other, the space of meanings.  It does 
not assume that this mapping is one-to-one; otherwise we would already 
have solved the interpretation task.  But it does assume that it is, in 
principle, fully computable and, for mathematical reasons which will be 
made clear below, that the mapping is fairly simple.

Thus, given a linguistic form, it should be possible to apply
lexical-semantic correspondences and grammatico-semantic relations to yield a 
subset of the space of possible meanings: the set of meanings that could 
ever lead to utterance of the linguistic form.  In practice it will be 
important to avoid explicitly constructing this subset, but conceptually a 
particular subset exists and can be computed in principle.

The central claim of the meaning distributions approach is this: 
whichever meaning within the subset of meanings allowed by the 
lexico-grammatical mapping is the most expected, is the intended meaning.  
Thus, the task of interpreting an utterance becomes one of finding the 
single most expected meaning out of those allowed by the 
lexico-grammatical knowledge.
\begin{examples}
    \item Moles feed on insects.  \label{eg:md_revising_moles}
\end{examples}
To illustrate, consider the sentence in example~\ref{eg:md_revising_moles}.
Suppose that we have a very simple lexical-semantic mapping as follows.
\begin{itemize}
\item \lingform{moles} corresponds to two concepts: 
\concept{mammal\_mole} and \concept{chemical\_mole}.
\item \lingform{insects} is unambiguously the concept \concept{insect}.
\item \lingform{feed} can be mapped to the predicate \predicate{eat/2} 
or the concept \concept{food}.  
\end{itemize}
Any serious grammar would find several parses (or partial parses at any 
rate) of the sentence, but for simplicity assume that only one syntactic 
reading is allowed, so that only two possible meanings are permitted by 
lexical and syntactic knowledge.  These are: 
\begin{itemize}
\item \predicate{eat}(\concept{mammal\_mole}, \concept{insect}), and 
\item \predicate{eat}(\concept{chemical\_mole}, \concept{insect}).
\end{itemize} 
Normally, in a large scale system, the set of meanings allowed by the 
lexico-grammatical mapping would be much larger than in this 
illustrative example.  

Now semantic expectations will serve to select the first of these 
meanings for two reasons.  First, in any reasonable occurrence of this 
sentence, the topic of discussion will be animals and probably moles, so 
that there is a high semantic expectation for the appearance of the 
concept \concept{mammal\_mole}.  Second, even ignoring context, 
there is a general expectation that the agent of the predicate 
\predicate{eat/2} will be an animal or person, making this reading preferable 
to the one in which the agent is a chemical unit.

So, in summary, the meaning distributions approach starts with semantic 
expectations which place a preference ordering on the space of possible 
meanings.  A mapping provided by the lexicon and grammar then
creates a set of constraints on the intended meaning derived
from the utterance.  Finally, selecting the 
most highly expected meaning within these constraints yields the desired 
interpretation.  In practice, the existence of the expectations is intended 
to make computation of the mapping efficient, avoiding the need to 
construct unnecessary pieces of the map.

In this sketch, I have referred often to the notion of a semantic 
expectation.  I have not given any notion of what form these expectations 
might take, save that they provide an ordering on the space of 
meanings.  In the next section I will take up this task by casting semantic 
expectations in the form of statistical distributions.  

\section{Meaning Expectations as Prior 
Probabilities}   \label{sec:md_priors}  

\subsection*{Probabilistic expectations}

The representation for semantic expectations in the meaning distributions 
approach is a probabilistic distribution.  Formally, we say that there is a 
random variable $M$ ranging over the space of possible meanings, $\cal 
M$, whose distribution models the probability of all possible intended 
meanings of all possible utterances.  It is important to emphasise that the 
distribution is context-sensitive.  Given a certain context, a distribution is 
computed that gives the likelihood of the source having each possible 
meaning in mind.    

To take an example, consider the context in which a speaker 
has just been asked the question \scare{How are you?}   
The utterance made by the speaker in response to this question 
can be intended to mean a variety of things, and each of these possible 
meanings has some probability.  For example, it is highly likely that the 
response will convey the fact that the speaker is quite normal healthwise.  
Responses conveying great happiness or depression are also likely, while less 
likely are those explaining complex problems and it is only remotely 
possible that the response will be irrelevant to the question.  In principle, 
the distribution of $M$, while potentially very complex, always exists 
whenever the speaker is about to make an utterance.

It is vital to distinguish here between the probability distribution of {\em 
utterances\/} and that of {\em intended meanings}.  There might be any 
number of different utterances which could be used to convey a certain 
meaning, or even several possible interpretations of one utterance.  
Meaning distributions are distributions over the space of {\em 
meanings}, $\cal M$; they are {\em not\/} distributions over possible 
utterances.  In the example above, the proposition that the speaker is quite 
normal healthwise could be conveyed by any of: 
\begin{examples}
\item 
    \begin{subexamples}
	\item I'm fine.
	\item Well, thank you.
	\item Not bad.
	\item Pretty good.
    \end{subexamples}
\end{examples}
or any of many other variations.

While the probability of each of these utterances is perfectly well-defined, 
such probabilities are not {\em semantic\/} expectations. Meaning 
distributions represent semantic expectations and measure the likelihoods 
of different intended meanings, not different utterances.  Every one of 
the above responses has essentially the same meaning.

It is also vital to distinguish between the probability distribution of 
intended {\em senses\/} and that of intended {\em referents}.  
The world contains many objects and, were it possible to randomly 
sample the world, different kinds of objects would appear with 
different probabilities.  However, this distribution is not the same as the 
distribution of intended meanings.  For example, although unique gifts 
are extremely rare, department store catalogs refer to them frequently.  
Similarly, the distribution of 
commands issued to a database management system is different from the 
distribution of information stored in it.  A newspaper report of a death is 
most likely to refer to a murder, even though far more people die from 
illnesses.  Meaning 
distributions {\em depend\/} on referent distributions, but the latter are 
only one factor influencing the former.

The idea that semantic expectations can be represented by probability 
distributions is closely related to information theory (developed by 
Shannon and Weaver,~1949).  According to information theory, the 
amount of information contained in a message is measured by the change 
in entropy between the prior and posterior probability 
distributions.\footnote{The entropy of a discrete probability distribution
over $X$ is given by $\sum_{x \in X} \Pr(x) \log \Pr(x)$.}  In 
the case of successful communication of a single proposition (one point 
in the meaning space), the posterior probability distribution has 
zero entropy; the recipient is certain of the intended 
meaning.  Therefore, the amount of information conveyed by the 
message is simply the negative of the entropy of the prior probability 
distribution.

Information theory says that 
if we wish to minimise the length of the message needed to convey
this information then we 
must know the probability distribution of possible communications.  If 
the recipient has strong semantic expectations, the entropy of the prior 
distribution will be close to zero, and therefore the size of the message 
can be small and the utterance short.  If there are only weak semantic 
expectations, the utterance will need to be longer.  
This is in keeping with the position taken above on language 
comprehension: to adhere to the Gricean maxim of concision requires 
both the source and recipient to have detailed knowledge of the prior 
probability distribution, that is, semantic expectations.

%
So, to recap, the approach advocated here represents semantic 
expectations as probability distributions over the space of possible 
meanings, $\cal M$, which we call meaning distributions.  These 
distributions are necessarily highly context sensitive, since semantic 
expectations must vary with context.  In what follows I will suggest that 
these distributions, in combination with lexical and grammatical 
constraints, may be used to build natural language understanding 
systems.  So far, though, I haven't said anything about how these 
meaning distributions might be constructed.  Certainly the meaning 
space $\cal M$ is extremely complex, and the computation of the 
probability distribution of the random variable $M$ for even the most 
simple of contexts might seem like an even more complicated problem 
than the natural language understanding task itself.  Therefore I will now 
turn to the topic of defining models for such distributions.

\subsection*{Conceptual models}

Meanings have structure, a fact I have ignored up to this point.
Furthermore, we don't generally have an explicit construction 
of the set of all possible meanings.  
Any \nlp\ system that did would have to be extremely simple.  
Since the meaning distributions approach involves assuming access to a 
probability distribution over this set, it is necessary to justify how this 
might be possible, even in principle.  The question of how it might be 
possible in practice I will delay until section~\ref{sec:md_linking}.

The problem of representing meaning has been the subject of decades of 
research and even a brief overview is beyond the scope of this thesis.  
Instead, I will assume the properties listed below of the underlying 
knowledge representation.  These properties do not impose any 
constraints of importance, being exhibited by most popular knowledge 
representation schemes, and by first-order predicate calculus.  It is true 
that the particular choice of knowledge representation will have a 
significant effect on the resulting probabilistic model (but then choosing 
the correct knowledge representation has long been recognised as crucial 
throughout artificial intelligence).  My aim here is to justify how, in 
principle, a meaning distribution could be intensionally constructed.

The knowledge representation scheme must:
\begin{enumerate}
\item be symbolic;
\item generate a recursively enumerable set of expressions;
\item build complex expressions from simpler ones by well-defined 
methods of composition; and
\item have canonical forms.
\end{enumerate}

These are precisely the properties possessed by a context free grammar 
which allow the construction of probabilistic context free grammars.
As we have 
seen, such grammars assign a probability to every possible derivation 
generated by the grammar by assuming that the probability of a 
derivation is the product of the probabilities of the steps in the 
derivation.  That is, each rule application is assumed conditionally 
independent of other rule applications.  This assumption is simply 
untrue: for example, noun phrases in subject position are likely to have 
very different structures to those in object position.  It is simply 
impossible for a probabilistic context free grammar to assign 
probabilities to parse trees that are accurate reflections of the probability 
of observing those parse trees.  But as long as the probability 
assignments are sufficiently accurate to distinguish highly unlikely 
structures from plausible ones, such grammars are useful.

By the same reasoning, given a knowledge representation scheme with 
the properties above, it is possible to construct a probabilistic conceptual 
model which assigns a computable probability to each element of $\cal 
M$.  Just as in the context free grammar case, to do this it is necessary to 
give a method of calculating the probability of a complex expression in 
terms of the probability of the simpler expressions that compose it.  This 
will involve making assumptions about the relationships between 
distributions of different meaning elements, assumptions which may in 
reality be false.  Nonetheless, if the structure of the knowledge 
representation scheme broadly reflects the relationships exhibited by 
meanings, then the probabilities assigned by the conceptual model 
should be useful in as much as they distinguish highly unlikely meanings 
from plausible ones.

In the experimental work in chapter~\ref{ch:experimental},
a complete example of a 
probabilistic conceptual model will be given for the semantics of 
compound nouns.  The form of such a model is heavily dependent on the 
particular knowledge representation scheme on which it is based.  Since 
it is difficult to explore the construction of conceptual probabilistic 
models further without making a commitment to a particular knowledge 
representation scheme, I will only give a rough example here.  The 
example uses predicate calculus, but it should be clear that a similar 
model could be given for any scheme meeting the conditions above.

We wish to assign a probability to each possible expression in the 
meaning representation.  For this example, we will suppose that 
meanings are generated top-down.  That is, a predicate is selected and 
then the arguments of the predicate are chosen, conditionally dependent 
on the predicate.  Since arguments may themselves be predicates, the 
process may recurse.  While I have used formal notation in what follows, 
I must emphasise that this is only a rough sketch.  There are many 
problems to be addressed when building a probabilistic 
conceptual model (consider quantification for instance) and it is not my 
intent to build a rigorous model here.

Let $P$ be the set of predicates, $A$ the set of atoms and 
$\mbox{arity}(p)$ be the number of arguments of a predicate $p \in P$.  
Call the disjoint union of predicates and atoms, $U$.  The model then 
has the following sets of parameters:
\begin{enumerate}
\item one parameter for each element of $U$ giving its probability 
in isolation, $\Pr(x \mid \bot)$; and
\item one parameter for each triple $\langle p, i, x \rangle$ 
where $p \in P$, $1 \le i \le \mbox{arity}(p)$ and $x \in U$, 
giving the probability that $x$ is the $i$th argument of $p$, 
$\Pr(p(\ldots, X_i=x, \ldots) \mid p)$.
\end{enumerate} 
For this to be a probabilistic model, we also require the following 
constraints:
\begin{equation}
\sum_{x \in U} \Pr(x \mid \bot) = 1
\end{equation}
\begin{equation}
(\forall p \in P) (\forall i:1\le i\le \mbox{arity}(p)) \sum_{x \in U} 
\Pr(p(\ldots, X_i=x, \ldots) \mid p) = 1
\end{equation}

Now by assuming conditional independence between arguments of a 
predicate and making several similar assumptions, we can arrive at a 
formula for computing the probability of an arbitrary expression.
\begin{equation}
\Pr(\lambda) = \left\{ \begin{array}{ll} 
	\Pr(\lambda \mid \bot) & \mbox{if $\lambda$ is an atom} \\
	\Pr(\rho \mid \bot) \prod_{p(\ldots, X_i=x, \ldots) \in \lambda} 
	\Pr(p(\ldots, X_i=x, \ldots) \mid p) & \mbox{if 
$\lambda$ has top predicate, $\rho$} 
\end{array} \right.
\label{eq:pred_calc_model}
\end{equation}
In order to use the model in practice, it is also necessary to have a 
method for acquiring estimates of these parameters.  A general approach 
to this problem will be suggested below in section~\ref{sec:md_linking}.

To illustrate the model, consider the example involving 
\predicate{eat}(\concept{mammal\_mole}, \concept{insect}) above.  
It is simple to apply equation~\ref{eq:pred_calc_model}.
\begin{eqnarray*}
\lefteqn{ 
\Pr(\mbox{\predicate{eat}(\concept{mammal\_mole}, \concept{insect})}) = 
	} \\
	& & \Pr(\mbox{\predicate{eat/2}} \mid \bot) \times \\
	& & \Pr(\mbox{\predicate{eat}(\concept{mammal\_mole}, $\ldots$)} 
	\mid \mbox{\predicate{eat/2}}) \times \\
	& & \Pr(\mbox{\predicate{eat}($\ldots$, \concept{insect})} 
	\mid \mbox{\predicate{eat/2}})
\end{eqnarray*}
Intuitively, we would expect all three probabilities on the right-hand side 
to be relatively high.  Moles are likely consumers, insects are good 
eating for many animals and, in the context of encyclopedia articles 
(from which this example is drawn), we expect propositions describing 
diets.  Therefore, the model correctly assigns a high probability to this 
meaning.  

This example was intended to show that in principle it is possible to 
construct probabilistic versions of any knowledge representation scheme 
which has the properties listed above.  In fact, some work already exists
on building probabilistic conceptual models for story understanding.
For example, Charniak and Goldman~(1989) 
describe a probabilistic quantifier-free 
first order semantic representation that places logical formulae at the 
nodes of a Bayesian network.\footnote{Charniak and Goldman's model 
is not a pure conceptual model in the sense required by the meaning 
distributions theory since it also includes lexical and syntactic nodes.  It 
therefore merges linguistic and world knowledge into one representation, 
a move explicitly rejected by the meaning distributions theory.  Also, 
while they are careful to point out the distinction between senses and 
referents, their probabilistic conceptual model makes the simplifying 
assumption that these two distributions are the same.  The most important
differences between Charniak and Goldman's work and this thesis are that
probabilities in this work are estimated from large corpora (rather than being
chosen by hand) and the resulting performance is evaluated on a statistically
significant test set.}  Algorithms 
operating over these networks furnish the probabilities assigned to each 
formula.  Goldman and Charniak~(1990) describe a method to 
dynamically construct such a probabilistic conceptual model so as to 
avoid the impracticality of constructing the entire meaning distribution.
So not only is the notion of a probabilistic conceptual model possible in 
principle, but it has already appeared in the context of at least one natural 
language task and some work has been done toward making such models 
efficiently computable.

\subsection*{Theoretical advantages}

In the remainder of this section I would like to consider what makes the use of 
meaning distributions and conceptual models different from similar work 
in statistical language learning and why we might expect the meaning 
distributions approach to be better.   Though I take the position that 
different approaches to statistical language learning must be evaluated on 
experimental evidence, it is worthwhile justifying the theory in 
theoretical terms because of the insights this yields into the experimental 
data.

Statistical language learning has its roots in speech recognition where it 
turns out that dependencies between elements of the speech stream are 
virtually all local.  Under these conditions, simple Markov models work 
well.  There is no need to construct more complicated representations 
than sequences.  Similar conditions held in the earliest successes of 
statistical language learning, part of speech tagging (De~Rose,~1988), 
where sequences of tags formed the primary representation.  Most later 
advancements of this work retained the sequence representation 
(Weischedel~\etal,~1993; Merialdo,~1994) and sequences of words were 
the foundation of many other systems, including the distituents of 
Marcus~(1991) and the collocations of Smadja~(1993).

Despite successes with sequence-based models, the use of more 
structured representations is 
inevitably necessary for language understanding, a fact that is reflected 
in a wide range of research.  Many probabilistic models based on 
structured representations have been constructed and used as the basis of 
experimental work.  Probabilistic context free grammars stand as the 
obvious example (Jelinek~\etal,~1992).  Others include the 
subject-verb-object statistics collected by 
Church and Hanks~(1989), the sense 
disambiguation method of Hearst~(1991), the prepositional phrase 
attachment information gathered by Hindle and Rooth~(1993), and 
Grefenstette's~(1994) thesaurus construction statistics.  However, all of 
these use only grammatical relationships to structure their statistical 
models.\footnote{In information retrieval there is also work on 
representing grammatical relations using Bayesian networks (Bruza and 
Ijdens,~1994).}

The linguistic events captured by all these models are relatively 
superficial ones in the language process as a whole: syntactic 
configurations and word and tag sequences.  This may be likened to the 
earliest machine translation efforts of the~1950s.  Translations 
performed word-by-word with surface syntactic rewriting were quickly 
found to be too primitive to model the complex structures 
of language because they were driven only by superficial elements.  
In the same way, the use of superficial elements as the basis of statistical 
language learning is a restriction that deserves some consideration.
%
It has been easier to use corpora to compute statistics about word 
sequences and syntactic configurations than about anything deeper, but 
if statistical \nlp\ is to go beyond its beginnings, the models must be 
extended to incorporate deeper notions.

Importantly, the inaccessibility of deeper linguistic elements does not 
preclude them from being included in statistical models.  The states 
in a hidden Markov model need not be available in 
the corpus.  So too can semantic elements, which are not directly 
available via annotations of the corpus, be included in more sophisticated 
statistical models of language interpretation.  

So I have argued that, in principle, since language has deeper structural 
elements than those that current statistical models can capture, current 
models should be extended.  But to give this argument any weight I also 
need to show that these deeper structural elements play a significant role 
in language understanding.  Let us therefore consider two such elements: 
sense distinctions and predicate-argument relations.

Polysemy is probably the most obvious challenge posed by natural 
language.  It emerges in some form in all natural language applications, 
even simple spelling checkers.  Individual words express multiple senses, 
each differing in their linguistic behaviour.  This variation in behaviour 
is not limited only to the meaning of the word.  Syntactic, morphological 
and phonological properties of a word can vary with its sense.  In 
practice, no matter what \nlp\ task is attempted, word senses are going to 
make a difference.  

Statistical models that do not individuate word senses cannot 
(directly) represent these differences.
In such models, information about a word is a 
heterogeneous conglomerate of information about each of its senses, 
typically combined by arithmetic addition.  
The resulting average gives only an 
approximate picture of the behaviour of any one sense of the word.  
Since word senses usually vary in their frequency, this picture will 
typically be dominated by the behaviour of the most frequent sense.  
No matter how 
much data is gathered and analysed, the model will never have accurate 
information about any of the less frequent senses.  This is rather like 
trying to build an \nlp\ system without ever putting more than one sense 
of a word into the lexicon.  One would imagine that such a system would 
reach a certain limit and then become unworkable.  A similar fate is in 
store for statistical language models which do not incorporate a notion of 
word senses.

Similar arguments hold for predicate-argument structure.  Selectional 
restrictions (placed by a predicate on its arguments) have long been 
recognised as crucial to many \nlp\ tasks (see for example, Wilks,~1975, 
Hirst,~1987, and Fass,~1991).  These relationships will be reflected in 
even simple statistical data concerning word 
associations.\footnote{Although note that probabilistic context free 
grammars specifically eliminate this information from the model and thus 
cannot capture selectional associations.}  But if the model does not 
explicitly include predicate-argument relations, this information must be 
distributed.  A given underlying predicate-argument 
structure can be expressed on the surface in several ways.  If the model is 
based only on superficial textual elements, each of these surface 
expressions will be treated distinctly.  The same selectional restriction 
will be represented by several different pieces of information.  
This multiplicity is detrimental to the success of
statistical language learners since it requires the system to learn
more than is necessary.

So, on theoretical grounds, the step to explicitly semantic statistical 
models is a very promising one (the view adopted in Lauer,~1993).  
Very recently, a few others have expressed the same view.  
Miller~\etal~(1994) describe a conceptual model based on frames for 
the limited domain of air travel queries.  Unfortunately, their system 
requires training data to be annotated with semantic representations 
which limits the applicability of the technique outside narrow domains.  
Alshawi and Carter~(1994) use the same air travel corpus to experiment 
with statistical metrics over a shallow semantic representation called 
quasi logical form (\acronym{qlf}), but again training is supervised.  

Alshawi~(1994) gives a mathematical treatment of the incorporation of 
semantics into speech translation models, although the conceptual model 
is based only on lexical relations and does not explicitly represent word 
senses.  Eizirik~\etal~(1993) propose a Bayesian network representation 
which incorporates both syntactic and semantic representations into one 
network.  However, parameters of the model arise from the {\em 
structure\/} of the network; there is no parameter estimation from training 
data, and furthermore, detailed hand-coding is required to build the 
semantic portion of the network.  Finally, a proposal for incorporating 
semantics into Data-Oriented Parsing (see section~\ref{sec:sn_grammars}) is 
also made by van~der~Berg~\etal~(1994), but it too assumes supervised
learning.

Apart from the discussion of context that I promised earlier, the 
remainder of 
this chapter is devoted to fleshing 
out the details of the meaning distributions approach.  

\section{A View of Context Dependence}   
\label{sec:md_context}  


The meaning distributions approach places strong emphasis on 
expectations, that is, on knowledge about possible meanings of a 
sentence that is available before considering the sentence itself.  I have 
avoided using the term {\em context} until now because 
I intend to interpret the notion of context very broadly.  
I define context to be the complete set of {\em non-linguistic} 
facts that might be relevant to interpreting an arbitrary 
sentence in a given circumstance.\footnote{There is some looseness
in this definition derived from what is taken to be linguistic.
For example, in systemic functional grammar, much
of what I call context is taken to be linguistic (Halliday,~1985:xiii-xiv).}
This includes information conveyed by the cotext 
(the text or dialogue accompanying the sentence), 
information about the goals and intentions of the writer or 
speaker, encyclopedic world knowledge and more besides.  
The meaning distributions theory holds that context, in the 
broad sense, deserves a more central place in the language understanding 
architecture.

%
The importance of context to language interpretation raises the 
question of how the majority of \nlp\ systems thus far built have escaped 
the need to represent context explicitly.  Many treat sentences in 
isolation, ignoring any cotext that might have surrounded the sentence 
and therefore discarding discourse and topic cues.  It is therefore 
unsurprising that ambiguities abound.\footnote{This is not to say that a 
practicable theory of the effects of discourse and topic is available, 
merely that, lacking one, ambiguities are to be expected.  There is a large 
research effort toward such a theory, including some statistical work 
(Reithinger and Maier,~1995).}  Other elements of context are usually 
treated as fixed by limiting the kinds of communicative goals or by 
operating in a limited domain.  Both of these tactics sidestep the need to 
capture the effects of context, but result in limitations on the usefulness 
of the resulting systems.  But if \nlp\ technology is to become more 
flexible and scale up, models of language 
must pay more attention to context of various kinds.

While methods for deriving expectations from context lie beyond the 
scope of the meaning distributions theory, it is instructive to look at the 
kinds of context that might be used for this purpose in order to illustrate 
how the theory might be developed in this area.

\subsection*{Sources of context}

I will briefly consider five areas of context, these being:
\begin{enumerate}
\item topic,
\item register,
\item author models,
\item discourse structure, and
\item world knowledge.
\end{enumerate}

Topic is an 
important contributor to understanding and a powerful source of 
expectations.  A text will generally focus on a specific topic and even when 
topic shifts occur the new topic can be expected to be related to the 
previous one.  The information retrieval community has devoted much of 
its efforts to identifying topic, although here again the models are 
couched in terms of surface linguistic elements, typically stemmed 
content words (although see Mauldin,~1991, who advocates the use
of a conceptual dependency representation for information retrieval).
But while merely identifying topic is useful for information retrieval, 
little work has been done on deriving semantic expectations from 
knowledge of the topic, a task which the meaning distributions theory 
regards as necessary for language understanding.  

To illustrate how a representation of topic might be used to guide
language understanding, consider Yarowsky's~(1992) sense disambiguation 
system.  By collecting all the words in a~101 word window around 
the target word, he has built a representation of topic.  The co-occurrence 
statistics relate the topic to expectations about the senses of words.  
His system then chooses the most expected 
sense based on this representation of topic, exactly in the way that
the meaning distribution theory suggests.
A more general implementation of the meaning distributions theory
could incorporate a system like this as one 
component of the conceptual model.

The second area of context is register.  Biber~(1993a) explores many 
linguistic variations across registers, demonstrating enormous differences 
amongst them.  In terms of semantic expectations, register is clearly 
important.  In newspaper articles, descriptions of the details of recent 
events are to be expected; in advertisements on the same page, emotive 
suggestions of the usefulness of products will most likely appear.  
Biber's work goes some of the way towards identifying register, but once 
again we lack methods for deriving semantic expectations from this 
knowledge.  Ideally, statistics in the conceptual model should be 
conditioned on register, though this is impractical at this stage.  In 
section~\ref{sec:md_register}, I will outline a more practical position on 
the use of register in statistical language learning.

Another contextual factor is knowledge of the writer or speaker.  This is 
less important in analysing written language than in recognising
speech, so while speaker dependence is of fundamental importance
in the speech recognition community, it not been seen as a problem in \nlp.
However, consider an interactive technical support dialogue system:  the 
queries entered by novice users will be significantly different from those 
of advanced users.  Therefore our semantic expectations should differ in 
each case.  In order to make \nlp\ systems sensitive to these differences, 
techniques from the user modelling community will be needed (for an 
overview see Kass and Finin,~1988).  It is worth noting that author 
modelling shares many elements with register dependence and the 
position I take below on register could also be applied to author 
modelling.  

Discourse structure is an aspect of context that has received quite a bit of 
{\em attention\/} (theoretical treatments include Grosz and Sidner,~1986, 
and Mann and Thompson,~1987),
although in 
practice corpus-based work mainly only addresses the task of identifying 
discourse segments (Hearst,~1994; Reynar,~1994; Litman and 
Passonneau,~1995).  One work where discourse information is applied in 
a statistical model is Hirschberg and Litman~(1993), although this deals 
only with prosodic predictions.  

Interestingly, Charniak and Goldman~(1989) discovered that their 
probabilistic conceptual model for story understanding needs to be 
augmented to allow for inter-sentential coherence.  The augmentation 
takes the form of special conditioning variables representing 
spatio-temporal locality.  
The effect is that the model assigns higher probability 
to logical formulae that share terms with other formulae from the same 
story.  This is a promising first step toward modelling text cohesion 
probabilistically, although as they point out, it is insensitive to discourse 
state.

Semantic expectations do vary with discourse state, chiefly in connection 
with topic.  This fact is exploited by Morris and Hirst~(1991).  In their 
system, thesaurus categories provide a conceptual representation which 
is used to identify coherence relations in a text and thus extract the 
discourse structure.  The fact that topic varies with discourse structure 
and the fact that semantic expectations depend on topic, together, allow 
this scheme to work.  The meaning distributions theory suggests that this 
dependence be utilised in the other direction: discourse structure is used 
to derive semantic expectations.  

A relatively crude way to do this would involve building a statistical 
model to identify discourse structures from cues and then using this 
information to mediate the statistical model of topic outlined above.  
Whenever a discourse boundary was encountered, the contributions 
made by topic to the expected meaning would be attenuated.  A more 
refined system could use a statistical model of rhetorical structure to 
make predictions about the meaning of a sentence.  For instance, in a 
circumstance where a justification link (from the preceding rhetorical 
segment) was likely, a meaning conveying a causal relationship should 
be assigned high probability.  In any event, the use of discourse structure 
in assisting natural language interpretation is a very promising line of 
research and fits directly into the meaning distributions theory.

Finally, under the definition of context that I have adopted, there is world 
knowledge, both of generic and domain specific kinds.  This knowledge 
is the primary concern of the semantic components of all \nlp\ systems 
and its necessity has spurred many of the efforts in knowledge acquisition 
described in section~\ref{sec:sn_motivations}.  I count it as a form of 
context because it is clearly dynamic.  That is, within the course of 
interpreting a text, world knowledge can be augmented, revised or even 
temporarily suspended, as in the case of counterfactuals (or science 
fiction stories for that matter).  I see no motivation for distinguishing 
between world knowledge that is fixed and that which is not, since one 
can always construct a text in which an arbitrary item of world 
knowledge is revoked.

Of course, in the practical construction of a probabilistic conceptual model,
it is likely to be necessary to assume that a great deal of world knowledge is 
fixed.  For example, ontological assumptions of the meaning 
representation scheme will fix some facts, and taking taxonomic 
classifications to be unchangeable is likely to confer necessary 
computational efficiencies.  However, the meaning distributions theory 
leaves this up to the conceptual model; it views world knowledge as just 
another source of semantic expectations.

An example of a statistical model designed to capture domain knowledge 
is that of Fisher and Riloff~(1992).  Their system attempts to identify the 
antecedents of relative pronouns using verb-complement selectional 
restrictions that are specific to the domain.  These selectional restrictions 
are extracted statistically from a corpus of terrorist activity reports and 
include such niceties as priests being one of the most likely things to be 
killed.  

Of all the sources of semantic expectations, world knowledge is the most 
critical to natural language understanding for two reasons.  First, it is by 
far the most voluminous resource, thus representing a large portion of the 
development effort if automatic acquisition is not undertaken.  Second, 
the expectations resulting from it are especially powerful instruments for 
disambiguation.  Therefore, a representation of world knowledge should 
be a central part of any probabilistic conceptual model.
However I have endeavoured to show that at least four other kinds of
semantic expectations are also important and that these can also
be accommodated by the theory.

\subsection*{The ideal of context}

While there are many kinds of contextual information, the meaning 
distributions theory treats them all homogeneously.  They all contribute 
to the collection of semantic expectations that are brought to bear by the 
reader: the prior meaning distribution.  One way to view this is to regard 
context as a summary of the reader's experience.  Therefore, not only 
does the context include the local short-term history provided by the text 
in which a sentence appears (the previous sentence, paragraph or even 
chapter), but it extends beyond this to events experienced by the reader 
long before: books read at an earlier time, statements made by others and 
inferences made from them, and observations of real world events.  
Potentially, all the information that the reader has ever been exposed to 
is relevant to the interpretation of an arbitrary sentence.  

From this perspective it follows that context, in the broad sense used 
here, varies from person to person.  We all bring different experiences to 
the task of interpreting a sentence, and in principle these differences can 
affect our interpretation.  Nonetheless, the intuition that most people will 
share the same interpretation of a given sentence
is still justified by two other 
intuitions.  First, within a single culture most people share a great deal 
of experience: the world presents itself to everyone in much the same 
way.  Second, language is designed to rely heavily on just those 
experiences that are most likely to be shared: speakers and writers avoid 
phrasings that require specialised knowledge unless they know that their 
intended audience shares that knowledge.

It also follows from this perspective that context is dynamic.  The 
information conveyed by each sentence can potentially contribute to the 
semantic expectations for the succeeding one.  In fact, amongst prior 
experiences of the reader, the preceding sentence is one of the most 
likely sources of useful semantic expectations.\footnote{Note, however, 
that there is only one preceding sentence and a vast array of earlier 
experiences, so that the expected utility of the preceding sentence is 
somewhat less than that of all earlier experiences.}  Therefore, in an 
ideal system, probabilities assigned by the conceptual model would be 
constantly updated to reflect the text being read.  This would require a 
detailed theory of the interactions of text and cotext meaning, including 
cohesion, rhetorical structure, attitude and many other supra-sentential 
linguistic processes, something that is almost certainly a long way off.  

Still, in the meantime, we can incorporate those elements for which we 
do have adequate theories into the probabilistic conceptual model.  In the 
experimental work of chapter~\ref{ch:experimental} the conceptual 
models developed ignore almost all elements of context, capturing 
primarily world knowledge.  There remain many further areas 
for exploration within the framework provided by the meaning 
distributions theory.   

\section{Linking Syntax and Semantics} 
\label{sec:md_linking}   

I have argued for the possibility and the necessity of 
using probability distributions to represent semantic expectations; let's 
now consider how the meaning distributions theory proposes that such 
expectations should be utilised.  It starts by assuming that we have at our 
disposal a conceptual model that assigns probabilities to all the meanings 
in $\cal M$, and aims to furnish a method for interpreting natural 
language.  Between these two endpoints there will need to be machinery 
for dealing with language, including substantial lexical and grammatical 
information.

The crux of the meaning distributions theory is a mapping between 
lexico-syntax and semantics.  Intuitively, there are things to say and ways 
to say them, and interpretation requires at least some information about 
the correspondence between the two.  The theory represents this 
correspondence as a mapping between, on the one hand,
superficial forms incorporating word sequences and, on the other, 
terms of the meaning representation language.  
For example, common nouns are typically used to 
indicate concepts, a fact that would be reflected by mapping these nouns 
onto generic symbols.  Verb complementation is typically used to 
indicate the arguments of a predicate and, similarly, the mapping will 
reflect this too.  

This is nothing new.  It has always been assumed that semantic 
interpretation rules could operate on the output of parsers to produce 
meanings; otherwise, why use parsers at all?  What is different here is 
that the meaning distributions theory treats all stages of linguistic 
analysis uniformly.  Semantic interpretation rules, grammatical rules and 
even morphology are simply considered to define a correspondence 
(mathematically speaking, a relation) between word strings and 
meanings.  

Just as the theory does not rely on one particular meaning representation 
(leaving this up to the conceptual model), it also avoids commitment to 
the means by which the mapping is specified.  Traditionally, this 
specification is provided by the combination of a lexicon, a grammar and 
a collection of semantic interpretation rules.  Given the enormous 
research efforts which have been devoted to these components, it would 
seem sensible in practice to use one of the multitude of existing grammar 
formalisms and corresponding parsers.  But the theory does not require 
this.  The only requirement is that computation of the correspondence 
between word strings and meanings be relatively simple.  I will return to 
the reasons for this at the end of this section.  

What is important here is that the theory does not attempt to provide this 
correspondence, nor even a formalism for specifying it.  Instead, it 
merely assumes that a mapping exists, relying on existing \nlp\ 
techniques to provide it.  The theory certainly does not reject the 
necessity of grammar in natural language interpretation.  However, it 
does assert that the grammar can afford to be far simpler if automatically 
acquired probabilistic representations of semantic expectations are 
utilised along with it.

\subsection*{Syntax as constraint}

Let the set of word strings (utterances) be denoted by $\cal W$ and
the set of syntax trees over these strings be denoted by $\cal S$.
The theory regards the mapping between syntactic forms and meanings
as a source of constraints {\em in both directions}.  
On one hand, meanings are expressed by syntax trees, which 
contain word strings.  Each meaning can potentially be 
expressed by many word strings, 
as shown in figure~\ref{fig:md_linking_m2w} (synonymy in the broader 
sense of the term).  
\begin{figure}
\centering
%
%
%
{\tt    \setlength{\unitlength}{0.92pt}
\begin{picture}(500,199)
\thinlines    
\put(311,43){\vector(1,0){17}}
\put(311,85){\vector(1,0){17}}
\put(311,127){\vector(1,0){17}}
\put(147,73){\vector(2,-3){19}}
\put(147,73){\vector(4,3){19}}
\put(147,73){\vector(1,3){19}}
\put(328,31){\framebox(140,23){}}
\put(328,74){\framebox(140,23){}}
\put(328,115){\framebox(140,23){}}
\put(0,61){\framebox(147,23){}}
\put(5,70){\small \predicate{eat}(\concept{MAMMAL\_MOLE},$\ldots$)}
\put(173,124){\tt \small ((Insects (are eaten)) by $\ldots$)}
\put(173,81){\tt \small (Moles (eat insects))}
\put(173,37){\tt \small ((Moles feed) on insects)}
\put(226,8){:}
\put(380,8){:}
\put(332,123){\lingform{Insects are eaten by $\ldots$}}
\put(332,82){\lingform{Moles eat insects.}}
\put(332,38){\lingform{Moles feed on insects.}}
\put(67,169){$\cal M$}
\put(228,166){$\cal S$}
\put(378,164){$\cal W$}
\put(167,115){\framebox(144,23){}}
\put(167,74){\framebox(144,23){}}
\put(167,31){\framebox(144,23){}}
\end{picture}}
\caption{Meanings generate many word strings}
\label{fig:md_linking_m2w}
\end{figure}

On the other hand, word strings admit parses which can be interpreted to 
yield meanings.  Each word string can potentially express many 
meanings, as shown in figure~\ref{fig:md_linking_w2m} (polysemy, in 
the corresponding sense of the term).  It is this latter ambiguity that poses 
such great difficulties for natural language understanding.\footnote{I 
should emphasise here that the concept of mapping word strings onto 
parse trees and thence onto meaning representations is not intended to be 
a contribution of the theory, being hardly novel.}
\begin{figure}
\centering
%
%
%
{\tt    \setlength{\unitlength}{0.92pt}
\begin{picture}(465,150)
\thinlines    
\put(173,76){\vector(-1,-2){15}}
\put(173,76){\vector(-4,3){15}}
\put(339,70){\vector(-4,-3){30}}
\put(339,70){\vector(-3,1){30}}
\put(10,36){\framebox(148,24){}}
\put(10,75){\framebox(148,24){}}
\put(173,64){\framebox(136,23){}}
\put(16,45){\small \predicate{eat}(\concept{MAMMAL\_MOLE},$\ldots$)}
\put(16,83){\small \predicate{eat}(\concept{UNIT\_MOLE},$\ldots$)}
\put(380,125){\bf $\cal W$}
\put(228,125){\bf $\cal S$}
\put(75,125){\bf $\cal M$}
\put(345,66){\lingform{Moles eat insects.}}
\put(75,17){:}
\put(228,45){:}
\put(192,73){\tt \small (Moles (eat insects))}
\put(339,57){\framebox(118,23){}}
\end{picture}} 
\caption{A word string can express many meanings}
\label{fig:md_linking_w2m}
\end{figure}

So lexico-syntax (henceforth I shall abuse terminology and say simply 
syntax) can be seen as a bidirectional many-to-many mapping between 
the space of meanings expressible by the meaning representation scheme 
and the set of word strings.  The theory takes the position that this 
mapping is a hard constraint. There are no soft edges in syntax: either a 
word string can express a meaning (in at least one context), or it never 
can.  

Thus, syntax is viewed as delineating what is possible in any context.  It 
follows that syntax is context-independent and that therefore all 
variations in language use across different contexts must arise from 
variations in meaning distributions.  Like many other assumptions 
commonly made in statistical models, it is easy to find convincing 
counter-examples to this implication.  However, the ultimate usefulness 
of the theory only depends on the performance of implementations of it.  
The approximation we make by accepting a false assumption may be 
accurate enough to justify the gains it provides.

I have emphasised the bidirectional nature of the mapping because it is 
necessary for the statistical model.  To 
apply the theory to the process of interpretation, it is not only necessary 
to be able to compute the set of possible meanings allowed by a word 
string, but it is also necessary to compute the set of word strings that 
allow a given meaning.  This is a necessary (if unusual) consequence of 
the mathematical formulation; however, assuming that the mapping is 
computable in both directions has the advantage of potentially making 
unsupervised training possible, as we shall soon see.  

\subsection*{Arriving at an interpretation}

Up until now I have been avoiding introducing formal mathematical 
notation, but to make the theory precise we will need a few definitions.
First, the goal is to derive intended meanings from the word strings 
used to express them.  Formally, we wish to define a function $\tau_C: 
{\cal W} \rightarrow {\cal M}$ denoting the best interpretation for each 
word string in the context, $C$.

Given a meaning (an element of $\cal M$), 
syntax provides a set of word strings, each of which can 
express that meaning in some context.  
Let $\theta: {\cal M} \rightarrow~2^{\cal W}$ denote the map from 
meanings to sets of word strings representing this.  
Figure~\ref{fig:md_linking_image} depicts the following definition.  
\begin{figure}
\centering
%
%
%
{\tt    \setlength{\unitlength}{0.92pt}
\begin{picture}(315,139)
\thinlines    
\put(35,75){Meaning}
\put(235,42){Image}
\put(70,87){\vector(4,-1){156}}
\put(251,45){\circle{46}}
\put(70,87){\circle*{4}}
\put(246,69){\oval(118,118)}
\put(242,103){$\cal W$}
\put(69,69){\oval(118,118)}
\put(65,103){$\cal M$}
\put(154,35){\shortstack{\large S\\ \large Y\\ \large N\\
	\large T\\ \large A\\ \large X}}
\end{picture}}
\caption{The image of a meaning is a set of word strings}
\label{fig:md_linking_image}
\end{figure}
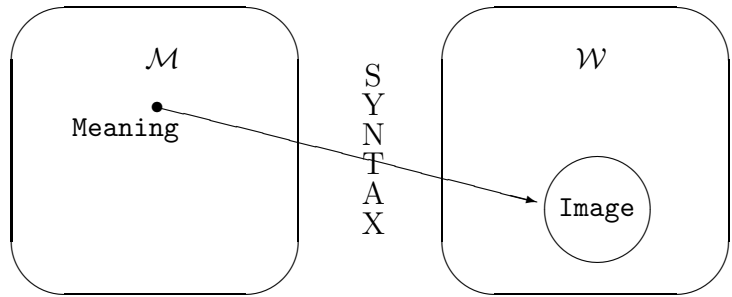
\begin{definition}[Image]
The {\em image\/} of a meaning, $m \in {\cal M}$, is the set of word 
strings that can express $m$, that is, $\theta(m)$.
\end{definition}

Each $m \in {\cal M}$ has, in a given context, a probability associated 
with it by the probabilistic conceptual model.  Let $\Pr_C(m)$ denote this.
In general, $m$ can be expressed in more than one way (that is, 
$|\theta(m)|>1$).
We are interested in the probability of $m$ being expressed in 
these different ways.  The theory takes the stance that 
syntax is purely a constraint. 
Therefore the syntactic mapping does not distinguish 
between the different elements of the image of $m$.  Either a word string 
is able to express $m$ (that is, it is in the image of $m$) or it is 
unable to do so (it is not in the image).  
Each element of the image is equally likely to be 
used to express $m$.
Let $\Pr_C(m,w)$ be the probability of a sentence in 
context $C$ expressing the meaning $m$ using the word string $w \in W$.
According to the theory, this probability is the same for all word
strings in the image of a given $m$.  This assumption is captured
by equation~\ref{eq:md_linking_contrib}.
\begin{equation}
\Pr_C(m,w) = \left\{ 
\begin{array}{ll} 
\frac{\Pr_C(m)}{\mid \theta(m) \mid} & \mbox{if $w \in \theta(m)$} \\
0 & \mbox{otherwise}
\end{array}
\right.
\label{eq:md_linking_contrib}
\end{equation}

We also need corresponding definitions in the other direction so that we 
can do interpretation from word strings.  Let $\phi: {\cal W} 
\rightarrow~2^{\cal M}$ denote the map from word strings to meanings 
defined by syntax.  Given a word string, it provides a set of a meanings, 
each of which could be expressed by that word string in some context.  
Figure~\ref{fig:md_linking_source} depicts the following definition.  
\begin{figure}
\centering
%
%
%
{\tt    \setlength{\unitlength}{0.92pt}
\begin{picture}(315,139)
\thinlines    
\put(242,73){String}
\put(45,42){Source}
\put(243,87){\vector(-4,-1){156}}
\put(64,45){\circle{46}}
\put(243,87){\circle*{4}}
\put(246,69){\oval(118,118)}
\put(242,103){$\cal W$}
\put(69,69){\oval(118,118)}
\put(65,103){$\cal M$}
\put(154,35){\shortstack{\large S\\ \large Y\\ \large N\\
	\large T\\ \large A\\ \large X}}
\end{picture}}
\caption{The source of a word string is a set of meanings}
\label{fig:md_linking_source}
\end{figure}
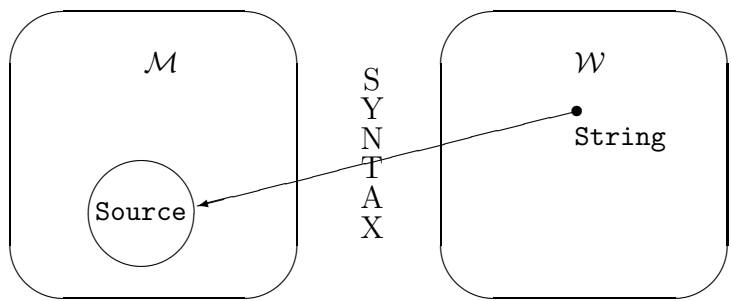
\begin{definition}[Source]
The {\em source\/} of a word string, $w \in {\cal W}$, is the set of 
meanings that could be expressed by $w$, that is, $\phi(w)$.
\end{definition}

To translate probabilities in the meaning space into probabilities of word 
strings, the probability of each meaning must be divided up amongst 
each of the word strings lying in its image.  Since, in general, word 
strings lie in the image of multiple meanings, each word string will 
receive multiple contributions to its probability.  The size of each 
contribution will depend not only on the probability of the corresponding 
meaning, but also on the number of other word strings lying in the image 
of that meaning.  Equation~\ref{eq:md_linking_total} gives the 
resulting probability distribution over word strings.
\begin{equation}
\Pr_C(w) = \sum_{m \in \phi(w)} \Pr_C(m,w) \label{eq:md_linking_total}
\end{equation}

We can now precisely state the central tenet of the meaning distributions 
theory.  
\begin{hypothesis}[Interpretation theory]
The best interpretation for a given sentence is that meaning which makes the 
greatest contribution to the probability of the sentence in the context.  
That is, $\tau_C(w) = \mbox{argmax}_{m \in \phi(w)} \Pr_C(m,w)$.
\end{hypothesis}

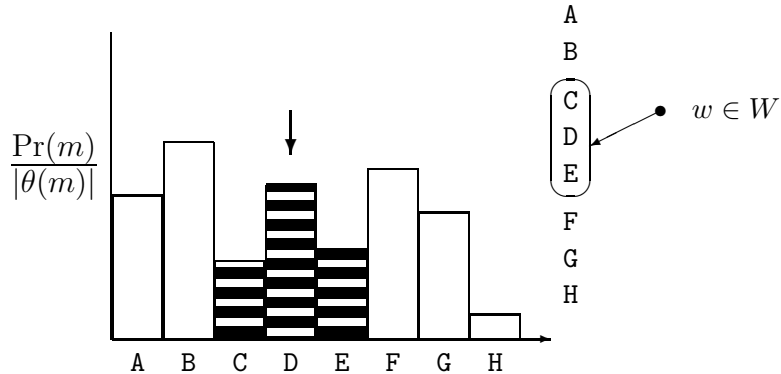
\begin{figure}
\centering
%
%
%
{\tt    \setlength{\unitlength}{0.92pt}
\begin{picture}(368,174)
\thicklines   
\put(0,95){\LARGE $\frac{\Pr(m)}{|\theta(m)|}$}
\put(281,119){$w \in W$}
\put(85,53){\rule{19pt}{4pt}}
\put(85,45){\rule{19pt}{4pt}}
\put(85,37){\rule{19pt}{4pt}}
\put(85,29){\rule{19pt}{4pt}}
\put(106,28){\rule{19pt}{1pt}}
\put(106,33){\rule{19pt}{4pt}}
\put(106,41){\rule{19pt}{4pt}}
\put(106,49){\rule{19pt}{4pt}}
\put(106,57){\rule{19pt}{4pt}}
\put(106,65){\rule{19pt}{4pt}}
\put(106,73){\rule{19pt}{4pt}}
\put(106,81){\rule{19pt}{4pt}}
\put(106,89){\rule{19pt}{3pt}}
\put(127,29){\rule{19pt}{4pt}}
\put(127,37){\rule{19pt}{4pt}}
\put(127,45){\rule{19pt}{4pt}}
\put(127,53){\rule{19pt}{4pt}}
\put(127,61){\rule{19pt}{4pt}}
\put(116,122){\vector(0,-1){18}}
\thinlines    
\put(228,43){H}
\put(228,58){G}
\put(228,73){F}
\put(228,93){E}
\put(228,108){D}
\put(228,123){C}
\put(228,143){B}
\put(228,158){A}
\put(268,122){\vector(-2,-1){29}}
\thicklines   
\put(268,122){\circle*{4}}
\thinlines    
\put(197,15){H}
\put(176,15){G}
\put(155,15){F}
\put(134,15){E}
\put(113,15){D}
\put(92,15){C}
\put(71,15){B}
\put(50,15){A}
\put(190,28){\framebox(20,10){}}
\put(169,28){\framebox(20,52){}}
\put(148,28){\framebox(20,70){}}
\put(127,28){\framebox(20,36){}}
\put(106,28){\framebox(20,63){}}
\put(85,28){\framebox(20,32){}}
\put(64,28){\framebox(20,81){}}
\put(43,28){\framebox(20,59){}}
\put(43,28){\vector(1,0){180}}
\put(42,154){\line(0,-1){126}}
\put(231,111){\oval(16,48)}
\end{picture}}  
\caption{The interpretation theory selects the most probable allowed 
meaning}
\label{fig:md_linking_best}
\end{figure}

This is shown diagrammatically in figure~\ref{fig:md_linking_best}.  
Syntax constrains the set of possible meanings of $w$ 
($\phi(w) = \{C, D, E\}$), as shown at the right hand side.
Once the source of a sentence has been identified (the filled columns of 
the figure), the highest probability meaning within the source is the best 
interpretation.  This is indicated by an arrow.

Apart from constraining the possible interpretations, syntax
does not contribute to the selection.  This is done by the
probabilistic conceptual model.
Syntactic knowledge cannot, according to the theory, specify that one 
syntactic structure is more likely than another; lexical knowledge cannot 
specify that one word is twice as likely as another.  The probability of a 
parse tree comes directly from the probability of its possible meanings, 
unmodulated by syntactic knowledge in any way other than as constraint.

Another way to say this is there are no independently represented 
probability distributions over $\cal S$ and $\cal W$.  The distributions 
of syntactic structures and word strings are derived solely from those of 
meanings using only the structural information from the 
context-independent syntactic constraints.  Much work has focussed on 
representing the distributions on $\cal W$ (for example, Marcus,~1991) 
and on $\cal S$ (for example, Jelinek~\etal,~1992), but assumed that the 
distributions on $\cal M$ weren't necessary.  The meaning distributions 
theory takes the opposite tack: model the distribution on $\cal M$ and 
assume that independent modelling of distributions on $\cal S$ and $\cal 
W$ are unnecessary.  

%
Naturally, if it is discovered that some specific element of lexico-syntax 
violates the assumptions of the theory in such a way as to significantly 
degrade performance, then it would be reasonable to extend the theory to 
incorporate the relevant distributions of that element.  However, until 
such elements are shown to be problematic, it is reasonable to proceed 
assuming that they are not.

Before I turn to some more practical aspects, let's look at one more 
formulation of the theory.  We can use the notion of conditional 
probability to represent the relationship between meanings and word 
strings.  Thus, $\Pr_C(m \mid w)$ represents the 
probability that $m$ is the intended meaning conveyed by the word 
string $w$ in a context $C$.  To pick the most likely 
allowed interpretation of the string, it is necessary to find whichever $m$ 
maximises this probability.  Using Bayes' theorem, this can be 
reformulated as in equation~\ref{eq:md_linking_bayes}.
\begin{equation}
\Pr_C(m \mid w) = 
\frac{\Pr(w \mid m) \Pr_C(m)}{\Pr_C(w)}
\label{eq:md_linking_bayes}
\end{equation}
Notice that the first factor of the numerator is independent of the context, 
$C$, by virtue of the assumption that syntax is context-independent.  
Also, the denominator is constant for a given word string, so to maximise 
the left-hand side, it is only necessary to maximise the numerator of the 
right-hand side.

Comparing equation~\ref{eq:md_linking_bayes} with 
equation~\ref{eq:md_linking_contrib} above we see that the 
interpretation theory is equivalent to the following assumption.
\begin{equation}
\Pr(w \mid m) = \left\{ 
\begin{array}{ll}
\frac{1}{\mid \theta(m) \mid} & \mbox{if $w \in \theta(m)$} \\
0 & \mbox{otherwise}
\end{array}
\right.
\end{equation}
Once again, it is easy to find counter-examples to this assumption.  
However, it is just as easy to find counter-examples to the Markov 
assumption in tag sequences, and yet hidden Markov models perform 
tagging very accurately.  Only empirical investigation can determine 
whether this assumption justifies the time and space savings it yields.

\subsection*{Some practical implications}

A key premise of the theory is the claim that, in constraining the 
interpretation, the burden placed on lexico-grammar may be made 
significantly lighter by the use of semantic expectations.  The grammar 
can widely underconstrain the set of possible interpretations and 
semantic expectations will still recover the intended meaning.  This 
accords with the intuition that semantics requires distinctions that syntax 
does not make.  The practical consequence of this for \nlp\ systems is 
that less effort need be expended on grammar development.  A relatively 
simple over-generating grammar will suffice.

This can be compared to the position taken on generation in Knight and 
Hatzivassiloglou~(1995).  Their system is provided with a set of 
generation rules mapping semantic elements into word sequences.  These 
rules generate an enormous number of sentences for any given content, 
many of them ungrammatical and most undesirable.  The resulting 
sentences are evaluated by a word bigram model and the most probable 
one is chosen to be generated.  The development effort required to 
construct the generation rules was greatly reduced because grammatical 
agreement and other syntactic complexities were unnecessary.  These 
phenomena were recovered by the statistical model.  The presence of a 
statistical filter at the {\em output\/} side of the system resulted in 
relatively weak lexico-grammatical knowledge being sufficient.

The meaning distributions theory can be seen as the mirror image of 
Knight and Hatzivassiloglou's system.  Instead of over-generating word 
strings and then filtering these statistically in order to perform 
generation, it over-generates meanings and filters these statistically in 
order to perform interpretation.  These two proposals share further 
properties, one of which I will mention in a moment.  

I have claimed earlier that the meaning distributions theory not only 
allows the syntactic component to be simpler, but that it {\em requires\/} 
the syntactic mapping to be relatively simple in a certain mathematical 
sense.  While in principle, the only restriction on the mapping is that it be 
computable, two practical considerations limit its complexity.
\begin{enumerate}
\item In order to control the computational complexity of parsing, it is 
necessary to translate probabilities on $\cal M$ into probabilities on 
$\cal S$.
\item To allow unsupervised training of the conceptual model, elements 
of the meaning representation must map to relatively simple word 
patterns.
\end{enumerate} 

The first of these is caused by the certain impracticality of first 
generating all possible parses of a sentence and then selecting the most 
semantically acceptable one.  Even using efficient parsing techniques and 
packed parse forest representations, the processing times to compute the 
syntactic structure alone are very large.  In any case, the original 
motivation for the meaning distributions theory was to give semantic 
information greater control over the interpretation process.  

Therefore, given a syntactic mapping, it is necessary to translate the 
probabilities given by the meaning distribution into probabilities relevant 
to computing the syntactic mapping.  For example, if this mapping is 
provided by a grammar and semantic interpretation rules (as in a 
traditional \nlp\ system), a means of calculating the likelihood of 
grammatical constituents from the meaning distribution is needed.  These 
probabilities can then be used to control parsing, just as the Inside 
algorithm is used to guide a Viterbi parser for context free grammars 
(see Jelinek~\etal,~1992).  

An analogous procedure is utilised by Knight and Hatzivassiloglou 
(1995).  Bigram probabilities are referred back to compute probabilities 
for the generation rules and these are compiled into the generation 
process.  Thus their system avoids the impossible task of generating the 
millions of possible word strings for a given content.  A further example 
of such probability translation is provided by Lafferty~\etal~(1992) for 
Link Grammar, a syntactic formalism that is somewhat closer to semantics 
than probabilistic context free grammars.

Since the meaning distributions theory does not commit to a particular 
syntactic formalism, optimisation procedures of this kind are beyond the 
scope of the theory; their development must be left for future work.  
However, the important point here is that the syntactic mapping must be 
sufficiently direct to support a mathematical translation such as that 
provided by the Inside algorithm.  It is in this sense that the syntactic 
mapping must be relatively simple.

The second consideration lies with the need to train the conceptual 
model.  If the meaning space is non-trivial, then a large amount of 
training data will be required.  Since it is impractical to annotate such 
large amounts of text with meaning representations, the availability of 
unsupervised training methods is crucial to the success of the theory.  A 
key difference between this work and that of Miller~\etal~(1994) is their 
reliance on supervised training data.  

The bidirectionality of the syntactic mapping is useful here.  In order to 
train the conceptual model it is necessary to estimate the probability of 
the different units that make up the the meaning representation.  To 
gather evidence about these units, we can use the syntactic component to 
map them into simple word patterns and count occurrences of these 
patterns in the corpus.  These counts represent a sample of the space 
defined by the conceptual model.  

However, the observations made in this way are unreliable.  Any simple 
word pattern will be semantically ambiguous to some degree, so that 
what we take to be evidence about one meaning element may actually be 
evidence for another.  Even if a boot-strapping scheme (such as the 
cyclic refinement used by Hindle and Rooth,~1993) is pursued, the data 
used for parameter estimation will always be noisy.  

As Yarowsky's~(1992) training algorithm demonstrates, a sufficiently 
simple, bidirectional, ambiguous mapping can be used to overcome noisy 
data if it has appropriate properties.  Evidence about an unseen event (in 
his case the word sense) can be extracted from multiple, ambiguous sets 
of corresponding seen events (in his case the co-occurrence data for each 
word in a thesaurus category).  Even though no single set of observations 
is unambiguous, the combination of the sets gives a clear indication of 
the properties of the unseen events.  

For this to work in the case of meaning distributions, the syntactic 
mapping must have certain properties.  To start with, each meaning must 
have a large {\em image}, so that evidence is combined from many 
different observed word strings.  Further, the {\em source\/} of each word 
string in the image should differ, so that noise is scattered thinly over 
different meanings.  It is an empirical question as to whether this is the 
case for suitable syntactic mappings.  What is clear is that the syntactic 
mapping must be relatively simple.  The units which make up the 
conceptual model must be easily mapped to corresponding word patterns 
which can be counted.  

So for two reasons the syntactic mapping must be relatively simple.  An 
important consequence of this is that the meaning representation scheme 
used in the conceptual model must be relatively shallow.  While this 
might be viewed as a disadvantage, in the next section I will discuss the 
recent increase in the use of shallow semantic representations, an 
increase that suggests we might be better off with them anyway.

\subsection*{Shallow semantic representations} 

Knowledge representation has been the subject of voluminous research 
(see for example, Brachman and Levesque,~1985) and is seen as an end 
in itself for much of this research, being applicable to many other
tasks than natural language processing.  
Much of the emphasis in knowledge representation research is 
on formal properties such as expressiveness, soundness and 
completeness (see for example, Sowa,~1984).  The computational 
complexity of inferencing also receives attention and is somewhat closer 
to the practical task of engineering \nlp\ systems, but generally the 
problems being addressed are motivated by examples invented for the 
purpose and are, therefore, a long way from the details of real text.

Still, in order to take advantage of the work in knowledge representation, 
many \nlp\ researchers have opted for a knowledge representation 
scheme so produced (for example, McDonald,~1982, uses 
\progname{netl}; Hirst,~1987, uses \progname{frail}; 
Hobbs~\etal,~1993, use first-order predicate calculus).  There is a cost, 
however.  Since these schemes were developed without particular regard 
to natural language understanding, they place emphasis on deep 
distinctions.  That is, they postulate many meaning differences that are 
not expressed on the surface --- differences which \nlp\ systems based on 
these schemes will have to recover.  Metaphorically speaking, we might 
say that there is a long distance between the natural languages of texts 
and knowledge representation languages.  This increases the difficulty of 
the interpretation task, perhaps making it more difficult than is necessary 
for the task at hand.

The alternative is to use more superficial representations of meaning, 
representations that are more text-like, but sufficiently less ambiguous 
than text to support useful \nlp\ tasks.  The idea is to sacrifice 
fully general logical inference (and many other knowledge representation 
niceties) with the goal of making interpretation easier.  
After all, for many tasks, 
computing all the logical consequences of a sentence will rarely (if ever) 
be necessary.  This idea has been around since at least 
Alshawi~(1983:23), for whom it is a key step in building a practical 
system.

According to this point of view, choosing a meaning representation 
which is both simpler and shallower than the full-blown schemes 
advocated by the knowledge representation community makes 
developing \nlp\ systems easier.  This argument is closely related to the 
one I have made already regarding grammatical knowledge: by relying on 
statistically acquired semantic expectations, the grammatical component 
can be simplified, resulting in lesser development costs.  Similarly, 
aiming at a shallower meaning representation again reduces the 
development cost.  And this is the \foreign{raison d'\^{e}tre} of 
statistical language learning: to bypass laborious human effort.

The cost of taking this position is the risk of failing to make a distinction 
in the meaning representation that turns out to be necessary for the task 
at hand.  By reducing the development effort, we've also reduced the 
discriminatory power of the system.  Therefore, without evidence to 
suggest that the reduction in power is insignificant in comparison to the 
reduction in development effort, we can't conclude that shallower 
representations are better, only that they aren't necessarily worse.  

However, a number of recent corpus-based works provide suggestive 
evidence in favour of the shallower path by utilising semantic 
representations that are quite superficial, even in comparison to 
Alshawi's~(1983) scheme.  Apart from the conceptual association 
proposed by Resnik and Hearst~(1993), which I have already cast
as employing a shallow semantic representation in 
section~\ref{sec:sn_conceptual}, I will consider four of these.

Basili~\etal~(1993) propose the use of selectional constraints associated 
with syntactic relationships.  In particular, they focus on prepositional 
modifiers, for which conceptual relationships like 
[\concept{go}]~$\rightarrow
$~(\semrel{instrument})~$\rightarrow
$~[\concept{vehicle}] are to be used in disambiguating 
the prepositional attachment in example~\ref{eg:md_linking_basili}.
\begin{examples}
\item John goes to Boston by bus    \label{eg:md_linking_basili}
\end{examples}
This is a fairly shallow semantic representation, if shown only by the fact 
that their mapping from syntactic relations to conceptual relationships is 
a direct and simple one.  But the representation they use in the 
experimental work is even shallower.  There, triples consisting of (word, 
preposition, semantic~tag) are used to represent conceptual relationships 
and the number of semantic tags for their domain is only twelve.  The 
performance results suggest that shallow semantic representations can be 
very useful.

Resnik~(1993) is similar, considering the selectional restrictions placed 
on the direct object of verbs.  The semantic representation consists of 
pairs of the form (predicate, concept), where the concept is taken to be 
an argument of the predicate~(Resnik,~1993:54).\footnote{The specific 
argument (subject, direct object, etc.) varies with the application.}  
This representation is again rendered shallower in practice, with the set of 
predicates being taken to be simply the transitive verbs. The concepts are 
provided by the synonym sets in the WordNet thesaurus.  The result is a 
very shallow form of predicate calculus, where WordNet categories 
provide the complete set of logical atoms and predicates are one-to-one 
with English transitive verbs.

Justeson and Katz~(1995) consider the sense discrimination of 
adjectives using the nouns that they modify.  
Their representation consists of pairs, (adjective, 
noun~feature), where the noun features are semantic markers like 
\marker{+concrete}.  This representation captures only a narrow
slice of the meaning of nouns, but is sufficient to select the 
correct adjectival sense with an accuracy of~97\% (at~50\% coverage).
Once again, shallow semantic representations prove useful for 
building a statistical model.

As already mentioned, Alshawi and Carter~(1994:637) use 
\acronym{qlf}s, which \shortquote{express 
semantic content, but are derived compositionally from complete 
syntactic analyses of a sentence and therefore mirror much syntactic 
structure as well}.  While their main purpose is to compare statistical 
metrics, they do sketch a representation that comes close to being a 
probabilistic conceptual model in the sense described earlier in this 
chapter, based on quasi logical forms.  These shallow semantic forms 
play the part of the simplified meaning representations advocated in 
Alshawi~(1983) and support several disambiguation tasks within their 
spoken language translation system.

All four of these studies use shallow semantic representations to 
facilitate statistical language learning.  Furthermore, while the 
promise of automatic language acquisition is an attractive enough lure by 
itself, even the Core Language Engine, \shortquote{a primarily 
rule-based natural language-processing system} (Alshawi and 
Carter,~1994:636), uses quasi logical forms.  This suggests that 
shallower semantic representations ought to have a place in natural 
language understanding, even if only as an intermediate representation, 
and that therefore, the requirements placed upon the semantic 
representation by the meaning distributions theory are not as limiting as 
they might first appear.


\section{A Note on Register} 
\label{sec:md_register} 

Finally, before turning from the topic of meaning distributions to more 
general reflections on statistical \nlp, I want to consider the notion of 
register.  The register of a text is a categorisation according to the 
medium and situation in which it appears, and the importance of register 
to linguistic phenomena did not escape the earliest corpus builders.  
When Francis and Ku\v{c}era~(1982) built the Brown corpus in the 
early sixties, they were very careful to compose it from a wide range of 
sources.  Since then a very large range of corpora of different kinds has 
become available, some \scare{balanced} (for example, Brown and 
\acronym{lob}), others pure (including AP Newswire and Grolier's encyclopedia) 
and still others collected without particular regard to balancing sources
(the Penn Treebank is an instance, see Marcus~\etal,~1993).  

Biber~(1993a) has made extensive study of the cross-register variation in 
language.  For example, he has shown that within two subsets of the 
\acronym{lob} corpus, one collected from fictional works and the other from 
expositions, there are marked differences in the distribution of 
grammatical categories of many words.  The word \lingform{given} is 
used passively in~71\% of its uses within exposition, but only in~19\% of 
its uses within fiction.
As Biber argues, this has \shortquote{important implications for 
probabilistic taggers and parsing techniques, which depend on accurate 
estimates of the relative likelihood of grammatical categories in 
particular contexts}~(Biber,~1993a:222).  He also argues that 
\shortquote{analyses must be based on a diversified corpus representing 
a wide range of registers in order to be appropriately generalised to the 
language as a whole, as in \dots a general purpose tagging 
program}~(Biber,~1993a:220).  For linguistic and lexicographic purposes 
it seems clear that so-called balanced corpora are required for 
completeness; however, for the purposes of statistical language learning, 
the moral is not so clear.  

It certainly appears that cross-register variation is significant enough to 
influence the performance of probabilistic taggers, so it is crucial that we 
pay attention to it.  However, it is not sufficient merely to use a balanced 
corpus.  Consider, for example, training a unigram tagger on a corpus 
consisting of two parts, one fictional and the other from expositions.  The 
probability estimate for the passive tagging of the word \lingform{given} 
will be based on training examples in both registers.  If the word is 
equally frequent in each half, then the expected estimate of this 
probability is~0.45 (using the percentages above).  This estimate is 
wildly incorrect for both halves of the training corpus, and will likely 
result in poor performance on this word.  While an accuracy rate 
of up to~76\% (on tagging the word \lingform{given}) is possible 
if the unigram tagger is trained and then applied to each half 
of the corpus separately, the best possible performance when 
trained on the whole corpus is only~55\%.\footnote{These figures
are only upper bounds because they (falsely) assume that \lingform{given}
has only two possible tags.}

This example demonstrates that the notion of an average English may 
well be a chimera and those who attempt to acquire it should 
beware.\footnote{Even linguists should take some care.  Clear~(1992) 
argues a similar point for corpus-based studies in a lucid paper on the 
abstract notion of the composition of language.}  An examination of various 
research work supports this view, with most of the highly successful 
statistical language learning results having been achieved 
using register-pure corpora.  We must recognise that the parameters 
of probabilistic models are generally dependent on the type of text 
being modelled. If statistical models are to yield an accurate 
picture of language, then separate distributions must be maintained 
for different registers.  In practice, however, this is going to 
be difficult.  We are already struggling to find sufficient data to train 
probabilistic models; dividing the corpus further will only exacerbate the 
problem.  

In the short term, there is a better way to proceed: choose one particular 
register, train using data only from that register, and accept that the 
resulting systems are only (reliably) applicable to the same register.  For 
this, we need large register-pure corpora, which, luckily, are currently 
available, at least for research.  In the current work, encyclopedia entries 
are used for both training and testing.  In my view the use of a corpus of 
uniform register is the correct method for applying probabilistic 
modelling.  Naturally, it is not possible to guarantee that the results will 
be equally useful for processing a different register.

\section{Summary} 
\label{sec:md_summary} 

In this chapter I have proposed an architectural theory of statistical \nlp.
It defines a novel class of designs for statistical language learners 
and thus expands the collection of proposed language models 
for \sll\ systems.
The architecture advocated by the theory includes three components:
\begin{itemize}
\item a shallow semantic representation;
\item a conceptual probabilistic model; and
\item a lexico-syntactic module.
\end{itemize}
According to the theory, natural language analysis is driven 
by semantic expectations that are provided by the conceptual
probabilistic model.  A clearly separate syntactic representation 
is maintained; however, lexico-grammar plays a passive role, 
acting only as a constraint.

This architecture places more emphasis on context than
syntactically driven \nlp\ approaches because context 
can be directly represented in the probabilistic conceptual model.
In principle, this provides for sensitivity to topic, register, 
author and discourse structure, as well as world knowledge.
While the architecture requires that the semantic representation
be relatively shallow, this does not immediately appear detrimental
and has the side-effect of facilitating unsupervised learning.

Finally, the aim of the theory is to suggest new designs for
language processing systems.  As such, the ultimate evaluation 
of the theory must be empirical.  This is the concern of the 
work in chapter~\ref{ch:experimental}.  I am not claiming that
the designs advocated by the theory are inherently superior.
Rather, such designs form promising ground for exploration.
The contribution of the theory is the identification 
of a sizable new area within the world of possible statistical 
models of language.

\cleardoublepage
 
\chapter{Data Requirements}
\label{ch:dr}

 

\section{Introduction}

The theory presented in chapter~\ref{ch:md} offers a new 
approach to the design of statistical language learners.
As such it offers one avenue in the space of possible designs.  
However, this space is both large and complex.  Furthermore,
the constraints governing the design of \sll\ systems 
are still only poorly understood.  
We need more informed methods for exploring the design space.

The subject of this chapter is the prediction of 
training data requirements for \sll\ systems.  Researchers
in the area have been using increasingly large training corpora 
for their experiments and what was once regarded as a large
corpus is nowadays relatively small.
Yet despite this enormous increase in supply, data sparseness 
remains a common phenomenon and
this raises the question of how much data is enough.  
For example, Brown~\etal~(1992a) use 583 million words of training
data, yet it is still possible that training on a billion
words would yield significantly better results.  What is needed
is a theory that can predict how much training data is
expected to yield a given performance.

Because training data is crucial to the performance of \sll\ systems,
availability of sufficient training data is a key design issue.  
Different designs often require dramatically different amounts of
training data.  Therefore predicting the data requirements of a given
model is an important part of informed design.  
A predictive theory of these requirements would be an invaluable 
navigational aid in the exploration of the \sll\ design space.
While the goal of predicting data requirements for an arbitrary
statistical language learning algorithm is still some way from being
realised, this chapter begins the development of just such a tool.

In section~\ref{sec:dr_probabilistic}, I advocate the use
of explicit probabilistic models because, not only do they support
reasoning about data requirements, but they have several other
independent motivations.

Section~\ref{sec:dr_sparseness} will explore the phenomenon
of data sparseness and consider the existing ways of dealing with it.
While techniques are available to alleviate data sparseness
in specific instances, they do not remove the need for a means
of predicting data requirements in advance.  In section~\ref{sec:dr_need},
I will argue for the necessity of such a theory and indicate why
existing statistical theories do not give the desired information.

To provide some data points that the theory might be
expected to explain, section~\ref{sec:dr_tour} examines six
\sll\ experiments that have been reported in the literature,
paying particular attention to their training data volumes.
From this survey, it appears that significantly less training
data than might be expected from a simple rule of thumb is
sufficient to give successful results.

In order to move toward a general theory, I have defined a framework
for viewing \sll\ systems that can be used to reason about their
data requirements.  The framework is defined in 
section~\ref{sec:dr_framework} and forms the notational foundation
for the results in later sections.  Section~\ref{sec:dr_beginning}
defines a simple statistical learning algorithm within this
framework and gives some results regarding the expected accuracy
of this algorithm as a function of the training set size.

To arrive at a bound on the expected accuracy of this
learning scheme in terms of the volume of training data, it is
necessary to make some simplifications.  This is done in 
section~\ref{sec:dr_global}, leading to a method of predicting
data requirements under certain limited conditions.  Finally,
section~\ref{sec:dr_simulations} reports a series of simulations
to verify these results and to explore an alternate and more
realistic set of conditions.

\section{Reasons for Probabilistic Models}  
\label{sec:dr_probabilistic}
 
It is fundamental to the approach that I will take that  
statistical language learning systems be based on probabilistic models.   
Therefore, before I turn to the theory, this section will consider some  
independent reasons for wanting such a model. 
 
The probabilistic taggers and grammars described earlier (see  
sections~\ref{sec:sn_taggers} and~\ref{sec:sn_grammars}) are all based on  
clearly defined probabilistic models. 
%
%
Having at least a rudimentary probabilistic model should be regarded as a  
necessity for any system that claims to be {\em statistical}, since 
statistical methods make some assumptions about the probability  
distributions from which samples are taken.  Currently, however, all \nlp\  
systems that derive information from a large corpus are referred to as  
statistical.  Of these, only a subset are based on an explicit probabilistic  
model. That is, only some describe linguistic events of interest by 
means of random variables with fixed ranges and 
precise (if unknown) probability distributions, 
where the evaluation of possible analyses follows the laws of  
probability theory. 
 
Others use statistical methods like the $t$-test (for example, Fisher and  
Riloff,~1992 and Resnik and Hearst,~1993) without specifying an  
underlying probabilistic model, and often invalidly.\footnote{Fisher and  
Riloff rank candidates by their $t$-score (thus choosing the candidate least  
likely under the (false) null model rather than the most likely candidate  
under the true model); Resnik and Hearst apply the paired $t$-test to  
obviously non-independent samples.}  In reality, these systems rely on an  
implicit model and, as Dunning~(1993) shows, these implicit models can be  
extremely ill-fitted to the particular characteristics of language.  In addition,  
there are still further \nlp\ systems that are construed as statistical, but which  
have little to do with statistics, being merely corpus-based (for example,  
Bourigault,~1993).  While many of these approaches succeed at their chosen  
tasks, there are several disadvantages inherent in failing to build upon an  
explicit probabilistic model. 
 
Mathematics often makes dull reading and probabilistic models are no  
exception.  It is natural then to ask why a precise model is worthwhile, or  
even useful.  To demonstrate the advantages of such a model, let us consider  
a corpus-based algorithm which is not based on one.  As already mentioned  
in section~\ref{sec:cn_statistical}, 
Pustejovsky~\etal~(1993) describe a heuristic  
for bracketing compounds using a corpus.  In chapter~\ref{ch:experimental}, I  
will develop a probabilistic model of this problem in detail; however,  
solving the problem of bracketing compounds is not their main aim and the  
performance of the heuristic is apparently sufficient for their purposes.  
Recall that it works as follows: 
 
\begin{citedquote}{Pustejovsky~\etal,~1993:341} 
\ldots to bracket each compound that includes more than two nouns, we  
test whether possible subcomponents of the phrase exist on their own  
(as complete noun compounds) elsewhere in the corpus.  
\end{citedquote} 
 
Thus, given the compound \lingform{database management system}, the  
analysis is left-branching if the compound \lingform{database management}  
appears in the corpus and right-branching if the compound 
\lingform{management system} appears in the corpus.  
Note that the former is most usually the correct analysis. 
 
There are a number of notable disadvantages with this algorithm. 
\begin{enumerate} 
\item It assumes that sufficient subcomponents will appear elsewhere in the  
corpus. 
\item It assumes that no spurious subcomponents also appear. 
\item It has no way to indicate the degree of confidence associated with an  
analysis. 
\item It does not take into account the frequency of words, so that it is biased  
towards bracketing common words together. 
\end{enumerate} 
 
All these difficulties must be explicitly addressed by any system which is  
based on a probabilistic model, either in the model itself, or in the parameter  
estimation procedures.  Also, in the example above, there appears to be no  
obvious reason why \lingform{management system} should not appear  
elsewhere in the corpus while \lingform{database management} does  
appear.  The development of a probabilistic model allows the intuitive  
motivations for an algorithm to be made clear. 
 
Of greater importance is the fact that a probabilistic model defines a precise  
semantics for numerical values in the analysis procedure. 
This supports rigorous reasoning about the results of the procedure.  
Also, since the model forces us to make assumptions explicit, it allows  
reasoning about the conditions necessary for the technique in question.  At  
least in principle, we know in which circumstances the method is applicable.   
Further, if an extension to a probabilistic model is proposed, 
the denotations provided  
by the model allow us to infer under what conditions  
such an extension will work. 
 
%
All these properties
allow us to generalise in a principled way from the success of a  
given statistical language learner.  Rather than concluding only that  
following a specific procedure will solve a particular problem under just  
certain conditions, we can see (precisely) the principles by which the  
procedure works, and thus reason about the application of the same  
principles and the same knowledge resources under different circumstances. 
If natural language processing techniques are to be applied in environments  
beyond the research laboratory, it is crucial that the applicability of a  
technique to a given task can be assessed accurately and easily.  Making  
assumptions explicit is an essential part of such assessment. 
 
Yet another advantage is the natural provision within probability theory for  
representing confidence in an outcome.
The existence of confidence measures facilitates the  
combination of multiple knowledge sources.  When relevant information  
suggests conflicting analyses, taking confidence into account provides a  
natural way to resolve the question: pay more attention to information that is  
known with high confidence.  Since one of the primary motivations for using  
statistical methods is as a tool for solving the combination problem (as I  
suggested in section~\ref{sec:sn_motivations}), this is an important advantage. 
 
Finally, and this is the whole point of the present discussion, probabilistic  
models form a foundation from which data requirements might be predicted.   
Only if given precisely defined probability spaces and the mathematical  
relationships between them, can we apply statistical reasoning to predict in  
advance the training set sizes for which we can expect 
the desired performance.   
Even though probabilistic models are typically based on only approximately  
correct assumptions, these models are sufficient to produce approximately  
correct results.  
Similarly, we can expect statistical reasoning 
about sample sizes based on these  
models to also be approximately correct.  For the purpose of evaluating the  
engineering viability of a design, even approximations as coarse as orders of  
magnitude will be useful. 
 
\section{Defining Data Sparseness and Statistical  
Zeroes} 
\label{sec:dr_sparseness}
 
The goal of this chapter
is the prediction  
of data requirements.  In the section following this one
I will be putting forward some  
arguments justifying the need for predictions of this kind, but before that we  
should consider in more detail just what the problem of data sparseness is.   
Readers familiar with \sll\ or related machine learning topics could skip  
this section; 
it serves mainly to introduce the concept of data requirements at  
a technical level. 
 
\sll\ techniques are obviously dependent on their training data.  Their  
performance must vary with both the volume of training data and the quality  
of it.  To understand this relationship better, 
a more technical examination is  
necessary.  How does training data drive learning?  What does this mean for  
relating the volume of training data to the success of learning?  For this  
discussion I will assume that training is supervised.  This is a working  
assumption of the theory that I will present later, 
and is stated more formally in section~\ref{sec:dr_framework}. 
 
Consider a system based on a probabilistic model, prior to any learning  
taking place: 
it contains a set of unknowns, called parameters, that need to be  
estimated from training data.  Perhaps the model supplies default values for  
these unknowns, but if the system is going to learn from its training data, it  
will need to use that data to establish and refine the values for these  
unknowns.   
 
We call the smallest unit of data from which the learning algorithm can  
update the system's parameters a \newterm{training instance}.  As more  
training instances are examined, the learning algorithm has more information  
and can make more accurate estimates for the parameters.  In the limit, after  
a sufficiently large number of training instances have been examined, the  
algorithm will have accurate estimates of the unknown parameters and thus  
be said to have learnt from the training data. 
 
A crucial  
element of statistical learning systems is the existence of {\em conflicting}  
training data.  Different training instances will result in adjustments of the  
parameter estimates in different directions.  Sometimes this will make them  
more accurate, sometimes less; learning relies on there being more of the  
former than of the latter in the training set.  
Thus, learning is non-monotonic,  
now moving forward, now backward, shakily progressing.  
For almost any reasonable distribution of the training data, 
the eventual convergence to correct parameter settings is guaranteed 
by a result that is fundamental to statistics called the 
central limit theorem (a counter-example is the Cauchy distribution).
The more training instances are examined, the greater is the likelihood
that the parameter estimates are accurate.
 
Each training instance contributes to the estimation of the parameters,
but usually each training instance is only relevant to some of
the parameters, so not every parameter is updated with each
training instance.  For example, if one of the model parameters
is of the form $\Pr(X=x|Y=y)$, then only training instances 
for which the random variable $Y$ is equal to $y$, are relevant
to the estimation of this parameter.
It should be clear that the accuracy of the system's estimate of a  
parameter depends on the number of training instances examined that are  
{\em relevant} to that parameter.  So now we can say 
what data sparseness is.  It is the condition of having too few relevant  
training instances for a parameter estimate to be as accurate as required for  
the purpose at hand. 
 
The most obvious and dire case of data sparseness is when there are no  
training instances relevant to a parameter.  It is impossible to 
say anything about the probability $\Pr(X=x|Y=y)$ if we have
never seen $Y$ take on the value $y$.  In such cases, the model
can only provide defaults.

More common and equally  
problematic is the case when none of the relevant training instances support  
a non-zero parameter estimate.  For example, if all the training instances
with $Y=y$ do not have $X=x$, then we have no evidence that $\Pr(X=x|Y=y)$ 
is non-zero.  In other words, there is a linguistic event  
which we want to estimate the probability of, but which does not occur in the  
training set.  This unseen event may actually be impossible; that is, the  
corresponding parameter might be zero.  
But since the training set is finite, it may  
be that it has low probability and that we just haven't seen it yet.  We call  
this latter occurrence a \newterm{statistical zero}.  It is impossible for the  
learning algorithm to distinguish real zeroes from statistical zeroes.
However, it is possible to get arbitrarily high confidence that 
the value of a parameter corresponding to an unseen event is below
a given threshold if the number of training instances can be
increased arbitrarily.

Statistical zeroes cause major technical difficulties.  
For example, the mutual information between two events 
which have never been observed to co-occur is negative infinity.  
This means that if the probabilistic model adds  
together unadjusted mutual information values, the appearance of statistical  
zeroes will force it to conclude that all analyses are impossible.  
No matter what other information the model incorporates, 
it will never be able to  
distinguish any analysis as better or worse than another in the presence of  
statistical zeroes. 
 
Low counts other than zero are also unreliable.  They reflect random  
fluctuations as much as underlying relationships.  One approach is to ignore  
counts below a certain threshold.  For example, Church~\etal~(1991b)  
choose a cut-off of three, below which counts are ignored, \shortquote{to  
alleviate the fact that mutual information values are misleading when the  
frequency counts are small}~(Church~\etal,~1991b:145).  
But having seen something twice does tell us more than having  
never seen it, so this strategy wastes the precious data that we are trying to  
conserve. 
 
There are a multitude of methods to alleviate data sparseness.   
Smoothing the observed counts is a common approach. 
Recall from section~\ref{sec:sn_taggers} that a common method 
is deleted interpolation, where the results of multiple probabilistic models 
(usually a series of successively simpler ones) are added together 
(the weight given to each model being optimised empirically).  
If a probabilistic model is smoothed with
a constant function then the result is equivalent  
to adding a constant to each count.  When this constant is one, this is called  
the Add-One method.  Gale and Church~(1994) show that this can yield very  
poor estimates for English word bigrams, 
worse even than using unsmoothed counts. 
 
Another approach, called the Good-Turing method, is to adjust the observed  
counts to allow for the fact that unseen events might yet be seen if a larger  
sample were examined (see Katz,~1987).  Church and Gale~(1991) show  
that this results in estimates that very accurately match empirically measured  
values.  Katz~(1987) also suggests a back-off method where a cruder model  
is used to estimate parameters in the Good-Turing method when there is 
insufficient relevant training data. 
 
These smoothing methods are powerful instruments for tackling data  
sparseness and thus reducing the effective data requirements of a given  
probabilistic model.  But, no matter how much smoothing you do, accurate  
estimates of the parameters of a model will always depend on having some  
minimum amount of training data.  This amount may be reduced by  
smoothing, but the model will still 
require a certain amount of training data.   
 
Another approach to data sparseness is to change the model, moving to a  
less data hungry one.  To reason about this it is useful to refer to the number  
of unknown parameters as the size of a model.  It should be intuitively clear  
that larger models generally require more data, though the exact relationship  
is somewhat more complicated. 
 
It can sometimes be advantageous  
to reduce the size of a model by taking a coarser  
view of the event space.  For example, if a hidden Markov model tagger  
distinguishes~80 different part of speech tags, then it includes~6400  
transition probabilities, each specifying the probability of one tag following  
another.  But if these~80 tags can be mapped onto a smaller set of~10 tag  
types, then the model could be modified to assume that the transition  
probabilities only depend on the {\em type} of the tag.  
This revised model only  
contains~800 transition parameters and thus should require less data to train  
accurately.  The new model distinguishes less linguistic detail than the  
original, but given the same amount of training data can derive more precise  
parameter estimates.  This may or may not be an advantage, depending on  
how valid the assumption about tag types is. 
 
This idea is closely related to the principle behind
the conceptual associations proposed by Resnik and Hearst~(1993).  
In the conceptual association schemes proposed so far there have been fewer  
concepts than words (exploiting synonymy), so that concepts are coarse  
representations of words (in the strict sense of coarse used above).  Using a  
coarse view of the event space is really just another form of smoothing, one  
in which data about events that are no longer distinguished is smoothed  
together.  However now smoothing has become an integral part of model design.
 
%
%
%
This illustrates that there is an important relationship  
between model design and training data requirements.  In the next section I  
will argue that a theory describing this relationship in a way that enables us  
to predict data requirements is unavailable and highly desirable.
 
\section{The Need for a Theory of Data  
Requirements} \label{sec:dr_need} 
 
This section is dedicated to justifying 
the search for a predictive theory of data requirements. 
First I will outline the possible uses of such a theory in both scientific  
and commercial endeavours and then I will briefly relate several existing  
methods that might be expected to fulfil our needs, but which do not for one  
reason or another.  
 
\subsection*{Research prototype design}

All \sll\ systems need large electronically stored text corpora to train on.  
For research prototypes, such as those appearing so frequently now in \nlp\  
laboratories, the cost of these corpora is not currently the  
limiting factor.  While acquiring, storing and maintaining these resources  
does cost money, it is primarily electronic availability that determines the  
size of training texts.  Generally a specific research project is begun with a  
particular training corpus already chosen, and if not, then this choice is made  
quite early.  The amount of training data is more usually a design constraint  
than an experimental variable. 
 
One consequence of this is that the researcher must be careful to design the  
experiment with this constraint in mind.  In particular, the design of the  
probabilistic model needs to be done with reference to the amount of  
available training data.  For example, Magerman~(1992) advises that  
\shortquote{if there is not enough data to estimate a joint probability  
distribution function for two events which are not independent, the best  
alternative is to pick a different model}~(1992:14).  He does not say exactly  
how to determine if this is the case (primarily because it is beyond the scope  
of his discussion).  Without explicitly computing estimates
of the joint distribution  
from the available training data, determining whether there is  
sufficient data is a non-trivial task.  Similarly, Merialdo~(1994) suggests  
that \shortquote{the amount of training data needed to properly estimate the  
model increases with the number of free parameters $\dots$ In the case of  
little training data, adding reasonable constraints \dots reduces the  
number of free parameters and should improve the quality of the  
estimates}~(1994:164).   
 
An important component of model design is choosing what linguistic  
distinctions to incorporate.  All models must be sensitive to linguistic  
patterns that allow the right distinctions to be made, but they may vary in  
their sensitivity (for example, by taking a more or less coarse 
view of events).  Linguistically simplistic systems will treat two  
linguistically different inputs in the same way because their linguistic  
knowledge fails to distinguish them, and this will lead to errors.  But while  
this suggests that it is desirable to incorporate as much linguistic  
sophistication as possible, there is a problem with data requirements.   
 
Consider the effect of introducing one additional random variable on which  
all outputs are conditioned.  The number of distinctions which the system  
can make will be increased by this addition, but at the same time, the number  
of parameters in the system will increase too.  If every existing parameter is  
to be further conditioned on the new random variable, then the size of the  
model will be multiplied by the number of values available to the new  
variable. 
 
For example, Resnik and Hearst's~(1993) prepositional phrase attachment  
system takes into account the head of the noun phrase 
that is the object of the preposition, thus distinguishing  
cases that Hindle and Rooth's~(1993) did not.  As a result, the number of  
parameters required for lexical association is multiplied by the number of  
nouns, an increase in model size of~3--4 orders of magnitude. Conceptual  
association is then used to try to overcome the resulting data sparseness, but  
it fails to reduce the size of the model back down to that of Hindle and  
Rooth's. 
 
If the training set is not hand annotated, the effect of the new random  
variable is even greater than what is suggested by the increase in model size.   
This is because introducing the new variable creates additional ambiguity in  
the training set; the value of this new variable must be determined for each  
training example by whatever mechanism is used to allow unsupervised  
training.  This effectively decreases the number of training instances  
resulting in greater data requirements. 
 
As argued in Lauer~(1995a), adding linguistic distinctions to the  
probabilistic model presents a trade-off between language sensitivity and  
data sparseness.  It is a balance between poor modeling of the language and  
insufficient data for accurate statistics.  If researchers are to find a  
satisfactory compromise in this trade-off, they will need a theory of data  
requirements. 
 
Furthermore, it is important that this theory be {\em predictive}.  We want  
to use the training data itself as little as possible in establishing the  
plausibility of different models.  If a series of models is constructed and each  
applied, in a trial and error fashion, to the training data to establish whether  
there is enough data for that model, then we risk overfitting the model to the  
training data.  Sound scientific methodology requires that the experimental  
design, including the development of the probabilistic model, be done  
independently of the training data.  While we may have enough test data to  
develop on one set and evaluate on another, frequently
we need all the training data we can get.  
If the training data is used during the model development, it  
becomes less valid to generalise from the resulting model.  Researchers need  
a predictive theory of data requirements. 
 
\subsection*{Commercial applications}

While electronic texts are becoming increasingly available, they can be  
expensive.  Even unannotated text is limited and the cost of storage media  
alone guarantees that it is never free.  The cost of marked up text is  
considerably more, and the price of any reasonably large amount of richly  
annotated text will stand out on all but the largest budgets.  Many corpora  
are available only for research purposes, or at a reduced price for
such purposes, so where as
researchers may pay little attention to financing their corpus, 
these considerations  
are paramount in a commercial endeavour. 
 
Consider the decisions facing the \sll\ technology consumer, that is, the  
architect of a planned commercial \nlp\ system.  A range of \sll\ techniques  
are available to her.  For example, she might opt for a bigram tagging model  
with an expected accuracy of~94\% or for a trigram model with~97\%.  Why  
wouldn't she halve the error rate by choosing the latter?  Because these  
accuracy figures are the performance of these models in the limit.  
The trigram  
model requires more training data to reach its maximum performance than  
the bigram model, probably a lot more, and every bit of data is going to cost  
money.   
 
The situation can get much more complicated than this example.  In this  
case, exactly the same type of training data is required for the two different  
models.  If one is building a trigram tagger, it is trivial to implement a  
bigram tagger at the same time.  So one strategy would be to build both and  
see which one performs best on the data available (or better still use deleted  
interpolation to smooth the two models together).  But this relies on having a  
fixed amount of data available, an unrealistic assumption.  Instead, the  
commercial system architect will be trying to minimise the cost of data and  
still attain a certain performance level.  She needs to estimate the  
price-performance trade-offs at design time. 
 
In more complicated cases, the different \sll\ techniques will require  
different types of data.  The option to build both models is not feasible  
because this would require purchasing two complete sets of training data,  
one of each type, not to mention the doubled model development costs.   
Informed design choices require advance knowledge of the trade-off  
between volumes of training data and resulting performance. 
 
The need for a predictive theory of data requirements doesn't end with  
system design either.  Suppose the architect settles on a particular design and  
hands over to the development team.  They buy one batch of training data,  
implement the model and come up with a system that performs at something  
like the expected level.  Can they improve the performance by purchasing  
further batches of training data?  Will the improvement be worth the cost?   
A theory of data requirements could be used to make such predictions.   
Without such a theory, the team must rely on extrapolating learning curves, a  
tricky business in the presence of the noise that results from randomness  
under data sparse conditions. 
 
\subsection*{What is available already?}

Thus, for both research and commercial applications, a predictive theory of  
data requirements would be extremely useful.  In the remainder of this  
section I will briefly explain why existing techniques are not sufficient and  
thus why we require a new theory especially devoted to \sll\ data  
requirements. 
 
The field of statistics is concerned with establishing how large a sample  
needs to be in order to have a given confidence in a measurement.  
It is reasonable  
to ask why standard statistical theory doesn't provide the answers we seek.   
The reason is that language exhibits some peculiar characteristics which  
make it different from the more traditional areas of application for statistics  
such as the physical sciences. 
 
First, obvious elements of language are all discrete and non-ordered.  There  
is no way to calculate the mean of a set of words, nor even the median.  This  
type of data is called \newterm{categorical data} in statistics.  It presents  
some special properties that make standard statistical models inappropriate  
and the statistical theory for categorical data is usually relegated to more  
advanced texts specifically devoted to it (see for example, Agresti,~1990). 
 
Within categorical statistics, it is usual to make assumptions about the  
distributions of events (that is, statistics are parametric).  The distributions  
found in language often violate these assumptions quite dramatically (see for  
example Church and Gale,~1995). Furthermore, the probability  
distributions of words are typically severely non-uniform.  Zipf's law states  
that the probability of a word is inversely proportional to its frequency rank,  
so a small number of frequent words carry much of the probability  
distribution, while a vast host of rarer words are left to carry the remainder.   
 
These properties make traditional statistical reasoning difficult.  For many  
promising statistical methods, one often finds that language exhibits exactly  
those conditions under which the method is inapplicable.  While there may  
be lesser-known statistical methods that are appropriate, we still await their  
transfer from the statistical disciplines to computational linguistics.  
If this thesis encourages or facilitates that transfer, 
then it has met one of its goals. 
 
Modern statistics does, however, provide a technique which allows the  
calculation of error bounds on any computable statistic called the \newterm{ 
bootstrap method} (see Efron and Tibshirani,~1991, for an  
introduction).
The method makes no distributional assumptions,  
relying on computation power to generate large numbers of samples from  
the training data.  Unfortunately, the resulting variation estimates are valid  
only for the available training data.  The method cannot be used to predict  
the effects of adding more data, nor used to predict requirements prior to  
access to the training data. 
 
 
Somewhat closer to computational linguistics than theoretical statistics is the  
field of machine learning; there are results in the area of probably  
approximately correct learning (or \acronym{pac}-learning) that are  
relevant.  A typical example is the work by Abe and Warmuth~(1992).
Their work concerns the estimation of parameters in any  
probabilistic finite state automaton.\footnote{The set of probabilistic finite  
state automata contains all the models and training algorithms considered  
later in this chapter and is likely to contain any proposed probabilistic  
models of language.}  Their result bounds the sample complexity of this  
estimation task, that is the number of training instances required to achieve a  
given accuracy with a given confidence.   
 
The result makes few assumptions, making it suitable for application to  
language models, and it doesn't even require that the language data obey the  
assumptions of the probabilistic model.\footnote{A weakness of the results  
appearing later in this chapter is that they do inherit the assumptions of the  
probabilistic model; but if the probabilistic model is to be useful, it must be  
approximately correct in any case.}  It also allows arbitrary constraints on  
the model, such as would be entailed by requiring each word to be tagged  
only with those parts of speech listed for it in the lexicon. 
 
The difficulties with this result are three-fold.  First, Abe and Warmuth  
define accuracy to be relative entropy between the actual distribution and  
that defined by the trained probabilistic automaton.  A theory of data  
requirements for \sll\ needs to predict the accuracy of the language analysis  
system after training.  While a low relative entropy makes accurate  
predictions likely, it is possible to have arbitrarily large 
differences between  
two relative entropies, but no resulting difference in the error rate of the  
language analysis system. 
 
Second, the bound allows for arbitrarily large confidence in the accuracy of  
the trained model.  The bound guarantees to be correct in $(100-\alpha)$\% of  
training runs where $\alpha$ is arbitrary.  In practice, we don't need to be  
highly confident that a given performance will be attained; an approximate  
estimate of what to expect is more valuable.  One could say that Abe and  
Warmuth give a worst-case bound, while an average-case bound is of more  
interest. 
 
The third problem, the crucial one, arises because of the first two.  The  
bound that they arrive at is extremely weak.  It will never predict that less  
than~64$t$ training instances are needed, where $t$ is the size of the model.   
In fact, for reasonable choice of other parameters (maximum state sequence  
length, desired accuracy and desired confidence level) the bound will be  
very much higher (these arguments are derived from 
Lemma 3.1 of Abe and Warmuth,~1992:219).
However, their main purpose in establishing the  
bound is to show that the sample complexity is polynomial rather than  
exponential, and for this purpose it is quite a strong bound.  For the purpose  
of predicting data requirements in real \sll\ systems it is far too weak. 
 
Where statistics and machine learning research run out of steam, intuition  
and heuristic take over.  The final technique I will mention here is an  
intuitive approximation with no firm theoretical basis.  The reason for  
including it is that it appears to be used as a rough guide in practice.  
Weischedel~\etal~(1993) state  
that one \shortquote{rule of thumb suggests that the training set needs to be  
large enough to contain on average ten instances of each type of tag  
sequence that occurs}~(1993:363).  In other words, the rule
suggests an expected data requirement of~10$t$, 
where $t$ is the size of the model.  As we shall see in  
section~\ref{sec:dr_tour}, this prediction does not appear to be borne out by  
experiments.  One reason for this is that model size isn't the only factor, as  
Weischedel~\etal\ also point out. 
 
So theories that are useful for predicting data requirements are not only  
necessary for building \sll\ systems, but are currently unavailable.  Despite  
this need, very little attention has been paid to the problem.\footnote{See de  
Haan~(1992) for an investigation of sample sizes for linguistic studies.}  In  
fact, training set sizes have generally received little investigation.   
Weischedel~\etal\ state that they \shortquote{are not aware of any other  
published studies documenting empirically the impact of training set size on  
performance}~(1993:364).\footnote{A detailed study of the learning rate of  
a hidden Markov model tagger has been given since then by Merialdo~(1994).} 
 

In summary, training data will always be limited and thus reasoning about  
training data requirements is an important part of system design.  At the  
moment, the field lacks a predictive theory of data requirements.  In the  
remainder of this chapter I will describe some initial results in the search  
for such a theory. 
 
\section{A Tour of Training Set Sizes}   
\label{sec:dr_tour} 
 
Statistical language learning systems are varied in both goals and  
construction and virtually the only thing common to all is the use of a  
training corpus.  Any theory which captures the data requirements  
characteristics of all of them will have to be quite sophisticated, and may  
need to be composed of a number of heterogeneous parts.   
 
In this section, I will analyse six \sll\ systems with particular  
reference to their training set sizes.  
These systems are typical subjects of the kind of theory required.   
Each can be seen as a data point that needs to be explained by the theory.   
They are: a sense disambiguator, two taggers, two prepositional 
phrase attachment systems and a compound noun parser.  
Putting them side-by-side should help make the domain of the theory more  
concrete. 
 
My analyses will involve some guesswork.  Even in those systems that are  
based on explicit probabilistic models, it is rare that sufficient information is  
given to evaluate the volume of training data used.  Most commonly the size  
of the training corpus is given in words rather than in number of training  
instances.  In most cases I have managed to arrive at rough estimates of  
important values with a little detective work, putting together facts from  
various parts of the paper describing a system.  I have not included any  
probabilistic grammar systems because it is difficult to find published  
reports that give enough detail to allow even this kind of guesswork. 
 
For each system I will pay attention to the size of the probabilistic model  
and the number of training instances.  Often the models contain two or more  
qualitatively different types of parameters.  A bigram tagger is the canonical  
example: it has transition probabilities and word emit probabilities.   
Parameters of each type are estimated from different types of training  
instances.  Thus, the training instances fall into classes, effectively forming  
multiple distinct training sets.  For a bigram tagger, pairs of consecutive tags  
are used to estimate the transition parameters while the word emit  
parameters are estimated from pairs composed of a word and its tag. 
 
Since each type of parameter is estimated independently from a distinct  
training set, the degree of data sparseness affecting the different types of  
parameters will vary.  If there is a large amount of training data of one type,  
estimates of the corresponding type of parameter will be very accurate.  In  
evaluating the degree of data sparseness it is a reasonable working  
assumption to ignore parameter types with large training sets.  In the  
analyses below, those parameter types that I have judiciously discounted for  
this reason will be noted.  Also, where no probabilistic model is given I will  
attempt to reconstruct a plausible one that approximates the statistics  
computed by the system. 
 
\subsection*{Six SLL systems}

Yarowsky's~(1992) sense disambiguation system (refer back to  
section~\ref{sec:sn_supervised} for more detail) uses a~101 word window  
of co-occurrences.  The model maintains estimates of the relative likelihood  
of words in each category of a thesaurus 
appearing within~50 words  
of every other word.  It chooses a sense by adding together the logarithms of  
probabilities $\Pr( \mbox{Cat} | w)$ for each word, $w$, in the window,  
which is equivalent to assuming that the words in the window
are all independent random variables.  To arrive at  
the estimates, it relies on the following application of Bayes' rule. 
\begin{equation} 
\Pr( \mbox{Cat} | w) = \frac{\Pr(w | \mbox{Cat})  
	\Pr(\mbox{Cat})}{\Pr(w)} 
\end{equation} 
Since the probabilities $\Pr(\mbox{Cat})$ are assumed constant, no data is  
required to estimate them.  This leaves two types of parameters, those of the  
form $\Pr(w)$ (word probabilities) and those of the form $\Pr(w |  
\mbox{Cat})$ (co-occurrence probabilities).  The former type cancels out  
from the probabilistic model, but the smoothing process uses them (see  
Gale~\etal,~1994, for details of the smoothing) so we still need to take  
account of them in evaluating the degree of data sparseness.   
 
I will guess that the vocabulary size is on the order of~100 thousand since it  
is derived by taking all words from Grolier's encyclopedia.\footnote{Some  
stemming is performed, so it is the number of stems in the vocabulary that  
we want.}  Thus there are about~100 thousand word parameters.  Since  
there are~1042 categories (the thesaurus used is the 1977 Roget's),
there are about~104 million co-occurrence parameters.
 
The training corpus is Grolier's encyclopedia, which contains on the order  
of~10 million words.  Each of these is used as a training instance for  
estimating the word probabilities (so there are about~100 training instances  
per parameter for these).  Further, each of the nouns appearing in the corpus  
provides~100 (ambiguous) training instances for estimating the  
co-occurrence probabilities.  Assuming that about every fifth word is a noun,  
there are about~200 million training instances for these (and thus around~2  
training instances per parameter).  Since the latter type of parameter has far  
less data, I will ignore the former type in the summary below.  The average  
accuracy reported is~92\%.  More recent work has shown that similar sense  
disambiguation tasks can be performed via supervised training algorithms  
with an accuracy of~96\% using the same (or less) context (see column~5 of  
the results in Yarowsky,~1995:194), so presumably human performance  
would be at least as good as this given the same test conditions. 
 
Weischedel~\etal~(1993) describe \progname{post}.  It is a trigram Markov  
model part of speech tagger (refer back to section~\ref{sec:sn_taggers}) and  
thus maintains two sets of parameters: word emit probabilities ($\Pr(w | t)$)  
and transition probabilities ($\Pr(t_3 | t_1, t_2)$).  To choose a tag for a  
word, it considers all possible tag sequences for the sentence and computes  
their probabilities by multiplying the estimates for each tag triple and     
word-tag pair in the sequence.   
 
It uses~47 part of speech tags, so the number of free transition parameters is  
a little over~100 thousand 
($47 \times 47 \times 46$).\footnote{The requirement 
that all probabilities in  
a given distribution must sum to~1 means there is one less free parameter  
than there are probabilities to be estimated in each distribution.} 
They only allow a  
word to be tagged with a tag which has been observed to be possible for that  
word.  Thus, there are far fewer than~47 free word emit parameters per  
word.  Since \shortquote{the average number of parts of speech per word  
[token] is approximately two}~(1993:361), we can guess that each word  
type has an average of roughly two possible tags and that therefore there are
only two free parameters per word type.  Given that the text is from a  
restricted domain (\publicationname{Wall Street Journal}), I will guess that  
the vocabulary is about~10 thousand words. 
 
The training data is one million words from the University of Pennsylvania  
Treebank.  Each of these yields one training instance for each of the word  
emit and transition parameter sets.  Since there are far fewer of the former, I  
will ignore them in the summary below.  For the transition parameters, there  
are about~10 training instances per free parameter, and the accuracy is  
reported to be~97\%, which is approximately the accuracy of human taggers  
using the whole context.  A second data point for the theory is provided by  
their study, since they also report the accuracy of the system when trained on  
only~64 thousand words of text.  This is only~0.6 training instances per  
parameter with a reported accuracy of~96\%. 
 
Merialdo's~(1994) tagger uses an almost identical model, but has a larger  
tag set of~76 parts of speech.  Thus, there are about~430 thousand transition  
parameters to be estimated.  Again, the words are restricted only to take tags  
which they have been observed to take in training, and the number of tags  
per word token is less than two.  Thus, there are relatively few word emit  
parameters and these can be ignored on the same grounds as above.  The  
training data is only ever measured in sentences, but since the first~2,000 
sentences contain \shortquote{less than~50,000 words}~(Merialdo,~1994:162), 
I will assume that the training set of~40 thousand sentences contains 
about one million words.  This leads to an estimate of~2.3 training 
instances per parameter, which is sufficient to achieve an accuracy of~97\%.  
Using only the first 2,000 sentences (and thus~0.1 training instances per 
parameter) yielded~95\% accuracy.

Hindle and Rooth's~(1993) system to syntactically disambiguate  
prepositional phrase attachments (refer back to  
section~\ref{sec:sn_specialised}) is based on a probabilistic model of  
associations between prepositions and the nouns or verbs that they attach to.   
The probability of a nominal attachment is represented by parameters of the  
form $\Pr(p | n)$, where $p$ is the preposition and $n$ the noun (noun  
associations). Verbal attachment probabilities are computed from two sets of  
parameters: $\Pr(p | v)$ (verb associations) and $\Pr(\nullsym | n)$ (null  
associations), where $v$ is the verb and $n$ is the head noun of the verb's  
object. 
 
The training data is extracted from~13 million words of newswire text by a  
partial parser.  I will guess that there are about~50 thousand nouns,~10  
thousands verbs and~100 prepositions in the vocabulary of this text.   
Therefore, there are around~5 million noun associations,~1 million verb  
associations and~50 thousand null associations.  The parser extracts  
around~1.9 million noun phrases which are not the object of a verb, so there  
is a great quantity of data for estimating null associations and I will therefore  
ignore these.  It also extracts over~750 thousand noun-preposition pairs  
(used to estimate the noun associations) including over~220 thousand  
verb-object-preposition triples (used to estimate the verb associations).  A  
process of cyclic refinement is used to generate training data from  
ambiguous examples.   
 
Both these kinds of association have about the same number of training  
instances per parameter, so for simplicity I will combine these: around~6  
million parameters are estimated from around~1 million training instances  
(0.2 training instances per parameter).  The resulting accuracy is close  
to~80\%, with human subjects achieving~85--88\% on the same information. 
 
Resnik and Hearst~(1993) aim to enhance Hindle and Rooth's~(1993) work  
by incorporating information about the head noun of the prepositional  
phrase in question (refer to section~\ref{sec:sn_conceptual}).  There is no  
probabilistic model given; instead a set of frequency weighted mutual  
information scores is submitted to a $t$-test to distinguish nominal from  
verbal attachment.  A reconstruction of the probabilistic model would  
involve parameters of the form $\Pr(c_1, p, c_2)$ and $\Pr(v, p, c_2)$  
(corresponding to the mutual information scores between these elements),
where $v$ is a verb, $p$ a preposition and $c$ a concept.   
 
WordNet contains over~70 thousand concepts, but the $t$-test and  
frequency weightings employed by Resnik and Hearst ensure that only those  
concepts high up in the taxonomy will be used in making the attachment  
decision.  I will guess that one thousand concepts is a more reasonable figure  
(this is the number of Roget's categories, for example) and leave the number  
of prepositions at~100.  Since the corpus is smaller I will assume
5 thousand verbs, which leads to
around~600 million parameters (100 million concept associations and
500 million verb associations).  Their training  
corpus \shortquote{is an order of magnitude smaller than the one used by  
Hindle and Rooth}~(Resnik and Hearst,~1993:61), but is parsed, avoiding  
the need for cyclic refinement.  Assuming that the distribution of  
prepositional phrases is similar in both corpora, the number of training  
instances is therefore about~100 thousand.  This is a ratio of 
two training instances per 10 thousand parameters. 
 
Even given the additional information about the head noun of the  
prepositional phrase, the accuracy reported fails to improve on that of  
Hindle and Rooth, being~78\%.  It is seems likely that insufficient training  
data is the cause of this shortfall.  Dras and Lauer~(1993) report that humans  
achieve up~92\% accuracy given the same information (that is, the verb,  
head of the object, preposition and head of the prepositional noun phrase). 
 
Finally, Lauer~(1994) describes a system for syntactically analysing  
compound nouns (details of the probabilistic model and results of this  
experiment are given below in the first part of  
chapter~\ref{ch:experimental}).  Two word compounds extracted from  
Grolier's encyclopedia are used to estimate an association between every  
pair of thesaurus categories and the results used to  
select a bracketing for three word compounds.  The associations measure the  
probability of a word in one thesaurus category pre-modifying a word in  
another category.  Thus, the free parameters are of the form $\Pr(t_1  
\rightarrow t_2 | t_2)$.  Since there are~1043 categories (the
thesaurus used is the 1911 Roget's), the model  
size is a little over one million free parameters.  
The training set of two word  
compounds consists of just over~24 thousand noun pairs, giving only~2.4
training instances per 100 parameters.  Lauer and Dras~(1994) report the
accuracy to be~77\% (various related results appear in 
chapter~\ref{ch:experimental} below).  Section~\ref{sec:cy_human}
below reports an experiment in which 
human subjects, given only the compounds, get an average of~81.5\%
accuracy.

\subsection*{Summary}
 
\begin{table*}  
\begin{center} 
\begin{tabular}{|c|c|c|c|c|c|c|}\hline  
System & Training & $m$ & $t$ & $m:t$ & 
		Accuracy & Humans \protect\footnotemark \\ 
\hline  
\hline 
Weischedel~\etal\ tags & Supervised & 1M & 100k & 10 & 97\%  
		& ($\leq$ 97\%) \\ 
Merialdo tags & Supervised & 1M & 430k & 2.3 & 97\%  
		&  ($\leq$ 97\%) \\ 
Yarowsky senses & Unsupervised & 200M & 104M & 2 & 92\%  
		& ($\geq$ 96\%) \\ 
Weischedel~\etal\ tags & Supervised & 64k & 100k & 0.6 & 96\%  
		& ($\leq$ 97\%) \\ 
Hindle \& Rooth PPs & Auto-supervised & 6M & 1M & 0.2 & 80\%  
		& 85--88\% \\ 
Merialdo tags & Supervised & 50k & 430k & 0.1 & 95\%  
		&  ($\leq$ 97\%) \\ 
Lauer CNs & Auto-supervised & 24k & 1M & 0.02 & 77\%  
		& 82\% \\ 
Resnik \& Hearst PPs & Supervised & 100k & 600M & 0.0002  
		& 78\% & ($\leq$ 92\%) \\ 
\hline 
\end{tabular} 
\end{center} 
\caption{Summary of a tour of training set sizes for statistical \nlp\ systems} 
\label{tb:dr_tour_summary} 
\end{table*} 
\footnotetext{Brackets indicate measured on different data 
and/or under different conditions.} 
 
Table~\ref{tb:dr_tour_summary} shows a summary of the above systems.   
The column labelled $m$ records the number of training instances, the  
column labelled $t$ shows the size of the model (number of free parameters)  
and the last two columns show the performance results of the system and  
that of humans given the same information.  Also listed is
the type of training used.  While the problems addressed  
by each of these systems are of differing difficulties, a correlation between  
success and data sparseness (as measured by the ratio $m:t$) is evident.   
 
It is also clear that far less than ten training instances per parameter are
necessary to achieve accurate performance.  As Weischedel~\etal~(1993)  
point out, there are many three tag sequences that never occur in their  
training data.  Thus, there are more training instances available for  
estimating the probabilities of the three tag sequences that do appear.  The  
fact that there is no data available for the non-occurring sequences does not  
affect performance because exactly these sequences do not occur in testing  
either.   
 
This characteristic is likely to be observed in a wide range of linguistic  
phenomenon because of the severely non-uniform distributions that occur in  
language (as mentioned in the last section).  This alone makes it clear that  
the size of the model is not the only factor in determining data requirements.   
In section~\ref{sec:dr_simulations} below I will consider the problem of    
non-uniform distributions further. 
 
Another observation is that all \sll\ systems have optimal performance rates  
that are less than~100\%.  Human judges cannot perform most of these tasks  
with complete accuracy when given only the information available to the  
system.  Even if the probabilistic model is perfect, there is still 
an upper limit to the performance which is independent of the volume of  
training data.  The test examples appear to exhibit dependence on context  
that is unavailable and so systems limited to 
the same contexts cannot perform with~100\% accuracy.  Also, the  
models make faulty assumptions, so their performance will be limited even  
further, regardless of how accurate their parameter estimates are. 
 
In the theory of data requirements to be developed below, the optimal error  
rate will play an important role.  System accuracy will be measured not in  
absolute terms but rather relative to this optimal accuracy.  The work  
presented forms only the beginnings of a complete theory of the subject and  
is too simplistic to explain all of the observations made during the tour  
above.  While its assumptions are limiting enough to prevent it being  
directly applicable to even one of the above systems, it does expose issues  
that are important to all of them. 
 
\section{A Simple Framework for Statistical NLP} 
\label{sec:dr_framework} 
 
In the previous section I began with the observation that \sll\ systems are 
widely varied in both goals and construction.  But if a theory of data  
requirements is going to provide insights into many of them, then it must  
begin on common ground.  In this section I will construct a framework for  
viewing \sll\ systems based on their commonalities.  Results appearing in  
later sections will concern one simple statistical language learning model  
within this framework.  The level of mathematical detail given is necessary  
to support those results, but, before the formal mathematics,
I will informally sketch the framework's main elements. 
 
\subsection*{Informal sketch}

The first commonality amongst all \sll\ systems is that they are designed to  
make choices on the basis of certain inputs.
The input  
information must have some form of representation, typically a string of  
words, and be of finite (if arbitrary) size.  In general,
the input representation will not fully represent all possible
linguistic events.  Instead, it will divide the space
of linguistic possibilities into a set of different cases.
The set of different cases distinguished by the input representation
will be called the \newterm{bins} in the formal framework defined
below.

Since language is highly context  
dependent, it is unlikely that the input representation will 
always provide sufficient information to determine the analysis uniquely. 
For example many sentences given to a tagger will have different readings  
in different contexts.  If these readings involve different parts of speech  
assignments, then the tag sequence will be dependent on more than just the  
words in the sentence.  Since the tagger is only given these words, it has  
insufficient contextual information. 
 
Another way to look at it is this: consider all of the sentences ever written.
By this I mean sentence tokens, so two occurrences of  
the same sentence count as two distinct elements.  Each element of  
this set has, in principle, a unique analysis (parts of speech assignment,  
parse, etc).  By looking at only the words in the sentence, the analyser is  
collapsing multiple distinct linguistic events into one representation.  
It has a  
coarse view of the linguistic possibilities.  No matter how sophisticated we  
make natural language understanding systems, there will always be  
contextual factors that the system does not account for.  Therefore there will  
always be linguistically distinct events that are treated identically by the  
system.  The degree of coarseness is determined by the input representation.   
Thus, the input representation plays an important role, defining the set of  
test cases that can be distinguished.  
When this set is larger, the system can be more  
linguistically sensitive, but it also has more to learn. 
 
All \sll\ systems have an output representation too.  This representation  
defines the set of choices available to the system.  To be successful, there  
must be a correlation between the input set and the output set.  The inputs  
might not fully determine the outputs, but they must partially determine  
them.  The goal of the system is to take test cases and predict analyses.  The  
system must compute a function from its input set to its output set, usually  
by computing probabilities for each possible output.  However,  
regardless of how the function is computed, the accuracy of the system  
depends only on the particular function produced.  Once each element of the  
input set is mapped to the analysis it should receive, the details of how the  
mapping was performed do not bear on the accuracy.  Therefore highly  
accurate parameter estimates are only valuable to the extent that they  
contribute to making correct predictions. 
 
So far we have included an input set, an output set and a function from  
inputs to outputs representing the learnt analysis procedure, with an  
associated accuracy.  We still need a training algorithm.  The goal of such an  
algorithm is the acquisition of the analysis function.  Given some training  
data, this algorithm will furnish parameter estimates,
that is, the information necessary to turn the  
probabilistic model into an analysis function.  If  
given very little data, we expect the resulting analysis function to be  
inaccurate.  As more data is provided, the analysis function should improve.   
If the training algorithm is well-behaved, then given unlimited data it should  
eventually result in the analysis function converging to an optimal accuracy.   
Our main interest here is the rate of this convergence and we will continue  
that investigation once all these elements have been made formal. 
 
\subsection*{Formal framework} 

Formally, there is a set of linguistic events $\Omega$ from which every test  
instance will be drawn, the universe of language.
Let $V$ be a  
countable set of values that we would like to assign to any given linguistic  
input.  This defines the range of possible answers that the analyser can make.   
Let $J:\Omega \rightarrow V$ be the random variable describing the  
distribution of values that linguistic events take on.  We also require a  
countable set of inputs, $B$, to use in selecting a value for a given linguistic  
event.  I will refer to each element of $B$ as a {\em bin}.  Let $I:\Omega  
\rightarrow B$ be the random variable describing the distribution of bins  
into which linguistic events fall.\footnote{I assume 
without loss of generality  
that $|B| > 0$, $|V| > 1$ and $(\forall b \in B) \Pr(I=b) > 0$ 
in all following results.} 
 
The task of the system is to choose which value is the most likely given only  
the available inputs.  Therefore, it requires an analysis function taking inputs  
and yielding values. 
 
\begin{definition}[Analysis Function] 
An {\em analysis function\/} is a function $A:B \rightarrow V$, 
used by an \sll\  
system to predict analyses in the basis of available information. 
\end{definition} 
 
The task of the learning algorithm is to acquire such a function by  
computing statistics on the training set.  Putting these components together,  
we can define a statistical processor. 
 
\begin{definition}[Statistical Processor] 
A {\em statistical processor\/}
is a tuple $ \langle \Omega , B , V , A \rangle $,  
where:  
\begin{itemize}  
\item $\Omega$ is the set of all possible linguistic events; 
\item $B$ and $V$ are countable sets, the bins and values respectively; and 
\item $A$ is the trained analysis function. 
\end{itemize}  
\end{definition} 
 
Amongst all such statistical processors we are interested in 
those using a probabilistic model to rank possible analyses. 
 
\begin{definition}[Probabilistic Analyser] 
A {\em probabilistic analyser\/}
is a statistical processor which computes a function  
$p:B \times V \rightarrow [0,1]$ such that $\sum_{v \in V} p(b,v) = 1$ for  
all $b \in B$ and then computes $A$ as:  
\begin{equation}  
A(b) = \argmax{v \in V} p(b,v)  
\label{eq:dr_framework_pa} 
\end{equation} 
\end{definition} 
 
The problem of acquiring $A$ is thus transformed into one of estimating the  
function $p$ using the training corpus. 
The intention is to view $p(b,v)$ as an  
estimate of $\Pr(J=v | I=b)$.  By trivial construction, every  
analysis function corresponds to some function $p$ of a probabilistic  
analyser, so for every statistical processor there is an 
equivalent probabilistic  
analyser.  For probabilistic analysers to be well-defined, the operator  
$\argmax{}$ must return a unique value.  In the reasoning to follow, I will  
assume that when more than one element is greater than or equal to all  
others, $\argmax{}$ returns any one of these at random. 
 
While the framework so far is sufficiently abstract to capture most \sll\  
systems, there is one important limitation.  Currently it does not permit the  
system to refuse to answer.  It requires~100\% coverage of the input  
set.\footnote{This entails~100\% recall in information retrieval terms.   
However, coverage and recall are generally not the same and in  
particular~100\% recall does not entail~100\% coverage.  This fact  
sometimes escapes attention in computational linguistics.}  One could, in  
principle, augment the definition to include in $V$ a value representing  
refusal to answer and then extend all the results below to predicting accuracy  
and coverage.  I leave this for future work. 
 
\subsection*{Formal accuracy} 

Once trained, a statistical processor can be put to use analysing test cases.   
Test cases are drawn from $\Omega$ according to the distributions of the  
random variables $I$ and $J$, so we can express the expected accuracy of an  
analysis function $A$ in terms of these.
First we define correctness as whether or not the  
prediction made by the function matches that observed for the test case. 
\begin{definition}[Correctness] 
The {\em correctness\/} of prediction made by a statistical processor using an  
analysis function $A$ is a random variable $C_A$ defined by the equation: 
\begin{equation} 
C_A \stackrel{\rm def}{=} \delta(J, A(I))  
\end{equation} 
where $\delta(x, y)$ is~1 if $x = y$ and~0 otherwise. 
\end{definition} 
 
The expected accuracy is then the mathematical expectation of the  
correctness. 
\begin{definition}[Expected Accuracy] 
The {\em expected accuracy\/} of an analysis function $A$ is given by 
\begin{eqnarray} 
\alpha(A) & \stackrel{\rm def}{=} & \mbox{E} [C_A] \nonumber \\ 
	& = & \sum_{b \in B;v \in V} \Pr(I=b, J=v) \delta(v, A(b))  
\nonumber \\ 
	& = & \sum_{b \in B} \Pr(I=b, J=A(b)) \nonumber \\ 
	& = & \sum_{b \in B} \Pr(I=b) \Pr(J=A(b) | I=b) 
\label{eq:dr_framework_ea} 
\end{eqnarray} 
\end{definition} 
This is the probability of the analyser being correct on a randomly selected  
element of $\Omega$. 
 
For any non-trivial statistical processor the input set used cannot  
perfectly represent the entire linguistic event space and so 
in general there exist values  
$v_{1}, v_{2} \in V$, for which both $\Pr(J=v_{1}, I=b) > 0$ and  
$\Pr(J=v_{2}, I=b) > 0$ for some $b \in B$.   Suppose without loss of  
generality that $A(b) = v_{1}$.  The analyser will be in error with  
probability at least $\Pr(J=v_{2}, I=b)$.  This means, as we have seen  
above, that typically a statistical language learner cannot achieve~100\%  
accuracy, even with unlimited training data.  This is the root of an interesting  
problem in \sll\ because in practice, no matter how inaccurate a trained  
statistical processor is, the inaccuracy may be due to the imperfect  
representation of $\Omega$ by $B$.\footnote{Unless a more accurate  
statistical processor based on the same input set already exists.}  If this is the  
case, acquiring more training data will have no effect. 
 
Theoretically though, it is simple to define the maximum possible  
performance for any statistical processor on unlimited data, given a         
well-behaved training algorithm.  I will call this the optimal accuracy. 
\begin{definition}[Optimal Accuracy] 
The {\em optimal accuracy\/}
of a statistical processor is dependent only on the bins $B$,
values $V$ and their distributions $I$ and $J$, and is defined by 
\begin{equation} 
\alphaopt \stackrel{\rm def}{=} \max_{A':B \rightarrow V} \alpha(A') 
\label{eq:dr_framework_oa} 
\end{equation} 
\end{definition} 
Given $B$ and $V$, $\alphaopt$ is the greatest possible accuracy rate we  
can expect during testing.  Any probabilistic analyser that achieves an  
accuracy close to this is not going to benefit 
significantly from further training data. 
 
\begin{definition}[Optimal Analysis Function] 
An {\em optimal analysis function\/} is any analysis function $A_{\opt}$ with  
$\alpha(A_{\opt}) = \alphaopt$. 
\end{definition} 
By inspection, $A$ is an optimal analysis function if and only if the  
following constraint holds: 
\begin{equation} 
(\forall b \in B) \Pr(J=A(b) | I=b) = \max_{v \in V} \Pr(J=v | I=b) 
\label{eq:dr_framework_oaf} 
\end{equation} 

\subsection*{Errors}
 
The shortfall between the expected accuracy and~100\% accuracy
(the optimal error rate) reflects the imperfection of $B$ and $V$ 
as representations of the  
linguistic universe $\Omega$.  However, the framework does not distinguish  
whether this error rate arises because of representational limits of $B$ and  
$V$ or because the random variables $I$ and $J$ do not accurately follow  
the true distribution of linguistic phenomena.  Poor modelling and noise  
can be treated identically as far as the framework is concerned. 
 
Consider the tagger example once more.  Suppose that the words have been  
incorrectly transcribed occasionally, so that the input to the tagger contains  
words not actually present in the original linguistic utterance.  This noise  
will cause errors.  Also, as we've seen above, extra-sentential context can  
make an unlikely reading the preferred one, resulting in an optimally trained  
tagger making an error.  Both these kinds of errors are treated identically by  
the framework and all the results below hold whether the optimal error rate  
arises from noise or violated model assumptions.  There is no need to extend  
the framework to handle noisy conditions. 
 
The framework does distinguish two error components: 
\begin{itemize} 
\item {\em optimal errors\/} caused by modelling limitations or by noise; and 
\item {\em statistical errors\/} caused by insufficient training data. 
\end{itemize} 
 
The importance of distinguishing these two components should be obvious.   
An illustration of them is provided by the errors that Brent's~(1993) verb  
frame acquisition system makes.   
\begin{citedquote}{Brent,~1993:256} 
Three of the five \dots violate the model's assumptions$\dots$and  
highlight the limitations of the model. The remaining [two] \dots would be  
much rarer in more mundane sources of text \dots than in the diverse  
Brown Corpus. 
\end{citedquote} 
Because of the great variety of language found in the Brown Corpus,  the  
amount of training data required to overcome statistical fluctuations is very  
large.  It is statistical errors that are the concern of a theory of data  
requirements. 
 
In the results below, the goal will be to relate the expected accuracy to the  
volume of training data.  In doing so, we will be evaluating the expected  
accuracy {\em relative} to the optimal accuracy.  It is only the error  
component due to statistical fluctuation that will be predicted.  This raises a  
practical issue: How do we predict the optimal accuracy?  Unless large  
volumes of manually annotated data exist, measuring $\alphaopt$ for a  
proposed statistical processor presents a difficult challenge.   
 
Hindle and Rooth~(1993) have used human judges to measure the  
context-dependence of prepositional phrase attachment.  Judges were given  
only the preposition and the preceding verb and noun, just the information  
available to their statistical processor.  The judges could only perform the  
attachment correctly in around~87\% of cases.  If we assume that the judges  
incorrectly analysed the remaining~13\% of cases because these cases  
depended on knowledge of the wider context, then any statistical learning  
algorithm based on the same input information cannot do better than~87\%.   
Of course, if there is insufficient training data the system may do  
considerably worse because of statistical errors. 
 
Unfortunately, this approach to estimating $\alphaopt$ is expensive to apply  
and makes a number of questionable psychological assumptions.  For  
example, it assumes that humans can accurately reproduce parts of their  
language analysis behaviour on command.  It may also suffer when  
representational aspects of the analysis task cannot be explained easily to  
experimental subjects.  A worthwhile goal for future research is to establish  
a statistical method for estimating or bounding $\alphaopt$ using small  
amounts of language data.   
 
\subsection*{Training} 

There is still one vital component missing from the formal framework and  
that is training data.  After all this is the requirement that the theory is  
designed to predict.  I will assume hereafter that training is supervised, and  
that training data and testing data come from the same  
distributions.  Note that if part of the optimal error rate is due to  
noise, this assumption implies that training and 
testing data are subject to the same noise.   
\begin{assumption}[Training and Test Set Equivalence] 
Test and training data are drawn from the same distributions. 
\end{assumption} 
Unsupervised learning algorithms are usually driven by assumptions that are  
not made explicit and therefore represent a substantially more difficult  
problem. 
\begin{assumption}[Supervised Training] 
All training data is annotated with values, allowing supervised learning. 
\end{assumption} 
 
Training proceeds on the basis of training instances extracted from a corpus,  
each of which contributes to the statistical measures collected.  Formally, I  
will define a training set as a collection of pairs of bins and values. 
\begin{definition}[Training Set] 
A {\em training set\/} of $m$ instances 
is an element of $(B \times V)^{m}$  
where each pair $(b,v)$ is sampled according to the random variables $I$  
and $J$ from $\Omega$. 
\end{definition} 
It follows from this definition that each training instance is independent of  
all the others.
While this assumption is made almost universally by \sll\ models,
it is often violated.  Work on adjusting models to allow for such
violations is reported in Church and Gale~(1995).   

\begin{assumption}[Training Sample Independence] 
Training instances are statistically independent of one another. 
\end{assumption} 
When a pair from a training set $c$ has bin $b$ as its first component, we say  
that the training instance has {\em fallen into\/} bin $b$.  
The corresponding value $v$ (the second component of the pair) is 
also said to have fallen into bin $b$. 
 
Recall that a probabilistic analyser includes a function $p:B \times V  
\rightarrow [0,1]$.  There are a variety of methods by which an appropriate  
function $p$ can be estimated from the training set.  Regardless of the  
learning algorithm used, each possible training corpus, $c$, results in the  
acquisition of some function, $\pfromc$, and consequently in an analysis  
function $A_c$.  Our aim is to explore the dependence of the expected  
accuracy $\alpha(A_c)$ on the magnitude of $m$, the size of the 
training set.    
 
If the learning algorithm is well-behaved in the following sense, then the  
expected accuracy of the \sll\ system is guaranteed to converge to the  
optimal accuracy rate as larger amounts of training data are used.  The proof  
of this is straight-forward.   
\begin{definition}[Well-behaved Learner] 
A learning algorithm is defined to be {\em well-behaved\/} when 
\begin{equation} 
(\forall b \in B) (\forall v \in V)  
	\lim_{m \rightarrow \infty} p(b,v) = Pr(J=v | I=b) 
\end{equation} 
where $p(b,v)$ is the function acquired by the learner after $m$ training  
instances of an infinite training set have been examined. 
\end{definition} 
This completes the framework. 
 
\section{Towards a Theory of Data  
Requirements} \label{sec:dr_beginning} 
 
While the framework provides a notational foundation for a theory of data  
requirements, it cannot be used by itself to make predictions about data  
requirements.  The relationship between training data volumes and expected  
accuracy depends heavily on both the structural characteristics of the  
probabilistic model and the learning algorithm used to estimate the model  
parameters.  It would be a mammoth task to construct mathematical  
representations of all the probabilistic models and training methods  
commonly used in statistical language learning. 
 
Instead, the remainder of this chapter focuses on a very simple case, in fact,  
the simplest well-behaved one.  To instantiate the framework we need to  
provide both a probabilistic model and a training algorithm to estimate the  
parameters of that model, thus yielding an appropriate function $\pfromc$  
for any given training set $c$. 
 
\subsection*{Choosing a model and training algorithm}

The probabilistic model I will examine is the trivial one, the complete joint  
distribution over $B \times V$.  This model has a set of parameters  
$\hat{p}(b, v)$, one for each pair composed of a bin and a value,  
representing an estimate of the probability $\Pr(J=v | I=b)$.  As long as the  
learning algorithm ensures that $\sum_{v \in V} \hat{p}(b,v) = 1$ for all $b  
\in B$, we can let $p(b,v) = \hat{p}(b,v)$ and we have a probabilistic  
analyser.  Since this constraint removes one free parameter for each bin, the  
size of the model is given by $t = |B| (|V|-1)$. 
 
The training algorithm is the simplest well-behaved one given this model:  
the maximum likelihood estimate assuming a multinomial distribution.  Let  
$\countfn_c(b, v)$ be the number of pairs in the training sequence $c$  
whose first component is bin $b$ and second component is value $v$.  Also,  
let $\countfn_c(b) = \sum_{v' \in V} \countfn_c(b, v')$, 
the number of training  
instances falling into bin $b$.  The training algorithm estimates the  
parameters of the model using 
\begin{equation} 
\hat{p}(b,v) = \left\{  
\begin{array}{cl} 
\frac{1}{|V|}  
	 & \mbox{if $(\forall v' \in V) \countfn_c(b, v') = 0$} \\ 
\frac{\countfn_c(b, v)}{\countfn_c(b)} 
	 & \mbox{otherwise} 
\end{array} 
\right. 
\label{eq:dr_beginning_mle} 
\end{equation} 
The fact that this model and training algorithm are well-behaved follows  
from fundamental results in statistics, but no proof will be given here.  Since  
the functions $p$ and $\hat{p}$ are the same I will use only $p$ hereafter. 
 
Maximum likelihood estimates are commonly used in \sll\ systems and other  
estimation methods are sufficiently similar in nature that reasoning about  
\acronym{mle} applies at least qualitatively to other methods  
too.\footnote{To be well-behaved other methods must at least share the  
convergence properties of \acronym{mle}.}  The limitation to complete joint  
distribution models is more restrictive since it results in two common \sll\  
strategies falling outside the domain of reasoning. 
 
First, by assuming the model is the complete joint distribution, we disallow  
the use of linguistic constraints to assist the analysis.  In practice, it is  
common to clamp certain parameters to zero, thus avoiding the need to  
estimate these and lowering data requirements.  Weischedel~\etal~(1993)  
and Merialdo~(1994) both assume that only those tags which have been  
observed for a word are possible.  This reduces the number of free  
parameters by an order of magnitude.  Yarowsky~(1992) only permits his  
noun sense disambiguator to assign a thesaurus category if the noun appears  
in that category.  These strategies are not available in the case I will  
investigate, although they are certainly possible within the framework given  
above. 
 
Second, by assuming a complete joint distribution, we prohibit the  
probabilistic model from combining simpler events to form more complex  
ones.  This is crucial to many \sll\ systems, including Markov model taggers,  
where tag $n$-grams are combined to form sentence length sequences.  The  
analysis chooses the sentence-length sequence which has maximum  
probability, rather than choosing the tag for each word individually.  In  
practice it is usually impractical to estimate the complete joint distribution  
between inputs and outputs.  Instead, assumptions are used to cut down the  
space, creating an intensional representation of the joint probability space.   
This therefore represents the most important avenue for
extending the present work.
 
These two qualifications mean that the simple case I will consider is not  
directly applicable to other probabilistic models and learning algorithms.   
However, I do believe the results provide general insights into the  
phenomenon of data sparseness as it appears in \sll, and that their  
significance extends beyond the simple case on which they are based.  They  
raise important issues for those engaged in engineering these kinds of  
systems and support order-of-magnitude predictions for planning purposes. 
Putting these caveats aside then, we can begin investigation of the simple  
case.   
 
\subsection*{Optimal error rates}

Consider first the optimal accuracy.  From  
equation~\ref{eq:dr_framework_oaf}, an optimal analysis function $A_{\opt}$
must have $\Pr(J=A_{\opt}(b) | I=b) = \max_{v \in V} \Pr(J=v | I=b)$ for every  
bin $b$.  An optimal analysis function selects some optimal value
for each bin.  Let $\vopt(b) = A_{\opt}(b)$ for some optimal analysis function  
$A_{\opt}$ and let $q(b) = \max_{v \in V} \Pr(J=v | I=b)$, the probability  
of the optimal value. Thus $\Pr(J=\vopt(b) | I=b) = q(b)$. 
 
Using equation~\ref{eq:dr_framework_ea} the optimal error rate $r_{\opt}$  
can be expressed in terms of $q(b)$. 
\begin{eqnarray} 
r_{\opt} & \stackrel{\rm def}{=} & 1-\alphaopt \nonumber \\ 
	& = & 1-\alpha(A_{\opt}) \nonumber \\ 
	& = & 1-\sum_{b \in B} \Pr(I=b) \Pr(J=\vopt(b) | I=b) \nonumber \\ 
	& = & 1-\sum_{b \in B} \Pr(I=b) q(b) \nonumber \\ 
	& = & \sum_{b \in B} \Pr(I=b) (1-q(b)) 
\label{eq:dr_beginning_ropt} 
\end{eqnarray} 
The contribution to the optimal error rate from each bin $b$ is $\Pr(I=b) (1- 
q(b))$. 
 
Consider now the analysis procedure used by the probabilistic analyser.
Given a test case falling into bin $b$ it uses  
the estimates $p(b,v)$ to select a value.  It follows from  
equations~\ref{eq:dr_framework_pa} and~\ref{eq:dr_beginning_mle} that  
the analyser will choose that value which has occurred with the given bin  
most often in the training corpus.  That is, there is a distribution of values  
which have been observed to fall into each bin and the analyser chooses the  
mode of that distribution.  For this reason, I will call the 
combination of the complete joint distribution model and the maximum
likelihood training algorithm a  
\newterm{mode-based learner}.  According to the definition of a  
probabilistic analyser and the interpretation of $\argmax{}$ given there,  
when the distribution of values falling into a bin
is multi-modal, the learner chooses any one of the modes at random. 
 
It is useful to distinguish two cases, corresponding to the two
alternatives in equation~\ref{eq:dr_beginning_mle}.  When the analyser is  
presented with a test case in bin $b$, 
either there is at least one value $v$ for  
which the training set contains some occurrences of $(b, v)$, or there are no  
training instances falling into bin $b$.  The latter situation I will call an  
\newterm{empty bin} (that is, $count_c(b) = 0$).  The former is then a  
\newterm{non-empty bin}.   
 
In the remainder of this section I will consider each of these situations in  
turn.  The mathematical reasoning from this point down to 
equation~\ref{eq:dr_beginning_nonemptybound} was developed by Mark
Johnson of Brown University for odd values of $n$.
My contribution was to complete the proof of 
equation~\ref{eq:dr_beginning_emptybound} and extend the reasoning
to even values of $n$.  I would like to thank him again for
his kind permission to publish these results.

\subsection*{Empty bins}  
\label{pg:dr_beginning_MJstart}
 
First, how do empty bins affect the expected accuracy? Let $p_{b}$ denote  
$\Pr(I=b)$.  Since the training instances are drawn independently of one  
another according to the same distribution as test instances, the probability  
of bin $b$ being empty after training on a corpus of $m$ training instances  
is $(1-p_{b})^{m}$.  Thus the probability, over all test inputs, 
of there being no training data in the bin for that input is given by  
\begin{equation}  
\Pr(\mbox{\it empty}) = \sum_{b \in B} p_{b}(1-p_{b})^{m}  
\end{equation}  
Put another way, this is the probability of encountering an empty bin during  
testing. 
 
Clearly this will vary with the distribution $I$ over the bins.  But since  
$\sum_{b \in B} p_{b} = 1$, it is possible to show using partial derivatives  
and boundary conditions that the maximum for $\Pr(\mbox{\it empty})$  
occurs when $(\forall b \in B)$ $p_{b} = \frac{1}{|B|}$.   
Therefore  
\begin{equation}  
\Pr(\mbox{\it empty}) \leq (1-\frac{1}{|B|})^{m} \leq e^{-m/{|B|}} 
\label{eq:dr_beginning_emptybound} 
\end{equation}  
So for values of $m/{|B|}$ greater than~1, the probability that any  
given test sample falls into a bin for which we received no training samples  
is quite low.  For example, when $m/{|B|}\geq 3$, they occur in less  
than~5\% of inputs.  

On test cases which do fall into an empty bin,  
equation~\ref{eq:dr_beginning_mle} and the definition of $\argmax{}$  
dictate that any value in $V$ is selected at random.  So the expected  
accuracy on these test cases is $\frac{1}{|V|}$.   
But even if we assume that all these test cases are analysed incorrectly, the  
contribution of empty bins to the error rate is never larger 
than $e^{-m/{|B|}}$.  Recall from section~\ref{sec:dr_need} that
the rule of thumb suggested that $m/t$ should be~10, where $t$
is the size of the model.
If the shortfall between  
expected accuracy and optimal accuracy is primarily due to unseen events,  
then this rule is very generous for small $V$,  
predicting larger data requirements than necessary.  This point is  
not the same as that made by Weischedel~\etal~(1993).  Their argument is  
that non-uniform bin distributions reduce the data requirements --- we will  
return to that issue in section~\ref{sec:dr_simulations}.  The result given  
here holds even for uniform bin distributions. 
 
\subsection*{Non-empty bins}
 
How then do non-empty bins affect the expected accuracy?  Such bins  
contain at least one training instance from the training set.  Let the number  
of these in a particular bin $b$ be denoted by $n = \countfn_c(b)$. 
 
Since $r_{\opt}$ is the best possible error rate, it follows 
from equation~\ref{eq:dr_beginning_ropt} that $q(b)$ must close to 1
for most bins if the system is to work well.  Since this is the  
probability that a test instance falling into bin $b$ has value $\vopt(b)$, we  
can expect this value to be one of the more frequent in the bin.  If more than  
half of the training instances falling into bin $b$ have the value $\vopt(b)$,  
then this must be the most frequent value in the bin; that is, $\vopt(b)$ must  
be the mode of the observed distribution of values in bin $b$. 
 
Thus, if $\countfn_c(b, \vopt(b)) > \frac{n}{2}$, then $A_c(b)= \vopt(b) =  
A_{\opt}(b)$ and the trained analysis function has optimal accuracy on test  
cases falling into bin $b$.   So by computing the probability of  
$\countfn_c(b, \vopt(b)) > \frac{n}{2}$, we can obtain a lower bound for  
the accuracy on bin $b$ relative to the optimal accuracy. 
The probability that the trained analysis function
is optimal in bin $b$ is bounded as follows.
\begin{eqnarray} 
\Pr(A_c(b) = \vopt(b) | I=b)  
 & \geq & \sum_{i= \lceil {\frac{n+1}{2}} \rceil}^{n} 
		\Pr(\countfn_c(b, \vopt(b)) = i) \nonumber \\ 
 & = & \sum_{i= \lceil {\frac{n+1}{2}} \rceil}^{n}  
           {n \choose i} (1-q(b))^{n-i} q(b)^{i}  
 \label{eq:dr_beginning_trained}
\end{eqnarray} 
When $n$ is even and $\countfn_c(b, \vopt(b)) = \frac{n}{2}$, only one  
other value $v'$ can possibly have $\countfn_c(b, v') = \frac{n}{2}$.  Thus,  
the probabilistic analyser will set $A_c(b) = \vopt(b)$ with probability at  
least one half.  This fact will be used to make the bound tighter below. 
 
Equation~\ref{eq:dr_beginning_trained} bounds the probability
that training in a given bin is optimal for that bin.  Consider
now a test instance falling into this bin.  Since training and testing
are independent we have the following equality.
\begin{equation} 
(\forall v \in V) 
(\Pr(A_c(b)=v, J=v | I=b) = \Pr(A_c(b)=v | I=b) \Pr(J=v | I=b))   
\end{equation}
We can then proceed as follows to bound the probability of the
trained analysis function matching the test case.
\begin{eqnarray*} 
\Pr(J=A_c(b) | I=b) & = & \sum_{v \in V} \Pr(A_c(b)=v, v=J | I=b)  \\
	& \geq & \Pr(A_c(b) = \vopt(b) | I=b) \Pr(J=\vopt(b) | I=b)   
\end{eqnarray*}
So if all bins contain at least one training instance the overall expected  
accuracy can be bounded as follows using equation~\ref{eq:dr_framework_ea}.
\begin{eqnarray} 
\alpha(A_c) & = & \sum_{b \in B} \Pr(I=b) \Pr(J=A_c(b) | I=b) \nonumber \\ 
	& \geq & \sum_{b \in B} \Pr(I=b) \Pr(A_c(b) = \vopt(b) | I=b) q(b)  
\nonumber \\ 
	& \geq & \sum_{b \in B} \Pr(I=b) q(b)  
		\sum_{i= \lceil {\frac{n+1}{2}} \rceil}^{n}  
			{n \choose i} (1-q(b))^{n-i} q(b)^{i}  
\label{eq:dr_beginning_mode}
\end{eqnarray} 
Therefore the contribution of a bin with at least one training instance to the  
expected accuracy is bounded below by the expression contained
in the outer sum. 
 
Switching from accuracy rates to expected error rates,
define a bounding function as follows, where $\even{n}$ is~1 when $n$  
is even and~0 otherwise. 
\begin{equation} 
U_n(b) \stackrel{\rm def}{=} 1-q(b) \left( \begin{array}{c}
\frac{1}{2} \even{n} {n \choose {\halfn}} (1-q(b))^{\halfn} q(b)^{\halfn} 
+ 
\sum_{i= \lceil {\frac{n+1}{2}} \rceil}^{n}  
	{n \choose i} (1-q(b))^{n-i} q(b)^{i} 
\end{array} \right) 
\end{equation} 
Examination of equation~\ref{eq:dr_beginning_mode} along with the
argument above for the case of even $n$ shows that
this is an upper bound on the contribution to the error rate from bins with  
$n$ training instances.  So the bound on the overall expected accuracy rate  
when all bins contain at least one training instance can be made slightly  
tighter and re-expressed as follows.  Note that $n$ varies with $b$. 
\begin{eqnarray} 
\alpha(A_c) & \geq & \sum_{b \in B} \Pr(I=b) (1-U_n(b)) \nonumber \\ 
	& = & 1- \sum_{b \in B} \Pr(I=b) U_n(b) 
\label{eq:dr_beginning_nonemptybound} 
\end{eqnarray} 
Comparing this to the optimal accuracy taken from  
equation~\ref{eq:dr_beginning_ropt} 
\begin{equation} 
\alpha(A_{\opt}) = 1- r_{\opt} = 1 - \sum_{b \in B} \Pr(I=b) (1-q(b)) 
\end{equation} 
we see that the error rate contributed by a bin is never more than  
$\frac{U_n(b)}{1-q(b)}$ times the optimal error rate for that bin.  This  
provides quite tight bounds.  For example, when $q(b) \geq 0.9$, 
$\frac{U_3(b)}{1-q(b)} \leq 1.26$ and $\frac{U_5(b)}{1-q(b)} \leq 1.08$. 
%
\label{pg:dr_beginning_MJfinish}

Unfortunately these bounds require knowing how many training instances
fall into a bin.  However,
if all bins contain at least one training instance, 
then we can loosely bound the overall expected accuracy as follows.  
The first step is a corollary of the  
half-binomial result given in appendix~\ref{appendix:halfbinomial}.
\begin{eqnarray}  
U_{n}(b) & \leq & U_{1}(b) \nonumber \\  
	& = & 1 - q(b)q(b) \nonumber \\
	& = & (1+ q(b))(1 - q(b)) \nonumber \\  
	& \leq & 2 (1-q(b)) 
\label{eq:dr_beginning_twiceoptimal} 
\end{eqnarray} 
Thus in all bins which have training instances in the corpus, the expected  
error rate for the bin never exceeds twice the optimal error rate for that bin  
and this yields the desired overall bound. 
\begin{equation} 
\alpha(A_c) \geq 1-\sum_{b \in B} \Pr(I=b) 2(1-q(b)) = 1 - 2r_{\opt}
\end{equation} 
This result is quite useful since we expect the optimal error rate to be quite  
low.  It is directly applicable if training data is collected in a way that  
ensures at least one training instance per bin.  If the optimal predictions  
are~90\% accurate, then a mode-based learner will be at least~80\% accurate  
after learning on just one instance per bin. 
 
 
So summarising the two main results so far. 
\begin{itemize} 
\item Empty bins occur fairly rarely --- less than~5\% of test instances will  
fall into an empty bin when there is an average of~3 training instances per  
bin. 
\item Non-empty bins exhibit quickly converging error rates --- 
their expected error rate is always less then double the optimal error rate
and with~5 training instances it is only~8\% over the optimal error rate.
\end{itemize} 
So in general it appears that~3--5 instances per bin will be sufficient. 
 
\section{Predicting the Global Expected Accuracy} 
\label{sec:dr_global}
 
 
 
Unfortunately, we cannot normally guarantee that no bins will be empty,  
since the corpus is typically a random sample. In order to combine  
equations~\ref{eq:dr_beginning_emptybound}  
and~\ref{eq:dr_beginning_twiceoptimal} to arrive at a bound for the overall  
expected accuracy after training on a random sample, we need to make  
further assumptions.  Two possibilities are: assume $\Pr(I=b)$ is constant  
across all bins or assume that $\Pr(J=\vopt(b) | I=b)$ is constant.  Since the  
former is a very poor assumption for reasons already given, I will opt for the  
latter. 
\begin{assumption}[Uniform Optimal Probability] 
The probability of the most likely value in each bin is constant:  
$\Pr(J=\vopt(b) | I=b) = p$. 
\end{assumption} 
Note that this does not require that the most likely value be the same value in  
each bin.  $\vopt(b)$ can vary with $b$ as long as its probability remains  
constant.  This assumption is not too drastic since if the analyser is to have   
reasonable optimal performance, most $\Pr(J=\vopt(b) | I=b)$ must be close  
to~1. 
 
Now equation~\ref{eq:dr_beginning_ropt} can be simplified, since $q(b) =  
\Pr(J=\vopt(b) | I=b) = p$, to obtain $r_{\opt} = 1-p$, where the  
contribution to the optimal error rate from each non-empty bin $b$ is $(1-p)  
\Pr(I=b)$.  So equation~\ref{eq:dr_beginning_twiceoptimal} tells us that the  
contribution to the expected accuracy rate from these bins is at least $(1-2(1- 
p)) \Pr(I=b)$.  It is also easy to show that this contribution must always be  
greater than $\frac{1}{|V|} \Pr(I=b)$.  Since empty bins result in the  
analyser making a random choice, the contribution to the expected accuracy  
rate from an empty bin $b$ is $\frac{1}{|V|} \Pr(I=b)$.  Using  
equation~\ref{eq:dr_beginning_emptybound} we can combine these as  
follows. 
\begin{eqnarray} 
\alpha(A_c) & = & \sum_{b \in B} \Pr(I=b) \Pr(J=A(b) | I=b) \nonumber \\ 
	& \ge & (1-e^{-m/{|B|}}) (1-2(1-p))     
		+ \frac{1}{|V|}e^{-m/{|B|}} \nonumber \\ 
	& = & (1-e^{-m/{|B|}}) (2p-1) 
		+ \frac{1}{|V|}e^{-m/{|B|}}  
\label{eq:dr_global_oldbound} 
\end{eqnarray} 
This bound is relatively weak but at least the assumptions it relies on are  
also quite weak.  An example plot of this bound will appear in  
figure~\ref{fig:dr_simulations_results} under the name \scare{Old Bound}. 
 
\subsection*{An exact expression for the expected accuracy}

If we make some different assumptions, an exact expression for the global  
expected accuracy is possible.  First let's assume that $V$ is binary.
Without loss of generality, let $V = \{v_0, v_1\}$ with 
$(\forall b \in B) \vopt(b) = v_0$.  
Let $B = \{b_1, b_2, \ldots \}$ and define a set of random variables  
over $V$, $V_i = (J|I=b_i)$.  
That is, $(\forall v \in V) \Pr(V_i=v) = \Pr(J=v | I=b_i)$.  
We can revoke the uniform optimal probability assumption by  
introducing a set of probabilities $v_i = \Pr(J=v_0 | I=b_i)$. 
 
It is useful to define the following quantity, about which some results  
appear in appendix~\ref{appendix:halfbinomial}. 
\begin{definition}[Half-binomial] 
The {\em half-binomial\/} is given by the following expression 
\begin{equation} 
\halfbin(p, n) \stackrel{\rm def}{=}  
\frac{1}{2} \even{n} {n \choose \halfn} p^{\halfn} (1-p)^{\halfn} 
+ 
\sum_{i=\halfnincup}^{n} {n \choose i} p^i (1-p)^{n-i} 
\end{equation} 
which is that part of the binomial probability distribution on the upper side  
of the midpoint. 
\end{definition} 
 
Define a further set of random variables $\yaftern_i$ on $V$, describing the  
distribution of the mode observed in bin $b_i$ after $n$ training instances 
have fallen into that bin.  This is expressed by the following equation
where $c_i(n)$ is a training set containing $n$ instances falling into bin  
$b_i$.  
\begin{equation}
\Pr(\yaftern_i=v_0) \stackrel{\rm def}{=} \Pr(A_{c_i(n)}(b_i)=v_0)
\end{equation}
Application of the binomial theorem leads to 
\begin{equation} 
\Pr(\yaftern_i=v_0) = \halfbin(v_i,n) 
\end{equation} 
 
Let $g_i(n)$ be the accuracy on test cases in bin $b_i$ after $n$ training  
instances have fallen into that bin.  
Using the definition of correctness, we get 
\begin{eqnarray} 
g_i(n) & = & \Pr(\yaftern_i = V_i) \nonumber \\ 
	& = & v_i \halfbin(v_i, n) + (1-v_i) \halfbin(1-v_i,n) \nonumber \\ 
	& = & 1-v_i+(2v_i-1) \halfbin(v_i,n) 
\end{eqnarray} 
The last step uses the fact that $\halfbin(1-p,n) = 1-\halfbin(p,n)$,
which can easily be verified by expansion.
 
The training set is allocated to the bins according to the random  
variable $I$ and so the number of training instances having fallen into bin  
$b_i$ is distributed binomially.  
That is, when $c$ contains $m$ training instances, 
$\Pr(\countfn_c(b_i) = n) = {m \choose n} {p_i}^n (1-p_i)^{m-n}$ 
where $p_i = \Pr(I=b_i)$.  So the expected accuracy for test cases in  
bin $b_i$ can be written as 
\begin{eqnarray} 
\mbox{E}_n [ g_i(n) ] & = & \sum_{n=0}^{m}  
		\Pr(\countfn_c(b_i) = n) g_i(n) \nonumber \\
	& = & \sum_{n=0}^{m}  
		{m \choose n} {p_i}^n (1-p_i)^{m-n}  
		(1-v_i+(2v_i-1) \halfbin(v_i,n)) \nonumber \\ 
	& = & 1-v_i+(2v_i-1) \sum_{n=0}^{m}  
		{m \choose n} {p_i}^n (1-p_i)^{m-n} \halfbin(v_i,n))  
\nonumber \\ 
	& = & 1-v_i+(2v_i-1) G(m, p_i, v_i) 
\end{eqnarray} 
where 
\begin{displaymath}  
G(m,p_i,v_i) \stackrel{\rm def}{=} \sum_{n=0}^{m}  
		{m \choose n} {p_i}^n (1-p_i)^{m-n} \halfbin(v_i,n)) 
\end{displaymath}  
 
Thus the global expected accuracy can be reformulated as 
\begin{eqnarray} 
\alpha(A_c) & = & \mbox{E}_i [ \mbox{E}_n [ g_i(n) ] ]  \nonumber \\ 
	& = & \sum_{i=1}^{|B|} p_i \mbox{E}_n [ g_i(n) ]  \nonumber \\ 
	& = & 1 - \sum_{i=1}^{|B|} p_i v_i  
			+ \sum_{i=1}^{|B|} (2v_i-1) p_i G(m, p_i, v_i) 
\end{eqnarray} 
This is an exact expression for the expected accuracy that does not
depend on assuming uniform bins nor uniform optimal probabilities.
However it is dependent on complete knowledge of the bin and
value distributions. 

We can simplify this further by reinstating the uniform optimal 
probability assumption, that is $(\forall i) v_i = p$.  
The expected accuracy then reduces to
\begin{equation} 
\alpha(A_c) = 1 - p + (2p-1) \sum_{i=1}^{|B|} p_i G(m, p_i, p) 
\end{equation} 
 
We still need to know both $p$ and the $p_i$.  
Also, the function $G$ is expensive to compute.  The  
remainder of this section will show how to arrive at a cheaply computable  
expression for the expected accuracy when the $p_i$ are assumed uniform  
and $p$ is known.  In the following section I will report some simulations to  
explore the effect of the uniform bins assumption. 
 
 
 
If the bin distribution is uniform ($(\forall i) p_i = \frac{1}{|B|}$) 
then we can simplify further as follows. 
\begin{equation} 
\alpha(A_c) = 1 - p + (2p-1) \frac{1}{|B|} G(m, \frac{1}{|B|}, p) 
\label{eq:dr_global_newbound} 
\end{equation} 
The main computational difficulty with the  
function $G$ is the appearance of ${m \choose n}$.  Most corpus-based  
language learners use large corpora, so we expect the number of training  
instances, $m$, to be very large.  So we need a more easily computable  
version of $G$.  The following argument leads to a fairly tight lower bound  
to $G$ for suitably chosen values of $k_j$ (see below).  For simplicity I  
will ignore the extra term that appears in  
$\halfbin(p,n)$ for even $n$.  It is straightforward to prove that this term  
carries through in the expected way. 
 
\begin{eqnarray*} 
G(m,r,p) & = & \sum_{n=0}^m \mbox{binomial}(n;m,r) 
                       \sum_{i=\halfnincup}^{n}  
 \mbox{binomial}(i;n,p)   \\ 
	& = & \sum_{n=0}^m \sum_{i=\halfnincup}^{n} 
                      \mbox{binomial}(n;m,r)  
\mbox{binomial}(i;n,p)   \\ 
	& = & \sum_{j=0}^{\lceil \frac{m}{2} \rceil} \sum_{n=2j+1}^m 
                      \mbox{binomial}(n;m,r)  
\mbox{binomial}(n-j;n,p)   \\ 
	& = & \sum_{j=0}^{\lceil \frac{m}{2} \rceil} \sum_{n=2j+1}^m 
                      {m \choose n} r^n (1-r)^{m-n} {n \choose j}  
p^{n-j} (1-p)^j   \\ 
	& = & \sum_{j=0}^{\lceil \frac{m}{2} \rceil} (1-r)^m (\frac{1- 
p}{p})^j  
                      \sum_{n=2j+1}^m \frac{m!}{(m-n)!} (1- 
r)^{-n}  
                          \frac{p^n}{n!} {n \choose j} r^n   \\ 
	& \ge & \sum_{j=0}^{\lceil \frac{m}{2} \rceil} (1-r)^m (\frac{1- 
p}{p})^j  
                      \sum_{n=2j+1}^{k_j} \frac{m!}{(m-n)!} (1- 
r)^{-n}  
                          \frac{p^n}{n!} {n \choose j} r^n 
\end{eqnarray*} 
 
The second step rearranges the order of the two sums.  
The final step introduces a  
series of variables which limit the number of terms in the inner sum.  The  
inequality holds for all $k_j \le m$.  Notice that the $k_j$ may vary for each  
term of the outer sum.  Since $n \le k_j \le m$ we can use the following  
relation:  
 
\begin{equation}  
\frac{m!}{(m-n)!} \ge (m-k_j)^n           
\label{eqn_factorial_approx}  
\end{equation}  
 
Letting $x_j \stackrel{\rm def}{=} r p \frac{(m-k_j)}{(1-r)}$ 
we can simplify as follows:  
 
\begin{eqnarray*} 
G(m,r,p) & \ge & \sum_{j=0}^{(m-1)/2} (1-r)^m (\frac{1- 
p}{p})^j  
                      \sum_{n=2j+1}^{k_j} {n \choose j}  
\frac{m!}{(m-n)!} 
                          \frac{(1-r)^{-n} r^n p^n}{n!}   \\ 
              & \ge & \sum_{j=0}^{(m-1)/2} (1-r)^m (\frac{1- 
p}{p})^j  
                      \sum_{n=2j+1}^{k_j} {n \choose j} (m- 
k_j)^n 
                          \frac{(1-r)^{-n} r^n p^n}{n!}   \\ 
               & = & \sum_{j=0}^{(m-1)/2} (1-r)^m (\frac{1- 
p}{p})^j  
                      \sum_{n=2j+1}^{k_j} {n \choose j}  
\frac{x_j^n}{n!}  \\ 
               & \ge & \sum_{j=0}^{g} (1-r)^m (\frac{1-p}{p})^j  
                      \sum_{n=2j+1}^{k_j} {n \choose j}  
\frac{x_j^n}{n!} 
\end{eqnarray*} 
 
The last step introduces $g$ and holds for all $g \le (m- 1)/2$.  This is  
because in practice only the first few terms of the outer sum are significant.   
Thus for suitably chosen $g, k_j$ this is a cheaply computable lower bound  
for $G$.  A program to compute this to a high degree of accuracy has been  
implemented. 

We now have a computable expression that closely approximates the
expected accuracy of a mode-based learner assuming a uniform bin
distribution and uniform optimal probabilities.  
An example plot of this expression appears in  
figure~\ref{fig:dr_simulations_results} under the name \scare{New Bound}. 
 
\section{Simulations of a Mode-based Learner} \label{sec:dr_simulations}  
 
The assumption of uniform bin probabilities significantly simplifies the  
analysis, but in most cases is drastically violated by the data. This is  
especially worrying because non-uniform bin distributions can have a strong  
affect on the training process. The expected number of relevant training  
instances when bins are logarithmically distributed is many times greater  
than that when bins are uniformly distributed.  When bins are uniformly  
probable, the expected number of training instances in the same bin as a  
random test instance is $\frac{m}{|B|}$.  But Zipf's law states that word  
types are distributed logarithmically (the $n$th most frequent word has  
probability proportional to $\frac{1}{n}$).  When this is true the expected  
number of training instances in the same bin as a random test instance is  
approximately $\frac{1.6 m}{\log{(0.56 |B|)}^2}$ ($ \gg \frac{m}{|B|}$).   
Thus we can expect much more information to be available about {\em  
typical} test cases. 
 
This idea is discussed by Weischedel~\etal~(1993).  They argue that this  
phenomenon is responsible for the near optimal accuracy of their tagger with  
only~64 thousand words of training data, rather than the~1 million they  
predict by parameter counting.  This is a highly plausible explanation and so  
in this section a series of simulations are reported to explore the effects of  
non-uniform distributions.  The simulations also serve to validate the results  
above for uniform distributions. 
 
The simulations use a fixed set of~10 thousand bins, allocating $m$  
training instances to the bins randomly according to either a uniform or  
logarithmic distribution.  Each training instance is randomly assigned one of  
two values, with the optimal value having probability $p = 0.9$.   
Simulations with other values of $p$ did not differ qualitatively.  The  
optimal accuracy rate is therefore~90\%.  For each value of $m$,
the correctness of the mode-based  
learner on~1000 randomly generated test instances is computed to arrive at  
an observed correctness rate. 
 
This process (training and testing) is repeated~30 times for each run, with  
the mean being recorded as the observed accuracy.   The standard deviation  
is used to estimate a~95\% $t$-score confidence interval ($t$=2.045). 
 
\begin{figure}  
\centering \input{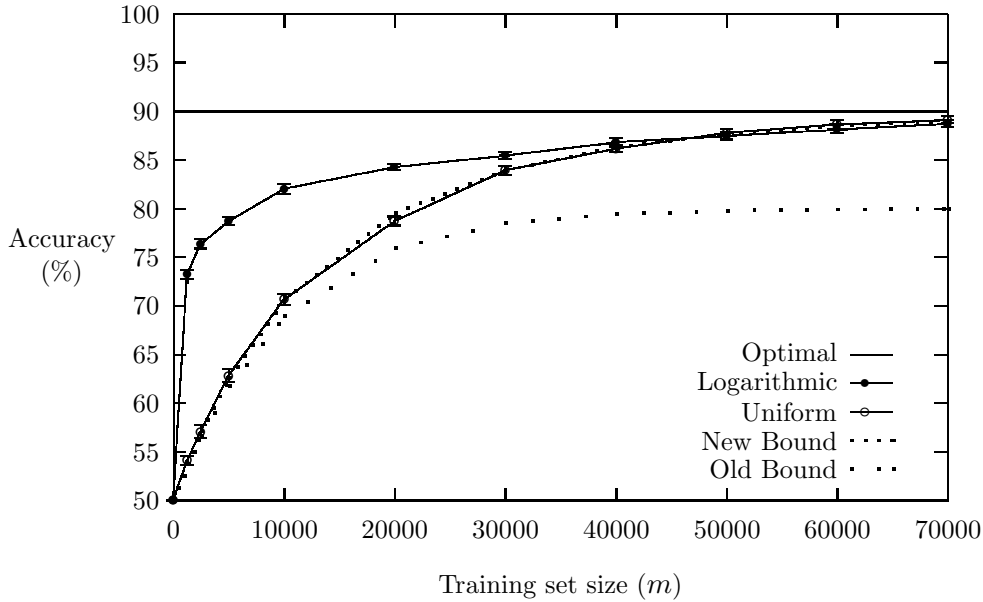}  
\caption{Simulation results and theoretical predictions for 10,000 bins}  
\label{fig:dr_simulations_results}  
\end{figure} 
 
 
Figure~\ref{fig:dr_simulations_results} shows five traces of accuracy as the  
volume of training data is varied.  The lowest curve shows the bound given  
by equation~\ref{eq:dr_global_oldbound}, the bound formed by combining  
the empty and non-empty bin results.  The other dotted curve shows the  
expected accuracy predicted using the exact expression for the accuracy  
assuming uniform bins and uniform optimal probability given by  
equation~\ref{eq:dr_global_newbound}, as approximated by the program  
described in the previous section.  The two further curves (with 
95\% confidence interval bars) then show the results of simulations, 
using uniform and logarithmic bin distributions. 
 
As can be seen, the approximation given for $G$ is quite accurate.
The match between the \scare{New Bound} and the uniform bin simulation
results is not surprising; it only serves to verify that the
algebra given in section~\ref{sec:dr_global} is correct.
The new bound can be used to predict data requirements 
when bins are uniformly distributed.   
It is far superior to the old bound beyond about one training instance
per bin ($m = 10$ thousand).  

However, when the bins are logarithmically distributed, learning converges  
significantly more quickly, as suggested by the reasoning about expected  
number of relevant training instances.  
Non-uniformity of the bin distribution accelerates the learning process.
Any practical theory of data requirements must incorporate some measure
of non-uniformity which can be related to the increase in learning rate.
Perhaps surprisingly though, the  
logarithmic distribution appears to eventually fall behind the uniform one  
once there is plenty of data.  This might be explained by the presence of very  
rare bins in the logarithmic distribution which thus take longer to learn.   
 
\section{Summary}

I have argued for the need for a theory of data requirements in \sll\
and provided a framework for reasoning about it.  Within this framework
I have given some results regarding the relationship between 
expected accuracy and training data volumes.  This includes a
computable expression for the expected accuracy under the assumption
of uniform optimal probabilities and uniform bin distribution.
In addition, I have reported on some simulations showing that
learning converges more quickly with non-uniform bin distributions.

While all these results 
provide insights into the nature of data requirements in \sll\ systems, 
there is a long way to go towards a general theory.  I hope that these  
first steps will form a reasonable foundation for future research, with the  
eventual outcome that the badly needed predictive theory will become  
available, making statistical language learning research less of an art form  
and statistical language learning technology more accessible to industrial  
application. 
 

\cleardoublepage

\chapter{Experiments on Noun Compounds}
\label{ch:experimental}


Since compound noun understanding involves a great deal of knowledge, the 
statistical acquisition of information for compound disambiguation  
is a promising task for the application of the meaning distributions theory.  
The experimental work in this thesis is aimed at two tasks.  
First, the syntactic analysis of noun compounds 
and, second, the semantic analysis of them.  
Accordingly, this chapter
is divided into two parts, each devoted to one of these tasks.
Where components have been used both in the experimental work 
on syntax and that on semantics, details are only given in the first half.

\section{Noun Compound Parsing} 


The first half of this 
chapter describes experiments on parsing compound nouns using the 
statistical techniques suggested by the meaning distributions theory.  

The goal here is the syntactic analysis of any noun compound, although 
for the purposes of the experiment I will only consider minimally
ambiguous noun compounds.  The problem and the precise scope of the 
experiments is defined in section~\ref{sec:cy_problem}.

In section~\ref{sec:cy_model}, I will use the meaning
distributions theory to develop a probabilistic model
of noun compound syntax.  This model is based on dependency relations 
between concepts, which provide a shallow semantic representation.
The section also shows that 
these dependency structures are isomorphic to constituency based 
parse trees.

An overview of the components of the experimental setup is given
in section~\ref{sec:cy_design}, including the corpus, the lexicon,
the analyser and the test set.  Full details of the experimental
method follow in section~\ref{sec:cy_method}.  These include
the method used to collect test and training data from the corpus,
the concepts used, the parameter estimation strategy and the
analysis procedure.

Section~\ref{sec:cy_results} gives the resulting performance
measures for the parsing model.  This is followed 
in section~\ref{sec:cy_comparisons} by a series of
empirical comparisons, including one between the new parsing model
and an equivalent adjacency algorithm (recall
from section~\ref{sec:cn_statistical} that all previous
proposals use the latter method).  

To establish how close the performance of the model
is to the best possible under these conditions, an experiment
with human judges has been conducted.  This is reported in
section~\ref{sec:cy_human}.  Finally, some conclusions are drawn
and limitations discussed in section~\ref{sec:cy_discuss}.  

\subsection{Defining the Parsing Problem} 
\label{sec:cy_problem}

In all the experimental work, I will only consider English compound nouns.  
Nonetheless, compounds appear in many other languages (a list of research 
work on compounds in other languages was given in 
section~\ref{sec:cn_nature}) and there seems no reason why the same 
techniques would work less well in these.  

I shall also assume that the compound has been recognised from the 
surrounding text, so that the system is presented with a sequence of nouns 
known to be a compound (see the description of the identification problem 
in section~\ref{sec:cn_computational}).  

Given an identified compound, it is simplest to define the parsing task as one 
of bracketing.  That is, the system must select the most likely binary 
bracketing of the noun sequence, assuming that it is a compound noun.  For 
example, \lingform{animal cruelty committee} would usually be analysed as 
shown in 
example~\ref{eg:cy_problem_brack_1}\ref{eg:cy_problem_brack_acc}
whereas \lingform{woman customs official} would be assigned that
shown in 
example~\ref{eg:cy_problem_brack_1}\ref{eg:cy_problem_brack_wco}.

\begin{examples}
   \item	\label{eg:cy_problem_brack_1}
     \begin{subexamples}
        \item{[}[animal cruelty \nn] committee \nn{]}   
\label{eg:cy_problem_brack_acc}
        \item{[}woman [customs official \nn] \nn{]}   
\label{eg:cy_problem_brack_wco}
        \item{[}[chocolate [[birthday party \nn] cake \nn] \nn] 
		obsession \nn{]}   \label{eg:cy_problem_cbpco}
\end{subexamples}
\end{examples}

I refer to the former as left-branching, and the latter as right-branching.  
Note that bracketing is well-defined for longer compounds too, as shown
in example~\ref{eg:cy_problem_brack_1}\ref{eg:cy_problem_cbpco}.

This representation is slightly weaker than most syntactic representations, in 
the same way that dependency grammar is (a detailed theory of this 
grammatical formalism is given in Mel'cuk,~1988).  The 
difficulty lies in determining the scope of modifiers 
(for a detailed description, see Covington,~1994).  
For example, in \lingform{prototype hydrogen balloon} 
it is clear that the syntax is right-branching, yet there is an 
ambiguity as to whether the object being described is a prototype for 
all balloons that happens to use hydrogen, or is a prototype for 
specifically hydrogen balloons.  For the purposes of this 
investigation, I shall assume that scoping ambiguity is negligible, and 
accept bracketing as sufficient syntactic analysis.

When analysing compounds, I take the correct bracketing to be that which 
reflects the compositional structure of the {\em meaning} of the compound.  
This is demonstrated in example~\ref{eg:cy_problem_brack_1}.  The 
meaning of \lingform{animal} is related to that of \lingform{committee} 
only through the meaning of \lingform{animal cruelty}.  Similarly, the 
meaning of \lingform{customs} is only related to that of \lingform{woman} 
through the meaning of \lingform{customs official}.  While this is the 
usual analysis, it is conceivable that a syntactic theory might require a 
different interpretation of bracketings (for example, in order to explain some 
observed grammatical alternations).  The interpretation of syntactic analysis 
used here is aligned with the ultimate goal of understanding compound 
nouns.

According to most views of compounding, the composition of two nouns 
yields an element with essentially the same syntactic behaviour as the 
original nouns.  A two word compound noun acts exactly like a single noun, 
as do three word compounds and so forth.  It is this recursion that allows 
arbitrarily long compounds.  Only one theory of compounding, 
Marcus~(1980), posits any change in the syntactic properties of compounds 
as they get longer, and counter-examples to this theory are well-known 
(Finin,~1980).  It follows that all the qualitative syntactic properties of 
compounds of any length are exhibited by three word compounds (shorter 
compounds lack internal syntactic ambiguity).

In the experimental work I will adopt this view as an assumption and only 
investigate three word compounds.  However, the probabilistic model of 
compound noun syntax to be developed in section~\ref{sec:cy_model} is not 
limited to three word compounds.  It assigns probabilities to arbitrarily long 
compounds (modulo the assumption that syntactic behaviour is fully 
recursive).  As we've seen, for three word compounds the syntactic 
ambiguity is binary.

So now I can precisely state the problem addressed in the experimental work 
below.

\begin{description}
\item[Problem Statement:] Given a three word English compound noun, 
predict whether the most likely syntactic analysis is left-branching or 
right-branching.
\end{description}

\subsection{A Model Based on Dependency 
Grammar} 
\label{sec:cy_model}

In section~\ref{sec:cn_statistical} I described a number of corpus-based 
approaches to precisely this problem.  Recall that 
all of these use an adjacency algorithm; that is, they
select the analysis whose innermost constituent is most acceptable.  
In this section I will develop a probabilistic
conceptual model of noun compounds.
The model assigns probabilities to shallow representations of
noun compound meanings.  In conjunction with the meaning distributions
theory, this leads to a new algorithm.

The section is composed of five parts.  First, I will discuss the
principle differences between the new algorithm and existing ones.
Second, I will formally define the structures upon which the model
is based and note the relationship of these structures to parse
trees.  
Third, the event space and assumptions of the model will be
given.  
Fourth, I will give the equations for applying the model to parsing
noun compounds and finally I will discuss a general prediction
of the model that is consistent with empirical data.

\subsubsection*{Two different analysis methods} 

There are three key differences between the new algorithm and the 
existing ones.  
First, and most significant of all,
the new algorithm selects the analysis incorporating the most 
acceptable dependency relations (instead of the most acceptable 
constituents).  Second, conceptual association is used rather than lexical 
association.  Third, the statistics used by previous algorithms are 
not derived from explicit probabilistic models.

Before beginning the mathematical development of the model, I will
describe informally the most important difference, the
distinction between analysis based on innermost constituents and that based 
on dependency relations.  The other two 
differences are sufficiently self-evident to stand without additional 
explanation.

Consider the algorithms based on innermost constituents.  As mentioned 
in section~\ref{sec:cn_statistical}, 
these are all variants of an algorithm proposed 
by Marcus~(1980), which makes calls to an oracle to determine how 
acceptable a two word compound is.  This is reproduced again here.
\begin{citedquote}{Marcus,~1980:253}
Given three nouns $n_1$, $n_2$ and $n_3$:
\begin{itemize}
\item If either [$n_1$ $n_2$] or [$n_2$ $n_3$] is not
semantically acceptable then build the alternative structure;
\item otherwise, if [$n_2$ $n_3$] is semantically
preferable to [$n_1$ $n_2$] then build [$n_2$ $n_3$];
\item otherwise, build [$n_1$ $n_2$].
\end{itemize}
\end{citedquote}
Notice that only adjacent nouns are ever given to the oracle for acceptability 
judgements.  I have called algorithms that select analyses 
on the basis of the acceptability of the innermost constituent 
adjacency algorithms.

An algorithm that captures dependency relations must allow for longer 
distance dependencies.  Such an algorithm was first reported in 
Lauer~(1994).\footnote{This algorithm can be viewed as a 
specialisation of the probabilistic grammar that conditions on 
syntactic relations and heads, which is proposed in 
Charniak~(1993) as described earlier in section~\ref{sec:sn_grammars}.}
This algorithm might be written as follows.
\begin{quote}
Given three nouns $n_1$, $n_2$ and $n_3$:
\begin{itemize}
\item Determine how acceptable the structures
[$n_1$ $n_2$] and [$n_1$ $n_3$] are;
\item if the latter is more acceptable, build [$n_2$ $n_3$] first;
\item  otherwise, build [$n_1$ $n_2$] first.
\end{itemize}
\end{quote}
This algorithm pays no attention to the acceptability of [$n_2$ $n_3$], 
focusing instead on two constructions involving $n_1$.  I will call any 
algorithm that maximises the acceptability of dependency relations within 
the compound a \newterm{dependency algorithm}.

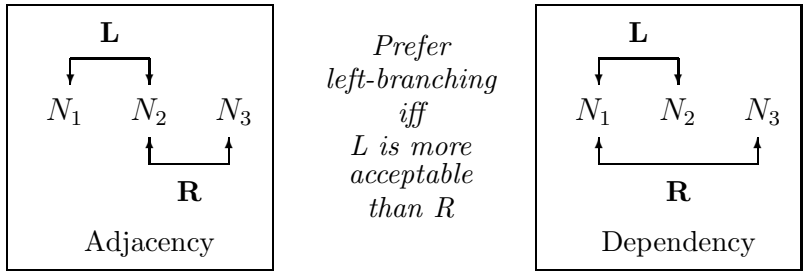
\begin{figure*}
\centering


\begin{picture}(300,100)
\put(200,0){
\begin{picture}(100,100)
\put(0,0){\framebox(100,100){}}
\put(50,60){
   \makebox(0,0){
      \large $N_1$ \hspace{10pt} 
      \large $N_2$ \hspace{10pt} 
      \large $N_3$
   }
}
\put(50,10){
   \makebox(0,0){Dependency}
}
\put(15, 70){
   \begin{picture}(40, 30)(-5,0)
     \put(15,20){\makebox(0,0){\bf L}}
     \put(0,10){\vector(0,-1){10}}
     \put(30,10){\vector(0,-1){10}}
     \put(0,10){\line(1,0){30}}
   \end{picture}
}
\put(15, 20){
   \begin{picture}(70, 30)(-5,0)
      \put(30,10){\makebox(0,0){\bf R}}
      \put(0,20){\vector(0,1){10}}
      \put(60,20){\vector(0,1){10}}
      \put(0,20){\line(1,0){60}}
   \end{picture}
}
\end{picture}
}

\put(120, 20){
   \shortstack{\em Prefer \\ \em left-branching \\ \em iff \\
       \em L is more \\ \em acceptable \\ \em than R }
}

\put(0,0){
\begin{picture}(100,100)
\put(0,0){\framebox(100,100){}}
\put(50,60){
   \makebox(0,0){
      \large $N_1$ \hspace{10pt} 
      \large $N_2$ \hspace{10pt} 
      \large $N_3$
   }
}
\put(50,10){
   \makebox(0,0){Adjacency}
}
\put(15, 70){
   \begin{picture}(40, 30)(-5,0)
     \put(15,20){\makebox(0,0){\bf L}}
     \put(0,10){\vector(0,-1){10}}
     \put(30,10){\vector(0,-1){10}}
     \put(0,10){\line(1,0){30}}
   \end{picture}
}
\put(45, 20){
   \begin{picture}(40, 30)(-5,0)
      \put(15,10){\makebox(0,0){\bf R}}
      \put(0,20){\vector(0,1){10}}
      \put(30,20){\vector(0,1){10}}
      \put(0,20){\line(1,0){30}}
   \end{picture}
}
\end{picture}
}
\end{picture}
\caption{Two analysis models and the associations they compare}
\label{fig:cy_model_2ways}
\end{figure*}

For example, when \lingform{backup compiler disk} is
encountered, the dependency analysis will be:
\begin{examples}
   \item	\label{eg:cy_model_dependency}
     \begin{subexamples}
        \item{[}[backup compiler \nn] disk \nn{]} 
	when \lingform{backup compiler} is more acceptable  
	\label{eg:cy_problem_brack_bcd_l}
        \item{[}backup [compiler disk \nn] \nn{]}   
	when \lingform{backup disk} is more acceptable
	\label{eg:cy_problem_brack_bcd_r}
\end{subexamples}
\end{examples}
%
%
This should be compared to example~\ref{eg:cn_statistical_adj} in 
section~\ref{sec:cn_statistical}.  The difference between these two types of 
algorithms is illustrated graphically in figure~\ref{fig:cy_model_2ways}.

I claim that the dependency algorithm makes more
intuitive sense for the following reason.  Consider
the compound \lingform{calcium ion exchange}, which is
typically left-branching.  There does not seem to be
any reason why \lingform{calcium ion} should be any
more frequent than \lingform{ion exchange}.  Both are
plausible compounds and regardless of the
bracketing, \lingform{ions} are the object of an 
\lingform{exchange}.  Instead, the correct parse depends on
whether \lingform{calcium} characterises the \lingform{ions}
or mediates the \lingform{exchange}.

The model given below follows the meaning distributions theory.  It 
involves a shallow representation of compound noun semantics, the structure 
of which is a homomorphic image of their syntactic structure.  It therefore 
satisfies the requirements of the meaning distributions theory.  The basic unit 
of this representation is the notion of modification.  The model uses 
\newterm{modificational structures}, a representation that can be 
constructed directly from a bracketing, but which reflect the semantics of the 
compound more directly.  Since the notion of modification is made explicit 
in dependency grammar, the model is akin to a probabilistic dependency 
grammar for compound nouns.  The resulting compound noun parsing 
algorithm is a dependency algorithm for this reason.

\subsubsection*{Modificational structure and semantic 
classes}

To build a dependency-based model we require a representation
that captures the dependency relations within a compound.
I will therefore now give a definition of modificational structures
and show how the compound noun parsing problem can be reduced 
to the selection of these structures.

According to the general syntax rules of compound nouns, every 
binary tree with $n$ leaves is a possible parse of a compound noun 
$w_1 w_2 \ldots w_n$ where the $w_i$ are all nouns.  In each case,
we say that the leaves of the parse tree are labelled by the words
in the string.  Each such parse incorporates a modificational structure.  
This is defined by assigning one 
modification relationship for each interior node in the binary tree.  In 
particular, for each interior node, we assert that the rightmost leaf of the left 
child is a \newterm{modifier} of the rightmost leaf of the right child.  

\begin{definition}[Modifier] 
Given a binary branching parse tree with leaf nodes $x_1, x_2, \ldots, x_n$ 
and an interior node, $t$, define $x_i$ to be a {\em modifier\/}
of $x_j$ through $t$, 
where $x_i$ is the rightmost leaf of the left child of $t$ and $x_j$ is the 
rightmost leaf of the right child of $t$.
\end{definition} 

This results in $n-1$ modification relationships, one for each word except 
the last.  The set of modification relationships defined by a parse tree is 
called the modificational structure of that tree.

\begin{definition}[Modificational Structure] 
Given a binary branching parse tree with leaf nodes $x_1, x_2, \ldots, x_n$,  
the {\em modificational structure\/} of that tree is 
the set of ordered pairs of leaf 
nodes $\{ (x_i, x_j) | x_i \mbox{ is a modifier of } x_j\}$.
\end{definition} 

\begin{figure*}
\centering
\input{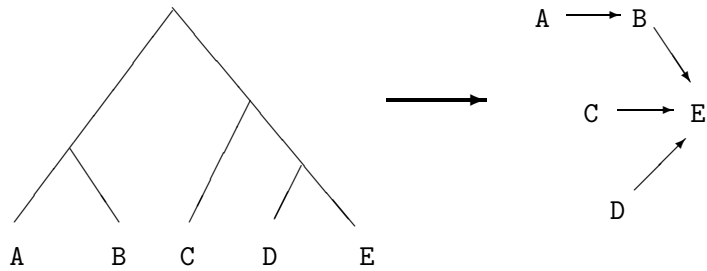}
\caption{Parse trees define modificational structures}
\label{fig:parse2modstruct}
\end{figure*}

Modificational structures can be displayed graphically by means of one 
point for each leaf node and one arrow for each modification relationship, 
pointing from the modifier to the modified leaf node.  An example showing 
how a parse tree yields a modificational structure is shown in 
figure~\ref{fig:parse2modstruct}.  While the children at a node
of a parse tree are ordered (left child and right child),
modifiers of a node in a modificational structure are unordered.
 
This definition of modificational structure follows the general 
properties of compound noun interpretations in which the rightmost 
noun below a given node is the head and carries the semantic class of the 
interpretation of that node.  I am not claiming that modificational 
structures fully represent the meaning of compounds.  Rather, they create 
divisions along the lines of such meanings, which are useful for syntactic 
predictions.  We can think of the modificational structure as a shallow 
representation of the meaning of the compound, from which the syntactic 
structure is generated in a productive process.  


A corollary of the definition of modifier is that every word $w_i$ 
($1 \leq i < n$) labels a leaf that is a modifier of a unique other 
leaf further to its right.  
Hence every modificational structure forms a tree 
with the rightmost word $w_n$ labelling the root.  Whenever $x_i$ is a
modifier of $x_j$, $x_j$ is closer to the root than $x_i$.  We say
that $x_j$ is above $x_i$ whenever $x_j$ lies on the path from $x_i$
to the root.

Since the modifiers of a node in a modificational structure are unordered, 
there is generally more than one parse tree for a single modificational
structure.  It is therefore useful to define the possible
orderings of the nodes of a modificational structure.
\begin{definition}[Consistent Ordering]
Given a rooted tree with $n$ nodes $x_a, x_b, \ldots, x_z$, 
any ordering of these nodes $x_1, x_2, \ldots x_n$ is defined 
to be {\em consistent\/} 
with the tree if every node of the tree has the property that
the set of nodes below it forms a complete consecutive subsequence
$x_i, x_{i+1},\ldots, x_j$ of the ordering for some $1 \leq i \leq j \leq n$.
\end{definition}

It is easy to show that whenever $x_1, x_2, \ldots x_n$ is consistent
with a directed tree, that tree is the modificational 
structure of exactly one parse of the string ``$w_1 w_2 \ldots w_n$''
where each $x_i$ is labelled $w_i$ for $1 \leq i \leq n$.
It is therefore sufficient to choose the correct 
modificational structure in order to determine the correct parse.  Henceforth, 
I will thus consider the goal to be selection of modificational structures 
rather than bracketings.  

Given a modificational structure with nodes labelled 
$w_1, w_2, \ldots, w_n$, every 
postorder traversal of the tree generates a string ``$w_{q(1)} w_{q(2)} 
\ldots w_{q(n)}$'' where $q$ is a permutation consistent with the 
modificational structure.  This string is a syntactically legal compound (any 
string of nouns is under the rule \={N} $\rightarrow$ \={N} \={N}), and in 
each case (that is, each possible postorder traversal), the modificational 
structure must correspond to exactly one parse of ``$w_{q(1)} w_{q(2)} 
\ldots w_{q(n)}$''.

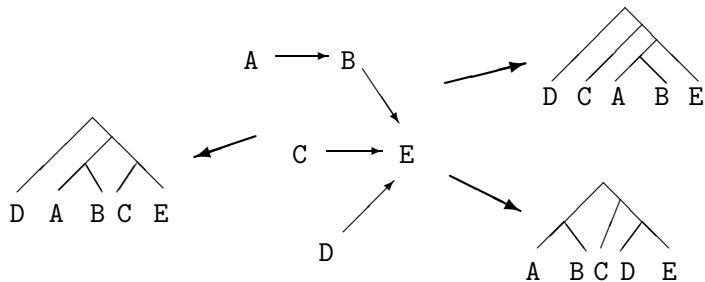
\begin{figure*}
\centering
%
%
%
%
%
%
{\tt
\setlength{\unitlength}{0.92pt}

\begin{picture}(306,134)

\put(86,36){
\begin{picture}(136,65)
\thicklines
\put(25,33){\vector(-3,-1){25}}
\put(103,55){\vector(4,1){33}}
\put(105,17){\vector(2,-1){29}}
\end{picture}
}

\put(10,32){
\begin{picture}(66,46)
\thinlines
\put(59,2){E}
\put(44,2){C}
\put(33,2){B}
\put(16,2){A}
\put(0,2){D}
\put(31,26){\line(3,-5){7}}
\put(42,37){\line(-1,-1){22}}
\put(52,26){\line(-2,-3){8}}
\put(34,45){\line(1,-1){29}}
\put(34,45){\line(-1,-1){31}}
\end{picture}
}

\put(222,10){
\begin{picture}(63,42)
\thinlines
\put(56,0){E}
\put(39,0){D}
\put(28,0){C}
\put(18,0){B}
\put(0,0){A}
\put(39,33){\line(-2,-5){8}}
\put(48,24){\line(-3,-4){9}}
\put(32,40){\line(1,-1){27}}
\put(16,24){\line(3,-4){9}}
\put(32,40){\line(-1,-1){27}}
\end{picture}
}

\put(229,80){
\begin{picture}(67,44)
\thinlines
\put(60,2){E}
\put(46,2){B}
\put(28,2){A}
\put(14,2){C}
\put(0,2){D}
\put(40,22){\line(1,-1){11}}
\put(47,29){\line(-1,-1){17}}
\put(41,35){\line(-1,-1){23}}
\put(34,42){\line(1,-1){30}}
\put(34,42){\line(-1,-1){30}}
\end{picture}
}

\put(106,16){
\begin{picture}(71,91)
\thinlines
\put(41,14){\vector(1,1){21}}
\put(49,81){\vector(2,-3){15}}
\put(34,47){\vector(1,0){23}}
\put(13,86){\vector(1,0){22}}
\put(64,42){E}
\put(31,2){D}
\put(20,42){C}
\put(40,81){B}
\put(0,81){A}
\end{picture}
}

\end{picture}
}
\caption{Modificational structures correspond to one parse
of each of several different strings}
\label{fig:cy_model_modstruct2parse}
\end{figure*}

Figure~\ref{fig:cy_model_modstruct2parse} shows how one modificational 
structure yields multiple word strings, but in each case only one parse of the 
string incorporates that modificational structure.  In summary, giving any 
two of the string, the parse or the modificational structure, uniquely defines 
the third. 
 
In addition to the use of dependency relations,
another novel aspect of the model is the incorporation of conceptual 
association (see section~\ref{sec:sn_conceptual}).  Rather than treating each 
word individually, words are gathered into semantic classes.  This has a 
practical motivation in reducing the number of parameters (and thus also the 
data requirements), since the number of semantic classes is smaller than the 
number of words.  On the theoretical side, this is in keeping with the 
meaning distributions theory.  Probabilities are assigned to 
semantic elements, arrangements of semantic classes, rather than to surface 
elements, the word strings.  All the definitions regarding modificational 
structures above hold equally well for nodes labelled by 
semantic classes as for nodes labelled by words.  
In so much as semantic classes can be viewed as semantic 
primitives, the resulting modificational structures form a shallow semantic 
representation of the meanings of compounds.  

Following the meaning distributions theory, the model is based on 
events involving these 
classes, rather than events involving words.  This is based on the assumption 
that all words that realise a given semantic class have roughly the same 
properties.  

\subsubsection*{Some notation and assumptions for the 
model}

I now turn to the formal specification of the model and its assumptions.
Let $W$ be the set of all words in compound nouns (each of 
which will have at least one noun sense).  Let $S$ be a set of semantic 
classes, which are themselves represented by sets of words.  That is, every 
class $s \in S$ is a subset of $W$.  

Each instance of a compound noun is considered an event.  We 
denote the occurrence of a compound whose nouns are $w_1, w_2, \ldots 
w_n$ in that order by ``$w_1 w_2 \ldots w_n$''.  We also assume that when 
a word appears in a compound, it is used in a sense that corresponds to one 
of the semantic classes.  Since words are polysemous, the sense being used 
is not explicit and different sense possibilities must be accounted for by the 
model.  We denote the (unknown) semantic class of a particular instance of a 
word, $w_i$, by $\mbox{\it sense\/}(w_i) \in S$.  To allow all word senses in a 
compound to be considered together, we take $s_1 s_2\ldots s_n$ to denote 
the occurrence of a compound ``$w_1 w_2 \ldots w_n$'' wherein 
$\mbox{\it sense\/}(w_i) = s_i$ for all $0 < i \leq n$.  

Assume that the entire vocabulary has been assigned to semantic classes.
\begin{assumption}[Lexical Meaning]
Every word is a member of some semantic class:
\begin{equation}
(\forall w \in W) (\exists s \in S \mbox{ such that } w \in S)
\end{equation}
\end{assumption}
It follows that $(\bigcup_{s \in S} s) = W$.  
Since words have multiple senses, each word may appear in 
several semantic classes.  
Define $\mbox{\it cats\/}(w)  \stackrel{\rm def}{=} \{ 
s \in S | w \in s \}$, the set of semantic classes that may be realised by a 
word.  The lexical meaning assumption guarantees that this is non-empty.  
Define $\ambig(w) \stackrel{\rm def}{=} | \mbox{\it cats\/}(w) |$,
the number of senses of the word $w$..  

We are interested in modificational structures.  Let $M(X)$ denote the set of 
possible rooted trees whose nodes are labelled by elements of the set $X$  
(where the arcs leading to each node are unordered).
The event $m$ where $m \in M(W)$ denotes the 
occurrence of a compound noun, whose modificational structure is the 
tree $m$.  When $m \in M(S)$, $m$ denotes the event $s_1 s_2\ldots s_n$ 
with the additional information that the modificational structure of the 
corresponding compound ``$w_1 w_2 \ldots w_n$'' is the tree $m$.  
Also, let $s_i \rightarrow s_j$ denote the occurrence of a compound whose 
modificational structure includes the link $x_i$ is a modifier of $x_j$ 
where $x_i$ is labelled $w_i$, $x_j$ is labelled $w_j$, 
$\mbox{\it sense\/}(w_i) = s_i$ and $\mbox{\it sense\/}(w_j) = s_j$.
Finally, the model uses probabilities of the form $\Pr(s_i \rightarrow s_j | 
\exists s: s \rightarrow s_j)$, which it will be useful to abbreviate as $\Pr(s_i 
\rightarrow s_j | s_j)$.



We are now in a position to state the principal modelling assumption 
underpinning the probabilistic conceptual model.  This will be used in 
addition to the meaning distributions theory,
and is of particular interest because it makes an assertion about the space of 
possible meanings, rather than the space of possible syntactic structures as is 
done in previous models (see section~\ref{sec:md_priors}).  


It is a novel assumption of the model that all modificational structures 
that are composed of modificational links with the same labels, are 
equi-probable.\footnote{Since 
the algorithm derived from the model only ever compares the probabilities 
of modificational structures possible for a given string, only structures that 
generate the same string need meet this requirement.}  In fact, we assume 
that the probability of any complex modificational structure is derived from 
the probabilities of its links by multiplication.
\begin{assumption}[Modifier Independence]
Given a modificational structure $m \in M(S)$, its probability is 
proportional to the product of the probabilities of its individual modification 
relationships; that is,
\begin{equation}
\Pr(m) \propto \prod_{c \mbox{ is a modifier of } x \mbox{ in } m}
	\Pr(c \rightarrow x | x) 
\label{eq:cy_model_link_product}
\end{equation}
\end{assumption}

This assumption differs substantially from other probabilistic grammar 
models proposed in the past, which typically assume that all parse trees 
involving the same rewrite rules are equi-probable.  
The modifier independence assumption makes an assertion 
about the distribution of possible 
meanings rather than about the distribution of possible syntactic structures.  
This difference becomes important because some modificational structures 
are possible interpretations of several different compound nouns (as is the 
case in figure~\ref{fig:cy_model_modstruct2parse}), 
while parse trees are always an analysis of a unique compound.  

Intuitively, when a speaker wishes to refer to an entity, she may 
choose among different possible orderings of the modifiers.  For 
example, suppose an object $A$ has two associated objects $B$ and $C$, 
used to identify it.  The speaker may use either ``$w_B w_C w_A$'' or 
``$w_C w_B w_A$''.  In contrast, if object $A$ is associated with object 
$B$, which is associated with object $C$, the speaker must use ``$w_C w_B 
w_A$''.  Therefore, assuming \foreign{a priori} equi-probable 
modificational structures, and given the compound ``$w_C w_B w_A$'', the 
probability of the first structure is half that of the second structure.

To capture this imbalance between the possible modificational structures, we 
define the degree of choice available to the speaker when expressing a given 
modificational structure $m$ as follows.  This is simply the number of 
postorder traversals of the modificational tree.

\begin{definition}[Choice]
Given a modificational structure $m$, the number of distinct strings with 
parses for which $m$ is a valid modificational structure is called the 
{\em degree of choice\/} of $m$, and is given by
\begin{equation}
\mbox{\it choice\/}(m) = \prod_{x \in \mbox{NodesOf}(m)} 
			\mbox{ChildrenOf}(x)!
\label{eq:cy_model_choice}
\end{equation}
\end{definition}
Note that this measure is independent of the type of node labels in the 
tree and applies with $m \in M(W)$ or $m \in M(S)$.  


In addition to the modifier independence assumption, I will also adopt the 
meaning distributions theory.  In this case, the space of possible meanings is 
$M(S)$ and the mapping between word strings and this space is determined 
by the grammar of compound nouns.  Given a string of elements of $W$, 
``$w_1 w_2 \ldots w_n$'', there is a set of modificational structures over 
senses, $\Psi_{w_1, w_2, \ldots, w_n} \subseteq  M(S)$, which can generate 
the string (precisely those with nodes 
$s_i \in \mbox{\it cats\/}(w_i)$ and with 
which the ordering $s_1,s_2, \ldots s_n$ is consistent).  Consider any $m \in 
\Psi_{w_1, w_2, \ldots, w_n}$.  The meaning distributions theory states in 
equation~\ref{eq:md_linking_contrib} (in section~\ref{sec:md_linking}) 
that all strings that can be generated by $m$ are equi-probable given $m$.  
I will adopt that assumption here.
\begin{assumption}[Meaning Distributions]
Given a modificational structure $m$ that is an allowable interpretation of 
the word string ``$w_1 w_2 \ldots w_n$'' $(m \in \Psi_{w_1, w_2, \ldots, 
w_n})$, every string generated by $m$ has an equal 
contribution to its probability from $m$.  That is,
\begin{equation}
(\forall q \mbox{ consistent with } m)
(\Pr(``w_{q(1)} w_{q(2)} \ldots w_{q(n)}\mbox{''} | m) 
	= \Pr(``w_1 w_2 \ldots w_n\mbox{''} | m) )
\label{eq:cy_model_uniform_generation}
\end{equation}
\end{assumption}

There are exactly $\mbox{\it choice\/}(m)$ sense sequences consistent with 
$m$.  For each of these sense sequences, the number of possible word 
sequences is given by $\prod_{i=1}^{n} |s_i|$.  Therefore the total number 
of word strings generated by $m$ is the product of these two factors, and so 
equation~\ref{eq:cy_model_uniform_generation} dictates that the 
probability that $m$ will generate exactly the string ``$w_1 w_2 \ldots 
w_n$'' is given by
\begin{equation}
\Pr(``w_1 w_2 \ldots w_n\mbox{''} | m) = 
	\frac{1}{\mbox{\it choice\/}(m) \prod_{i=1}^{n} |s_i|}
\label{eq:cy_model_prob_generation}
\end{equation}
Note that $m \in M(S)$ and so the values of the $s_i$ are determined by 
$m$.

\subsubsection*{Analysis}

Having developed this machinery, we are now in a position to apply 
it to the problem of analysing compound nouns.  The choice facing the 
system is to select some $m_w \in M(W)$ which is the correct 
modificational structure for a given compound ``$w_1 w_2 \ldots w_n$''.  
In the terminology of the data requirements theory 
of chapter~\ref{ch:dr}, the bins are word strings and the values
are modificational structures.
Thus it must compare the various probabilities $\Pr(m_w | ``w_1 w_2 \ldots 
w_n\mbox{''})$ for each possible $m_w \in M(W)$.  In order to do so, the model 
uses estimates of a set of parameters
$\Pr(s_1 \rightarrow s_2 | s_2 )$, the probability that 
semantic class $s_1$ modifies semantic class $s_2$, 
given that $s_2$ is modified.  We therefore need an expression for the
probability of each parse in terms of these parameters.
%

First, since the probabilistic conceptual model assigns probabilities to 
elements of $M(S)$, it will be convenient to let $m_s \in M(S)$ stand for the 
modificational structure formed from $m_w$ by assigning the sense $s_i$ to 
each word $w_i$ in $m_w$, where the values of $s_i$ will be fixed below.  
From the lexical meaning assumption, $S^n$ is a partition of $W^n$ so 
probability theory gives the following equality.
\begin{eqnarray}
\Pr(m_w | ``w_1 w_2 \ldots w_n\mbox{''})
  & = & \sum_{
	s_1 \in \mbox{\it cats\/}(w_1),\ldots,s_n \in \mbox{\it cats\/}(w_n)}
	   	\Pr(m_s | ``w_1 w_2 \ldots w_n\mbox{''})   \\
  & = & \sum_{
	s_1 \in \mbox{\it cats\/}(w_1),\ldots,s_n \in \mbox{\it cats\/}(w_n)}
		\frac{\Pr(m_s) \Pr(``w_1 w_2 \ldots w_n\mbox{''} | m_s)
			}{\Pr(``w_1 w_2 \ldots w_n\mbox{''})}
\end{eqnarray}
where the second step applies Bayes' rule.

Equation~\ref{eq:cy_model_prob_generation} allows us to simplify further:
\begin{eqnarray}
\lefteqn{\Pr(m_w | ``w_1 w_2 \ldots w_n\mbox{''})} \nonumber \\  
  & = & \sum_{
	   s_1 \in \mbox{\it cats\/}(w_1),\ldots,s_n \in \mbox{\it cats\/}(w_n)}
		\frac{Pr(m_s)} 
		{\mbox{\it choice\/}(m_s) \prod_{i=1}^{n} |s_i| 
			\Pr(``w_1 w_2 \ldots w_n\mbox{''})}   \\
  & = & \frac{1}
	{\mbox{\it choice\/}(m_w) \Pr(``w_1 w_2 \ldots w_n\mbox{''})}
	\sum_{
	   s_1 \in \mbox{\it cats\/}(w_1),\ldots,s_n \in \mbox{\it cats\/}(w_n)}
		\frac{Pr(m_s)}{\prod_{i=1}^{n} |s_i|} 
\label{eq:cy_model_analysis}
\end{eqnarray}
The second step follows from the fact that $\mbox{\it choice\/}(m_w) = 
\mbox{\it choice\/}(m_s)$ because $m_s$ has the same structure as $m_w$.

Since $\Pr(``w_1 w_2 \ldots w_n\mbox{''})$ is constant during the analysis of a given 
compound, the only probability needed is $\Pr(m_s)$, which is given in 
terms of the model parameters by the modifier independence assumption.  
Thus, given the set of possible analyses, the most probable can be found by 
computing the above function for each and taking the largest.  We now have 
a dependency algorithm.  

The model given can be applied to compound nouns of any length.  In the 
experimental work described below, it will be applied to three word 
compounds.  Before turning to that work, one implication of the model is 
worth emphasising.

\subsubsection*{A prediction of the model}

While the various adjacency algorithms are not based on probabilistic 
models, the statistics that they compute can, in most cases, be interpreted as 
probabilities.  In this way inferences can be made about what any 
probabilistic model for them must look like.  In particular, they exhibit a 
reflectional symmetry through the middle word, which can tell us something 
about their assumptions.  For them, left-branching and right-branching 
parses are reflections of one another and nothing about the statistics 
computed or the decisions based on those statistics breaks that symmetry.  
Therefore, given no information about the words, any reconstructed 
probabilistic model for an adjacency algorithm must predict that 
left-branching and right-branching compounds are equally likely.  Half of all 
three word compounds should be left-branching and the other half
right-branching.

A significant prediction of the dependency model is that the proportion of
left- and right-branching three word compounds should differ.  Suppose a 
word string ``$w_1 w_2 w_3$'' allows two equally likely modificational 
structures over $S$: a left-branching $m_L$ and a right-branching $m_R$ 
with $\Pr(m_R) = \Pr(m_L)$.  According to 
equation~\ref{eq:cy_model_analysis}, the posterior likelihoods of the two 
structures differ exactly by the term $\frac{1}{\mbox{\it choice\/}(m)}$.  We 
have $\mbox{\it choice\/}(m_L) = 1$ and $\mbox{\it choice\/}(m_R) = 2$, so 
left-branching analyses are twice as likely as right-branching analyses given 
equally likely meanings.  If we, not unreasonably, assume that across the 
meaning space $M(S)$ left-branching and right-branching modificational 
structures are equally likely, then the proportion of left-branching 
compounds will be twice that of right-branching ones.  That is, two-thirds of 
compounds should be left-branching.

It turns out that in the test set used here 
(see table~\ref{tb:cy_method_test_dist} in 
section~\ref{sec:cy_method} below) and in that of Resnik~(1993), the 
proportion of left-branching compounds is 67\% and 64\% respectively, well 
within statistical variation of the two-thirds proportion predicted by the 
dependency model.  Thus, the dependency algorithm has the advantage
of correctly predicting the observed distribution where the adjacency
algorithm does not.  It is therefore all the more surprising that
all previous proposals involved adjacency algorithms.

Since the dependency model was first proposed in Lauer~(1994)
a similar model has also been developed, apparently independently,
by Kobayasi~\etal~(1994) for analysing Japanese 
noun compounds.  Using a corpus to acquire associations, they bracket 
sequences of Kanji with lengths four to six (roughly equivalent to two or 
three words).  Unfortunately, as a simple calculation shows, using their own 
preprocessing heuristics to guess a bracketing provides a higher accuracy on 
their test set than their statistical model does.  
This renders their experiment 
inconclusive.  In the experimental work below the accuracy results are 
carefully compared to the baseline accuracy of two-thirds achieved by 
guessing.

\subsection{Experimental Design} 
\label{sec:cy_design}

We now have a probabilistic model for parsing noun compounds based
on dependency relations.
The remainder of the first half of this chapter describes a range of 
experimental work aimed at evaluating the model and the resulting parsing 
strategy.  The first experiment aims to 
measure the accuracy of the preferences suggested by the parsing strategy.
Subsequently, a range of 
variations on the algorithm will be compared, including a comparison 
between the adjacency and dependency algorithms.

In this section, the architecture of the experimental system is described.  In 
order to discover how accurate the parsing strategy is, the experimental
method includes implementation of a program to apply the strategy,
plus a performance test of the program.  The result of the experiment 
is a measure of the accuracy of the strategy in predicting the appropriate 
parses from compounds in the test set.  

First, the architecture of the experimental system will be 
outlined.  The parsing strategy requires a large corpus, a part-of-speech 
lexicon and a thesaurus; therefore, I will then give
a brief description of each of these and finally cover the 
other input to the experimental system: the test set.

The experimental method will be given in much more detail in 
section~\ref{sec:cy_method} below, including the processing of 
the corpus to acquire training data, the generation of a test set 
and the algorithm used to analyse three word compounds using 
information derived from the training data.  

\subsubsection*{Architecture}

The overall structure of the experimental setup is determined
by the model given in section~\ref{sec:cy_model}.  In order to use the 
model, values for the parameters $\Pr(s_1 \rightarrow s_2 | s_2 )$ are 
required.  These will be estimated by assuming that the distribution of 
dependency relations in two word compounds is representative of those in 
longer compounds.  Therefore, we need only extract examples of two word 
compound nouns from a corpus to provide evidence for the parameter 
estimates.  I will call each of these parameters the \newterm{affinity} 
between $s_1$ and $s_2$.

To evaluate the performance of the model, we therefore need to compose the 
following elements:
\begin{itemize}
\item a training corpus,
\item a means of extracting two word compound nouns from 
that corpus,
\item a data structure for recording the resulting evidence,
\item a program to make inferences from the evidence, and
\item a set of test examples with the correct parse for each.
\end{itemize}

In what follows, all evidence used to estimate the parameters of 
the model is collected in one pass over the corpus and stored in 
a fast access data structure.  Evidence is gathered across the 
entire vocabulary, not just for those words necessary for 
analysing a particular test set.  Once trained in this way, the 
program can quickly analyse any compound (restricted only by 
the lexicon and thesaurus).  This demonstrates that the parsing strategy can 
be directly employed using currently available hardware in broad coverage 
natural language processing systems.  

It is often the case with research prototypes for statistical 
language learning techniques that experiments are conducted by 
collecting only those corpus statistics necessary for the test set in 
hand (see for example, Yarowsky,~1992, Resnik and Hearst,~1993).  
While such a methodology establishes the accuracy of these systems, 
many of the proposed techniques could not be applied across a 
reasonable-sized vocabulary because of computational limits, both in 
processing time and storage space.  The program described here runs on a 
desktop computer and, once the data structure has been set up, will analyse 
compounds with no perceptible delay.

The program also starts with raw text and is designed to use only freely 
available resources.  For this reason, it should be possible for anyone to 
cheaply reimplement the parsing strategy that I have used.  Of particular 
importance is the fact that no manual annotation of the training data is 
necessary.  This is a result of the assumption that inferences about longer 
compounds can be made on the basis of observations of the behaviour of 
shorter compounds.

\subsubsection*{The corpus}

The corpus that is used here is derived from an on-line 
multimedia encyclopedia called \publicationname{The New Grolier 
Multimedia 
Encyclopedia}~(Grolier Inc.,~1992).  This work is mass-produced 
on \acronym{cd-rom} and contains text, graphics, sound and video.  The 
textual component, comprising approximately 8 million words, 
was written by a variety of expert authors.  It covers an 
extremely wide variety of topics, so that it contains a broad 
vocabulary.  The subject matter is clearly not from a limited 
domain, though attention is concentrated on a few themes 
(geography, history, art, biology), with others appearing less 
frequently.

Despite the unconstrained domain, the writing style is relatively 
uniform, adopting a register designed for clarity and 
informativeness.  As I have argued in section~\ref{sec:md_register}, a 
uniform register is precisely what we desire.  Language properties are 
dramatically affected by register; any system which ignores
changes in register will be more prone to error, since important
information is discarded.  A model trained on text from several
registers should not be expected to accurately capture the behaviour 
of any one of them.

Given the rapidly increasing volume of electronically stored 
text, we can expect unannotated corpora of this size to be available 
everywhere quite soon.  By contrast, the availability of 
manually annotated training data is much less probable.  Even if large 
research efforts such as the Penn Treebank project (Marcus~\etal,~1993) 
succeed in producing low cost high quality parsed corpora 
containing millions of words, the likelihood that these corpora will have 
characteristics that are appropriate to a given task (for example,
appropriate topic, matching register and required annotations) is low.

\subsubsection*{The lexicon}

The program requires independent lexical knowledge for two tasks:
identifying noun compounds and grouping synonymous nouns together.
The first task is necessary in order to extract examples of
compounds.  This is done by means of a simple heuristic that is described by 
equation~\ref{eq:cy_method_trainset} in section~\ref{sec:cy_method} 
below.  The only linguistic resource it requires is a list of nouns, and the 
assumption that such a list exists is not a highly demanding one.  For this 
experiment, a large lexical database from the University of Pennsylvania 
morphological analyser was used.  This is freely available over the Internet 
for research purposes (see address in section~\ref{sec:cy_method} below).
The program used to implement the heuristic takes raw text as input and 
produces a stream of two word compounds.  

These compounds form the evidence to be stored in a data 
structure for use by the program.  This structure has one element 
for each parameter of the probabilistic model.  Since the 
probabilistic model is based on semantic classes rather than on words a 
mapping is needed to associate words with such classes.  Once this is 
provided, observations of the behaviour of words will yield inferences about 
the behaviour of concepts represented by semantic classes.  

The assignment of words to semantic classes involves some 
rather deep linguistic knowledge and the availability of this kind 
of resource is less assured.  For anything larger than a toy 
domain, the development of a lexical-semantic ontology 
involves substantial effort.  In this work, a machine-readable 
thesaurus is used as a rough conceptual system.  While the kind 
of synonymy captured by thesaurus categories is unlikely to be 
ideal, it is easily available.  The thesaurus used is the 
1911 version of Roget's Thesaurus, which is freely 
available over the Internet (see address in section~\ref{sec:cy_method}).

If no well-motivated conceptual lexicon is available 
for the target text type, the trivial conceptual system in which 
there is exactly one concept for each word could be used.  
There are a couple of drawbacks to this 
lexically parametrised scheme.  In a conceptual lexicon, words can
be assigned to multiple semantic classes, representing different
senses of the word.  The lexically parametrised scheme cannot
represent polysemy in this way.  Also, for a large vocabulary,
the number of parameters of the lexical scheme would become impractical. 
Nonetheless, where it is practical, it provides an alternative
when a conceptual lexicon is unavailable.

\subsubsection*{The test set}

To generate the test set, a list of compounds, three words in length, were 
extracted from the corpus.  A similar heuristic to that used for finding 
training instances was employed and the results then analysed by hand to 
assign the correct answers.  The extraction heuristic is described by 
equation~\ref{eq:cy_method_testset} in section~\ref{sec:cy_method} 
below.  

After discarding some problematic test cases (again see 
section~\ref{sec:cy_method}), each of 244 test compounds was assigned a 
reading of either left-branching or right-branching by the author.  This 
annotation was done before any testing of the program was conducted.

The program was then given these compounds and asked to 
supply one or other of the two possible readings.  It was required 
to give an answer in every case, so that all accuracy figures 
below represent the performance at 100\% coverage.  The accuracy 
was then computed as the proportion of cases in which the 
reading selected by the program matched that assigned 
manually.  Since the ambiguity is binary, the expected accuracy 
of a uniform random selection is 50\%.

\subsection{Experimental Method} 
\label{sec:cy_method}

In this section, the experimental setup will be presented in some detail.  The 
reader who is uninterested should skip to section~\ref{sec:cy_results}, 
noting two equations:  Equation~\ref{eq:cy_method_assoc_metric} 
describing the measure used to associate concepts and 
equation~\ref{eq:cy_method_dep_ratio} giving the means by 
which analyses are made.  

\subsubsection*{The training set}

Grolier's encyclopedia contains over 30,000 text articles, each 
with a title and a variable length body containing on average 
around 12 sentences.  An example sentence is:\footnote{This 
extract is taken from the \articlename{Aardvark} article by Everett 
Sentman; The Academic American Encyclopedia (Electronic 
Version), Copyright 1992 Grolier, Inc., Danbury, CT.}

\begin{examples}
\item 
The aardvark (or ant bear), Orycteropus afer, is the only species 
in the MAMMAL family Orycteropodidae, order Tubulidentata.
\label{eg:cy_method_groliers_sentence}
\end{examples}
Here, the word \lingform{mammal} is capitalised because it anchors a 
hyperlink.  In the on-line version, the user may click on this to 
view the article on mammals.  

To create the corpus, each of the articles was extracted and concatenated.  
After discarding some of the fields associated with each article (the titles, the 
pronunciation codes and some bibliographic information), the 
remaining text formed the corpus with no further preprocessing.  
In particular, no tagging, sentence boundary 
identification or stemming is needed before 
the training algorithm begins.\footnote{The training algorithm
does include a couple of simple rules for converting plural nouns 
to singular.  These are given in the
subsection on conceptual grouping below.} 
The applicability of the algorithm does not depend on 
having such tools available.  

Following the probabilistic model, training proceeds by estimating affinities 
between semantic classes.  Since two word compounds are syntactically 
unambiguous, affinities can be estimated by counting such compounds.  The 
first step therefore is to identify two word compounds within the corpus.  
Since the corpus is untagged, a heuristic is used for this purpose.  Even 
though the heuristic is not perfect, useful training estimates can still be 
computed.  

The heuristic begins by tokenising the text into a stream of 
tokens of two types: words and punctuation.  For the purposes of 
the experiment, words are strings consisting entirely of a 
combination of hyphens, apostrophes and alphabetics.  All other 
characters are considered part of a punctuation token, including
numerics.  Thus, the string \lingform{during the 1930s Hitler (b.~1889, 
d.~1945) became Germany's} results in 14 tokens, including \token{1930},  
\token{s}, \token{b} and \token{Germany's}, where the first of these four is 
treated as punctuation.  Compounds are assumed to be consecutive strings of 
words, so that the inclusion of any kind of punctuation within a compound 
will cause the heuristic to ignore it.  This has the advantage that comma 
separated lists are not identified as compounds by the program.

The University of Pennsylvania morphological analyser (Karp~\etal,~1992) 
contains a large lexicon of parts of speech.\footnote{
Email Dania Egedi as egedi@ahwenasa.cis.upenn.edu, or anonymous ftp 
to ftp.cis.upenn.edu and retrieve /pub/xtag/morphology/morph-1.4.tar.Z.}  
Among the over 317,000 entries, there are some 90,000 which are listed as 
always being nouns.  From these, a short stoplist of 9 words that 
are entered incorrectly as always being nouns, and also all 
one-letter words, have been removed to yield a set of sure nouns 
(call this set, $\cal N$).\footnote{The nine words are 
\lingform{baroque}, \lingform{characteristic}, 
\lingform{en}, \lingform{epic}, \lingform{full-scale}, 
\lingform{full-length}, \lingform{incident}, \lingform{laissez-faire} 
and \lingform{mere}.}

Since the heuristic is intended to find two word compounds it 
looks for pairs of consecutive sure nouns.  For example, when 
scanning \lingform{The mountain goat is found in} the heuristic 
returns \lingform{mountain goat} as a two word compound.  To 
prevent it from selecting pairs within a larger compound that 
might be syntactically ambiguous, the heuristic requires that the 
tokens immediately to the left and immediately to the right not 
be nouns (either punctuation or words not in $\cal N$).  So in 
\lingform{Baking caramel muesli cookies} the heuristic will not identify 
\lingform{caramel muesli} as a compound.  Thus, the training pattern 
is given by:

\begin{equation}
T_{\mbox{train}} = \{ (w_1, w_2) \mid w_1 w_2 w_3 w_4; w_1,w_4 \notin 
{\cal N}; w_2, w_3 \in {\cal N} \}
\label{eq:cy_method_trainset}
\end{equation}
Here , $w_1 w_2 w_3 w_4$ denotes the occurrence of four tokens
in sequence in the corpus.

Since $\cal N$ contains only words which are always nouns, this 
is quite a conservative procedure.  Many common nouns, like 
\lingform{bear}, have verbal usages as well, so that the occurrence of 
\lingform{ant bear} in example~\ref{eg:cy_method_groliers_sentence}
above will not be extracted.  However, there is no guarantee that two 
consecutive nouns form a compound for two reasons.  

First, other grammatical constructions can result in two 
consecutive nouns.  For example, direct and indirect objects of a 
verb (as in \lingform{give every {\bf zebra gifts}}) or premodifying 
prepositional phrases (as in \lingform{In the third {\bf century 
peasants} farmed the valley}) both result in consecutive nouns.  
Second, because the heuristic ignores nouns like \lingform{bear}, longer 
compounds may only be partially extracted and the part extracted may not 
be a compound.  For example, the string \lingform{aluminium radiator
grill} produces \lingform{aluminium radiator} as a candidate 
(since \lingform{grill} is also used as a verb) when it is unlikely that 
this is a compound in this case.  

I manually checked the first 1000 compounds extracted by the 
heuristic and found only 21 of them did not form compounds.  
The causes of these errors are shown in 
table~\ref{tb:cy_method_train_errors}.
\begin{table*}
\centering
\begin{tabular}{|l|r|} \hline
Category & Number \\ \hline
PP boundary crossed & 5 \\
Class instance construction (`\lingform{the term earth}') & 5 \\
Relative clause  boundary crossed & 3 \\
Direct-indirect object boundary crossed & 2 \\
Foreign language & 1 \\
Sentence capitalisation & 1 \\
Elided relative pronoun & 1 \\
Name mistaken for noun & 1 \\
Participle mistaken for noun & 1 \\
Ungrammatical sentence & 1 \\  \hline
\end{tabular}
\caption{Training set error distribution from 1000 examples}  
\label{tb:cy_method_train_errors}
\end{table*}

The heuristic is also conservative about capitalisation, always 
making case sensitive comparisons.  This permits it to keep 
information about proper nouns separate from similar common 
nouns.  Unfortunately, this can result in the omission of 
compounds at the beginning of a sentence because the first
word may be capitalised when ordinarily it would not be.

This policy also obscures all compounds involving a hyperlink anchor.  
This results in the compound \lingform{MAMMAL family} in 
example~\ref{eg:cy_method_groliers_sentence} above being ignored.  
Many of the problems that I have identified with the heuristic in 
this section could be alleviated by using a sentence boundary 
detector and part of speech tagger.  

The training pattern was passed over the entire corpus.  
Table~\ref{tb:cy_method_noun_seqs} shows the number of noun pairs 
produced, along with equivalent counts for longer noun 
sequences.  It is worth noting that this distribution understates 
the actual distribution of compounds in the corpus by a great 
deal.  The degree of conservatism in the heuristic both reduces 
the total number of compounds identified and distorts the 
relative lengths towards shorter compounds.  This is because 
longer compounds are more likely to contain a noun that also 
has a verbal usage, somewhere within them.  Of the sequences 
extracted, the quads were ignored, the pairs used for training and 
the triples for testing.

\begin{table*}
\centering
\begin{tabular}{|l|r|c|}   \hline 
Type	&	Number	&	Specification \\
\hline
pairs	&	35343	&	$\{ (v, w) \mid u v w x; v,w \in 
{\cal N}; u,x \notin {\cal N}\}$  \\
triples	&	625	&	$\{ (v, w, x) \mid u v w x y; v,w,x \in 
{\cal N}; u,y \notin {\cal N}\}$  \\
quads	&	6	&	$\{ (v, w, x, y) \mid u v w x y z; v,w,x,y 
\in {\cal N}; u,z \notin {\cal N}\}$  \\
longer	&	0	&	$\{ (v, \ldots, w) \mid u v \ldots w x;
v, \ldots, w \in {\cal N}; u,x \notin {\cal N}\}$  \\
\hline
\end{tabular}
\caption{Noun sequences extracted heuristically from Grolier's 
encyclopedia}
\label{tb:cy_method_noun_seqs}
\end{table*}

\subsubsection*{Conceptual grouping}

To provide the concepts necessary for conceptual association, 
an electronic version of the 1911 Roget's 
thesaurus is used.  This is produced by Micra Inc. and is 
available via anonymous ftp from project Gutenberg.\footnote{
Anonymous ftp to Project Gutenburg: 
mrcnext.cso.uiuc.edu/etext/etext91/roget13a.txt.  Credit is due to 
Patrick Cassidy of Micra Inc. for making this resource publicly available.}  
While it is rather out of date, it is freely available.
The text in the thesaurus is freeform, being designed 
for a human reader.  Therefore, several preprocessing steps (a 
few days' development work) are required to arrive at a machine 
readable set of categories.  Readers interested in using the 
processed version of the thesaurus for research purposes should 
send me electronic mail.

The thesaurus contains 1043 categories.  These categories are 
intended to capture all possible concepts.  Viewed as elements
for conceptual association they embody something of the notion of semantic 
primitives (like the conceptual dependency theory of Schank,~1975, for 
example, but with many more primitives).  All words within a category are 
to be treated equally.  The hope is that differences in compound noun 
modification behaviours occur only across categories, not within a single 
category.  For example, \lingform{goldfish pond} and \lingform{trout pool} 
have similar behaviour because in each case the words used are drawn
from the same semantic classes.  If their behaviour differed from
one another, the assumptions of the probabilistic model
would be violated.

The size of thesaurus categories is widely variable, so that the 
degree of generalisation afforded by the concepts varies.  Eight of the 
categories only contain one noun, while one category (\tc{receptacle}{191}) 
contains 371.  The average number of words per category is 34.  
It is not immediately obvious what effect this has on the performance
of the model.
%
Also, the thesaurus is not as comprehensive as it could be.  
There are 20,445 distinct nouns in the thesaurus, of which 
11,716 are elements of $\cal N$.  For the purposes of the 
experiment, I discarded all sequences containing nouns not 
found in the thesaurus, unless they appeared to be plural forms 
whose singular was in the thesaurus (the rules used were: 
remove final 's', change final 'ses' to 's' and change final 'ies' 
to 'y').  This reduced the training set to 24,251 pairs.

\subsubsection*{Computing affinities}

To complete the training phase it was necessary to store all the 
evidence contained in this training set in a data structure for later 
use by the program.  Since the probabilistic model contains one 
parameter for each pair of concepts, the appropriate structure is 
a matrix whose rows and columns are indexed by thesaurus 
categories.  We interpret $A_{ij}$, the contents of the $i$th row 
and $j$th column, as the probability that the concept represented 
by the $i$th thesaurus category will be a compound noun 
modifier of the concept represented by the $j$th thesaurus 
category, that is $\Pr(t_i \rightarrow t_j \mid t_j)$.

For example, category \tc{fluidity}{333} represents liquids and 
category \tc{receptacle}{191} represents containers, so we'd 
expect the corresponding 
$A_{333\,\,191} =  \Pr(t_{333} \rightarrow t_{191} \mid t_{191})$ to 
be fairly high because of examples like \lingform{juice bottle}.  On 
the other hand, category \tc{liberation}{750} representing the 
concept of freedom is not usually an acceptable modifier of 
category \tc{receptacle}{191} so that $A_{750\,\,191}$ should be 
low.

The training set is a set of pairs of words and yet the 
model refers to pairs of concepts.  If each word always 
expressed only one concept, then each observation of a word 
would be an observation of the corresponding concept and 
counting words would be identical to counting concepts.  Due to 
polysemy, most words can be used to express several different 
concepts. Each noun in the 1911 Roget's Thesaurus is in an 
average of 1.7 distinct categories (the word \lingform{point} is in 
18).~\footnote{This average is is for types; that is, it
is not weighted for frequency.  If it
were we could expect a higher figure.}
Therefore, when a word is observed it can generally be an 
occurrence of any one of several distinct concepts.  

To allow for this, counts of concept occurrences were 
constructed by dividing the contribution from a word by the 
ambiguity of that word.  That is, when an occurrence of 
\lingform{corn} was observed, the counts for the categories 
\tc{convexity}{250} and \tc{food}{298} were each incremented by 
$\frac{1}{2}$, since \lingform{corn} is in both of them.  Joint counts 
(of pairs of concepts) were constructed by dividing the 
contribution from word pairs by the product of the words' 
ambiguities.  This procedure corresponds to Bayesian estimation 
where the prior probability of the different concepts is assumed 
uniform.

Given the counts of observed concept pairs, the desired 
probabilities were estimated using maximum likelihood assuming
a binomial distribution.  Since \acronym{mle} is well-known to be poorly 
behaved on small counts (see for example, Church~\etal,~1991b:127), this 
requires some justification.  Consider the four common alternatives:
\begin{itemize}
\item expected likelihood estimates (add 0.5 to each count),
\item backing-off to a smaller model (for example, deleted interpolation),
\item statistical hypothesis testing of various kinds, and
\item Good-Turing estimates.
\end{itemize}
While \acronym{ele} can be used to provide acceptable estimates in a 
$t$-test (Church~\etal,~1991b:127), any estimation method that simply adds 
a constant value to each count is significantly worse than \acronym{mle} for 
estimating probabilities themselves, as shown by Gale and Church~(1994).  
For the decision task addressed in this experiment, adding a constant to each 
count would have little effect (some additional weight would be given to the 
sense information provided by the known dependency link $n_2 \rightarrow 
n_3$; see equation~\ref{eq:cy_method_dep_ratio} below).

Adopting a back-off strategy requires having multiple models and choosing 
a criterion for either changing from one model to another or combining their 
predictions.  In an engineered system, making use of available alternative 
models is sensible practice.  From a scientific perspective the situation is 
more complex because the experiment effectively evaluates several models 
at once.  Choosing the back-off criterion forms an extra experimental 
variable.  Furthermore, Collins and Brooks~(1995) discovered that their 
prepositional phrase attachment system performed best when the cut-off 
frequencies were set to zero at every level.  That is, a model was used if just 
one occurrence of an event supported the necessary estimates for that model.  
In the experimental results below, zero counts occur in only a small 
proportion of the test cases, so these results suggest that a back-off strategy 
would be unlikely to help.

Hypothesis testing allows a model to check when it has sufficient data to 
have a given degree of confidence in its analysis.  Examples include the 
$t$-test (Fisher and Riloff,~1992), the likelihood ratio test (Dunning,~1993), 
the $\chi^2$ test (Alshawi and Carter,~1994), and the odds ratio test (Dagan 
and Itai,~1994).  The difficulty here is deciding what to do when the test is 
inconclusive.  Either the system can refuse to make an analysis (in which 
case, the coverage is below 100\%) or it can apply a different model (in 
which case, the system has the same difficulties mentioned for back-off 
strategies).  In either case, hypothesis testing does not change the analysis 
predicted by the model, it simply provides confidence estimates for the 
analysis that the model would have made in any case.

Good-Turing estimation is a powerful smoothing technique with good 
mathematical foundation and Gale and Church~(1994) have shown the 
resulting estimates to be highly accurate predictors of English bigram 
probabilities.  However, there are a few of difficulties in applying the 
Good-Turing method here.  First, the counts of concepts are not real counts 
because they are derived from counting words that express concepts only 
ambiguously, and are therefore not necessarily integers.  Since the method 
relies on construction of the frequency distribution (counts of the number of 
events which were observed to occur with each frequency), a direct 
application of the method is impossible.  Second, the method assumes that 
the frequency distribution is strictly decreasing (the number of types 
occurring $n$ times is greater than the number of types occurring $n+1$ 
times), otherwise it assigns non-monotonic probabilities (that is, events 
observed $n$ times get a higher probability than those observed $n+1$ 
times).  While word bigram frequencies fall off very rapidly, concept bigram 
frequencies have a much flatter frequency distribution.  Finally, reasonable 
results depend on smoothing the frequency distribution, a procedure 
involving further interference by the experimenter, and thus compromising 
the experimental methodology.

All this said, the most commendable characteristic of maximum likelihood 
estimation is the ease by which it can implemented.  For the purposes of this 
experiment it presents a practical compromise.  
Equation~\ref{eq:cy_method_assoc_metric} gives the maximum likelihood 
estimates for the model parameters in terms of the noun pair frequencies, 
$\countfn(w_i, w_j)$. It incorporates the division of counts for
ambiguous words across their different senses.
\begin{eqnarray}
\Pr(t_i \rightarrow t_j \mid t_j) & = &
\frac{1}{\eta_j}
\sum_{
\begin{array}{c}
\scriptstyle w_i \scriptstyle \in \scriptstyle t_i \\
\scriptstyle w_j \scriptstyle \in \scriptstyle t_j
\end{array}
}
\frac
{\countfn(w_i, w_j)}
{\ambig(w_i)\,\,\ambig(w_j)}
\label{eq:cy_method_assoc_metric}  \\
\eta_j & = &
\sum_{
\scriptstyle w_i \scriptstyle \in \scriptstyle {\cal N}
}
\frac
{\countfn(w_i, w_j)}
{\ambig(w_i)\,\,\ambig(w_j)} \nonumber
\end{eqnarray}
Recall from section~\ref{sec:cy_model} that $\ambig(w)$ is the 
number of categories in which $w$ appears.

An important case to notice is when $\countfn(w_i, w_j) = 
0$ for all $w_i \in t_i$ and $w_j \in t_j$.  Where this is the case, 
the probability estimate for $\Pr(t_i \rightarrow t_j \mid t_j)$ is 
zero.  This unfortunate side-effect of using maximum likelihood 
estimates means that the resulting model prohibits any analysis 
involving these relationships (that is, where a word used to 
express the concept represented by $t_i$ modifies a word used 
to express the concept represented by $t_j$).  Apart from the 
minor correction described in the next section, nothing has been 
done to address this difficulty.

It should also be clear from 
equation~\ref{eq:cy_method_assoc_metric} that the conceptual 
association used is asymmetric.  That is, it is generally true that 
$A_{ij} \neq A_{ji}$.  

\subsubsection*{Test set}

To generate the test set, a pattern like that used for training was
passed over the corpus.  It extracts sequences of three nouns:
\begin{equation}
T_{\mbox{test}} = \{ (w_2, w_3, w_4) \mid 
w_1 w_2 w_3 w_4 w_5; w_1,w_5 \notin {\cal N}; w_2, w_3, w_4 \in {\cal 
N} \}
\label{eq:cy_method_testset}
\end{equation}
As noted above, this yielded 625 triples.  Since the model can 
only analyse compounds containing words in the thesaurus, any 
sequences containing words unknown to the thesaurus were 
discarded.  This left 308 triples, which were then manually 
analysed by the author using the entire article in which they 
appeared as context (examples are shown in 
table~\ref{tb:cy_method_test_dist} below).  The entire test set
is given in appendix~\ref{appendix:cytest}.

As for the training set, not all sequences extracted in this way 
were compound nouns.  For the same reasons as described 
above (nouns adjacent across a constituent boundary and 
partially extracted longer compounds) many of the sequences 
were not valid compounds and were marked as errors.  An 
example appears in table~\ref{tb:cy_method_test_dist}.  These 
triples were not used in computing the accuracy of the model.

Another group of triples did form valid compounds, but could 
not be assigned a strictly left-branching or right-branching 
structure in the context.  In principle, the two possible syntactic 
readings of a three word compound noun correspond to distinct 
meanings.  Normally one or other of these meanings will be the 
intended one, even if the other happens to be true of the situation 
being described.  However, in certain contexts, both of these 
meanings appear to be part of the communicative intent of the 
writer.  An example appears in table~\ref{tb:cy_method_test_dist}, 
though one would need more context than I have provided here 
to verify that this is indeed an example.

Hindle and Rooth~(1993:113) observe a similar phenomenon in the 
attachment of prepositional phrases.  They call such examples 
\newterm{semantically indeterminate}.  An example adapted 
from theirs is \lingform{They mined the roads along the coast} in 
which \lingform{along} may attach to either \lingform{mined} or 
\lingform{roads} but means the same both ways.  They observe 77 cases 
out of 880 (although they also have a category called `other' 
containing a further 78 cases).  

Apart from the extraction errors and the semantically 
indeterminate compounds, the remaining 244 triples were all 
assigned a syntactic reading, either left-branching or right-branching.  
The distribution of all 308 triples is shown in 
table~\ref{tb:cy_method_test_dist}.
\begin{table*}
\centering
\begin{tabular}{|l|r|r|l|} \hline
Type  & Number & Proportion & Example \\ \hline \hline
Error & 29 & 9\% & $\begin{array}{l}
	\mbox{In {\em monsoon regions}} \\
	\mbox{{\em rainfall} does not \ldots }
	\end{array}$  \\ \hline
Indeterminate & 35 & 11\% & $\begin{array}{l}
	\mbox{Most advanced aircraft have } \\
	\mbox{{\em precision navigation systems}. }
	\end{array}$  \\ \hline
Left-branching & 163 & 53\% & $\begin{array}{l}
	\mbox{\ldots escaped punishment by } \\
	\mbox{the Allied {\em war crimes tribunals}. }
	\end{array}$  \\ \hline
Right-branching & 81 & 26\% & $\begin{array}{l}
	\mbox{Ronald Reagan, who won two } \\
	\mbox{{\em landslide election victories}, \ldots }
	\end{array}$  \\ \hline
\end{tabular}
\caption{Test set distribution}  \label{tb:cy_method_test_dist}
\end{table*}

An important aspect of choosing the syntactic structure of these 
compounds is their dependence on context.  When the test set 
was analysed, the best reading {\em within the context} was assigned 
to each compound.  However, such readings are not necessarily 
the most likely readings in all contexts, as I discussed in 
section~\ref{sec:cn_context}.  The program doesn't have access to the 
context because it is only given the compound.  It tries to give the best 
out-of-context response (that is, the one most likely across all different 
contexts), but even if it does, it may be scored as incorrect.

One example given in section~\ref{sec:cn_context} is \lingform{club cover 
charge}.  This typically has a right-branching analysis (a fee to 
cover drinks exacted by a nightclub upon entrance), but in a golfing shop 
may be left-branching (the price of a plastic jacket used to 
protect golf sticks).  If this were a test compound and it came from a text 
about golf shops, where it was left-branching, then the program would 
be marked incorrect if it chose a right-branching analysis, even though this is 
the best possible answer without context.

Before I turn to the analysis procedure used to predict the 
readings of compound nouns, two aspects of the test 
methodology need to be mentioned.  First, the test set was 
manually analysed prior to the application of the model described here (or 
indeed any automatic analysis procedure) and so the readings assigned are 
independent of possible biases caused by developing the model.  Second, the 
test and training sets are disjoint (in the sense that no individual occurrence 
of a word appears in both), even though they are taken from the same 
corpus.  This follows from an inspection of the two patterns used 
to extract the training and test sets.  Whenever 
$T_{\mbox{test}}$ applies (see equation~\ref{eq:cy_method_testset}), 
three elements of $\cal N$ appear consecutively.  This sequence is 
prohibited by $T_{\mbox{train}}$ (see 
equation~\ref{eq:cy_method_trainset}).

\subsubsection*{Analysis}  

Given the test set and the parameter estimates derived from the 
training corpus, all that remains is to specify the decision 
procedure used to analyse the test set.  It is motivated directly by 
the probabilistic model, which can be simplified for three word compounds.  
Since there are only two possible analyses, the decision procedure simply 
computes the probabilities of the two alternatives using 
equation~\ref{eq:cy_model_analysis} in section~\ref{sec:cy_model} and 
chooses that one with the greater probability.  

The equation giving the probability of each analysis involves a sum over the 
possible categories for each word.  This must be computed for each analysis 
so that all possible senses of a word contribute to the decision.  For example, 
if $w_1 \in t_1, w_2 \in t_2, w_3 \in t_{3a}$ and $w_3 \in t_{3b}$, then 
four conceptual structures are possible: 
\begin{itemize}
\item $t_1 \rightarrow t_2 \rightarrow t_{3a}$,
\item $t_1 \rightarrow t_{3a} \leftarrow t_2$,
\item $t_1 \rightarrow t_2 \rightarrow t_{3b}$ and
\item $t_1 \rightarrow t_{3b} \leftarrow t_2$.
\end{itemize}
Here, the first and third are left-branching and the second
and last right-branching.  Thus two affinities must be added to compute the 
probability of each analysis.

Notice that this strategy has the advantage that all evidence coming from 
the observation of a word in the training set is collected together 
when analysing that word in the test set, even though it may 
have been split up across several different parameters.  

It is more efficient if, instead of computing each of the two sums 
individually, their ratio is computed, so that some elements cancel out.  For 
the purposes of the experiment, I will assume that all semantic classes are 
the same size, thus allowing one further term to be cancelled.  In 
section~\ref{sec:cy_comparisons} below I will investigate empirically the 
effects of this approximation.  Having cancelled the common factors, we are 
left with the following equation for $R_{dep}$, the ratio of the two desired 
probabilities (under the dependency model).  

\begin{eqnarray} 
R_{\mbox{dep}} \,\, & \stackrel{\rm def}{=} \,\, &
\frac{
\Pr( w_1 w_2 w_3 \mbox{ is left-branching})
}{
\Pr( w_1 w_2 w_3 \mbox{ is right-branching})
}   \nonumber \\
& = &
\frac
{
\sum_{
	\scriptstyle t_i \scriptstyle \in
	    \scriptstyle \mbox{\it cats\/}(\scriptstyle w_i)
	}
\Pr(t_{1} \rightarrow t_{2} \mid t_{2}) \Pr(t_{2} \rightarrow t_{3} \mid 
t_{3})
}
{
\sum_{
	\scriptstyle t_i \scriptstyle \in
	   \scriptstyle \mbox{\it cats\/}(\scriptstyle w_i)
	}
\Pr(t_{1} \rightarrow t_{3} \mid t_{3}) \Pr(t_{2} \rightarrow t_{3} \mid 
t_{3})
}
\label{eq:cy_method_dep_ratio}
\end{eqnarray}
For the moment, I have omitted from this equation the factor of 2 resulting 
from the differences in $\mbox{\it choice\/}(m)$ between left-branching and 
right-branching modificational structures.  As observed in 
section~\ref{sec:cy_model}, right-branching modification structures have 
$\mbox{\it choice\/}(m) = 2$, while left-branching ones have 
$\mbox{\it choice\/}(m) = 1$.  This factor has been left out in order to make a 
fairer comparison between the adjacency and dependency methods, even 
though it is warranted by the assumptions underpinning the dependency 
model.  In this way, I avoid the criticism that this factor is an \foreign{ad 
hoc} bias toward the more common analysis in the test set and unfairly 
advantages the dependency model.  The factor will be restored in 
section~\ref{sec:cy_comparisons} under the designation tuned model.

Having computed $R_{\mbox{dep}}$, the decision procedure then assigns 
a right-branching analysis if and only if this ratio is less than 1.  When the 
ratio is exactly 1, the program assigns a left-branching analysis as the 
default.

One problem with the above formulation is the possibility of the 
denominator being zero.  This can easily occur, since many of 
the parameter estimates will be zero.  When this occurs, 
the decision procedure chooses a left-branching analysis.  The 
reasoning goes as follows:  there is no apparent evidence to 
support a right-branching analysis.  If there is evidence for a 
left-branching analysis, then we should choose that.  If there is 
no evidence for either, then the best we can do is take the 
default, which again is left-branching.

A particularly bad example of the effect of data sparseness 
when using maximum likelihood estimates appears in this 
equation.  Both numerator and denominator contain the 
multiplicand $\Pr(t_{2} \rightarrow t_{3} \mid t_{3})$.  This factor 
does not cancel out, because it varies across the terms of the 
sums.  If it happens to be zero in all of the terms of the sums, 
then both the numerator and denominator will be zero and a left-branching 
default will be selected.  Yet the test example itself 
provides the information that $w_2$ can be a modifier of 
$w_3$.  Thus, there must be some $t_{2}$ and $t_{3}$ for which 
$\Pr(t_{2} \rightarrow t_{3} \mid t_{3})$ is actually non-zero (even 
though it has been estimated to be zero).  

To avoid this problem, the implementation specifically checks for this 
occurrence.  When it is detected, the ratio is recomputed under the 
assumption that all of the parameters $\Pr(t_{2} \rightarrow t_{3} 
\mid t_{3})$ are non-zero and equal to each other.  Under this 
assumption, these terms do cancel out and the ratio only depends 
on the various $\Pr(t_{1} \rightarrow t_{3} \mid t_{3})$ and $\Pr(t_{1} 
\rightarrow t_{2} \mid t_{2})$.  This correction only applies to estimates for 
$\Pr(t_{2} \rightarrow t_{3} \mid t_{3})$.  No correction is made to the 
probability estimates for $\Pr(t_1 \rightarrow t_2)$ and $\Pr(t_1 \rightarrow 
t_3)$ in unseen cases; if unseen, their probability is estimated as zero.

Let's consider an example of the decision procedure.  Suppose 
the program is presented with the compound \lingform{highland 
outrigger tour}.  The word \lingform{highland} appears in exactly one
category, \tc{land}{342} ($\tla$), 
while \lingform{tour} appears in \tc{journey}{266} ($\tjo$).
Meanwhile \lingform{outrigger} appears in two categories: 
\tc{support}{215} ($\tsu$) and \tc{ship}{273} ($\tsh$).
Because there is just one, binary, sense ambiguity, 
each of the sums in equation~\ref{eq:cy_method_dep_ratio} has 
just two terms.  The ratio therefore becomes:
\begin{eqnarray}
\lefteqn{R_{\mbox{dep}} =} \label{eq:cy_method_rdep_hot} \\
 & & \frac{ 
 \Pr(\tla \rightarrow \tsu \mid \tsu) \Pr(\tsu \rightarrow \tjo \mid \tjo)
+ \Pr(\tla \rightarrow \tsh \mid \tsh) \Pr(\tsh \rightarrow \tjo \mid \tjo)
}
{
 \Pr(\tla \rightarrow \tjo \mid \tjo) \Pr(\tsu \rightarrow \tjo \mid \tjo)
+ \Pr(\tla \rightarrow \tjo \mid \tjo) \Pr(\tsh \rightarrow \tjo \mid \tjo)
} \nonumber
\end{eqnarray}

There are four possible conceptual structures modelled in the 
equation.  The two right-branching ones (in the denominator) 
include a modification relationship between the concepts 
represented by \tc{land}{342} and \tc{journey}{266}.  The first 
left-branching one involves a relationship between \tc{land}{342} 
and \tc{support}{215}, while the second contains a 
relationship between \tc{land}{342} and \tc{ship}{273}.  Since 
both \tc{land}{342} $\rightarrow$ \tc{ship}{273} and \tc{land}{342}
$\rightarrow$ \tc{support}{215} are unlikely 
relationships, the numerator is small.  The first term of the 
denominator is likewise small because \tc{support}{215} 
$\rightarrow$ \tc{journey}{266} is reasonably unlikely.  The 
second term however will be significantly larger because both 
\tc{land}{342} $\rightarrow$ \tc{journey}{266} and \tc{ship}{273} 
$\rightarrow$ \tc{journey}{266} are plausible compounds.  Therefore 
$R_{\mbox{dep}}$ is less than 1 and a right-branching analysis is 
preferred.

If both $\Pr(\tsh \rightarrow \tjo \mid \tjo)$ and 
$\Pr(\tsu \rightarrow \tjo \mid \tjo)$ were estimated to be 
zero (only possible because of data sparseness since \tc{ship}{273} 
$\rightarrow$ \tc{journey}{266} is quite plausible), 
then the ratio would be recomputed as:
\begin{equation}
R_{\mbox{dep}} = \frac
{\Pr(\tla \rightarrow \tsu \mid \tsu) + \Pr(\tla 
\rightarrow \tsh \mid \tsh)}
{2 \Pr(\tla \rightarrow \tjo \mid \tjo)}
\label{eq:cy_method_rdep_hot_sparse}
\end{equation}

\subsection{Results} 
\label{sec:cy_results}


Each of the 244 compounds from the test set that received either 
a left-branching or right-branching analysis were passed to a 
program which implemented the decision procedure just described.    
The distribution of responses is shown in 
table~\ref{tb:cy_results_all}.  

\begin{table*}
\centering
\begin{tabular}{|l|c|c|} \hline 
	&	Actual Left	&	Actual Right \\
\hline Response Left	&	132	&	24 \\
\hline Response Right	&	31	&	57 \\	
\hline
\end{tabular}
\caption{Response distribution}
\label{tb:cy_results_all}
\end{table*}

The proportion of correct responses was 77.5\%.  If we instead 
always chose a left-branching analysis, we would get 66.8\%.  An 
important question is whether the observed performance is really 
better than this simple strategy, allowing for statistical 
fluctuations.  The standard statistical test for comparing the 
difference of two means when the sample sizes are large 
involves computing a pooled estimate of the proportion in 
both samples for estimating the variance.  Since the samples are 
sufficiently large,  it is possible to use $z$-scores to estimate the 
probability of seeing the observed difference in proportions by 
chance.  In this case, the pooled estimate of the proportion is 
72.1\%, which yields a $z$-score of 2.64.  Therefore there is less 
than 0.5\% chance of observing this, or a greater, difference in 
proportions by chance alone (one-tailed test).  So the observed 
difference between the experimental unit's correctness and that
of a simple guessing strategy is significant at the 1\% level.


The decision procedure defaults to a left-branching response 
when it has no evidence to support either analysis.  This 
happened in only 9 test cases (3.69\%), so the choice of default 
did not influence the performance greatly.\footnote{Six of these nine are 
left-branching, so the expected accuracy of an algorithm using uniform 
random choice as the default would be just 0.6\% lower.}  A further 76 cases 
(31.1\%) had evidence for only one of the analyses, and the 
distribution of these cases is shown in 
table~\ref{tb:cy_results_sparse}.  The correctness in these 
cases was not appreciably different from the overall correctness.  
This suggests that parameter estimates of zero were not especially
less accurate than other parameter estimates.

\begin{table*}
\centering
\begin{tabular}{|l|c|c|} \hline 
	&	Actual Left	&	Actual Right \\
\hline Response Left	&	20	&	1 \\
\hline Response Right	&	17	&	38 \\
\hline
\end{tabular}
\caption{Response distribution for cases with evidence for only 
one analysis}
\label{tb:cy_results_sparse}
\end{table*}

Interestingly, there was only one right-branching compound that 
had evidence for just a left-branching analysis 
(the compound is \lingform{reservoir evaporation losses}).  A possible 
explanation is that writers try to avoid right-branching 
compounds unless there is either no possibility of a 
left-branching reading, or a strong association between the head 
and its more distant modifier.

A side effect of using conceptual association is that words must be
assigned to concepts in order to calculate the relative probabilities of the 
different analyses.  Each term in the sums appearing in 
equation~\ref{eq:cy_method_dep_ratio} represents an assignment 
of words in the compound to concepts.  By comparing the 
relative sizes of these terms, we can estimate the likelihood of 
different possible senses for a word.  For instance, in the
\lingform{highland outrigger tour} example at the end of 
section~\ref{sec:cy_method} above, the most significant
term in equation~\ref{eq:cy_method_rdep_hot} corresponds to
$ \tsh \rightarrow \tjo \leftarrow \tla$.  Therefore
the program has evidence that \lingform{outrigger} is used in the sense
of \tc{ship}{273} rather than \tc{support}{215}.
Of course, this evidence only reflects the information provided by other 
words in the compound (in fact, only those other words which are modifiers 
or heads of the word in question).  While this amount of context 
is admittedly very narrow, it is still useful, especially since 
providing sense cues is one of the linguistic functions of 
compounding (the reason for the first word in \lingform{golf club} is 
the ambiguity of the second).  However, I have not measured the 
performance of this strategy for sense disambiguation.

The initial experimental goal was to measure the accuracy of the
model in predicting the parses of three word noun compounds.
This has been achieved with the result being 77.5\%.  In the next
section I will describe additional experiments in which
different aspects of the model are varied.

\subsection{Comparisons} 
\label{sec:cy_comparisons}

In the experiments below comparisons are made across six dimensions.
\begin{enumerate}
\item Analysis method: The performance of the dependency model is 
compared to that of an equivalent adjacency algorithm.
\item Training pattern: A range of different training patterns are used
to estimate the model parameters, including windowed association.
\item Training symmetry: A symmetric training scheme is
compared to the asymmetric one.
\item Tuning factors: The models are applied both with and without 
the tuning factors suggested by the meaning distributions theory.
\item Parameterisation: The concept based model is compared to a lexical one.
\item Tagged data: Training data is gathered by relying on the predictions
of an automatic tagger, instead of on the heuristic used earlier.
\end{enumerate}
This section will detail each of these comparisons.

\subsubsection*{Dependency meets adjacency}

Since adjacency algorithms have been suggested a few times in the 
literature, the first comparison is between the dependency method 
and an equivalent adjacency method.
The decision procedure for the adjacency method differs from the 
dependency method only in the way in which the parameters are used.  

To obtain an adjacency algorithm, it is 
sufficient to rewrite equation~\ref{eq:cy_method_dep_ratio} to use only 
adjacent pairs of words.  That is, when given the compound ``$w_1 w_2 
w_3$'', the adjacency method computes the following ratio and assigns a 
right-branching structure if and only if it is less than 1.
\begin{equation} \label{eq:cy_adjacency_adj_ratio}
R_{\mbox{adj}}
\,\, = \,\,
\frac
{
\sum_{
	\scriptstyle t_i \scriptstyle \in
	   \mbox{\it \scriptsize cats($\scriptstyle
w_i\!$)}
	}
\Pr(t_1 \rightarrow t_2)
}
{
\sum_{
	\scriptstyle t_i \scriptstyle \in
	   \mbox{\it \scriptsize cats($\scriptstyle
w_i\!$)}
	}
\Pr(t_2 \rightarrow t_3)
}
\end{equation}
Notice that the correction applied in section~\ref{sec:cy_method} when the 
common factor in equation~\ref{eq:cy_method_dep_ratio} is zero, does not 
apply here because there is only one factor in each term of the sum.

The training pattern used earlier to estimate the parameters of the dependency 
model is derived from that model, so the experiment also uses a range of 
alternative training schemes, all unsupervised.  One of these results in the 
same training set as that proposed by Liberman and Sproat~(1992) for their 
adjacency algorithm.\footnote{Specifically, the windowed scheme with $n = 
2$.}

The alternative schemes use a window to collect training instances by 
observing how often a pair of nouns co-occur within some fixed number of 
words.  A variety of window sizes are used.  For each, the training set is 
given by 
\begin{equation}
T_{\mbox{train}} = \{ (w_1, w_i) \mid w_1 w_2 \ldots w_i; 1 < i \leq n; 
w_1,w_i \in {\cal N} \}
\label{eq:cy_adjacency_windowtrain}
\end{equation}
where $w_1 w_2 \ldots w_i$ denotes the occurrence of $i$ tokens
in sequence in the corpus, and $n \geq 2$ is the window width. Note that, 
just as for the pattern, windowed counts are asymmetric.

To ensure that the test set is disjoint from the training data, all occurrences 
of the test noun compounds have been removed from the training corpus.  
Each training scheme yields counts of noun pairs which are used to compute 
maximum likelihood estimates of the parameters as before using 
equation~\ref{eq:cy_method_assoc_metric} in 
section~\ref{sec:cy_method}.



Eight different training schemes have been used 
and each set of estimates used to analyse the test set under both the 
adjacency and the dependency method.  The schemes used are:
\begin{itemize}
\item the pattern given by equation~\ref{eq:cy_method_trainset} in section 
\ref{sec:cy_method}; and
\item windowed training schemes with window widths of 2, 3, 4, 5, 10, 50 
and 100 words.
\end{itemize}

\begin{figure*}
\centering
\input{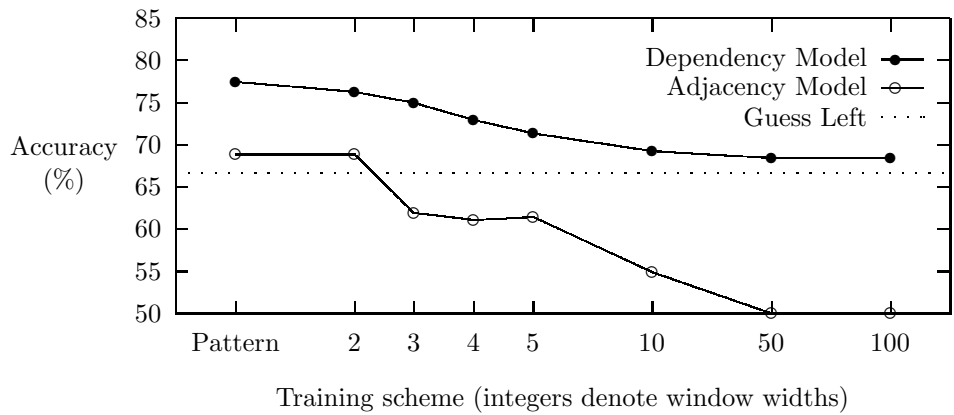}
\caption{Accuracy of dependency and adjacency method for
various training schemes} \label{fig:cy_adjacency_dva_accuracy}
\end{figure*}

The accuracy on the test set for all these
experiments is shown in figure \ref{fig:cy_adjacency_dva_accuracy}.
As can be seen, the dependency model is more accurate than the 
adjacency model.  This is true across the whole spectrum of training 
schemes.  In the case of the pattern training scheme, the
difference between 68.9\% for adjacency and 77.5\% for dependency results 
in a pooled $z$-score of 2.14, which is statistically significant at the 5\%
level ($p = 0.0162$; one-tailed test), demonstrating the superiority
of the dependency model, at least for the compounds within Grolier's 
encyclopedia.

In no case do any of the windowed training schemes outperform 
the pattern scheme.  It seems that additional instances admitted
by the windowed schemes are too noisy to make an improvement.  
While using windowed co-occurrence did not help here, it is possible
that under more data sparse conditions better performance could
be achieved by this method.

\begin{figure*}
\centering
 


 \setlength{\unitlength}{0.240900pt}
 \ifx\plotpoint\undefined\newsavebox{\plotpoint}\fi
 \sbox{\plotpoint}{\rule[-0.200pt]{0.400pt}{0.400pt}}%
 \begin{picture}(1500,600)(0,0)
 \font\gnuplot=cmr10 at 10pt
 \gnuplot
 \sbox{\plotpoint}{\rule[-0.200pt]{0.400pt}{0.400pt}}%
 \put(220.0,113.0){\rule[-0.200pt]{292.934pt}{0.400pt}}
 \put(220.0,113.0){\rule[-0.200pt]{4.818pt}{0.400pt}}
 \put(198,113){\makebox(0,0)[r]{0}}
 \put(1416.0,113.0){\rule[-0.200pt]{4.818pt}{0.400pt}}
 \put(220.0,206.0){\rule[-0.200pt]{4.818pt}{0.400pt}}
 \put(198,206){\makebox(0,0)[r]{1}}
 \put(1416.0,206.0){\rule[-0.200pt]{4.818pt}{0.400pt}}
 \put(220.0,299.0){\rule[-0.200pt]{4.818pt}{0.400pt}}
 \put(198,299){\makebox(0,0)[r]{2}}
 \put(1416.0,299.0){\rule[-0.200pt]{4.818pt}{0.400pt}}
 \put(220.0,391.0){\rule[-0.200pt]{4.818pt}{0.400pt}}
 \put(198,391){\makebox(0,0)[r]{3}}
 \put(1416.0,391.0){\rule[-0.200pt]{4.818pt}{0.400pt}}
 \put(220.0,484.0){\rule[-0.200pt]{4.818pt}{0.400pt}}
 \put(198,484){\makebox(0,0)[r]{4}}
 \put(1416.0,484.0){\rule[-0.200pt]{4.818pt}{0.400pt}}
 \put(220.0,577.0){\rule[-0.200pt]{4.818pt}{0.400pt}}
 \put(198,577){\makebox(0,0)[r]{5}}
 \put(1416.0,577.0){\rule[-0.200pt]{4.818pt}{0.400pt}}
 \put(220.0,113.0){\rule[-0.200pt]{0.400pt}{4.818pt}}
 \put(220.0,557.0){\rule[-0.200pt]{0.400pt}{4.818pt}}
 \put(314.0,113.0){\rule[-0.200pt]{0.400pt}{4.818pt}}
 \put(314,68){\makebox(0,0){Pattern}}
 \put(314.0,557.0){\rule[-0.200pt]{0.400pt}{4.818pt}}
 \put(501.0,113.0){\rule[-0.200pt]{0.400pt}{4.818pt}}
 \put(501,68){\makebox(0,0){2}}
 \put(501.0,557.0){\rule[-0.200pt]{0.400pt}{4.818pt}}
 \put(594.0,113.0){\rule[-0.200pt]{0.400pt}{4.818pt}}
 \put(594,68){\makebox(0,0){3}}
 \put(594.0,557.0){\rule[-0.200pt]{0.400pt}{4.818pt}}
 \put(688.0,113.0){\rule[-0.200pt]{0.400pt}{4.818pt}}
 \put(688,68){\makebox(0,0){4}}
 \put(688.0,557.0){\rule[-0.200pt]{0.400pt}{4.818pt}}
 \put(781.0,113.0){\rule[-0.200pt]{0.400pt}{4.818pt}}
 \put(781,68){\makebox(0,0){5}}
 \put(781.0,557.0){\rule[-0.200pt]{0.400pt}{4.818pt}}
 \put(968.0,113.0){\rule[-0.200pt]{0.400pt}{4.818pt}}
 \put(968,68){\makebox(0,0){10}}
 \put(968.0,557.0){\rule[-0.200pt]{0.400pt}{4.818pt}}
 \put(1155.0,113.0){\rule[-0.200pt]{0.400pt}{4.818pt}}
 \put(1155,68){\makebox(0,0){50}}
 \put(1155.0,557.0){\rule[-0.200pt]{0.400pt}{4.818pt}}
 \put(1342.0,113.0){\rule[-0.200pt]{0.400pt}{4.818pt}}
 \put(1342,68){\makebox(0,0){100}}
 \put(1342.0,557.0){\rule[-0.200pt]{0.400pt}{4.818pt}}
 \put(220.0,113.0){\rule[-0.200pt]{292.934pt}{0.400pt}}
 \put(1436.0,113.0){\rule[-0.200pt]{0.400pt}{111.778pt}}
 \put(220.0,577.0){\rule[-0.200pt]{292.934pt}{0.400pt}}
 \put(45,345){\makebox(0,0){\shortstack{Guess Rate\\(\%)}}}
 \put(828,-22){\makebox(0,0){Training scheme (integers denote window widths)}}
 \put(220.0,113.0){\rule[-0.200pt]{0.400pt}{111.778pt}}
 \put(1306,512){\makebox(0,0)[r]{Dependency Model}}
 \put(1328.0,512.0){\rule[-0.200pt]{15.899pt}{0.400pt}}
 \put(314,455){\usebox{\plotpoint}}
 \multiput(501.58,449.84)(0.499,-1.433){183}{\rule{0.120pt}{1.244pt}}
 \multiput(500.17,452.42)(93.000,-263.418){2}{\rule{0.400pt}{0.622pt}}
 \multiput(594.00,187.92)(0.618,-0.499){149}{\rule{0.595pt}{0.120pt}}
 \multiput(594.00,188.17)(92.766,-76.000){2}{\rule{0.297pt}{0.400pt}}
 \put(314.0,455.0){\rule[-0.200pt]{45.048pt}{0.400pt}}
 \put(1350,512){\makebox(0,0){$\times$}}
 \put(314,455){\makebox(0,0){$\times$}}
 \put(501,455){\makebox(0,0){$\times$}}
 \put(594,189){\makebox(0,0){$\times$}}
 \put(688,113){\makebox(0,0){$\times$}}
 \put(781,113){\makebox(0,0){$\times$}}
 \put(968,113){\makebox(0,0){$\times$}}
 \put(1155,113){\makebox(0,0){$\times$}}
 \put(1342,113){\makebox(0,0){$\times$}}
 \put(688.0,113.0){\rule[-0.200pt]{157.549pt}{0.400pt}}
 \put(1306,467){\makebox(0,0)[r]{Adjacency Model}}
 \put(1328.0,467.0){\rule[-0.200pt]{15.899pt}{0.400pt}}
 \put(314,455){\usebox{\plotpoint}}
 \multiput(501.58,449.84)(0.499,-1.433){183}{\rule{0.120pt}{1.244pt}}
 \multiput(500.17,452.42)(93.000,-263.418){2}{\rule{0.400pt}{0.622pt}}
 \multiput(594.00,187.92)(0.618,-0.499){149}{\rule{0.595pt}{0.120pt}}
 \multiput(594.00,188.17)(92.766,-76.000){2}{\rule{0.297pt}{0.400pt}}
 \put(314.0,455.0){\rule[-0.200pt]{45.048pt}{0.400pt}}
 \put(1350,467){\circle{18}}
 \put(314,455){\circle{18}}
 \put(501,455){\circle{18}}
 \put(594,189){\circle{18}}
 \put(688,113){\circle{18}}
 \put(781,113){\circle{18}}
 \put(968,113){\circle{18}}
 \put(1155,113){\circle{18}}
 \put(1342,113){\circle{18}}
 \put(688.0,113.0){\rule[-0.200pt]{157.549pt}{0.400pt}}
 \end{picture}
 
\caption{Guess rates of dependency and adjacency method for
various training schemes} \label{fig:cy_adjacency_dva_guess}
\end{figure*}
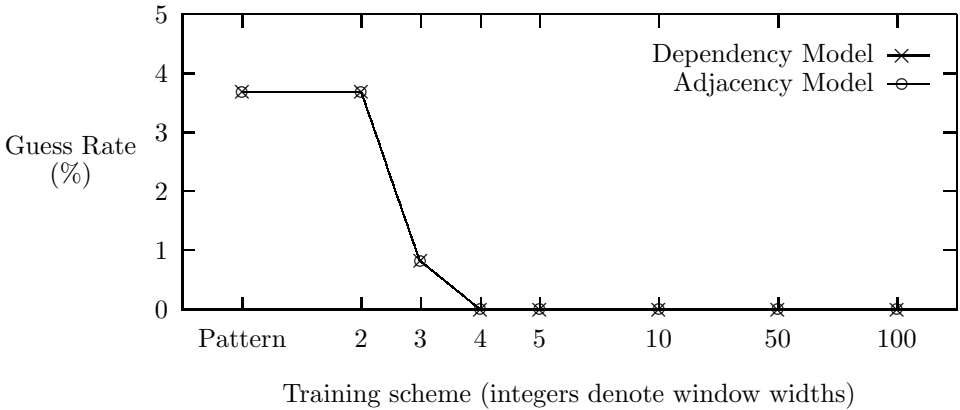

The proportion of cases in which the procedure was forced to 
guess, either because no data supported either 
analysis or because both were equally supported, is shown
in figure~\ref{fig:cy_adjacency_dva_guess} and is 
quite low for both methods.  For the pattern and two word window training
schemes, the guess rate is less than 4\% for both models.  
In the three word window training scheme, the guess
rates are less than 1\%.  For all larger windows,
neither method is ever forced to guess.
It is a coincidence of the particular test and training sets
that the guess rates of both models are always equal in this figure.

Initial results from applying these methods to the \acronym{ema} corpus on 
ceramic materials have been obtained by ter Stal~(1995), and support the 
conclusion that the dependency model is superior to the adjacency model.

\subsubsection*{Symmetric association}

The association used so far is asymmetric.  
It is possible that the windowed training schemes could benefit from a 
symmetric association, since that would permit paraphrases of compounds 
involving post-modifiers to be included as training instances.  
For example, observing \lingform{sports played in summer} would be used 
as evidence that \lingform{summer} is a likely modifier of \lingform{sports} 
(assuming that the window width was at least 4).  This would provide a way 
of reducing data sparseness too, since it would halve the number of free 
parameters.

\begin{figure}
\centering
%

\setlength{\unitlength}{0.240900pt}
\ifx\plotpoint\undefined\newsavebox{\plotpoint}\fi
\begin{picture}(1500,600)(0,0)
\font\gnuplot=cmr10 at 10pt
\gnuplot
\sbox{\plotpoint}{\rule[-0.200pt]{0.400pt}{0.400pt}}%
\put(220.0,113.0){\rule[-0.200pt]{292.934pt}{0.400pt}}
\put(220.0,113.0){\rule[-0.200pt]{4.818pt}{0.400pt}}
\put(198,113){\makebox(0,0)[r]{0}}
\put(1416.0,113.0){\rule[-0.200pt]{4.818pt}{0.400pt}}
\put(220.0,206.0){\rule[-0.200pt]{4.818pt}{0.400pt}}
\put(198,206){\makebox(0,0)[r]{1}}
\put(1416.0,206.0){\rule[-0.200pt]{4.818pt}{0.400pt}}
\put(220.0,299.0){\rule[-0.200pt]{4.818pt}{0.400pt}}
\put(198,299){\makebox(0,0)[r]{2}}
\put(1416.0,299.0){\rule[-0.200pt]{4.818pt}{0.400pt}}
\put(220.0,391.0){\rule[-0.200pt]{4.818pt}{0.400pt}}
\put(198,391){\makebox(0,0)[r]{3}}
\put(1416.0,391.0){\rule[-0.200pt]{4.818pt}{0.400pt}}
\put(220.0,484.0){\rule[-0.200pt]{4.818pt}{0.400pt}}
\put(198,484){\makebox(0,0)[r]{4}}
\put(1416.0,484.0){\rule[-0.200pt]{4.818pt}{0.400pt}}
\put(220.0,577.0){\rule[-0.200pt]{4.818pt}{0.400pt}}
\put(198,577){\makebox(0,0)[r]{5}}
\put(1416.0,577.0){\rule[-0.200pt]{4.818pt}{0.400pt}}
\put(220.0,113.0){\rule[-0.200pt]{0.400pt}{4.818pt}}
\put(220.0,557.0){\rule[-0.200pt]{0.400pt}{4.818pt}}
\put(372.0,113.0){\rule[-0.200pt]{0.400pt}{4.818pt}}
\put(372,68){\makebox(0,0){Pattern}}
\put(372.0,557.0){\rule[-0.200pt]{0.400pt}{4.818pt}}
\put(676.0,113.0){\rule[-0.200pt]{0.400pt}{4.818pt}}
\put(676,68){\makebox(0,0){2}}
\put(676.0,557.0){\rule[-0.200pt]{0.400pt}{4.818pt}}
\put(828.0,113.0){\rule[-0.200pt]{0.400pt}{4.818pt}}
\put(828,68){\makebox(0,0){3}}
\put(828.0,557.0){\rule[-0.200pt]{0.400pt}{4.818pt}}
\put(1132.0,113.0){\rule[-0.200pt]{0.400pt}{4.818pt}}
\put(1132,68){\makebox(0,0){5}}
\put(1132.0,557.0){\rule[-0.200pt]{0.400pt}{4.818pt}}
\put(1436.0,113.0){\rule[-0.200pt]{0.400pt}{4.818pt}}
\put(1436,68){\makebox(0,0){10}}
\put(1436.0,557.0){\rule[-0.200pt]{0.400pt}{4.818pt}}
\put(220.0,113.0){\rule[-0.200pt]{292.934pt}{0.400pt}}
\put(1436.0,113.0){\rule[-0.200pt]{0.400pt}{111.778pt}}
\put(220.0,577.0){\rule[-0.200pt]{292.934pt}{0.400pt}}
\put(45,345){\makebox(0,0){\shortstack{Guess Rate\\(\%)}}}
\put(828,-22){\makebox(0,0){Training scheme (integers denote window widths)}}
\put(220.0,113.0){\rule[-0.200pt]{0.400pt}{111.778pt}}
\put(1306,512){\makebox(0,0)[r]{Symmetric Dependency}}
\put(1328.0,512.0){\rule[-0.200pt]{15.899pt}{0.400pt}}
\put(372,379){\usebox{\plotpoint}}
\multiput(676.58,376.51)(0.499,-0.625){301}{\rule{0.120pt}{0.600pt}}
\multiput(675.17,377.75)(152.000,-188.755){2}{\rule{0.400pt}{0.300pt}}
\multiput(828.00,187.92)(2.006,-0.499){149}{\rule{1.700pt}{0.120pt}}
\multiput(828.00,188.17)(300.472,-76.000){2}{\rule{0.850pt}{0.400pt}}
\put(372.0,379.0){\rule[-0.200pt]{73.234pt}{0.400pt}}
\put(1350,512){\raisebox{-.8pt}{\makebox(0,0){$\Diamond$}}}
\put(372,379){\raisebox{-.8pt}{\makebox(0,0){$\Diamond$}}}
\put(676,379){\raisebox{-.8pt}{\makebox(0,0){$\Diamond$}}}
\put(828,189){\raisebox{-.8pt}{\makebox(0,0){$\Diamond$}}}
\put(1132,113){\raisebox{-.8pt}{\makebox(0,0){$\Diamond$}}}
\put(1436,113){\raisebox{-.8pt}{\makebox(0,0){$\Diamond$}}}
\put(1132.0,113.0){\rule[-0.200pt]{73.234pt}{0.400pt}}
\put(372,455){\usebox{\plotpoint}}
\multiput(372,455)(20.756,0.000){15}{\usebox{\plotpoint}}
\multiput(676,455)(10.298,-18.021){15}{\usebox{\plotpoint}}
\multiput(828,189)(20.136,-5.034){15}{\usebox{\plotpoint}}
\multiput(1132,113)(20.756,0.000){15}{\usebox{\plotpoint}}
\put(1436,113){\usebox{\plotpoint}}
\put(1306,467){\makebox(0,0)[r]{Symmetric Adjacency}}
\put(1328.0,467.0){\rule[-0.200pt]{15.899pt}{0.400pt}}
\put(372,341){\usebox{\plotpoint}}
\multiput(676.58,338.51)(0.499,-0.625){301}{\rule{0.120pt}{0.600pt}}
\multiput(675.17,339.75)(152.000,-188.755){2}{\rule{0.400pt}{0.300pt}}
\multiput(828.00,149.92)(4.032,-0.498){73}{\rule{3.300pt}{0.120pt}}
\multiput(828.00,150.17)(297.151,-38.000){2}{\rule{1.650pt}{0.400pt}}
\put(372.0,341.0){\rule[-0.200pt]{73.234pt}{0.400pt}}
\put(1350,467){\circle{18}}
\put(372,341){\circle{18}}
\put(676,341){\circle{18}}
\put(828,151){\circle{18}}
\put(1132,113){\circle{18}}
\put(1436,113){\circle{18}}
\put(1132.0,113.0){\rule[-0.200pt]{73.234pt}{0.400pt}}
\put(372,455){\usebox{\plotpoint}}
\multiput(372,455)(20.756,0.000){15}{\usebox{\plotpoint}}
\multiput(676,455)(10.298,-18.021){15}{\usebox{\plotpoint}}
\multiput(828,189)(20.136,-5.034){15}{\usebox{\plotpoint}}
\multiput(1132,113)(20.756,0.000){15}{\usebox{\plotpoint}}
\put(1436,113){\usebox{\plotpoint}}
\end{picture}
\caption{Guess rates of symmetric association for
various training schemes} \label{fig:cy_adjacency_sym_guess}
\end{figure}
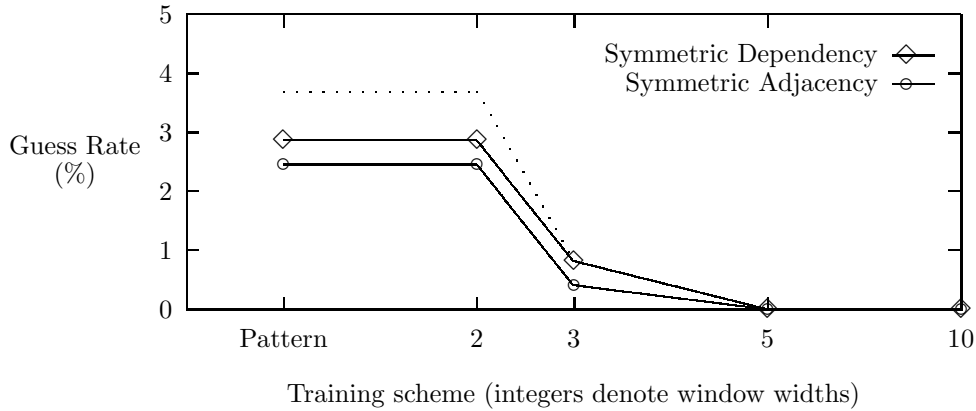

In order to test whether symmetric association schemes could make a 
significant difference to the performance of the program, five of the training 
schemes have been repeated using symmetric counts (that is, whenever 
$(w_1, w_2)$ is observed as a training instance increment the count for 
$(w_2, w_1)$ as well).  The results are shown in 
figures~\ref{fig:cy_adjacency_sym_accuracy} 
and~\ref{fig:cy_adjacency_sym_guess}.  The accuracy figures for 
asymmetric association are shown with a broken line.  As expected, the 
guess rates fall, but the accuracy also drops slightly.  Only for relatively 
large window widths does symmetric association outperform the asymmetric 
schemes.  The dependency model still displays a markedly
greater accuracy across all training schemes.
 
\begin{figure}
\centering

\setlength{\unitlength}{0.240900pt}
\ifx\plotpoint\undefined\newsavebox{\plotpoint}\fi
\begin{picture}(1500,600)(0,0)
\font\gnuplot=cmr10 at 10pt
\gnuplot
\sbox{\plotpoint}{\rule[-0.200pt]{0.400pt}{0.400pt}}%
\put(220.0,113.0){\rule[-0.200pt]{4.818pt}{0.400pt}}
\put(198,113){\makebox(0,0)[r]{50}}
\put(1416.0,113.0){\rule[-0.200pt]{4.818pt}{0.400pt}}
\put(220.0,179.0){\rule[-0.200pt]{4.818pt}{0.400pt}}
\put(198,179){\makebox(0,0)[r]{55}}
\put(1416.0,179.0){\rule[-0.200pt]{4.818pt}{0.400pt}}
\put(220.0,246.0){\rule[-0.200pt]{4.818pt}{0.400pt}}
\put(198,246){\makebox(0,0)[r]{60}}
\put(1416.0,246.0){\rule[-0.200pt]{4.818pt}{0.400pt}}
\put(220.0,312.0){\rule[-0.200pt]{4.818pt}{0.400pt}}
\put(198,312){\makebox(0,0)[r]{65}}
\put(1416.0,312.0){\rule[-0.200pt]{4.818pt}{0.400pt}}
\put(220.0,378.0){\rule[-0.200pt]{4.818pt}{0.400pt}}
\put(198,378){\makebox(0,0)[r]{70}}
\put(1416.0,378.0){\rule[-0.200pt]{4.818pt}{0.400pt}}
\put(220.0,444.0){\rule[-0.200pt]{4.818pt}{0.400pt}}
\put(198,444){\makebox(0,0)[r]{75}}
\put(1416.0,444.0){\rule[-0.200pt]{4.818pt}{0.400pt}}
\put(220.0,511.0){\rule[-0.200pt]{4.818pt}{0.400pt}}
\put(198,511){\makebox(0,0)[r]{80}}
\put(1416.0,511.0){\rule[-0.200pt]{4.818pt}{0.400pt}}
\put(220.0,577.0){\rule[-0.200pt]{4.818pt}{0.400pt}}
\put(198,577){\makebox(0,0)[r]{85}}
\put(1416.0,577.0){\rule[-0.200pt]{4.818pt}{0.400pt}}
\put(220.0,113.0){\rule[-0.200pt]{0.400pt}{4.818pt}}
\put(220.0,557.0){\rule[-0.200pt]{0.400pt}{4.818pt}}
\put(372.0,113.0){\rule[-0.200pt]{0.400pt}{4.818pt}}
\put(372,68){\makebox(0,0){Pattern}}
\put(372.0,557.0){\rule[-0.200pt]{0.400pt}{4.818pt}}
\put(676.0,113.0){\rule[-0.200pt]{0.400pt}{4.818pt}}
\put(676,68){\makebox(0,0){2}}
\put(676.0,557.0){\rule[-0.200pt]{0.400pt}{4.818pt}}
\put(828.0,113.0){\rule[-0.200pt]{0.400pt}{4.818pt}}
\put(828,68){\makebox(0,0){3}}
\put(828.0,557.0){\rule[-0.200pt]{0.400pt}{4.818pt}}
\put(1132.0,113.0){\rule[-0.200pt]{0.400pt}{4.818pt}}
\put(1132,68){\makebox(0,0){5}}
\put(1132.0,557.0){\rule[-0.200pt]{0.400pt}{4.818pt}}
\put(1436.0,113.0){\rule[-0.200pt]{0.400pt}{4.818pt}}
\put(1436,68){\makebox(0,0){10}}
\put(1436.0,557.0){\rule[-0.200pt]{0.400pt}{4.818pt}}
\put(220.0,113.0){\rule[-0.200pt]{292.934pt}{0.400pt}}
\put(1436.0,113.0){\rule[-0.200pt]{0.400pt}{111.778pt}}
\put(220.0,577.0){\rule[-0.200pt]{292.934pt}{0.400pt}}
\put(45,345){\makebox(0,0){\shortstack{Accuracy\\(\%)}}}
\put(828,-22){\makebox(0,0){Training scheme (integers denote window widths)}}
\put(220.0,113.0){\rule[-0.200pt]{0.400pt}{111.778pt}}
\put(1306,512){\makebox(0,0)[r]{Symmetric Dependency}}
\put(1328.0,512.0){\rule[-0.200pt]{15.899pt}{0.400pt}}
\put(372,428){\usebox{\plotpoint}}
\multiput(676.00,426.92)(4.850,-0.494){29}{\rule{3.900pt}{0.119pt}}
\multiput(676.00,427.17)(143.905,-16.000){2}{\rule{1.950pt}{0.400pt}}
\multiput(828.00,410.92)(9.716,-0.494){29}{\rule{7.700pt}{0.119pt}}
\multiput(828.00,411.17)(288.018,-16.000){2}{\rule{3.850pt}{0.400pt}}
\multiput(1132.00,394.92)(9.130,-0.495){31}{\rule{7.253pt}{0.119pt}}
\multiput(1132.00,395.17)(288.946,-17.000){2}{\rule{3.626pt}{0.400pt}}
\put(1350,512){\circle*{18}}
\put(372,428){\circle*{18}}
\put(676,428){\circle*{18}}
\put(828,412){\circle*{18}}
\put(1132,396){\circle*{18}}
\put(1436,379){\circle*{18}}
\put(372.0,428.0){\rule[-0.200pt]{73.234pt}{0.400pt}}
\put(372,477){\usebox{\plotpoint}}
\multiput(372,477)(20.727,-1.091){15}{\usebox{\plotpoint}}
\multiput(676,461)(20.627,-2.307){8}{\usebox{\plotpoint}}
\multiput(828,444)(20.502,-3.237){14}{\usebox{\plotpoint}}
\multiput(1132,396)(20.668,-1.904){15}{\usebox{\plotpoint}}
\put(1436,368){\usebox{\plotpoint}}
\put(1306,467){\makebox(0,0)[r]{Symmetric Adjacency}}
\put(1328.0,467.0){\rule[-0.200pt]{15.899pt}{0.400pt}}
\put(372,347){\usebox{\plotpoint}}
\multiput(372.00,345.93)(27.432,-0.482){9}{\rule{20.367pt}{0.116pt}}
\multiput(372.00,346.17)(261.728,-6.000){2}{\rule{10.183pt}{0.400pt}}
\multiput(676.00,339.92)(1.088,-0.499){137}{\rule{0.969pt}{0.120pt}}
\multiput(676.00,340.17)(149.990,-70.000){2}{\rule{0.484pt}{0.400pt}}
\multiput(828.00,269.92)(5.491,-0.497){53}{\rule{4.443pt}{0.120pt}}
\multiput(828.00,270.17)(294.779,-28.000){2}{\rule{2.221pt}{0.400pt}}
\multiput(1132.00,241.92)(4.032,-0.498){73}{\rule{3.300pt}{0.120pt}}
\multiput(1132.00,242.17)(297.151,-38.000){2}{\rule{1.650pt}{0.400pt}}
\put(1350,467){\circle{18}}
\put(372,347){\circle{18}}
\put(676,341){\circle{18}}
\put(828,271){\circle{18}}
\put(1132,243){\circle{18}}
\put(1436,205){\circle{18}}
\put(372,363){\usebox{\plotpoint}}
\multiput(372,363)(20.756,0.000){15}{\usebox{\plotpoint}}
\multiput(676,363)(17.756,-10.747){9}{\usebox{\plotpoint}}
\multiput(828,271)(20.751,-0.410){14}{\usebox{\plotpoint}}
\multiput(1132,265)(19.954,-5.711){16}{\usebox{\plotpoint}}
\put(1436,178){\usebox{\plotpoint}}
\end{picture}
\caption{Accuracy of symmetric association for
various training schemes} \label{fig:cy_adjacency_sym_accuracy}
\end{figure}
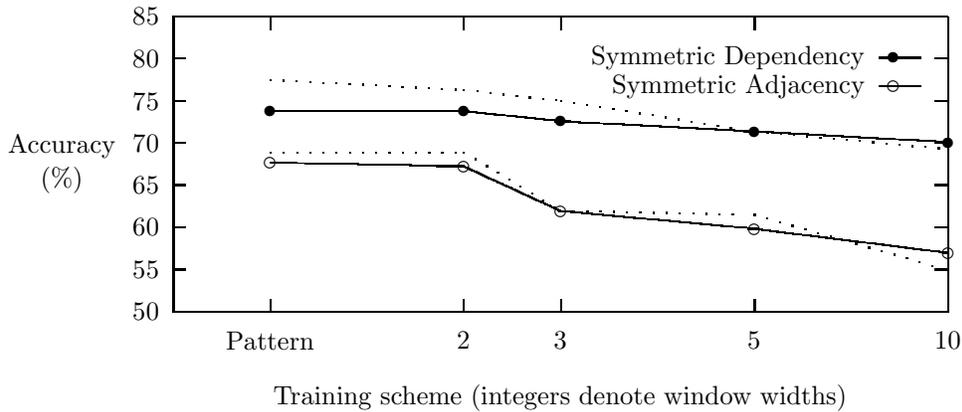

\subsubsection*{Tuned model}

As mentioned in section~\ref{sec:cy_method}, 
equation~\ref{eq:cy_method_dep_ratio} omits the factor of 2 resulting from 
the difference in $\mbox{\it choice\/}(m)$ between left-branching and 
right-branching modificational structures.  It also simplifies by assuming that 
the sizes of semantic classes are uniform.  I will now report the results of 
employing a model in which both of these deviations from the dependency 
model are corrected.  I call this the tuned model.

While these changes are theoretically motivated only in the dependency 
method, I have also applied them to the adjacency method for comparison.
To implement them, equations~\ref{eq:cy_method_dep_ratio} 
and~\ref{eq:cy_adjacency_adj_ratio} must be modified to incorporate a 
factor of $\frac{1}{\mid t_1 \mid \mid t_2 \mid \mid t_3 \mid}$ in each
term of the sum and the entire ratio must be multiplied by two.  

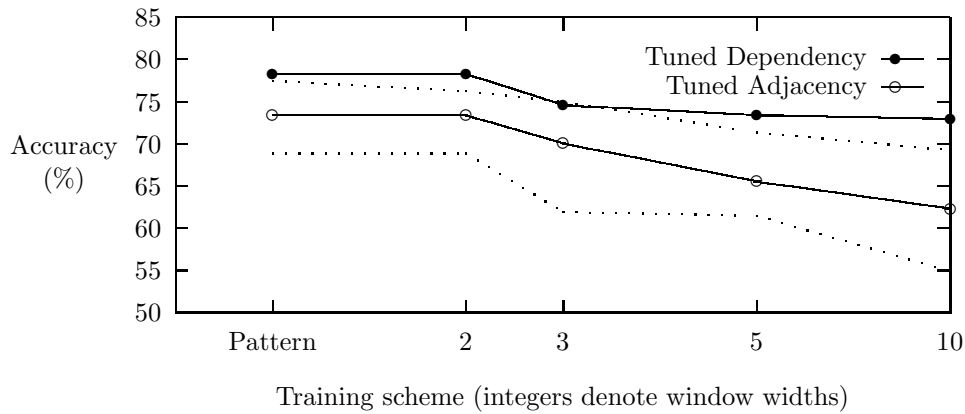
\begin{figure*}
\centering


\setlength{\unitlength}{0.240900pt}
\ifx\plotpoint\undefined\newsavebox{\plotpoint}\fi
\begin{picture}(1500,600)(0,0)
\font\gnuplot=cmr10 at 10pt
\gnuplot
\sbox{\plotpoint}{\rule[-0.200pt]{0.400pt}{0.400pt}}%
\put(220.0,113.0){\rule[-0.200pt]{4.818pt}{0.400pt}}
\put(198,113){\makebox(0,0)[r]{50}}
\put(1416.0,113.0){\rule[-0.200pt]{4.818pt}{0.400pt}}
\put(220.0,179.0){\rule[-0.200pt]{4.818pt}{0.400pt}}
\put(198,179){\makebox(0,0)[r]{55}}
\put(1416.0,179.0){\rule[-0.200pt]{4.818pt}{0.400pt}}
\put(220.0,246.0){\rule[-0.200pt]{4.818pt}{0.400pt}}
\put(198,246){\makebox(0,0)[r]{60}}
\put(1416.0,246.0){\rule[-0.200pt]{4.818pt}{0.400pt}}
\put(220.0,312.0){\rule[-0.200pt]{4.818pt}{0.400pt}}
\put(198,312){\makebox(0,0)[r]{65}}
\put(1416.0,312.0){\rule[-0.200pt]{4.818pt}{0.400pt}}
\put(220.0,378.0){\rule[-0.200pt]{4.818pt}{0.400pt}}
\put(198,378){\makebox(0,0)[r]{70}}
\put(1416.0,378.0){\rule[-0.200pt]{4.818pt}{0.400pt}}
\put(220.0,444.0){\rule[-0.200pt]{4.818pt}{0.400pt}}
\put(198,444){\makebox(0,0)[r]{75}}
\put(1416.0,444.0){\rule[-0.200pt]{4.818pt}{0.400pt}}
\put(220.0,511.0){\rule[-0.200pt]{4.818pt}{0.400pt}}
\put(198,511){\makebox(0,0)[r]{80}}
\put(1416.0,511.0){\rule[-0.200pt]{4.818pt}{0.400pt}}
\put(220.0,577.0){\rule[-0.200pt]{4.818pt}{0.400pt}}
\put(198,577){\makebox(0,0)[r]{85}}
\put(1416.0,577.0){\rule[-0.200pt]{4.818pt}{0.400pt}}
\put(220.0,113.0){\rule[-0.200pt]{0.400pt}{4.818pt}}
\put(220.0,557.0){\rule[-0.200pt]{0.400pt}{4.818pt}}
\put(372.0,113.0){\rule[-0.200pt]{0.400pt}{4.818pt}}
\put(372,68){\makebox(0,0){Pattern}}
\put(372.0,557.0){\rule[-0.200pt]{0.400pt}{4.818pt}}
\put(676.0,113.0){\rule[-0.200pt]{0.400pt}{4.818pt}}
\put(676,68){\makebox(0,0){2}}
\put(676.0,557.0){\rule[-0.200pt]{0.400pt}{4.818pt}}
\put(828.0,113.0){\rule[-0.200pt]{0.400pt}{4.818pt}}
\put(828,68){\makebox(0,0){3}}
\put(828.0,557.0){\rule[-0.200pt]{0.400pt}{4.818pt}}
\put(1132.0,113.0){\rule[-0.200pt]{0.400pt}{4.818pt}}
\put(1132,68){\makebox(0,0){5}}
\put(1132.0,557.0){\rule[-0.200pt]{0.400pt}{4.818pt}}
\put(1436.0,113.0){\rule[-0.200pt]{0.400pt}{4.818pt}}
\put(1436,68){\makebox(0,0){10}}
\put(1436.0,557.0){\rule[-0.200pt]{0.400pt}{4.818pt}}
\put(220.0,113.0){\rule[-0.200pt]{292.934pt}{0.400pt}}
\put(1436.0,113.0){\rule[-0.200pt]{0.400pt}{111.778pt}}
\put(220.0,577.0){\rule[-0.200pt]{292.934pt}{0.400pt}}
\put(45,345){\makebox(0,0){\shortstack{Accuracy\\(\%)}}}
\put(828,-22){\makebox(0,0){Training scheme (integers denote window widths)}}
\put(220.0,113.0){\rule[-0.200pt]{0.400pt}{111.778pt}}
\put(1306,512){\makebox(0,0)[r]{Tuned Dependency}}
\put(1328.0,512.0){\rule[-0.200pt]{15.899pt}{0.400pt}}
\put(372,488){\usebox{\plotpoint}}
\multiput(676.00,486.92)(1.558,-0.498){95}{\rule{1.341pt}{0.120pt}}
\multiput(676.00,487.17)(149.217,-49.000){2}{\rule{0.670pt}{0.400pt}}
\multiput(828.00,437.92)(9.716,-0.494){29}{\rule{7.700pt}{0.119pt}}
\multiput(828.00,438.17)(288.018,-16.000){2}{\rule{3.850pt}{0.400pt}}
\multiput(1132.00,421.93)(27.432,-0.482){9}{\rule{20.367pt}{0.116pt}}
\multiput(1132.00,422.17)(261.728,-6.000){2}{\rule{10.183pt}{0.400pt}}
\put(1350,512){\circle*{18}}
\put(372,488){\circle*{18}}
\put(676,488){\circle*{18}}
\put(828,439){\circle*{18}}
\put(1132,423){\circle*{18}}
\put(1436,417){\circle*{18}}
\put(372.0,488.0){\rule[-0.200pt]{73.234pt}{0.400pt}}
\put(372,477){\usebox{\plotpoint}}
\multiput(372,477)(20.727,-1.091){15}{\usebox{\plotpoint}}
\multiput(676,461)(20.627,-2.307){8}{\usebox{\plotpoint}}
\multiput(828,444)(20.502,-3.237){14}{\usebox{\plotpoint}}
\multiput(1132,396)(20.668,-1.904){15}{\usebox{\plotpoint}}
\put(1436,368){\usebox{\plotpoint}}
\put(1306,467){\makebox(0,0)[r]{Tuned Adjacency}}
\put(1328.0,467.0){\rule[-0.200pt]{15.899pt}{0.400pt}}
\put(372,423){\usebox{\plotpoint}}
\multiput(676.00,421.92)(1.736,-0.498){85}{\rule{1.482pt}{0.120pt}}
\multiput(676.00,422.17)(148.924,-44.000){2}{\rule{0.741pt}{0.400pt}}
\multiput(828.00,377.92)(2.545,-0.499){117}{\rule{2.127pt}{0.120pt}}
\multiput(828.00,378.17)(299.586,-60.000){2}{\rule{1.063pt}{0.400pt}}
\multiput(1132.00,317.92)(3.559,-0.498){83}{\rule{2.928pt}{0.120pt}}
\multiput(1132.00,318.17)(297.923,-43.000){2}{\rule{1.464pt}{0.400pt}}
\put(1350,467){\circle{18}}
\put(372,423){\circle{18}}
\put(676,423){\circle{18}}
\put(828,379){\circle{18}}
\put(1132,319){\circle{18}}
\put(1436,276){\circle{18}}
\put(372.0,423.0){\rule[-0.200pt]{73.234pt}{0.400pt}}
\put(372,363){\usebox{\plotpoint}}
\multiput(372,363)(20.756,0.000){15}{\usebox{\plotpoint}}
\multiput(676,363)(17.756,-10.747){9}{\usebox{\plotpoint}}
\multiput(828,271)(20.751,-0.410){14}{\usebox{\plotpoint}}
\multiput(1132,265)(19.954,-5.711){16}{\usebox{\plotpoint}}
\put(1436,178){\usebox{\plotpoint}}
\end{picture}
\caption{Accuracy of tuned dependency and adjacency methods
for various training schemes} \label{fig:cy_adjacency_tuned_accuracy}
\end{figure*}

Five training schemes have been applied with these extensions.  
The accuracy results are shown in 
figure~\ref{fig:cy_adjacency_tuned_accuracy}.  For comparison, the 
untuned accuracy figures are shown with dotted lines.  A marked 
improvement is observed for the adjacency model, while the
dependency model is only slightly improved.

\subsubsection*{Lexical association} 

The conceptual probabilistic model given uses $|S| \times |S|$ parameters.  If 
instead of using conceptual association, a lexical parameterisation were 
used, the probabilistic model would have $|{\cal N}| \times |{\cal N}|$ 
parameters for a total of 8.1 billion.  Apart from the probable data 
sparseness problem, manipulating such a model would involve huge 
computational expense.

However, it is possible to implement such a system for a small test set.  To 
determine the cost (if any) of assuming that words in the same category 
behave identically, I have done this for the test set used above.
The pattern training scheme was retrained using lexical counts for both the 
dependency and adjacency method, but only for the words in the test set.
Left-branching is favoured by a factor of two as described in the previous 
section, but category sizes were omitted (these being meaningless for the 
lexical association method).

\begin{figure*}
\centering
 


\setlength{\unitlength}{0.240900pt} 
\ifx\plotpoint\undefined\newsavebox{\plotpoint}\fi 
\begin{picture}(1730, 900)(0,0) 
\put(0,0){ 
\begin{picture}(840,900)(0,0) 
\font\gnuplot=cmr10 at 10pt 
\gnuplot 
\sbox{\plotpoint}{\rule[-0.200pt]{0.400pt}{0.400pt}}%
\put(220.0,68.0){\rule[-0.200pt]{4.818pt}{0.400pt}} 
\put(198,68){\makebox(0,0)[r]{50}} 
\put(756.0,68.0){\rule[-0.200pt]{4.818pt}{0.400pt}} 
\put(220.0,184.0){\rule[-0.200pt]{4.818pt}{0.400pt}} 
\put(198,184){\makebox(0,0)[r]{55}} 
\put(756.0,184.0){\rule[-0.200pt]{4.818pt}{0.400pt}} 
\put(220.0,299.0){\rule[-0.200pt]{4.818pt}{0.400pt}} 
\put(198,299){\makebox(0,0)[r]{60}} 
\put(756.0,299.0){\rule[-0.200pt]{4.818pt}{0.400pt}} 
\put(220.0,415.0){\rule[-0.200pt]{4.818pt}{0.400pt}} 
\put(198,415){\makebox(0,0)[r]{65}} 
\put(756.0,415.0){\rule[-0.200pt]{4.818pt}{0.400pt}} 
\put(220.0,530.0){\rule[-0.200pt]{4.818pt}{0.400pt}} 
\put(198,530){\makebox(0,0)[r]{70}} 
\put(756.0,530.0){\rule[-0.200pt]{4.818pt}{0.400pt}} 
\put(220.0,646.0){\rule[-0.200pt]{4.818pt}{0.400pt}} 
\put(198,646){\makebox(0,0)[r]{75}} 
\put(756.0,646.0){\rule[-0.200pt]{4.818pt}{0.400pt}} 
\put(220.0,761.0){\rule[-0.200pt]{4.818pt}{0.400pt}} 
\put(198,761){\makebox(0,0)[r]{80}} 
\put(756.0,761.0){\rule[-0.200pt]{4.818pt}{0.400pt}} 
\put(220.0,877.0){\rule[-0.200pt]{4.818pt}{0.400pt}} 
\put(198,877){\makebox(0,0)[r]{85}} 
\put(756.0,877.0){\rule[-0.200pt]{4.818pt}{0.400pt}} 
\put(220.0,68.0){\rule[-0.200pt]{0.400pt}{4.818pt}} 
\put(220.0,857.0){\rule[-0.200pt]{0.400pt}{4.818pt}} 
\put(405.0,68.0){\rule[-0.200pt]{0.400pt}{4.818pt}} 
\put(405,23){\makebox(0,0){Dependency\hspace{0.1in}}} 
\put(405.0,857.0){\rule[-0.200pt]{0.400pt}{4.818pt}} 
\put(591.0,68.0){\rule[-0.200pt]{0.400pt}{4.818pt}} 
\put(591,23){\makebox(0,0){\hspace{0.1in} Adjacency}} 
\put(591.0,857.0){\rule[-0.200pt]{0.400pt}{4.818pt}} 
\put(220.0,68.0){\rule[-0.200pt]{133.940pt}{0.400pt}} 
\put(776.0,68.0){\rule[-0.200pt]{0.400pt}{194.888pt}} 
\put(220.0,877.0){\rule[-0.200pt]{133.940pt}{0.400pt}} 
\put(45,472){\makebox(0,0){\shortstack{Accuracy\\(\%)}}} 
\put(220.0,68.0){\rule[-0.200pt]{0.400pt}{194.888pt}} 
 
\put(355,68){\rule{12.05pt}{157.55pt}} 
\put(405,68){\framebox(50,549){\hspace{0.1in}}} 
\put(541,68){\rule{12.05pt}{127.68pt}} 
\put(591,68){\framebox(50,512){\hspace{0.1in}}}

\put(646,812){\makebox(0,0)[r]{Conceptual}} 
\put(690,812){\raisebox{-.8pt}{\rule{5.5pt}{5.5pt}}} 
\put(646,767){\makebox(0,0)[r]{Lexical}} 
\put(690,767){\makebox(0,0){\hspace{5pt}$\Box$}} 
 
\end{picture} 
} 
\put(890,0){ 
\begin{picture}(840,900)(0,0) 
\font\gnuplot=cmr10 at 10pt 
\gnuplot 
\sbox{\plotpoint}{\rule[-0.200pt]{0.400pt}{0.400pt}}%
\put(220.0,68.0){\rule[-0.200pt]{133.940pt}{0.400pt}} 
\put(220.0,68.0){\rule[-0.200pt]{4.818pt}{0.400pt}} 
\put(198,68){\makebox(0,0)[r]{0}} 
\put(756.0,68.0){\rule[-0.200pt]{4.818pt}{0.400pt}} 
\put(220.0,203.0){\rule[-0.200pt]{4.818pt}{0.400pt}} 
\put(198,203){\makebox(0,0)[r]{5}} 
\put(756.0,203.0){\rule[-0.200pt]{4.818pt}{0.400pt}} 
\put(220.0,338.0){\rule[-0.200pt]{4.818pt}{0.400pt}} 
\put(198,338){\makebox(0,0)[r]{10}} 
\put(756.0,338.0){\rule[-0.200pt]{4.818pt}{0.400pt}} 
\put(220.0,473.0){\rule[-0.200pt]{4.818pt}{0.400pt}} 
\put(198,473){\makebox(0,0)[r]{15}} 
\put(756.0,473.0){\rule[-0.200pt]{4.818pt}{0.400pt}} 
\put(220.0,607.0){\rule[-0.200pt]{4.818pt}{0.400pt}} 
\put(198,607){\makebox(0,0)[r]{20}} 
\put(756.0,607.0){\rule[-0.200pt]{4.818pt}{0.400pt}} 
\put(220.0,742.0){\rule[-0.200pt]{4.818pt}{0.400pt}} 
\put(198,742){\makebox(0,0)[r]{25}} 
\put(756.0,742.0){\rule[-0.200pt]{4.818pt}{0.400pt}} 
\put(220.0,877.0){\rule[-0.200pt]{4.818pt}{0.400pt}} 
\put(198,877){\makebox(0,0)[r]{30}} 
\put(756.0,877.0){\rule[-0.200pt]{4.818pt}{0.400pt}} 
\put(220.0,68.0){\rule[-0.200pt]{0.400pt}{4.818pt}} 
\put(220.0,857.0){\rule[-0.200pt]{0.400pt}{4.818pt}} 
\put(405.0,68.0){\rule[-0.200pt]{0.400pt}{4.818pt}} 
\put(405,23){\makebox(0,0){Dependency\hspace{0.1in}}} 
\put(405.0,857.0){\rule[-0.200pt]{0.400pt}{4.818pt}} 
\put(591.0,68.0){\rule[-0.200pt]{0.400pt}{4.818pt}} 
\put(591,23){\makebox(0,0){\hspace{0.1in} Adjacency}} 
\put(591.0,857.0){\rule[-0.200pt]{0.400pt}{4.818pt}} 
\put(220.0,68.0){\rule[-0.200pt]{133.940pt}{0.400pt}} 
\put(776.0,68.0){\rule[-0.200pt]{0.400pt}{194.888pt}} 
\put(220.0,877.0){\rule[-0.200pt]{133.940pt}{0.400pt}} 
\put(45,472){\makebox(0,0){\shortstack{Guess Rate\\(\%)}}} 
\put(220.0,68.0){\rule[-0.200pt]{0.400pt}{194.888pt}} 
 
\put(355,68){\rule{12.05pt}{24.09pt}} 
\put(405,68){\framebox(50,696){\hspace{0.1in}}} 
\put(541,68){\rule{12.05pt}{24.09pt}} 
\put(591,68){\framebox(50,453){\hspace{0.1in}}}

\put(646,812){\makebox(0,0)[r]{Conceptual}} 
\put(690,812){\raisebox{-.8pt}{\rule{5.5pt}{5.5pt}}} 
\put(646,767){\makebox(0,0)[r]{Lexical}} 
\put(690,767){\makebox(0,0){\hspace{5pt}$\Box$}} 
\end{picture} 
} 
\end{picture}
\caption{Accuracy and guess rates of lexical and conceptual association}
\label{fig:cy_lexical_lex_accuracy}
\end{figure*}
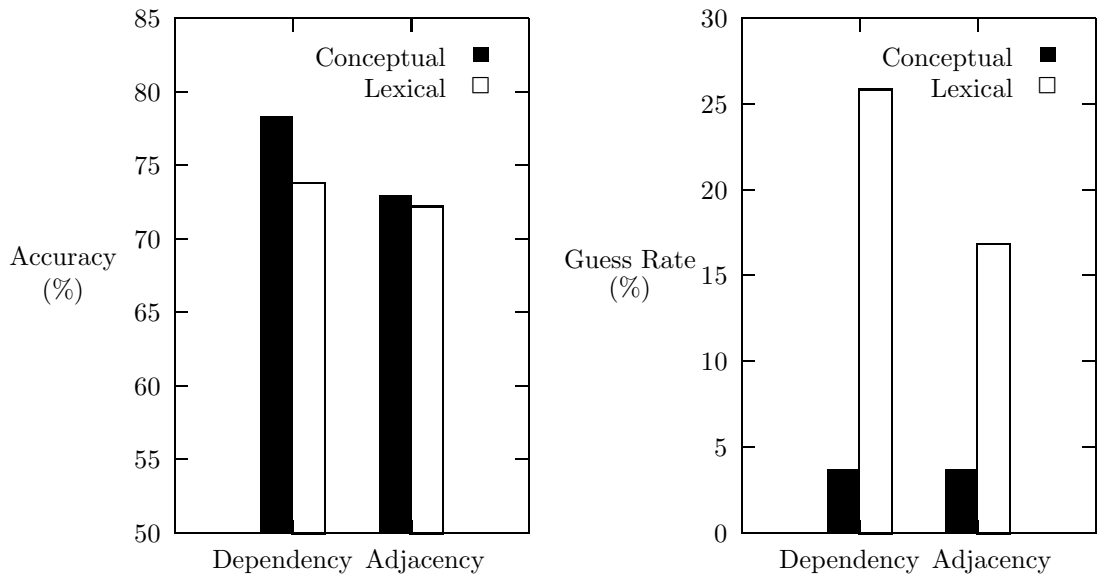

Accuracy and guess rates are shown in 
figure~\ref{fig:cy_lexical_lex_accuracy}.  Guess rates are, unsurprisingly, 
much higher for lexical association.  Probably because of this, the accuracy 
falls.  Therefore grouping words into concepts not only makes the model 
tractable, but improves the performance by allowing generalisation amongst 
similar words.

\subsubsection*{Using a tagged corpus} 

One problem with the training method given in
section \ref{sec:cy_method} is the restriction of
training data to nouns in $\cal N$.  Many nouns,
especially common ones, have verbal or adjectival
usages that preclude them from being in $\cal N$.
Yet when they occur as nouns, they still provide
useful training information that the system ignores.  
To test whether using tagged data would
make a difference, the freely available Brill tagger (Brill,~1993) 
was applied to the corpus.  Since no manually tagged 
training data is available for our corpus, 
the tagger's default rules were used (these
rules were produced by Brill by training on the 
Brown corpus).  This results in rather poor
tagging accuracy, so it is quite possible that a
manually tagged corpus would produce better
results.

\begin{figure*}
\centering


\setlength{\unitlength}{0.240900pt} 
\ifx\plotpoint\undefined\newsavebox{\plotpoint}\fi 
\sbox{\plotpoint}{\rule[-0.200pt]{0.400pt}{0.400pt}}%
\begin{picture}(1500,900)(0,0) 
\font\gnuplot=cmr10 at 10pt 
\gnuplot 
\sbox{\plotpoint}{\rule[-0.200pt]{0.400pt}{0.400pt}}%
\put(220.0,113.0){\rule[-0.200pt]{4.818pt}{0.400pt}} 
\put(198,113){\makebox(0,0)[r]{50}} 
\put(1416.0,113.0){\rule[-0.200pt]{4.818pt}{0.400pt}} 
\put(220.0,222.0){\rule[-0.200pt]{4.818pt}{0.400pt}} 
\put(198,222){\makebox(0,0)[r]{55}} 
\put(1416.0,222.0){\rule[-0.200pt]{4.818pt}{0.400pt}} 
\put(220.0,331.0){\rule[-0.200pt]{4.818pt}{0.400pt}} 
\put(198,331){\makebox(0,0)[r]{60}} 
\put(1416.0,331.0){\rule[-0.200pt]{4.818pt}{0.400pt}} 
\put(220.0,440.0){\rule[-0.200pt]{4.818pt}{0.400pt}} 
\put(198,440){\makebox(0,0)[r]{65}} 
\put(1416.0,440.0){\rule[-0.200pt]{4.818pt}{0.400pt}} 
\put(220.0,550.0){\rule[-0.200pt]{4.818pt}{0.400pt}} 
\put(198,550){\makebox(0,0)[r]{70}} 
\put(1416.0,550.0){\rule[-0.200pt]{4.818pt}{0.400pt}} 
\put(220.0,659.0){\rule[-0.200pt]{4.818pt}{0.400pt}} 
\put(198,659){\makebox(0,0)[r]{75}} 
\put(1416.0,659.0){\rule[-0.200pt]{4.818pt}{0.400pt}} 
\put(220.0,768.0){\rule[-0.200pt]{4.818pt}{0.400pt}} 
\put(198,768){\makebox(0,0)[r]{80}} 
\put(1416.0,768.0){\rule[-0.200pt]{4.818pt}{0.400pt}} 
\put(220.0,877.0){\rule[-0.200pt]{4.818pt}{0.400pt}} 
\put(198,877){\makebox(0,0)[r]{85}} 
\put(1416.0,877.0){\rule[-0.200pt]{4.818pt}{0.400pt}} 
\put(220.0,113.0){\rule[-0.200pt]{0.400pt}{4.818pt}} 
\put(220.0,857.0){\rule[-0.200pt]{0.400pt}{4.818pt}} 
\put(463.0,113.0){\rule[-0.200pt]{0.400pt}{4.818pt}} 
\put(463,68){\makebox(0,0){Pattern}} 
\put(463.0,857.0){\rule[-0.200pt]{0.400pt}{4.818pt}} 
\put(950.0,113.0){\rule[-0.200pt]{0.400pt}{4.818pt}} 
\put(950,68){\makebox(0,0){3}} 
\put(950.0,857.0){\rule[-0.200pt]{0.400pt}{4.818pt}} 
\put(1436.0,113.0){\rule[-0.200pt]{0.400pt}{4.818pt}} 
\put(1436,68){\makebox(0,0){5}} 
\put(1436.0,857.0){\rule[-0.200pt]{0.400pt}{4.818pt}} 
\put(220.0,113.0){\rule[-0.200pt]{292.934pt}{0.400pt}} 
\put(1436.0,113.0){\rule[-0.200pt]{0.400pt}{184.048pt}} 
\put(220.0,877.0){\rule[-0.200pt]{292.934pt}{0.400pt}} 
\put(45,495){\makebox(0,0){\shortstack{Accuracy\\(\%)}}} 
\put(828,-22){\makebox(0,0){Training scheme (integers denote window widths)}} 
\put(220.0,113.0){\rule[-0.200pt]{0.400pt}{184.048pt}} 
\put(950,288){\makebox(0,0)[r]{Tagged Dependency}} 
\put(972.0,288.0){\rule[-0.200pt]{15.899pt}{0.400pt}} 
\put(463,784){\usebox{\plotpoint}} 
\multiput(463.00,782.92)(1.604,-0.499){301}{\rule{1.382pt}{0.120pt}} 
\multiput(463.00,783.17)(484.132,-152.000){2}{\rule{0.691pt}{0.400pt}} 
\multiput(950.00,630.92)(6.811,-0.498){69}{\rule{5.500pt}{0.120pt}} 
\multiput(950.00,631.17)(474.584,-36.000){2}{\rule{2.750pt}{0.400pt}} 
\put(994,288){\circle*{18}} 
\put(463,784){\circle*{18}} 
\put(950,632){\circle*{18}} 
\put(1436,596){\circle*{18}} 
\put(463,730){\usebox{\plotpoint}} 
\multiput(463,730)(20.481,-3.364){24}{\usebox{\plotpoint}} 
\multiput(950,650)(20.724,-1.151){24}{\usebox{\plotpoint}} 
\put(1436,623){\usebox{\plotpoint}} 
\put(950,243){\makebox(0,0)[r]{Tagged Adjacency}} 
\put(972.0,243.0){\rule[-0.200pt]{15.899pt}{0.400pt}} 
\put(463,605){\usebox{\plotpoint}} 
\multiput(463.00,603.92)(3.395,-0.499){141}{\rule{2.806pt}{0.120pt}} 
\multiput(463.00,604.17)(481.177,-72.000){2}{\rule{1.403pt}{0.400pt}} 
\multiput(950.00,531.92)(2.739,-0.499){175}{\rule{2.284pt}{0.120pt}} 
\multiput(950.00,532.17)(481.259,-89.000){2}{\rule{1.142pt}{0.400pt}} 
\put(994,243){\circle{18}} 
\put(463,605){\circle{18}} 
\put(950,533){\circle{18}} 
\put(1436,444){\circle{18}} 
\put(463,623){\usebox{\plotpoint}} 
\multiput(463,623)(20.532,-3.036){24}{\usebox{\plotpoint}} 
\multiput(950,551)(20.346,-4.103){24}{\usebox{\plotpoint}} 
\put(1436,453){\usebox{\plotpoint}} 
\end{picture} 
\caption{Accuracy using a tagged corpus for various
training schemes}  \label{fig:cy_tagged_tag_accuracy}
\end{figure*}
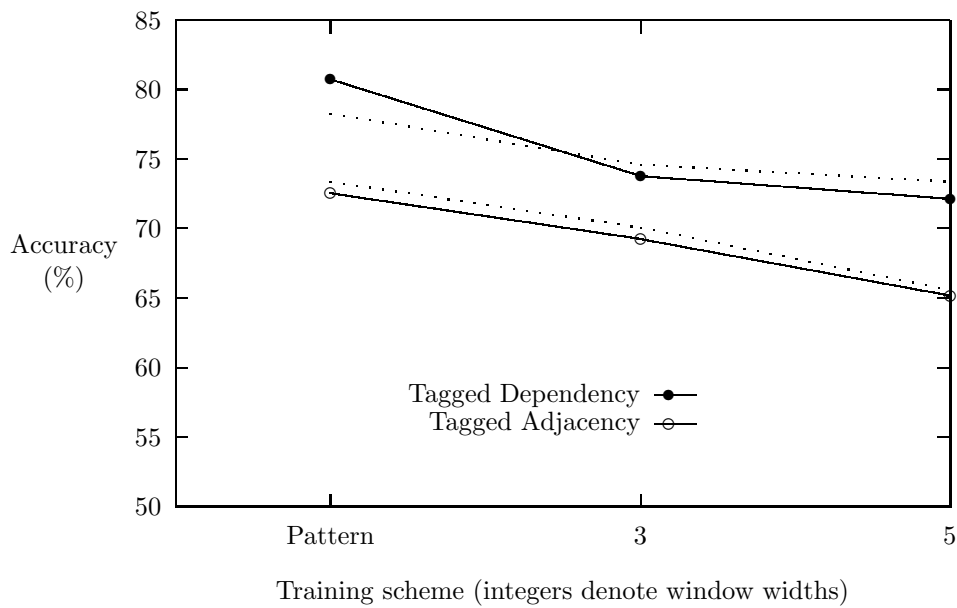

Three training schemes have been used and the
tuned model then applied to the test set.
Figure \ref{fig:cy_tagged_tag_accuracy} shows the resulting
accuracy, with accuracy values from figure
\ref{fig:cy_adjacency_tuned_accuracy} displayed with dotted lines.
If anything, admitting additional training data based
on the tagger introduces more noise, reducing the
accuracy.  However, for the pattern training scheme
an improvement was made to the dependency
model, producing the highest overall accuracy of 80.7\%.

\subsection{An Experiment with Human Judges} 
\label{sec:cy_human}

In 4 out of 5 three word compounds the parsing strategy suggested by the 
dependency model can predict the correct bracketing.  To achieve this, the 
model assigns probabilities to modificational structures by computing 
statistics from a corpus.  However, these probabilities are conditioned only 
on the compound in isolation; the model is incapable of varying its analysis 
in response to different contexts.\footnote{I should emphasise that the fault 
lies with the probabilistic model, not the meaning distributions theory.  In 
terms of section~\ref{sec:md_context}, the model entirely ignores three of 
the five types of context (topic, author and discourse structure) and only 
statically models the other two (register and world knowledge), making it a 
relatively crude model overall.}  

In section~\ref{sec:cn_context} I argued that the syntactic analysis of 
compounds can vary with context, so there must be an upper limit to the 
model's performance that is less than perfect.  Putting it in terms of the data 
requirements theory, we know that there is a non-zero optimal error rate.  In 
order to properly evaluate the accuracy results given above, it would be 
useful to have some idea of what this upper limit is.  In this section I will 
discuss some work toward estimating that limit.

\subsubsection*{Some important qualifications}

One approach to this is to ask human judges to perform the same task as the 
program does using exactly the same information.  Hindle and Rooth~(1993) 
pursue this approach for prepositional phrase attachment, treating the result 
as a significant indication of the optimal accuracy.  For each test case, the 
judge is given just those values upon which the probabilistic model is 
conditioned and required to assign their best guess as to the appropriate 
in-context analysis.  Resnik~(1993:128) has also conducted a similar 
experiment for parsing three word compound nouns.  It is worth considering 
some of the assumptions of such an experiment.  

To begin with, the task facing the judges is an artificial one --- it isn't a 
problem that humans would normally encounter.  So the assumption that 
human judges will achieve a near-optimal accuracy is questionable on 
psychological grounds.  Also, it is impossible to give the judges exactly the 
same information as the program is given.  For example, if the program is 
trained on a newswire corpus, the trained model will incorporate the 
information that test compounds come from that genre.  So to be fair, we 
must tell the human judge what kind of text the test set comes from.  But 
how much detail do we give?  In principle, the entire training corpus might 
be of use to the judge, but supplying this is clearly impractical.  Also, if a 
test case contains a word that the judge doesn't know, how is this to be 
treated?  She is unlikely to achieve optimal performance then.

These problems result in humans performing at below optimal accuracy.  
Might the experiment therefore provide a lower bound on the optimal 
accuracy?  In general, no.  Even if the probabilistic model conditions on 
exactly the same set of variables that are provided to the judges, any 
reasonable model (that is, every one bar the complete joint distribution 
discussed in section~\ref{sec:dr_beginning}) makes some assumptions that 
limit the possible answers.  For example, Hindle and Rooth's~(1993) model 
assumes that the evidence provided by the noun and verb can be combined 
independently.  Therefore, their program cannot ever respond with the 
following analyses:\footnote{I have simplified here by ignoring the effect of 
the parameters $\Pr(n, \nullsym)$; however, the basic point remains valid.}
\begin{itemize}
\item $\langle v_1, n_1, p \rangle$ gets verbal attachment, 
\item $\langle v_1, n_2, p \rangle$ gets nominal attachment,
\item $\langle v_2, n_2, p \rangle$ gets verbal attachment, and
\item $\langle v_2, n_1, p \rangle$ gets nominal attachment.
\end{itemize}
Human judges are under no such restriction, and therefore potentially have 
higher accuracy rates than are optimal for the program.  Likewise, in the 
model of compound nouns reported in this thesis, the probabilistic model 
conditions only on semantic classes.  It seems unfair in the extreme to 
require judges to bracket sequences of thesaurus categories simply because 
the model chooses this as its internal representation.  I am convinced that 
humans would do far worse than my program under such conditions.  Yet, if 
we give the judges word sequences, they will have more information at their 
disposal than the program does.

However, since the test set is far too small to provide any insight into the 
optimal accuracy, there appears to be little choice in the matter.  Therefore, 
keeping all these qualifications in mind, I will now report a human judges 
experiment for the compound noun parsing task.  The goal of the experiment 
is to estimate the optimal performance of any analysis system that, given a 
compound, predicts the bracketing assigned to it by the manual annotation 
regime that I used to mark up the test set.  Note that this differs from 
predicting the \scare{correct} parse of the compound, since it allows both 
for idiosyncrasies in my annotation of the test set and for supervision 
errors.\footnote{I am not yet aware of any supervision errors.}  Hindle and 
Rooth fail to mention these possible causes of error, ascribing the entire 
error rate to context dependence.

\subsubsection*{Experimental method}

Seven subjects volunteered to act as judges.  All came from the local 
research environment, although with varying backgrounds.  Two have 
doctoral degrees in computer science unrelated to \nlp\ and linguistics, four 
are postgraduate students in computer science (one of whom is working in 
\nlp), and one is an administrative assistant (no tertiary education).  All are 
native speakers of English.  The judges were randomly assigned to one of 
two blocks (3 judges in block 1 and 4 judges in block 2) prior to the 
experiment.  For comparison purposes, Hindle and Rooth~(1993) use two 
judges and Resnik~(1993) uses one.

A custom designed interface was built which presented each test compound 
out of context to the judge, along with the two possible readings.  In each 
case the judge was required to select either the left-branching or 
right-branching analysis, according to their best guess about the compound 
in the (unknown) context.  The interface allowed additional comments 
to be made on each compound.  The judges were told that the compounds 
came from an encyclopedia and instructed to take as long as necessary to 
maximise their performance.  The complete set of instructions given appears 
in appendix~\ref{appendix:humaninstructions}.

Each compound was presented only once (responses on compounds that 
appeared multiple times in the test set were weighted so that the results 
below represent performance by token frequencies).  The order in which the 
compounds were presented was chosen randomly, each block getting a 
different order.  The time taken to analyse each compound was recorded, 
although the judges were unaware of this.  The total time taken to analyse 
the 215 compounds ranged from 26 minutes to 72 minutes, with an average 
of 49 minutes.

\subsubsection*{Results}

The responses given by the judges were evaluated in the same manner as the 
computer program described above, with the resulting accuracy figures 
shown in table~\ref{tb:cy_human_results}.  The order of presentation did 
not appear to systematically alter the performance, since the block averages 
do not differ significantly.  The average accuracy is 81.5\%, showing a 
surprisingly large error rate.  
This is not significantly better than the 80.7\% 
achieved by the dependency model trained on tagged data.  The result also 
matches that found by Resnik~(1993:128) whose single judge scored only 
80.0\% (on around 160 compounds with 100\% recall).

\begin{table*}
\centering
\begin{tabular}{|l|c|c|c|c|c|c|c|c|} \hline
Block & \multicolumn{3}{|c|}{Block 1}  & 
\multicolumn{4}{|c|}{Block 2}  & Average \\
\cline{1-8} 
Judge & A & B & C & D & E & F & G &  \\
\hline
Accuracy & 78.7\% & 80.3\% & 86.1\% & 79.1\% & 79.9\% & 82.8\% & 
83.6\% & 81.5\% \\
\hline
\end{tabular}
\caption{Accuracy of human judges on parsing three word 
compounds out of context}
\label{tb:cy_human_results}
\end{table*}

These results suggest that the computer program may be performing at close 
to optimal.  However, it is also possible that neither the humans nor the 
computer are performing optimally.  One way to detect this is to examine 
how often the computer agrees with the human subjects.  If the humans 
make entirely different errors to those made by the computer, then it is likely 
that neither are making optimal predictions.  We still cannot make the 
converse conclusion if they agree well.

Table~\ref{tb:cy_human_agreement} shows the inter-human agreement 
rates and the human-computer agreement rates.  The inter-human agreement 
rate for a judge is derived by taking the proportion of matching analyses 
given by each of the other six judges and averaging these six proportions.  
The overall average is computed by averaging the seven agreement rates so 
derived.  The human-computer agreement rate for a judge is just the 
proportion of matching analyses given by the best dependency model (the 
one with 80.7\% accuracy).  The overall average is the mean of these seven 
agreement rates.  To dispel suspicion that one of the judges is pulling down 
all the other judges' averages, the highest agreement rate between any pair of 
judges is 84.4\%; the lowest is 73.0\%.

\begin{table*}
\centering
\begin{tabular}{|l|c|c|c|c|c|c|c|c|} \hline
Block & \multicolumn{3}{|c|}{Block 1}  & 
\multicolumn{4}{|c|}{Block 2}  & Average \\
\cline{1-8} 
Judge & A & B & C & D & E & F & G &  \\
\hline
Inter-human & 77.6\% & 78.6\% & 82.2\% & 79.2\% & 80.4\% & 82.1\% & 
81.7\% & 80.3\% \\
\hline
Human-computer & 73.4\% & 75.8\% & 77.5\% & 76.2\% & 71.3\% & 
73.4\% & 77.5\% & 75.0\% \\
\hline
\end{tabular}
\caption{Inter-human and human-computer agreement rates on compound 
parsing task without context}
\label{tb:cy_human_agreement}
\end{table*}

The computer agreement rates are clearly below the inter-human agreement 
rates, but not by a large margin.  In fact, the computer has better agreement 
with 6 out of 7 judges than that between the least agreeing pair of humans.  
The difference could be plausibly attributed to the lower accuracy rate of the 
computer (notice that the inter-human agreement rates are correlated with 
individual judge accuracy rates).  Therefore, the tentative conclusion of the 
experiment is that the optimal error rate is relatively large and the 
performance of the best dependency model is reasonably close to it.  This 
suggests that improvements in the performance of algorithm are unlikely 
without the addition of context sensitivity to the model.

\subsection{Discussion} 
\label{sec:cy_discuss}

\subsubsection*{Conclusions}

The experiments above demonstrate a number of important points.
The most general of these is that even quite crude corpus 
statistics can provide information about the syntax of compound
nouns.  At the very least, this information can be applied in
broad coverage parsing to assist in the control of search.
I have also shown that with a corpus of moderate size it is
possible to get reasonable results without using a tagger
or parser by employing a customised training pattern.  

The significance of the use of conceptual association deserves
some mention.  I have argued that without it a broad coverage system
would be impossible.  In this study, not only has the technique proved its 
worth by supporting generality, but through generalisation of 
training information it outperforms the equivalent lexical association 
approach given the same information.  This is in contrast to previous work
on conceptual association where it resulted in little
improvement on a task which could already be performed (Resnik and 
Hearst,~1993).

Amongst all the comparisons performed in these experiments one stands out 
as exhibiting the greatest contrast.  In all experiments the dependency 
method provides a substantial advantage over the adjacency method, even 
though the latter is more prevalent in proposals within the literature.  This 
result gives strong support to the meaning distributions theory.  The 
dependency model also has the further commendation that it predicts 
correctly the observed proportion of left-branching compounds found in two
independently extracted test sets.

In all, the most accurate technique achieved an accuracy
of 81\% as compared to the 67\% achieved by guessing left-branching.
Given the high frequency of occurrence of noun compounds in many texts, 
this suggests that the use of these techniques in probabilistic parsers will 
result in higher performance in broad coverage natural language
processing.

Furthermore, the human judgement experiment suggests that without context 
people achieve about 82\% accuracy on the same task.
If we assume that this figure represents the optimal accuracy,
then the model of noun compound syntax based 
on the meaning distributions theory is close to the best possible
without context.

While it is impossible to infer whether other \sll\ designs 
based on the meaning distributions theory will also exhibit 
high performance, the results here provide some empirical
evidence that the meaning distributions architecture is a fruitful
area of the design space.

\subsubsection*{Limitations}

I will review four of the limitations of this experimental work, three being 
limitations of the proposed parsing strategy, and the last of the experimental 
design.
\begin{enumerate}
\item The model is not sensitive to context.
\item Roget's categories do not always form appropriate semantic classes. 
\item Sense ambiguity can confuse the program.
\item The test given here is limited to one corpus.
\end{enumerate}

Possibly the most significant limitation of the parsing strategy is the 
failure to use contextual knowledge in making analyses.  The 
program is presented with each compound out of context and the 
model only uses information obtained from within the 
compound.  While there is no easy way to tell what part of the error rate is 
due to context dependence and what part is due to other model deficiencies 
(like violated assumptions in the model or poor parameter estimation), the 
human judgement experiment suggests that the correct analysis of a 
compound depends heavily on wider context.  Therefore any model which 
ignores context will exhibit a significant error rate.  

Another source of problems is the use of Roget's categories as conceptual 
groupings.  The intention of grouping words into concepts was to capture 
synonymy.  But Roget's thesaurus groups together words which are related 
in much more arbitrary ways.  For example, the category \tc{river}{348} 
contains all of \lingform{cataract}, \lingform{flux}, \lingform{fountain}, 
\lingform{force}, \lingform{hydraulics}, \lingform{rush}, 
\lingform{shower}, \lingform{syringe} and \lingform{tide}.  
The assumption that all these words (in their relevant senses) behave 
identically is clearly a poor one.  There are also many missing senses: 
\lingform{present} for instance only appears in the category 
\tc{giving}{784} when it should also appear in \tc{present time}{118} with 
\lingform{hour}, \lingform{nowadays} and \lingform{today}.  
Thus, the generalisations made by the model 
from words to concepts are both noisy and haphazard.  This 
problem might be overcome by the use of a large lexical 
ontology such as WordNet (Miller,~1990).

Even though the model performs a form of sense 
disambiguation, it can often go wrong here too.  In the test 
example \lingform{country music singer}, the program uses both of the 
possible senses of \lingform{country}, represented by the categories 
\tc{abode}{189} and \tc{government}{737a}.  Since 
\lingform{singer} is in the category \tc{musician}{416} and musicians can 
come from many different governed social units (villages, 
towns, cities, countries, etc) the program uses the 
\tc{government}{737a} sense of \lingform{country} to conclude 
incorrectly that a right-branching structure is more likely (that is, 
\lingform{country} modifies \lingform{singer} rather than 
\lingform{music}).  
If a practical sense disambiguation algorithm (see Yarowsky,~1995) 
were available then both the training and test sets could 
be preprocessed by this algorithm to yield sense-tagged data.  
There would then be no need to distribute the evidence arising 
from a word amongst its different possible concepts and, as long 
as the sense disambiguation worked accurately, these kinds of 
errors would be eliminated.

Finally, the results given here are specific to the style and size of 
corpus I have used.  It is not possible to make conclusions  
about the applicability of the technique to other types 
of text, or even about the behaviour on larger or smaller training 
corpora.  As in all statistical learning systems, it is important to 
ask whether sufficient training data was used.  Even though the corpus is 
fairly large,  there is a problem with data sparseness: a training 
set numbering tens of thousands has been used to estimate 
parameters numbering millions. However, the low guess rate 
(less than 4\%) and close to optimal accuracy suggests that the problem is 
not as severe as this ratio implies.  It seems likely that such performance 
under extremely data sparse conditions is due to the extreme non-uniformity 
of the bin distribution.  Also, a significantly larger corpus (written in a 
single register) is unlikely to be available to any but the largest 
of projects.  For the purposes of answering the scientific 
question of whether a practical statistical model can parse 
compound nouns, there appears to be sufficient training data.  But we 
cannot infer from the results given above whether using more 
data would help the correctness.


\section{Noun Compound Semantics} 



While the syntactic analysis of compound nouns explored in the first half of 
this chapter could be used to make parsing more efficient, it is only the first 
step in understanding compound nouns.  Understanding them also entails 
being able to identify the semantic relationships implied by them.  
The remainder of this chapter will describe some experiments in applying \sll\ 
techniques to this problem.

Though the ultimate goal is the semantic analysis of arbitrary noun compounds,
the experiments here concern only the paraphrasing of two word compounds 
that have a prepositional paraphrase.
The scope of the experiments is discussed in section~\ref{sec:ce_problem},
leading to a precise task definition.

In section~\ref{sec:ce_model}, I will develop a probabilistic model
of noun compound paraphrasing based on the meaning distributions theory.  
The model combines information from both the head and the
modifier in order to determine the semantic relationship between
them.

An overview of the components of the experimental setup is given
in section~\ref{sec:ce_design}, many of which are shared with
the earlier parsing experiments (see section~\ref{sec:cy_design}).
Full details of the experimental
method follow in section~\ref{sec:ce_method}.  These include
the method used to collect test and training data from the corpus,
the concepts used, the parameter estimation strategy and the
analysis procedure.

Section~\ref{sec:ce_results} gives the resulting performance
measures for the paraphrasing model, both with and without
a simple smoothing technique.  Further results derived by 
applying restrictions to avoid data sparseness also
appear there.  Finally, some conclusions are drawn
in section~\ref{sec:ce_discussion}.

\subsection{Defining the Semantic Analysis Problem} 
\label{sec:ce_problem} 

\subsubsection*{Semantic classifications}

Before considering different available semantic classifications of compounds, 
I should mention one assumption that is made universally by constructive 
theories of compound noun semantics.  It is usual to assume that compounds 
longer than two words are built up by a series of compounding steps, each 
one incorporating its own implicit semantic relationship.  For example, 
\lingform{statistics tutorial room schedule} involves three separate implied 
semantic relationships: the tutorials are about statistics, the room is used for 
the tutorials and the schedule describes the usage of the room.  It is assumed 
that each of these relationships behaves exactly as it would in a two word 
compound.  That is, the semantics of \lingform{statistics tutorial}, 
\lingform{tutorial room} and \lingform{room schedule} can be combined by 
conjunction to obtain the semantics of the whole.

If this is true, then giving a theory of the semantics of two word compounds 
suffices to give a theory of all compounds.  Since this assumption is so 
common, I will adopt it here.  Even if it is not strictly true, it seems 
reasonable to expect that longer compounds will largely be driven by the 
same semantic rules as two word compounds, so that any solution for the 
latter will represent substantial progress on the former.  It is worth pointing 
out here that one exception to this assumption is proposed by 
McDonald~(1982).  His claim is that one noun cannot be modified by two 
different modifiers with both implying the same semantic relationship.  The 
example used is \lingform{computer database management}, where 
\lingform{database} specifies the patient role of \lingform{management}, 
forcing \lingform{computer} to have the agent role, even though 
\lingform{computer management} would ordinarily imply a patient role.  
This constraint also appears to be implied by Finin's~(1980) algorithm.

Having restricted attention to two word compounds, I will now consider 
possible formulations of the semantic analysis task.  To make this task more 
concrete, it will be necessary to adopt a theory of their semantics.  Unlike for 
the grammar of compounds, there is virtually no concensus amongst 
linguists on the possible semantics of compounds, which complicates the 
task of choosing any one theory.

Levi's~(1978) theory (see section~\ref{sec:cn_meaning})
is attractive for two reasons.  First, it is the most well-developed and 
explicit theory of compound semantics available.  Second, it 
postulates a small set of possible semantic relations, which can be taken to 
be the target set.  Recall that she divides compounds into two kinds: 
nominalisations and those with recoverably deleteable predicates.  She 
claims that only nine predicates are recoverably deleteable from all 
compounds in the latter group.  As one support for this claim, she allocates a 
couple of hundred example compounds to their associated predicates.

So a natural problem specification would set the goal to be recovery of the 
deleted predicate in compounds so formed.  Stated in terms of Levi's 
generative semantics view, this involves reversing the final step of the 
derivation of the compound to find the intermediate semantic form.  For 
example, given the compound \lingform{olive oil} the predicate recovered 
should be \semrel{from} because this compound is derived by 
transformation of the intermediate form 
\lingform{oil}~\semrel{from}~\lingform{olive}.  Similarly, 
\lingform{mountain lodge} should be analysed as 
\lingform{lodge}~\semrel{in}~\lingform{mountain}.

One problem with this specification is that a given compound can have 
several different, valid analyses.  For instance, one entire sub-group of the 
\semrel{make} predicate (mass noun modifiers) \shortquote{has a regular 
alternative analysis under \semrel{be}}~(Levi,~1978:281).  This multiplicity 
of analyses Levi calls analytic indeterminacy.  Since the allocation of test 
cases to one or other predicate is partially a matter of individual preference, 
evaluation is complicated, especially since the different predicates are 
subjective and highly abstract.

An alternative approach would be to examine a set of compounds and 
develop a new semantic classification for them, according to the data.  
This follows the convention among linguists, all of whom seem to have 
their own individual semantic divisions.  
Vanderwende~(1993) chooses this path (see section~\ref{sec:cn_knowledge}) 
arriving at a set of 11 semantic relations, based on \lingform{wh}-questions, 
derived from inspection of a development set of 59 compounds.  
For example, \lingform{bird sanctuary} involves a \semrel{purpose} 
relation because the pre-modifier answers the question \scare{What for?}

The difficulty with this approach is that the resulting classifications are 
unlikely to be comprehensive and have no independent justification.  
Furthermore, they remain highly abstract and subjective.  Since linguists 
who have specialised in the study of compounds disagree about even the 
broad semantic divisions, attempting to specify our own comprehensive, 
precisely defined classification is unlikely to result in a useful scheme.  In 
the absence of a specific application to provide the necessary semantic 
distinctions, inventing intuitive ones can only lead to \foreign{ad hoc} 
results.

Another alternative, the one I will select, is to classify compounds according 
to their different paraphrases.  So the problem is now cast as one of finding 
the most preferred paraphrase of a compound.  Taking this tack makes 
evaluation far more concrete.  While preferences for different paraphrases 
are still subjective, paraphrases are at least precisely defined, unlike abstract 
semantic relations.  Decisions about the acceptability of different 
paraphrases are on similar footing to grammatical acceptability judgements, 
which form the foundation of syntactic theory even though they vary 
between dialects and idiolects.  Abstract semantic relations are, by 
comparison, shadowy, elusive constructs.

Though Leonard~(1984) provides her own semantic classification of 
compounds, the measurable goal of her computer analysis system is 
paraphrasing, so in effect she adopts this approach.  Given the compound 
\lingform{mountain vista}, her system produces the paraphrase 
\lingform{vista of a mountain or mountains}.  
Likewise, \lingform{Monday morning} is rewritten as 
\lingform{morning in or on a Monday}.
But the system's inability to distinguish between the two locative 
prepositions in this second example exposes the underlying theory based on 
semantic relations.  All three prepositions, \lingform{in}, \lingform{on}
and \lingform{at} are subsumed under one locative 
relation (Leonard,~1984:93).  
Similarly, the semantic aspects of definiteness and number are not 
distinguished by the semantic theory, leading to a somewhat underspecified 
paraphrase.  So the position Leonard takes is a hybrid between defining the 
problem in terms of abstract semantic relations and defining it by 
paraphrase.

In this work, I will define the problem solely in terms of paraphrasing.  By 
doing so, I am addressing a different, though related, problem.  For example,  
one semantic relation can be expressed by different paraphrases for purely 
collocational reasons.  Consider the compound \lingform{evening ride}, 
whose most preferred paraphrase in my dialect is \lingform{ride in the 
evening}.  Semantically, the compound serves to place the riding event 
within a periodic time interval.  The semantics of \lingform{night ride} are 
as similar to those of \lingform{evening ride} as it is possible to get.  Yet, 
the most preferred paraphrase in my dialect for \lingform{night ride} is 
\lingform{ride at night}.

Conversely, one paraphrase scheme can express different semantic relations.  
A \lingform{one month vacation} has most preferred paraphrase 
\lingform{vacation for one month}.  Similarly, \lingform{defense outpost} 
has most preferred paraphrase \lingform{outpost for defense}.  Yet the 
former clearly implies the duration of the vacation, while the latter implies 
the purpose of the outpost.  It should be clear from these examples that 
defining the problem in terms of paraphrase means addressing a different 
problem; though paraphrasing is neither obviously easier, nor obviously 
more difficult, it is certainly more concrete.

\subsubsection*{Types of compound paraphrases}

By selecting paraphrases as the target of semantic analysis, the need to select 
a set of abstract semantic classes is avoided.  However, we still need to 
know what paraphrases are possible.  Some help in this endeavour comes 
from looking again at the linguistic theories of compound noun semantics.  
Those that make any attempt to define or characterise the elements of their 
particular set of semantic relations, usually do so by means of paraphrases.  
For example, Warren~(1978) explains virtually all her semantic classes with 
reference to the typical paraphrase, summarising these observations in the 
following passage.
\begin{citedquote}{Warren,~1978:47--48}
The semantic relation expressed by a compound is normally made overt in 
its prepositional paraphrase.  Source-Result allows an 
\lingform{of}-paraphrase \dots  Copula compounds, however, do not 
allow any prepositional paraphrase \dots  Purpose compounds allow a 
\lingform{for}-paraphrase \dots 
\end{citedquote}
By collecting together all such characteristic paraphrases, we can arrive at a 
set of paraphrase schemes which, while not independently constructed, does 
have some independent claim to being comprehensive, at least to the extent 
that paraphrases represent the semantics of compounds.

All theories of compound semantics that give paraphrases posit at least one 
of the following three paraphrase groups.
\begin{enumerate}
\item Copula compounds with paraphrases like \lingform{fish that is a tuna}.
\item Verbal-nexus compounds or nominalisations with paraphrases like 
\lingform{demonstrations by students} and \lingform{construction of 
buildings}.
\item Prepositional compounds with paraphrases like \lingform{bars of 
steel}, \lingform{stories about war} and \lingform{commercials on 
television}.
\end{enumerate}
There are a few scattered examples which do not fit into these classes, for 
example causation (\lingform{virus that causes colds}) and price 
(\lingform{jacket that costs \$50}).  However, these appear to be infrequent.  

I will assume that it is possible to allocate all 
compounds into these three paraphrase groups in an objective 
manner.\footnote{In section~\ref{sec:ce_method} below I give the criterion 
used to distinguish nominalisations from prepositional compounds in the 
experiments.}
There are good reasons to treat these three paraphrase groups separately.  
Copula compounds clearly require hypernymic knowledge such as might be 
found in a taxonomic structure like WordNet, and it is possible that this 
would be sufficient to analyse them.  Analysis of the other two groups might 
benefit from this kind of information, but hypernymy alone is certainly 
insufficient.  It is plausible that copula compounds could be analysed by an 
\sll\ system designed to learn hypernymic relations from a corpus (see for 
example, Hearst~1992).  However, since this constitutes a significant 
research goal by itself, I will exclude copula compounds from the scope of 
the experiment.

The semantics of verbal-nexus compounds are tied intimately to those of the 
verb in question.  For this reason, Warren~(1978) only considers 
non-verbal-nexus compounds.  For example, to interpret \lingform{garbage 
collection}, knowledge of the semantics of, and case roles for, the verb 
\lingform{collect} are needed.  In order to go beyond the merely syntactic 
observation that \lingform{garbage} plays the role of direct object of this 
verb, we require a semantics for the verb.  So to address these types of 
compounds it appears necessary to have both a theory of, and a way of 
acquiring knowledge about, the semantics of verb phrases.\footnote{One 
approach to this is the predicate argument statistics collected by 
Hindle~(1990), which might be used as the basis of a verbal-nexus 
compound interpreter.} 
In addition, the morphological characteristics of verbal-nexus compounds 
are crucial to their interpretation.  Nominalisations can be created by various 
suffixes, often indicating different roles.  An \lingform{employee} is the 
patient of an \lingform{employment} action.  The agent is an 
\lingform{employer} and all of their staff are in their \lingform{employ}.  
So any deep analysis of the semantics of nominalisations will involve 
both detailed verb semantics and a 
substantial morphological component.  

For these reasons I will restrict the scope of the experiments below to 
compounds in the prepositional paraphrase group.  Since it is not hard to 
identify verbal-nexus compounds (their heads are morphologically
related to a verb),  this does not 
present a methodological difficulty (although see the notes in 
section~\ref{sec:ce_method} below).  I will make the further simplification 
of taking the goal to be selection of the preposition appearing in the most 
preferred paraphrase of the compound.  This ignores several semantic 
distinctions necessary for faithful paraphrasing, such as the number and 
definiteness of the modifier.  

These restrictions have several advantages:
\begin{itemize}
\item The range of possible \scare{semantics} of prepositional compounds 
can be precisely identified and defined.
\item There are only a small set of possible analyses.
\item Most linguistic theories have made reference to prepositional 
paraphrases in characterising semantic classes, so these compounds form an 
important group.
\item The granularity of the semantic distinctions (see 
section~\ref{sec:cn_accommodation}) is naturally fixed by the range of 
prepositions found in English.
\item Since prepositions are overtly expressed in text, the possibility of 
acquiring knowledge about their properties by statistical means is made 
more likely.
\end{itemize}

\subsubsection*{Prepositions as the target}

To construct a list of possible prepositional paraphrases of compounds I 
have used Warren's~(1978) study, the largest corpus based study of 
compound noun semantics available.\footnote{Recall from 
section~\ref{sec:cn_meaning} that her study involved semantic analysis of 
4,557 compounds (types) extracted from 360 thousand words (tokens) of the 
Brown corpus.}  She excludes verbal-nexus compounds, but allows copula 
ones.  For each of her major semantic sub-classes, she gives the typical 
preposition used to paraphrase compounds in that sub-class (see
Warren,~1978:48, table~4).\footnote{These 
classes are tabulated later in her book 
with {\em type} frequency counts within her sample (Warren,~1978:229).}  
In the case of her Place and Time sub-classes, she gives three prepositions 
\lingform{in}, \lingform{at} and \lingform{on} because they are all 
common enough to warrant inclusion.  This yields seven possible 
prepositions.  In addition to these I will include one extra preposition, 
because one of her minor sub-classes (Subject-Matter-Whole) within the 
Constitute class makes up 3.6\% of all her prepositional compounds (types 
not tokens).  Since these compounds have a preferred paraphrase using 
\lingform{about} rather than \lingform{of}, I will also include the former.

This procedure produces the following possible prepositional paraphrases of 
compounds (excluding verbal-nexus and copula compounds).
\begin{description}
\item[of:] \lingform{state laws} means \lingform{laws of the state}.
\item[for:] \lingform{a baby chair} means \lingform{a chair for babies}.
\item[in:] \lingform{morning prayers} means \lingform{prayers in the 
morning}.
\item[at:] \lingform{airport food} means \lingform{food at the airport}.
\item[on:] \lingform{Sunday television} means \lingform{television on 
Sunday}.
\item[from:] \lingform{reactor waste} means \lingform{waste from a 
reactor}.
\item[with:] \lingform{gun men} means \lingform{men with guns}.
\item[about:] \lingform{war story} means \lingform{story about war}.
\end{description}

This list excludes one paraphrase scheme from Warren's~(1978) table~4, 
that being \lingform{like}-paraphrases.  I have chosen to exclude this 
because such compounds can be analysed as copula if metaphor is taken into 
account.  For example, although the compound \lingform{barrel chest} has a 
preferred paraphrase of \lingform{chest like a barrel}, the paraphrase 
\lingform{chest that is a barrel} also conveys the same meaning in the same 
contexts.  The former is preferred only because it is more explicit.  Since 
metaphor appears everywhere in language and any system which exhibits 
deep understanding must model it in some form, it is economical to analyse 
these compounds as metaphorical copula compounds.  It is also possible for 
prepositional paraphrases to make use of metaphor; for example, 
\lingform{steel father} can be paraphrased as \lingform{father of steel}, 
both of which are not usually intended literally. 

The example of \lingform{airport food} above raises another general issue.  
The relationships expressed by compounds are usually inherent or typical 
relationships.  As we've seen in section~\ref{sec:cn_meaning}, a 
\lingform{tree kangaroo} is {\em typically} found in trees.  It does not cease 
being a tree kangaroo when it hops down to earth.  In evaluating the possible 
paraphrases of compounds, typicality is implied, so that assigning the 
preposition \lingform{at} to the compound \lingform{airport food} 
represents the paraphrase \lingform{food typical at an airport}, rather than 
referring to any food that happens to be carried there.

So now I can precisely state the problem addressed in the experimental work 
below.

\begin{description}
\item[Problem Statement:] Given a non-copula, non-verbal-nexus 
compound, predict which of the following prepositions is most likely to be 
used in the preferred paraphrase of the compound, allowing for both 
metaphorical paraphrases and typicality: \lingform{of}, \lingform{for}, 
\lingform{in}, \lingform{at}, \lingform{on}, \lingform{from}, 
\lingform{with} and \lingform{about}.
\end{description}

\subsection{A Probabilistic Model of Noun Compound Paraphrasing} 
\label{sec:ce_model} 

To build a statistical language learner for this problem we begin
with a probabilistic model.  Using such a model we can compute the
probability of each of these 8 outcomes and then choose the most probable 
one.  Let $P$ be the set of prepositions above and $N$ the set of nouns. 
The appropriate analysis function can be written as
\begin{equation}
(\forall n_1,n_2 \in N) (A(n_1,n_2) = \argmax{p \in P} \Pr(p | n_1,n_2))
\end{equation}
where $\Pr(p | n_1,n_2)$ is given by the probabilistic model.  
In this section I 
will give a model using the meaning distributions theory.

Consider the complete joint distribution model for this problem.  It contains 
one parameter for every triple, $\langle p, n_1, n_2 \rangle$.  Assuming a 
vocabulary of 90 thousand nouns yields over 56 billion free parameters.  
Without hundreds of millions of training instances, it is not possible to train 
this model, even allowing for skewed distributions and techniques like 
smoothing.  Therefore we need to make some assumptions to reduce the data 
requirements.

The first step is to adopt the meaning distributions theory.  Let $C$ be a set 
of concepts, being the semantic representation of nouns, with $\phi_1: N 
\rightarrow 2^C$ giving the possible semantics of each noun.  Similarly, let 
$R$ be a set of roles, being the semantic representations of the eight 
prepositions, with $\phi_2: P \rightarrow 2^R$ giving the possible semantics 
of each preposition.  The meaning distributions theory requires a 
probabilistic model to give the set of probabilities $\Pr(r | c_1,c_2)$ where 
$r \in R, c_1,c_2 \in C$.  The syntactic component defined by the source 
maps $\phi_1$ and $\phi_2$ (and their corresponding image maps) then 
fixes the probabilities $\Pr(p | n_1,n_2)$.

The second step is to assume 
that the head and modifier of the compound contribute independent 
evidence.  This factors the model into two parts: one for representing the 
information contributed by the head (association between $r$ and $c_2$) 
and the other for information contributed by the modifier (association 
between $r$ and $c_1$).  

While this is a crude approximation, it finds some support in the literature.  
For instance, the idea is central to Ryder's~(1994) schema based theory of 
noun compound meanings.
\begin{citedquote}{Ryder,~1994:63}
\dots nouns are generally strongly autonomous, and so in a noun-noun 
compound both elements often have equal autonomy, rather than one being 
clearly dependent on the other.
\end{citedquote}
Accordingly, the process of interpreting a compound noun involves finding a 
relation in which both nouns are plausible participants.  In the model below, 
the probability distribution $\Pr(r | c_1)$ gives a profile of the relations in 
which the modifier is likely to participate, while the probability distribution 
$\Pr(r | c_2)$ gives a profile of the relations in which the head is likely to 
participate.  To match these profiles together, the model multiplies the 
probability distributions.

We can now formally define the model.
Recall that the goal is to 
assign probabilities to different possible paraphrases, so we need a simple 
expression for $\Pr(p | n_1,n_2)$ for each $p \in P$.  Proceed as follows.
\begin{eqnarray}
\Pr(p | n_1,n_2) & = & \sum_{c_1 \in \phi_1(n_1);c_2 \in \phi_1(n_2)}
		\hspace{-1cm} 
		\Pr(p | c_1, c_2, n_1, n_2) \Pr(c_1, c_2 | n_1,n_2) 	
						\nonumber \\
	& = & \sum_{c_1 \in \phi_1(n_1);c_2 \in \phi_1(n_2)}
		\hspace{-1cm} 
		\Pr(c_1, c_2 | n_1,n_2) 	
		\sum_{r \in \phi_2(p)} 
		    \Pr(p | r, c_1, c_2, n_1, n_2) 
		    \Pr(r | c_1, c_2, n_1, n_2) \nonumber \\
	& = & \sum_{c_1 \in \phi_1(n_1);c_2 \in \phi_1(n_2)}
		\hspace{-1cm} 
		\Pr(c_1, c_2 | n_1,n_2) 	
		\sum_{r \in \phi_2(p)} 
		    \Pr(p | r) \Pr(r | c_1, c_2) \nonumber \\
	& = & \sum_{r \in \phi_2(p)} \Pr(p | r) 
		\sum_{c_1 \in \phi_1(n_1);c_2 \in \phi_1(n_2)}
		    \hspace{-1cm} 
		    \Pr(r | c_1, c_2) 
		    \frac{\Pr(n_1,n_2 | c_1, c_2) \Pr(c_1, c_2)}{\Pr(n_1,n_2)}
\label{eq:ce_model_restate}
\end{eqnarray}
The third step uses $\Pr(p | r, c_1, c_2, n_1, n_2) = \Pr(p | r)$ 
and $\Pr(r | c_1, c_2, n_1, n_2) = \Pr(r | c_1, c_2)$, both
of which follow from the meaning distributions theory 
when $n_1 \in \theta_1(c_1)$, $n_2 \in \theta_1(c_2)$ and
$p \in \theta_2(r)$.
The other steps follow from probability theory and Bayes' rule.

To simplify further, it will be necessary to make some assumptions about the 
syntactic mappings $\phi_1$, $\phi_2$ and the corresponding images 
$\theta_1$ and $\theta_2$.
\begin{assumption}[Homogeneous Syntax]
Suppose that $\theta_1$ and $\theta_2$, are such that the following are each 
constant:
\begin{eqnarray*}
(\forall c \in C) (|\theta_1(c)| & = & k_1) \\
(\forall r \in R) (|\theta_2(r)| & = & k_2)
\end{eqnarray*}
\end{assumption}

Using the meaning distributions theory, these assumptions result in two of 
the probabilities above being constant.  In particular, $\Pr(p | r) = 
\frac{1}{k_2}$ and $\Pr(n_1,n_2 | c_1,c_2) = (\frac{1}{k_1})^2$.  This 
simplifies equation~\ref{eq:ce_model_restate} to
\begin{eqnarray}
\Pr(p | n_1,n_2) & = & \frac{1}{k_2} (\frac{1}{k_1})^2 
	\sum_{r \in \phi_2(p)}
	\sum_{c_1 \in \phi_1(n_1);c_2 \in \phi_1(n_2)}
	\frac{\Pr(r | c_1, c_2) \Pr(c_1, c_2)}{\Pr(n_1,n_2)} \nonumber \\
 & = & \frac{1}{k_2} (\frac{1}{k_1})^2 
	\sum_{r \in \phi_2(p)}
	\sum_{c_1 \in \phi_1(n_1);c_2 \in \phi_1(n_2)}
	\frac{\Pr(r, c_1, c_2)}{\Pr(n_1,n_2)} \nonumber \\
 & = & \frac{1}{k_2} (\frac{1}{k_1})^2 
	\sum_{r \in \phi_2(p)}
	\sum_{c_1 \in \phi_1(n_1);c_2 \in \phi_1(n_2)}
	\frac{\Pr(c_1 | r, c_2) \Pr( c_2 | r) \Pr(r)}{\Pr(n_1,n_2)} 
\label{eq:ce_model_eliminate}
\end{eqnarray}

The denominator will cancel during the analysis, so 
this expression depends only on 
the probabilistic conceptual model giving the distributions of 
roles and concepts.  This is based on two assumptions.
\begin{assumption}[Uniform Priors]
Each role is equally likely.  That is,
\begin{equation}
(\forall r \in R) \Pr(r) = k_3
\end{equation}
\end{assumption}
This means that we have no \latin{a priori} preference for one 
semantic relationship over another.  We also need the crucial independence
assumption.
\begin{assumption}[Head-Modifier Autonomy]
Let the probability of a modifier participating 
in a given implicit relationship be independent of the head.  
That is, 
\begin{equation}
(\forall c_1,c_2 \in C) (\forall r \in R) 
	(\Pr(c_1 | r, c_2) = \Pr(c_1 | r))
\end{equation}
\end{assumption}
This factorises the model into two parts: information from the head
and information from the modifier.\footnote{I have abused 
notation here since $\Pr(c_1 | r) \neq \Pr(c_2 | r)$ even when 
$c_1 = c_2$.  The former probability is the likelihood of concept $c_1$ 
appearing as the {\em modifier} of a relationship $r$, 
while the latter is that of 
concept $c_2$ appearing as the {\em head} of that relationship.}
Equation~\ref{eq:ce_model_eliminate} can now be reformulated as
\begin{equation}
\Pr(p | n_1,n_2) = \frac{k_3}{k_2 \Pr(n_1, n_2)} (\frac{1}{k_1})^2 
	\sum_{r \in \phi_2(p)}
	\sum_{c_1 \in \phi_1(n_1);c_2 \in \phi_1(n_2)}
	\Pr(c_1 | r) \Pr(c_2 | r) 
\label{eq:ce_model_factorise}
\end{equation}
and we have a probabilistic model with two sets of parameters $\Pr(c_1 | r)$ 
and $\Pr(c_2 | r)$ for a total of $2(|C|-1).|R|$ free parameters.

Although the model above is stated in terms of roles, in 
section~\ref{sec:ce_problem} I argued that the most concrete form of 
semantic representation is the prepositional paraphrase.  Therefore, in the 
experimental work below I take $R$ to simply be $P$, with $\phi_2$ and 
$\theta_2$ transparent maps (that is, they return singleton sets containing 
their arguments).\footnote{One half of the homogeneous syntax assumption 
now holds trivially.}  Since this removes the outer sum, 
and the outer factor is constant for any 
given analysis, the analysis process is now given by
\begin{equation}
(\forall n_1,n_2 \in N) (A(n_1,n_2) = \argmax{p \in P} 
	\sum_{c_1 \in \phi_1(n_1);c_2 \in \phi_1(n_2)}
		\Pr(c_1 | p) \Pr(c_2 | p))
\label{eq:ce_model_analysis}
\end{equation}

\subsection{Experimental Design} 
\label{sec:ce_design} 

In the experiments below, I will investigate the idea of using counts of 
prepositional phrases in a large corpus to estimate the free parameters of the 
model above.  That is, predictions about the semantics of noun compounds
will be made using observations of prepositional phrases.
The key assumption here is that semantically 
related pairs of concepts are expressed equally often by means of 
a compound noun and by means of a prepositional phrase.  

For example, when the meaning 
\lingform{haven} $\stackrel{\rm FOR}{\rightarrow}$ \lingform{reptiles}
is intended, the expressions \lingform{reptile haven} and
\lingform{haven for reptiles} will be used equally frequently.  Thus the
frequency of \lingform{reptile haven} (where the intended meaning can be 
paraphrased using \lingform{for}) is assumed to be the same as the 
frequency of \lingform{haven for reptiles}.  This means that $\Pr(c_1|r)$
and $\Pr(c_2|r)$ not only represent the probabilities of noun compound
meanings but also of prepositional phrase meanings.  The two forms
of expression are generated from the same underlying meaning distribution.

While this assumption follows from the meaning distributions theory, 
there is a good reason why it might be false.  
If a meaning can be expressed by a lexicalised compound, this form will 
be preferred because it is shorter than the prepositional paraphrase.
Therefore the prepositional form will occur only rarely.  
Conversely, if the compound form is ambiguous, the 
prepositional form will be preferred and the compound (with 
associated semantic relation) will occur only rarely.  This suggests that the 
probabilities of the two forms might be inversely related. However, even if 
strong preferences for one form over the other are generally observed, the 
model might still work well for the following reason.  If there are semantic 
relations that {\em never} hold between two concepts, the corresponding 
prepositional phrases should not occur at all, allowing the model to rule out 
at least these relations.  

Luckily, the effects of such frequency inversion are reduced by the 
head-modifier autonomy assumption.  By factorising the model into two 
parts, we can measure the frequency of \lingform{haven for reptiles} as a 
product of two different frequencies, those of \lingform{haven for} and 
\lingform{for reptiles}.  Now even if \lingform{reptile haven} were a 
lexicalised compound, thus making \lingform{haven for reptiles} a rare 
construction, we could still expect to find \lingform{haven for} and 
\lingform{for reptiles} with relatively high frequency.

Also, in these examples, I have been using word frequencies, but the model 
is stated in terms of concept frequencies.  For example, instead of counting 
occurrences of \lingform{haven for}, the model maps \lingform{haven} onto 
one or more concepts (say, \concept{safe\_place}) and then counts 
occurrences of \lingform{\concept{[safe\_place]} for}.  If the concepts 
capture appropriate generalisations, this will help reduce the frequency 
inversion problem because conceptually similar forms will have different 
lexicalisation patterns.

\subsubsection*{Choosing the concept set}

Before turning to an overview of the system architecture, I will
address the question of what to use for the set of concepts.
Using Roget's thesaurus is one possibility that I will explore; 
that is, let $C_a$ be the set of Roget's categories, 
and $\phi_{1a}(w)$ the set of categories in which $w$ 
appears.  The 1911 version used here is freely available and was described 
in section~\ref{sec:cy_design}.
However, the task I am addressing is paraphrasing, not semantic 
interpretation.  As I have argued in section~\ref{sec:ce_problem}, the 
paraphrasing problem is influenced by collocative patterns that depend on 
specific words.  Recall for instance, the example involving \lingform{in the 
evening} and \lingform{at night}, where semantically identical constructions 
have different paraphrases.  
Conceptual associations should be applied to prepositional roles, 
not to prepositions themselves (see Dras and Lauer,~1993, for
a more detailed argument of this point).

Therefore, because the model is based on prepositions rather than roles,  I 
will also investigate a lexical parameterisation; 
that is, let $C_b = N_{\mbox{test}}$, 
where $N_{\mbox{test}}$ is the set of nouns found in the test set, 
and $\phi_{1b}(w)$ be the singleton set containing $w$.  
Just as for 
the experiments on compound syntax, constructing a lexical model for the 
entire vocabulary would require very much larger amounts of storage space 
(see section~\ref{sec:cy_comparisons}).  
The portion built here is just sufficient to 
measure performance on the test set.  In contrast, the implementation of the 
conceptual parameterisation (using Roget's categories) is a complete 
compound noun semantics analyser.  Other than this difference though, the 
same architecture, training set and analysis procedure will be used for both 
parameterisations.

\subsubsection*{Architecture}

I will now give an overview of the architecture of the system.  Details of the 
various components will be given in section~\ref{sec:ce_method}.  The 
experiment involves four main steps.
\begin{enumerate}
\item Extract a test set of two word compounds from the corpus and 
manually assign the most preferred paraphrase to each.
\item Define a pair of patterns for extracting examples of the nouns 
modified by, and the nouns governed by, the eight prepositions.
\item Use these patterns to estimate the distributions $\Pr(c_1|p)$
and $\Pr(c_2|p)$ for each preposition 
across both Roget's categories ($C_a$) and nouns in the test set 
($C_b$).
\item Use the distributions and the probabilistic model above to analyse each 
element of the test set and record the accuracy.
\end{enumerate}

The corpus is the same as that used for the compound parsing experiments, 
\publicationname{The New Grolier Multimedia Encyclopedia}~(Grolier 
Inc.,~1992), containing approximately 8 million words (for more details see 
section~\ref{sec:cy_design}).  The test set is a collection of 400 two word 
noun sequences extracted from the corpus by means of a simple pattern 
given below, and which I have manually annotated with preferred 
paraphrases.

Since tagging the corpus with the Brill tagger (Brill,~1993) proved useful 
when training the models of compound syntax (see 
section~\ref{sec:cy_comparisons}), 
this strategy will be used here too.  Note that 
this tagger is freely available and has been applied using the rule set 
provided with it (trained on the Brown corpus).  
The patterns for extracting the 
noun modified by, and the noun governed by, a preposition rely only on the 
parts of speech assigned by the tagger, and so can be applied automatically 
to any text using only freely available resources.

Applying the first pattern to the corpus yields an observed distribution of 
concepts that are modified by each of the eight prepositions.  I 
will call this the \newterm{head distribution}.  
Applying the second pattern to the corpus yields an observed distribution of 
concepts that are governed by each of the eight prepositions, in 
a symmetrical way.  I will call this the \newterm{object distribution}.  
This distribution records the frequency of each concept  
appearing as the object of each preposition.

These two distributions are combined by the probabilistic model to give a 
ranking of the possible paraphrases of the compound.  The accuracy of the 
system is measured as the proportion of highest ranked paraphrases that 
match the most preferred one assigned manually.  Note that because the goal 
is to select the {\em most preferred} paraphrase, the system will score 
nothing for a merely acceptable paraphrase; the answer given must match the 
single most preferred paraphrase, not just have the same meaning.  Also, the 
system is required to give an answer to every test example, so that accuracy 
is measured at 100\% coverage.  

In the next section, the experimental method is given in detail, including the 
extraction patterns for training and testing and the precise analysis method.  
The reader may skip these details if desired, noting only the two training 
patterns in equations~\ref{eq:ce_method_headpattern} 
and~\ref{eq:ce_method_objpattern}, and 
equation~\ref{eq:ce_method_estimates} describing parameter estimation.

\subsection{Experimental Method} 
\label{sec:ce_method} 

\subsubsection*{Test set}

The first step is to extract a test set of two word compound nouns.  Recall 
from section~\ref{sec:cy_method} that the training data for the compound 
parsing experiments consisted of two word noun sequences collected by 
means of a simple pattern.  The pattern is reproduced here.
\begin{equation}
T_{\mbox{test}} = \{ (w_1, w_2) \mid w_1 w_2 w_3 w_4; w_1,w_4 \notin 
{\cal N}; w_2, w_3 \in {\cal N} \}
\end{equation}
Here, $w_1 w_2 w_3 w_4$ denotes the occurrence of four tokens
in sequence in the corpus, and $\cal N$ is the set of words listed in the 
University of Pennsylvania morphological analyser lexicon as being sure 
nouns (see section~\ref{sec:cy_method} for more detail).  After removing 
pairs containing a word not in Roget's thesaurus, this produces 24,251 two 
word noun sequences.  From these a random sample of 400 sequences was 
selected as the test set. 

The position within the corpus of each noun sequence was kept along with it, 
so that each of the sequences in the sample could be examined in context.  
By looking at the full context in each case, I assigned one of the following 
annotations to each test case using a specially developed 
annotation environment.
\begin{description}
\item[E] Extraction errors, where the noun sequence does not form a noun 
compound, plus proper names and foreign compounds.
\item[B] Copula compounds where the modifier and head both classify the 
indicated object (including some metaphorical readings).
\item[V] Verbal-nexus compounds where the modifier takes the part of 
either subject or object of a verb of which the head is a nominalisation.
\item[O,R,I,T,W,F,N,A] Prepositional compounds whose most preferred 
paraphrase involves \lingform{of}, \lingform{for}, \lingform{in}, 
\lingform{about}, \lingform{with}, \lingform{from}, \lingform{on} or 
\lingform{at} respectively.
\end{description}

Assignments of \scare{E} and \scare{B} are straightforward.
The noun pair \lingform{passengers hostage} is an extraction error (E)
since the two nouns are separate complements of \lingform{holding} 
in the context in which it appears.  
The compound \lingform{boyar duma} (a Russian 
aristocratic title abolished in the early 18th century) is also marked with 
\scare{E}.  The compound \lingform{patron goddesses} is copula (B),
as is \lingform{puppet government} based on metaphorical 
interpretation.  

Somewhat more difficult to distinguish are the verbal-nexus compounds (V).  
It is often not clear whether a noun is a nominalisation of a verb or not.  
The verb \lingform{colonise} is a \scare{verbalisation} of the noun 
\lingform{colony}, not the other way round.  Though this case is clear, an 
objective decision procedure is needed to define precisely which nouns are 
nominalisations.  For the purposes of annotating the test set, I have taken a 
noun to be a nominalisation only where the etymology marked in the Shorter 
Oxford English Dictionary~(Onions,~1973) shows it as being derived from 
the corresponding verb.  For example, \lingform{hull maintenance} is 
assigned \scare{V} because the head is derived from the verb 
\lingform{maintain}, but \lingform{plutonium theft} is considered a 
prepositional compound (\scare{O}) because the head is derived from the 
noun \lingform{thief} (the verb \lingform{thieve} is also from this noun).

A further condition of being marked as a verbal-nexus compound is the 
requirement that the modifier is semantically interpreted as playing the part 
of either the subject or object of the verb that corresponds to the head.  In 
the example above, the modifier plays the part of object: it is hulls that are 
maintained.  In \lingform{peasant rebellion}, the modifier plays the part of 
subject, and so this example receives a \scare{V} too.  In contrast, 
\lingform{city dwellers} is considered a prepositional compound (\scare{I}) 
because, although the head is derived from the verb \lingform{dwell}, the 
modifier does not act as subject or object of this verb.

The annotation environment presented the user with the complete paraphrase 
for all eight prepositions, along with the context surrounding the compound.  
The annotator could therefore carefully compare all of the allowed 
paraphrases with the given context and use the mouse to select the most 
preferred paraphrase before proceeding to the next test case.  In each case, 
one and only one prepositional paraphrase could be selected, so that every 
example in the test set was assigned a single correct answer.

All the examples given so far in this section are taken from the annotated test 
set used in the experiments.  The entire test set, along with annotations is 
given in appendix~\ref{appendix:cetest}.  The distribution of annotation 
types is given in table~\ref{tb:ce_method_testtypes}.
\begin{table*}
\centering
\begin{tabular}{|c|c|c|c|c||c|} \hline
Error & Be & Verbal$_{\mbox{subj}}$ & Verbal$_{\mbox{obj}}$ & 
Prepositional & Total  \\
\hline
15 & 41 & 19 & 43 & 282 & 400 \\
\hline
4\% & 10\% & 5\% & 11\% & 70\% & 100\% \\
\hline
\end{tabular}
\caption{Distribution of annotation types in paraphrasing test set}
\label{tb:ce_method_testtypes}
\end{table*}

Since the goal of the experiment was to paraphrase prepositional 
compounds, \scare{E}, \scare{B} and \scare{V} cases were eliminated, 
leaving a test set of 282 
compounds, each annotated with one of the eight target prepositions.  
Table~\ref{tb:ce_method_testdist} shows the distribution of answers across 
the test set.
\begin{table*}
\centering
\begin{tabular}{|c|c|c|c|c|c|c|c||c|} \hline
Of & foR & In & abouT & With & From & oN & At & Total \\
\hline
94 & 78 & 39 & 22 & 18 & 13 & 12 & 6 & 282 \\
\hline
33\% & 28\% & 14\% & 8\% & 6\% & 5\% & 4\% & 2\% & 100\% \\
\hline
\end{tabular}
\caption{Distribution of prepositions involved in paraphrases of test set}
\label{tb:ce_method_testdist}
\end{table*}
The distribution of different paraphrases shows a heavy bias towards three 
of the prepositions, \lingform{of}, \lingform{for} and \lingform{in}, which 
account for 75\% of the test cases.  These three also have the largest amount 
of training data available, a fact that I will exploit in 
section~\ref{sec:ce_results}.  The most important point to note from the 
table is that the simple strategy of always choosing \lingform{of} as the 
paraphrase achieves 33\% accuracy on this test set.  Therefore, any system 
that does worse than this is of no use.

\subsubsection*{Training data and parameter estimation}

Setting aside the test set, the second step is to extract relevant training 
information from the corpus.  According to the probabilistic model, the 
information required is a pair of distributions for each preposition.  I have 
used one pattern for the head distribution and one for the object distribution, 
which I will describe in turn.

The pattern for the head distribution is extremely simple.  It assumes 
that prepositions following a noun post-modify that noun.  
Whenever one of the eight target prepositions is observed to follow a word 
tagged as a noun, the noun is counted as an observation relevant to the head 
distribution of that preposition.  For example, \lingform{story}/\tag{nn} 
\lingform{about}/\tag{in} will generate a training instance for the head 
distribution of \lingform{about}, increasing the estimate of the
probability that concepts expressed by the word \lingform{story} are 
the head of an \lingform{about} prepositional phrase.
In Brill's tag set, I take the tags \tag{nn}, \tag{nns}, 
\tag{nnp}, \tag{nnps} and \tag{vbg} to be nouns for this 
purpose.\footnote{These are common singular nouns, common plural nouns, 
proper singular nouns, proper plural nouns and present participles, 
respectively.  \tag{vbg} is included in order to allow gerunds that have 
been incorrectly tagged; only words listed in the thesaurus as nouns are 
counted in the distribution, so few true present participles are likely to be 
counted.}

Formally, the pattern is
\begin{equation}
T_{\mbox{head}}(p) = \{ w_1 \mid w_1 w_2; w_2 = p; t_1 \in 
\{\mbox{\tag{nn}, \tag{nns}, \tag{nnp}, \tag{nnps}, \tag{vbg}}\} \}
\label{eq:ce_method_headpattern}
\end{equation}
where $w_1 w_2$ denotes the occurrence of two tokens in sequence in the 
corpus and $t_1$ is the tag assigned to $w_1$ by the tagger.

The pattern for the object distribution is more complex.  It assumes that the 
first noun following a preposition is the head of the preposition's object 
noun phrase.  A sequence of up to three noun modifiers are permitted to 
appear between the preposition and the noun, including determiners, 
adjectives and cardinals.  To avoid catching prepositions used as verb case 
role markers, the pattern also checks for a possible nominal attachment for 
the preposition.  This is achieved by checking that the word preceding the 
preposition is a noun, just as the pattern for the head distribution does.  
Thus, the object distribution pattern applies in a strict subset of the places 
where the head distribution pattern applies.  

Whenever one of the eight target prepositions follows a word tagged as a 
noun and is followed by up to three noun modifiers and then another word 
tagged as a noun, the second noun is counted as an observation relevant to 
the head distribution of that preposition.  For example, 
\lingform{tower}/\tag{nn} \lingform{on}/\tag{in} \lingform{the}/\tag{dt} 
\lingform{western}/\tag{jj} \lingform{horizon}/\tag{nn} will generate a 
training instance for the object distribution of \lingform{on}, increasing the 
estimate of the probability that concepts expressed by the word 
\lingform{horizon} are the object of an \lingform{on} prepositional phrase.
The same set of tags as above are taken to be nouns.  
I take Brill's tags \tag{jj}, \tag{dt}, \tag{cd}, \tag{prp\$} 
and \tag{pos} to be possible noun modifiers for this 
purpose.\footnote{These are adjective, determiner, cardinal, possessive 
pronoun and possessive markers, respectively.  An idiosyncrasy of the 
tagger is tagging nouns with possessive markers incorrectly as adjectives.  
The pattern exploits this to find the subsequent head noun by allowing 
possessive markers to be noun modifiers.}

Formally, the pattern is
\begin{eqnarray}
\lefteqn{T_{\mbox{object}}(p) = }  \nonumber \\
  & \{ &  w_{i} \mid w_1 w_2 \ldots w_{i}; 3 \leq i \leq 6; w_2 = p; 
\nonumber \\
  & &  t_1,t_{i} \in \{\mbox{\tag{nn}, \tag{nns}, \tag{nnp}, \tag{nnps}, 
\tag{vbg}}\}; \nonumber \\
  & &  t_3,\ldots,t_{i-1} \in \{\tag{jj}, \tag{dt}, \tag{cd}, \tag{prp\$}, 
\tag{pos}\} \} 
\label{eq:ce_method_objpattern}
\end{eqnarray}
where $w_1 w_2 \ldots w_{i}$ denotes the occurrence of $i$ tokens in 
sequence in the corpus and $t_j$ is the tag assigned to $w_j$ by the tagger.

Table~\ref{tb:ce_method_trainsizes} shows the number of training instances 
generated by these two patterns from the corpus.  The left side of the table 
shows the quantities extracted for training the model based on $C_a$, the 
Roget's categories.  This model has been trained on the full vocabulary of 
the thesaurus, covering 20,445 noun stems in 1043 categories.  The right 
hand side shows the quantities extracted for training the (partial) model 
based on $C_b$, the nouns in the test set.  It contains 462 noun stems.

\begin{table*}
\centering
\begin{tabular}{|r|r|c|r|r|} \hline
\multicolumn{2}{|c|}{Thesaurus} & \multicolumn{1}{c|}{Preposition} & 
\multicolumn{2}{c|}{Test Set Nouns} \\ 
\cline{1-2} \cline{4-5}
\multicolumn{1}{|c|}{Object} & \multicolumn{1}{c|}{Head} &    & 
\multicolumn{1}{c|}{Object} & \multicolumn{1}{c|}{Head} \\
\hline
112150 & 219353 & of & 26852 & 43037 \\
13402 & 21409 & for & 3254 & 3712 \\
27828 & 63627 & in & 7030 & 13352 \\
867 & 1801 & about & 215 & 375 \\
6833 & 14030 & with & 1375 & 2640 \\
5465 & 12948 & from & 1092 & 1981 \\
7457 & 12773 & on & 1359 & 1829 \\
3671 & 8527 & at & 469 & 1773 \\
\hline
\end{tabular}
\caption{Numbers of training instances generated by the paraphrasing 
patterns}
\label{tb:ce_method_trainsizes}
\end{table*}

The third step in the experiment is to use the evidence provided by the 
training set to estimate the parameters of the model.  These estimates are 
stored in a data structure for later use by the analyser.  The probabilistic 
model employs two parameters for each concept-preposition pair: the head 
and object probabilities, and so has two data structures.  Both of these are 
represented as a matrix with eight columns (one for each preposition), whose 
rows are indexed by concepts.  

We interpret $A_{ij}^{\mbox{\sc head}}$ as the probability that the 
concept represented by the $j$th thesaurus category will be the {\em head} 
of a compound noun whose most preferred paraphrase involves the $i$th 
preposition.  Likewise, we interpret $A_{ij}^{\mbox{\sc obj}}$ as the 
probability that the concept represented by the $j$th thesaurus category will 
be the {\em modifier} of a compound noun whose most preferred paraphrase 
involves the $i$th preposition.  Thus, when $r$ is the $i$th preposition, 
$A_{ij}^{\mbox{\sc head}} = \Pr(c_2 | r)$ when $c_2$ is the $j$th 
thesaurus category and $A_{ij}^{\mbox{\sc obj}} = \Pr(c_1 | r)$ when 
$c_1$ is the $j$th thesaurus category.

Because the nouns identified by the training patterns can appear in multiple 
Roget's categories (in the conceptual model, based on $C_a$), each of the 
training instances generally contributes to more than one probability 
estimate.  Just as was done for the experiments on parsing compounds, this 
is handled by dividing the counts contributed by each noun by the number of 
categories that the noun appears in.  So, when \lingform{web} is returned by 
one of the patterns, the counts for categories \tc{crossing}{219} and 
\tc{texture}{329} are both incremented by $\frac{1}{2}$, since 
\lingform{web} is in both of them.  This procedure corresponds to Bayesian 
estimation where the prior probability of the different concepts is assumed 
uniform.  This problem does not arise in the lexical model, based on $C_b$.

To begin with, the parameters will be estimated using maximum likelihood 
under the binomial distribution, as was done for the parsing experiments (see 
section~\ref{sec:cy_method}).  Given the frequency with which each noun 
was returned by each pattern for each preposition, say $\countfn_{\mbox{\sc 
head}}(w, p)$ and $\countfn_{\mbox{\sc obj}}(w, p)$, the parameter 
estimates 
are given by
\begin{eqnarray}
\Pr(c_x | r) & = &
\frac{1}{\eta_r}
\sum_{w \in \theta_1(c_x)}
\frac{\countfn_{\mbox{\sc d}}(w, r)}{\ambig(w)}  
\label{eq:ce_method_estimates} \\
\eta_r & = &
\sum_{w \in N}
\frac{\countfn_{\mbox{\sc d}}(w, r)}{\ambig(w)} \nonumber
\end{eqnarray}
where $\mbox{\sc d} = \mbox{\sc obj}$ when $x = 1$ and $\mbox{\sc d} = 
\mbox{\sc head}$ when $x = 2$.

Using maximum likelihood is more problematic with the current model than 
it was in the parsing experiments.  Since two independent estimates are 
multiplied, there are two opportunities for statistical zeroes to arise.  If 
$\countfn_{\mbox{\sc head}}(w, r) = 0$ for all $w \in \theta_1(c_2)$ then 
$\Pr(c_2 | r)$ will be estimated as zero.  Similarly, $\Pr(c_1 | r)$ will be 
estimated as zero, if $\countfn_{\mbox{\sc obj}}(w, r) = 0$ for all $w \in 
\theta_1(c_1)$.  If {\em either} of these conditions holds, then $\Pr(c_1, r, 
c_2)$ will be assigned the value zero, regardless of how large the counts 
in the alternate distribution are.  

This has the effect of hiding information.  Suppose, for example, that all the 
object distribution counts for concepts expressed by a given noun, $n_1$, 
were zero (that is, this noun was never identified by the pattern as the first 
noun following one of the eight prepositions, and neither were any of the 
other nouns expressing the same concepts).  This would be 
reasonably likely if $n_1$ is in only one small category.  
In the lexically parameterised model, every noun is in its own category,
so it is highly likely.  

Because the relevant object distribution counts are zero, all the
relevant parameter estimates $\Pr(c_1|p)$ will also be zero.
This means that regardless of the values of $\Pr(c_2|p)$,
every $\Pr(p | n_1,n_2)$ will be zero as well.
No matter what the head distribution counts for $n_2$ 
are, the model will be unable to select one preposition over another.
If $n_2$ expresses a concept with a strong preference for 
one of the prepositions, this will be ignored.  
No evidence from the head distribution will be used, even though the object 
distribution does not support any of the analyses.  A symmetric insensitivity 
to the object distribution is also possible.  

One method of bypassing this difficulty is to smooth counts away from zero.  
In the experiments reported below 
both maximum likelihood estimation and a simple smoothed version have 
been used and a comparison made.

\subsubsection*{Analysis procedure}

Having estimated the parameters of the model, we can now analyse the test 
set.  The decision procedure computes the probability of a paraphrase 
involving each preposition and chooses the highest.  Since the 
model only takes the two nouns into account, any broader context is ignored.  
If the most preferred paraphrase is highly variable across different contexts, 
then the accuracy of the predictions will be quite low.  
I have not conducted any 
investigation of the optimal error rate for this task.

Using the matrices, $A_{ij}^{\mbox{\sc head}}$ and $A_{ij}^{\mbox{\sc 
obj}}$, the analyser applies equation~\ref{eq:ce_model_analysis} eight 
times, once for each preposition, leading to eight probability scores, and thus 
a ranking of the possible prepositional paraphrases.  Ties are resolved by a 
fixed precedence ordering which follows the observed frequency ranking in 
a pilot set of 57 compounds collected prior to and independently of the main 
test set.  The precedence ordering is (from highest to lowest): \lingform{of}, 
\lingform{for}, \lingform{in}, \lingform{about}, \lingform{with}, 
\lingform{from}, \lingform{on}, \lingform{at}.  

This frequency ranking matches exactly that observed in the main test set, 
however the precedence was chosen before the test set was constructed; no 
information from the test set was used in defining either the 
analysis procedure or the probabilistic model.  In any case, on the test set 
used in this experiment, the conceptual model never produces a tie, so the 
precedence ordering has no effect.  The lexical model is only forced to rely 
on the precedence ordering in 10 out of the 282 test cases (3.5\%); in all of 
these it assigns zero probability to every preposition, which results in 
prediction of an \lingform{of}-paraphrase.

Let's consider an example of the decision procedure.  Suppose the system is 
presented with the compound \lingform{cask wine}.  The word 
\lingform{cask} appears in category \tc{receptacle}{191} and the word 
\lingform{wine} appears in category \tc{food}{298}.  In general, the words 
will be sense ambiguous (they will appear in more than one category), but for 
this example I have chosen monosemous words to simplify the description.

The analyser extracts one row of eight values from each distribution matrix: 
for each of the eight prepositions $p$, it extracts and multiplies $A_{298, 
p}^{\mbox{\sc head}}$ and $A_{191, p}^{\mbox{\sc obj}}$.  Suppose that 
probability estimates are as in table~\ref{tb:ce_method_example}.  The 
analyser multiplies the two probabilities in each row of the table.  Since the 
preposition \lingform{in} has the highest combination (medium head 
probability and high object probability), it is selected as the most likely 
analysis: \lingform{wine in a cask}.  Other probable analyses include 
\lingform{wine from a cask} and \lingform{wine for a cask}.

\begin{table*}
\centering
\begin{tabular}{|c|c|c|} \hline
 \tc{food}{298} as head & Preposition & \tc{receptacle}{191} as modifier 
\\
$A_{298, p}^{\mbox{\sc head}}$ & $p$ & $A_{191, p}^{\mbox{\sc 
obj}}$ \\
\hline
low & of & medium \\
high & for & low \\
medium & in & high \\
zero & about & (low) \\
medium & with & low \\
medium & from & medium \\
low & on & low \\
(low) & at & zero \\
\hline
\end{tabular}
\caption{Relevant parameter values for hypothetical example: 
\lingform{cask wine}}
\label{tb:ce_method_example}
\end{table*}

The appearance of zero probabilities causes the corresponding probability in 
the opposite column to be ignored.  The values given in brackets in the table 
cannot make any difference to the analysis, no matter how large they are 
relative to the other values in the table.

\subsection{Results}
\label{sec:ce_results} 

A program has been implemented to apply the decision procedure given 
above.  It has been run on the test set using both parameterisations ($C_a$ 
and $C_b$), with the results shown in table~\ref{tb:ce_results_mle8}.  
Each column of the table shows the performance on test cases whose
answer is a particular preposition.  The last column gives the
overall accuracy.  The table also shows the results of a simple 
guessing strategy (always choose \lingform{of}) as a baseline.

\begin{table*}
\centering
\begin{tabular}{|l|r|r|r|r|r|r|r|r||r|} \hline
Answer & \multicolumn{1}{c|}{of} & \multicolumn{1}{c|}{for} & 
\multicolumn{1}{c|}{in} & \multicolumn{1}{c|}{about} & 
\multicolumn{1}{c|}{with} & \multicolumn{1}{c|}{from} & 
\multicolumn{1}{c|}{on} & \multicolumn{1}{c||}{at} & 
\multicolumn{1}{c|}{Total} \\
\hline
Number of cases & 94 & 78 & 39 & 22 & 18 & 13 & 12 & 6 & 282 \\
\hline
\hline
$C_a$ accuracy & 19\% & 29\% & 33\% & 64\% & 17\% & 8\% & 
33\% & 33\% & 28\% \\
\hline
$C_b$ accuracy & 52\% & 47\% & 36\% & 14\% & 6\% & 23\% & 50\% 
& 17\% & 40\% \\
\hline
\scare{Guess \lingform{of}} accuracy & 100\% & 0\% & 0\% & 0\% & 0\% & 
0\% & 0\% & 0\% & 33\% \\
\hline
\end{tabular}
\caption{Maximum likelihood estimate results for compound paraphrasing}
\label{tb:ce_results_mle8}
\end{table*}

Apart from the high accuracy on predicting \lingform{about}-paraphrases, 
the conceptual model ($C_a$) does very poorly, worse even than 
guessing.\footnote{The high accuracy on \lingform{about} appears to be 
due to just the right categories appearing in Roget's.  The heads of 11 of the 
14 correct \lingform{about}-compounds fall into the two categories 
\tc{Knowledge}{490} and \tc{Description}{594}.}  While this is 
disappointing, it isn't particularly surprising for the reasons given in 
section~\ref{sec:ce_design}: prepositions have strong collocative 
preferences with nouns.  If we want the conceptual model to work well, the 
associations must be between concepts and roles, not concepts and 
prepositions.

The lexical model does somewhat better.  Though the improvement over 
guessing is not large, it is statistically significant.  Pooling the proportions of 
test cases correctly analysed by the lexical model and by guessing gives 
36.9\%, which can be used to estimate the variance of the difference.  Since 
the sample size is relatively large, the difference in accuracy is 
approximately normally distributed.  Therefore, the improvement of 7.1\% 
corresponds to a $z$-score of 1.75, which is significant at the 5\% level 
(one-tailed test, $p = 0.04$).  We therefore have some evidence that the 
distribution of prepositional phrases is correlated with the distribution of 
compound noun paraphrases.

\subsubsection*{Smoothing}

As shown above, there is a technical difficulty with using maximum 
likelihood estimates; unseen events are prohibited by the model leading to 
information hiding.  
One way to avoid this difficulty is to smooth counts away from zero.  The 
expected likelihood estimator (Church~\etal,~1991b:127) is computed by 
taking the same counts as the maximum likelihood estimator and adding 0.5 
to each count.  To use this method, equation~\ref{eq:ce_method_estimates} 
must be revised.  Parameter estimates under expected likelihood estimation 
are computed by
\begin{eqnarray}
\Pr(c_x | r) & = &
\frac{1}{\eta_r}
(\frac{1}{2}+
\sum_{w \in \theta_1(c_x)}
\frac{\countfn_{\mbox{\sc d}}(w, r)}{\ambig(w)})
\label{eq:ce_results_ele} \\
\eta_r & = &
\frac{|C|}{2} +
\sum_{w \in N}
\frac{\countfn_{\mbox{\sc d}}(w, r)}{\ambig(w)} \nonumber
\end{eqnarray}
where $\mbox{\sc d} = \mbox{\sc obj}$ when $x = 1$ and $\mbox{\sc d} = 
\mbox{\sc head}$ when $x = 2$.

High counts are less affected by this revision than low counts.  However, 
probability estimates corresponding to high counts are reduced because of 
the large increase in the value of the normaliser.  Probability estimates 
derived from low counts increase.  Meanwhile, zero counts 
now result in non-zero estimates for the corresponding events, but these are 
small --- less than half of the maximum likelihood estimate for events seen 
once.  Therefore, this simple technique prevents the information hiding 
exhibited by maximum likelihood estimates.  While this offers a cheap 
escape, a word of warning is mandated: expected likelihood estimation can 
result in very poor probability estimates.  Gale and Church~(1994) show that 
for models of English word bigrams, expected likelihood estimates are 
significantly worse predictors than maximum likelihood estimates.

After revising the parameter estimates to use expected likelihood, the 
analyser was re-run, once again using both parameterisations ($C_a$ and 
$C_b$).  The results are shown in table~\ref{tb:ce_results_ele8}.  
The conceptual model is slightly improved by the expected likelihood 
estimation (though none of the changes in accuracy are statistically 
significant).  
The generalisation allowed by the conceptual groupings means that counts 
are generally not low, so adding 0.5 does not have much effect.  In contrast, 
the lexical model has many low counts, so there is a greater change.  
The new estimates result in significantly degraded performance, 
demonstrating the dangers of applying simplistic smoothing techniques.  
%
\begin{table*}
\centering
\begin{tabular}{|l|r|r|r|r|r|r|r|r||r|} \hline
Answer & \multicolumn{1}{c|}{of} & \multicolumn{1}{c|}{for} & 
\multicolumn{1}{c|}{in} & \multicolumn{1}{c|}{about} & 
\multicolumn{1}{c|}{with} & \multicolumn{1}{c|}{from} & 
\multicolumn{1}{c|}{on} & \multicolumn{1}{c||}{at} & 
\multicolumn{1}{c|}{Total} \\
\hline
Number of cases & 94 & 78 & 39 & 22 & 18 & 13 & 12 & 6 & 282 \\
\hline
\hline
$C_a$ accuracy & 24\% & 28\% & 38\% & 64\% & 17\% & 8\% & 
33\% & 33\% & 30\% \\
\hline
$C_b$ accuracy & 31\% & 44\% & 28\% & 45\% & 6\% & 23\% & 50\% 
& 33\% & 34\% \\
\hline
\end{tabular}
\caption{Expected likelihood estimate results for compound paraphrasing}
\label{tb:ce_results_ele8}
\end{table*}

Interestingly the accuracy of the lexical model is not degraded
on five out of the eight prepositions.  
All of the performance decrease is sustained on just three
of the prepositions.  The three prepositions 
showing reduced accuracy are exactly those with the most training 
data: \lingform{of}, \lingform{for} and \lingform{in} (see 
table~\ref{tb:ce_method_trainsizes} in section~\ref{sec:ce_method}).
A possible interpretation of this is that smoothing yields an improvement in 
accuracy on cases suffering data sparseness, but it also forces the analyser to 
consider rare cases more often than it should.  Recall from 
section~\ref{sec:ce_design} that one of the intuitive motivations for using 
distributions of prepositional phrases to estimate those of compound noun 
paraphrases, was that impossible paraphrases would be ruled out.
Smoothing by 
expected likelihood estimates undermines the model's ability to rule out 
these rare cases, leading to lower performance on the more frequent cases.

\subsubsection*{Avoiding data sparse distributions}

This is exacerbated by the large number of data sparse alternatives.  Only 
three of the prepositions do not suffer from severe data sparseness.  Each of 
the other five is a potential distractor, and under noisy conditions, the 
chances are that one of them will be assigned high enough probability to be 
selected incorrectly.  Suppose, for example, that the analyser is faced with a 
compound having a \lingform{for}-paraphrase.  The relative abundance of 
data giving information about \lingform{of}- and \lingform{in}-paraphrases 
will prevent these from being incorrectly chosen above a 
\lingform{for}-paraphrase.  However, the distributions estimated for the 
other five prepositions are each subject to large purely statistical 
fluctuations, with a low signal-to-noise ratio.  While on average, they all 
have a low probability, chances are that one of them will be estimated to 
have high probability for the given compound because of noise.  Because of 
this, the analyser is frequently lead to a mistaken conclusion.  

Therefore, a better design for the decision procedure is to ignore information 
from the data sparse distributions.  The analyser computes only three 
probabilities, those of paraphrases involving \lingform{of}, \lingform{for} 
and \lingform{in}, and takes the maximum.  Since this design only ever 
selects one of these three prepositions, 
it cannot correctly analyse other types 
of compounds.  Because these three only make up three-quarters
of the test set, this restriction places a ceiling 
on the performance of 75\% accuracy.  However, 
so far the results have been much lower than this anyway.

Table~\ref{tb:ce_results_mle3} shows the performance of the analyser when 
restricted in this way.  Maximum likelihood estimates have been used for the 
conceptual model (the results for expected likelihood estimates do not differ 
significantly), while both estimation methods are shown for the lexical 
model.  
All three results show sizeable improvements over the corresponding models 
under the unrestricted design.  This is further evidence that consideration of 
the amount of available training data is of critical importance to statistical 
language learning systems.  It is encouraging to note that the expected 
likelihood estimation now yields a slight improvement to the lexical model.  
This is in stark contrast to the performance degradation observed in the 
unrestricted design.  The accuracy of 47\% is the best result obtained by the 
system under any configuration.
%
\begin{table*}
\centering
\begin{tabular}{|l|r|r|r|r|r|r|r|r||r|} \hline
Answer & \multicolumn{1}{c|}{of} & \multicolumn{1}{c|}{for} & 
\multicolumn{1}{c|}{in} & \multicolumn{1}{c|}{about} & 
\multicolumn{1}{c|}{with} & \multicolumn{1}{c|}{from} & 
\multicolumn{1}{c|}{on} & \multicolumn{1}{c||}{at} & 
\multicolumn{1}{c|}{Total} \\
\hline
Number of cases & 94 & 78 & 39 & 22 & 18 & 13 & 12 & 6 & 282 \\
\hline
\hline
$C_a$ accuracy (\acronym{mle}) & 43\% & 50\% & 62\% & 0\% & 0\% & 0\% & 0\% 
& 0\% & 37\% \\
\hline
$C_b$ accuracy (\acronym{mle}) & 65\% & 56\% & 67\% & 0\% & 0\% & 
0\% & 0\% & 0\% & 46\% \\
\hline
$C_b$ accuracy (\acronym{ele}) & 55\% & 69\% & 67\% & 0\% & 0\% & 
0\% & 0\% & 0\% & 47\% \\
\hline
\end{tabular}
\caption{Paraphrasing results when restricted to three most common 
answers}
\label{tb:ce_results_mle3}
\end{table*}

One final result deserves mention.  Since the prepositions \lingform{of} and 
\lingform{for} comprise more than 60\% of the test set, a further 
improvement might be made by restricting the design to just these two.  The 
results of this experiment are shown in table~\ref{tb:ce_method_mle2}.  
The lexical model makes no further gain from the restriction, although 
expected likelihood estimation gives slightly better results, confirming the 
contrast observed in the previous restricted design.  The conceptual model is 
improved to an accuracy of 40\%.  Using the pooled $t$-test as above gives 
a $z$-score of 1.66, showing that the improvement over guessing is 
statistically significant at the 5\% level (one-tailed test, p=0.049).
%
\begin{table*}
\centering
\begin{tabular}{|l|r|r|r|r|r|r|r|r||r|} \hline
Answer & \multicolumn{1}{c|}{of} & \multicolumn{1}{c|}{for} & 
\multicolumn{1}{c|}{in} & \multicolumn{1}{c|}{about} & 
\multicolumn{1}{c|}{with} & \multicolumn{1}{c|}{from} & 
\multicolumn{1}{c|}{on} & \multicolumn{1}{c||}{at} & 
\multicolumn{1}{c|}{Total} \\
\hline
Number of cases & 94 & 78 & 39 & 22 & 18 & 13 & 12 & 6 & 282 \\
\hline
\hline
$C_a$ accuracy (\acronym{mle}) & 66\% & 65\% & 0\% & 0\% & 0\% & 0\% & 0\% 
& 0\% & 40\% \\
\hline
$C_b$ accuracy (\acronym{mle}) & 84\% & 63\% & 0\% & 0\% & 0\% & 
0\% & 0\% & 0\% & 45\% \\
\hline
$C_b$ accuracy (\acronym{ele}) & 71\% & 79\% & 0\% & 0\% & 0\% & 
0\% & 0\% & 0\% & 46\% \\
\hline
\end{tabular}
\caption{Paraphrasing results when restricted to two most common answers}
\label{tb:ce_method_mle2}
\end{table*}

\subsection{Discussion}
\label{sec:ce_discussion} 

The overall best accuracy is 47\%; however, this is difficult to 
evaluate without more information.  
No other work in statistical language learning has attempted this task 
and there are few knowledge-based systems that have been quantitatively
evaluated.  Leonard's~(1984) accuracy of 76\% was measured
on the development set and was only achieved through labour intensive
coding of appropriate lexical information.  McDonald's~(1982) estimate
that 60\% of his test set could be handled by his program also
requires extensive knowledge to be provided.  

A more comparable result is that of Vanderwende~(1994).  She
reports that on a random test sample her program has an accuracy
of 52\%.  While it is based on knowledge extracted from 
Longman's dictionary, the development of the extraction patterns, 
and of the analysis heuristics and weights, 
again involves substantial effort (though admittedly far less 
than that required for the other two).  Also it assumes that an
appropriate dictionary is available (Longman's is not freely
available; research licences are considerably more costly than
those for any corpus).  

In contrast, my program is extremely
simple to implement, uses freely available resources and adapts
to the text type used for training.  In this light, 47\% accuracy
is a good result.  Since it is significantly better than guessing,
we can, at the very least, conclude that measurements of 
prepositional phrase distributions assist 
in predicting the paraphrases of noun compounds.

One significant problem with evaluation is that we don't know the optimal
accuracy.  For example, if compound paraphrases 
are highly dependent on context, we could already be close to the optimal 
accuracy.  Without some independent means of measuring the optimal 
accuracy, such as experiments with human subjects, it is impossible to draw 
conclusions about the quality of this performance.  The test set is too small 
to offer much help in estimating the optimal accuracy, since
there are only eleven duplicated test cases.
Of these, two have different preferred paraphrases on each occurrence.
This gives some evidence that noun compound paraphrases are
context dependent, but to what degree is impossible to say.

Even if we knew the optimal accuracy, there is still the problem of data 
requirements.  How much of the non-optimal error rate is due to faulty 
assumptions in the model, and how much to inadequate training data 
volumes?  Unless the performance closely approaches the optimal accuracy, 
and lacking an applicable theory of data requirements, 
the shortfall could be due to 
either data sparseness or poor modelling; the scientific implications of the 
experiment are uncertain.  

Still, we can make some qualitative conclusions about the different strategies 
that have been applied.  First, conceptual association did not work well 
because prepositions have purely lexical collocation properties.  The test 
compound \lingform{university education} has most preferred paraphrase in 
context involving \lingform{in}.  The conceptual model predicts a 
paraphrase involving \lingform{at}.  Likewise, \lingform{street scenes} has 
most preferred paraphrase in context involving \lingform{on}, but the 
conceptual model chooses \lingform{in}.  These examples suggest that 
concepts should be associated with roles rather than particular prepositions.  

Second, the assumption that the head and modifier lend independent 
evidence (the head-modifier autonomy assumption) leads the analyser into 
error in some test cases.  
For example, the lexical model assigns 
\lingform{computer catalog} an \lingform{of}-paraphrase, presumably 
because catalogs are typically composed of items and systems are often 
composed of computers.  However, in this case the paraphrase 
should involve \lingform{on}.  It is only through the interaction between the 
head and modifier that this can be determined.  
A similar circumstance leads 
to error on \lingform{welfare agencies}, which the analyser assigns an 
\lingform{on}-paraphrase to.  The fact that agencies are not usually the 
recipients of welfare is dependent on the nature of both agencies and 
welfare, and thus both the head and modifier.  These two examples are clear 
cases where the error is due to violated assumptions, but typically the reason 
for failure is not so obvious.  
Therefore, it is difficult to assess by inspection 
the degree of error resulting from modelling simplifications.

Finally, we've seen that a restricted model, which 
only considers the more frequent answers, is beneficial 
when training data is both sparse and noisy.
In contrast, a simple 
smoothing technique results in dramatically worse performance.  The moral 
is that smoothing is just one instrument in overcoming data sparseness; it 
must be coupled with an understanding of the effects of data sparseness, 
which is the kind of knowledge that should be provided by a theory of data 
requirements.  If there is any strong conclusion to be drawn from the 
experiments on compound paraphrasing, it is that the ability to reason about 
data requirements is vital to statistical language learning research.


\cleardoublepage

\chapter{Conclusion}
\label{ch:conclusion}


\section{Summary}

%
%
In \sll\ research, we are engaged in a search through a space of 
possible designs.  \sll\ provides an architecture for building 
corpus-based language learners using probabilistic models; but
what the architecture cannot tell us is which probabilistic models 
are most appropriate.  This we must discover by a process of exploration.

The search through design space is constrained by the need both for 
linguistic fidelity and for practical construction.  A good model must 
make accurate predictions about language and it must be simple to train 
and apply.  The search is therefore a complex one.  In the few years since 
\sll\ has become popular, we have only touched the surface of the design 
space.  This thesis is about exploring that design space further.  

\subsection{Theoretical Contributions}

In pursuit of that goal, the thesis makes two key theoretical 
contributions.  First, it identifies a new class of designs by providing an 
architectural theory of statistical natural language processing, the 
meaning distributions theory.  Second, it works toward the development 
of a vital tool that will assist in future navigation of the design space, a 
mathematical theory of data requirements.  I will discuss each of these in 
turn.

\subsubsection*{Meaning distributions}

In chapter~\ref{ch:md}, I developed the meaning distributions theory for 
statistical language learning.  According to this view, statistical models 
of language involve probability distributions over semantic 
representations.  A probabilistic conceptual model, whose parameters are 
estimated from a corpus, captures semantic expectations, which are then 
used to guide natural language processing tasks.  By emphasising 
semantic expectations, the theory is effectively 
a revised form of semantically-driven \nlp; however, it specifically 
avoids several failings of early work in that area.

The meaning distributions theory holds that a probabilistic conceptual 
model can be used in conjunction 
with an independent lexico-syntactic component to perform natural 
language analysis tasks.  
The lexico-grammar relates the semantic representation to corresponding 
syntactic and textual forms.  Rather than assigning probabilities to the 
syntactic representations independently, such probabilities are 
determined from the semantic distributions 
by viewing lexico-grammar as constraints.  

This theory is only useful in as much as it allows the construction of 
working natural language processors.  Evaluation of the theory must be 
empirical, determined by the performance of systems designed using it.  
While it is derived from reflections on human language processing, it is 
in no way intended to represent human behaviour.  Rather it represents 
new territory in the design space of \sll\ models.  Whether that territory 
turns out to be fertile is an empirical question.

One useful implication of the theory concerns the selection of corpus 
materials for \sll\ training.  The emphasis placed on context by the 
meaning distributions theory highlights the importance of register to 
probabilistic language models.  Since it is crucial that such models take 
account of register variations, and since we do not currently have 
sufficiently large corpora to condition on register, I have advocated the 
use of large register-pure corpora as the best short term compromise.

\subsubsection*{Data requirements}

While the meaning distributions theory opens up new territory
in design space, we still have only a limited understanding 
of one of the most fundamental limits to exploration 
of that space: training data requirements.
Chapter~\ref{ch:dr} has laid out a blueprint for developing a predictive 
theory of data requirements, a theory that will be an invaluable tool for 
further exploration.

It is obvious that the success of \sll\ systems depends on the volume of 
data used to train them; yet the exact relationship is very poorly 
understood.  Predictions of the amount of training data required for a 
given \sll\ model are not directly available from the statistics and 
machine learning disciplines, and yet such predictions are necessary to 
both scientific research and commercial applications.  

In this thesis, I have reviewed a range of statistical learning systems with 
the goal of identifying data points that the proposed theory might 
explain.  One outcome of this study is the observation that less than 1 
training instance per parameter is sometimes sufficient.  This appears 
to be due to the highly non-uniform distributions commonly found in 
language, suggesting that any theory of data requirements will have to 
account for this phenomenon.

I have also built a framework for reasoning about data requirements in 
\sll\ systems and given several results regarding the 
expected accuracy of a mode-based learner after a given volume of 
training data.  Further, under certain assumptions, I have derived a 
computable expression for the expected accuracy of such a learner in 
terms of training data volume.  This can be used to predict data 
requirements under limited conditions.  
Perhaps the most limiting assumption required by this result is that the 
bin distribution be uniform.  Therefore, I have also reported a series of 
simulations that both validate the result and 
explore the effects of non-uniform bin distributions.  
As predicted non-uniformity leads to much more rapid learning.  

These results lay a foundation for the development of a general 
predictive theory of data requirements, without which exploration of the 
design space will continue to involve substantial guesswork.

\subsection{Experiments on Noun Compounds}

To illustrate these two theoretical contributions, this thesis has addressed 
the problem of analysing noun compounds.  Noun compounds form a 
promising medium for exploring the design of statistical language 
learners because they are both common and highly ambiguous, thus 
posing a significant challenge to traditional \nlp\ techniques.  
Furthermore, the importance of semantic expectations in the 
interpretation of noun compounds highlights a weakness of existing \sll\ 
models, that being the lack of attention to semantics.

The fact that existing noun compound processing systems are 
primarily knowledge-driven, and have generally failed due to the 
enormous development effort required, makes noun compounds an ideal 
subject for the automatic knowledge acquisition afforded by the \sll\ 
approach.

In chapter~\ref{ch:experimental}, I have described work on two 
computational tasks that have traditionally been attempted 
using knowledge-based systems.  The first is grammatical analysis of 
syntactically ambiguous noun compounds, and the second semantic 
analysis of two word noun compounds.

\subsubsection*{Parsing}

The work on parsing noun compounds began with the development of a 
novel probabilistic model of noun compound syntax.  Following the 
meaning distributions theory, this model uses conceptual association to 
evaluate the dependency relations within a noun compound.  All 
previous statistical algorithms were based on adjacency.  The new model 
correctly predicts that 
two-thirds of all noun compounds 
should be left-branching.  This prediction requires the meaning 
distributions theory and is incompatible with adjacency algorithms.

The model has been implemented and evaluated on three word English 
noun compounds from an encyclopedia corpus.  The implementation 
uses unannotated training data and freely available lexical resources, 
runs on a desktop computer and can rapidly analyse any noun compound, 
not just those in the test set.

Experiments with the system show that the prediction accuracy reaches 
around 81\%.  This closely approaches the average performance rate of 
around 82\% achieved by human judges attempting the same task.
The experiments performed with human judges are also of 
independent value, since they provide quantitative evidence that noun 
compounds have significant context dependence.  This context 
sensitivity is sufficient to cause at least 14\% of compounds to be 
incorrectly parsed when context is not available.

Furthermore, comparison under identical experimental conditions shows 
that the new model is significantly better than one based on association 
between adjacent words, such as all those proposed previously are.  
In addition to supporting the use of this model for noun compound parsing, 
this result provides some evidence that designs based on the meaning 
distributions theory are better than existing ones.  The new territory of 
design space has some empirical commendation.

\subsubsection*{Semantic analysis}

This thesis has also presented a novel model for paraphrasing 
non-verbal-nexus noun compounds, the first probabilistic model of noun 
compound semantics.  As for the syntactic model, it is based on the 
meaning distributions theory.  It assigns probabilities to different 
paraphrases by combining information from the head with information 
from the modifier independently.

This model has been implemented and evaluated on two word English 
noun compounds.  Once again, the system uses only freely available 
lexical resources and unannotated training data.  The latter is made 
possible by using distributions of prepositional phrases to train a 
probabilistic conceptual model of noun compounds, a strategy suggested 
by the meaning distributions theory.

Experiments show that the accuracy of the trained model is up to 47\% 
and is significantly better than the baseline, demonstrating that observing 
the distributions of prepositional phrases can assist 
in paraphrasing non-verbal-nexus noun compounds.  No estimate 
of the optimal context-free accuracy is available and comparable 
performance tests of knowledge-based systems have not been conducted.  
Therefore, the result is difficult to evaluate.  The task is 
similar to that addressed by Vanderwende's~(1994) dictionary-based 
system which achieved 52\% accuracy.  
Advantages of the statistical algorithm reported 
here include that, by comparison, it is trivial to implement and that it 
adapts to the training text.

Data sparseness is a significant problem in the paraphrasing experiments.  
Not only is the volume of training data low, but there is significant noise 
as well.  Even if the accuracy of the system was known to fall short of the 
optimal accuracy, it would be impossible to conclude whether this was 
due to insufficient training data or a poor model.  Such questions require 
a predictive theory of data requirements.  Of particular interest is the 
observation that applying a simple smoothing technique 
drastically reduced the performance.  In contrast, a restricted model that is 
sensitive to training data volumes, gave the best performance.  These 
results show that a general, practical theory of data requirements is 
fundamentally important to the search for good \sll\ designs.  Without 
such a tool, exploration of the design space is necessarily haphazard.  

\subsection{What has been Achieved?}

In summary, there are three key achievements.  
\begin{enumerate}
\item The meaning distributions theory has opened up a new and 
promising area of the \sll\ design space, by proposing a probabilistic
representation of semantic expectations.
The success of the noun compound parsing model provides empirical
evidence that the new territory is a promising one.

\item The work on data requirements has laid the 
foundations for development of a vital tool to assist future explorations
of the design space.  The necessity of a theory of data requirements
is highlighted in the experiments on noun compound paraphrasing.

\item At least one \nlp\ problem has been solved: statistical 
parsing of noun compounds.  The accuracy of the model developed
is significantly better than previously proposed algorithms
and closely approaches that achieved by humans attempting the same task.

\end{enumerate}


\section{Looking Ahead}

This is the section usually headed with a quotation about 
the unbounded nature of science that typically implies one step has been 
made and the next should follow from it.  But I think computational 
linguistics as a science is relatively immature; we are still at the stage of 
exploring what can be done, especially in \sll, and we should expect just as 
many backward steps as forward ones.  For this reason, I will not 
optimistically present a bright way forward to the future.  
Instead, I will give 
a collection of suggestions for things to try.  

\subsection{Meaning Distributions}

\begin{description}
\item[Cross-register performance:] To substantiate the position taken in 
section~\ref{sec:md_register}, a study could be made of the performance of 
a Markov model tagger trained on one register and applied to another.  This 
might also lead to a measurement of the loss of accuracy associated with 
using cross-register training data.
\item[Probabilistic dependency grammar:] Probabilistic context free 
grammars have been investigated in great detail.  The experimental work in 
this thesis suggests that models based on probabilities of dependency 
relations, rather than of constituents, 
do better for parsing noun compounds.  Perhaps then a 
grammar-wide probabilistic model of dependency relations would prove 
more successful than the constituent based models tried so far.  As 
mentioned in section~\ref{sec:sn_grammars}, some proposals already exist, 
but these need to be implemented and tested.
\item[Meaning distributions:] The meaning distributions theory, recording 
probabilities of semantic representations and regarding grammar as a form 
of constraint, can be applied to many \sll\ tasks.  
As I have emphasised before, 
the value of the theory lies in the performance of systems designed 
according to the theory's dictates.  
Various \nlp\ tasks could be approached in this 
way, including anaphora resolution, prepositional phrase attachment and 
relative clause attachment.
\end{description}

\subsection{Data Requirements}

\begin{description}
\item[Statistical inheritance:] The probabilistic models used in this thesis 
employ a fixed set of thesaurus categories.  To reduce data sparseness,
words within a category are assumed to behave similarly.  However,
the cost of this assumption is
paid for all words, not just those affected by data sparseness.  A better 
alternative would be to vary the size of the thesaurus categories dynamically, 
responding to the degree of data sparseness appropriately.  
One approach would be to try successively larger category sizes.  First
a decision would be made using only data about the particular
word, and then the statistical confidence of the decision estimated.
If the confidence estimate failed to support the decision with 
high probability, then the process would be repeated, each time
using data about larger categories until the statistical test proved
significant.

A natural way to provide the various sized categories would be to 
employ a taxonomy like WordNet.  Classes higher up in the 
taxonomy will contain more words, so that association data for these 
classes will be much more {\em precisely} estimated.  However, parameter 
estimates for higher classes will less {\em accurately} reflect the
behaviour of individual words within the class, 
because the higher classes inevitably contain more 
variation within them.  The fewer words in a class, the more 
accurately the parameter estimates for the class will reflect the 
properties of each word in it.  The strategy then is to select 
the smallest class that contains the desired word for which the 
appropriate parameter can be estimated confidently.  A form of statistical 
inheritance has already been used for acquiring selectional restrictions by 
Ribas~(1995).
\item[Quantifying non-uniformity:]  One of the most severe limitations of the 
data requirements results in section~\ref{sec:dr_global} is the assumption 
that bins are equi-probable.  The simulations in 
section~\ref{sec:dr_simulations} are only a small step in exploring the 
effects of non-uniformity.  Ideally further simulations could lead to a 
quantification of non-uniformity.  The theory begs to be extended in many 
other ways too (consider weakening any of the assumptions).  For example, 
one direction worth pursuing would be predicting the data requirements for 
unsupervised algorithms such as expectation maximisation.
\end{description}

\subsection{Noun Compounds}

\begin{description}
\item[Context dependence of compounds:] The human judgement 
experiment in section~\ref{sec:cy_human} suggests that noun compounds are 
fairly context sensitive.  The proposals for dealing with this (see 
section~\ref{sec:cn_context}, especially Meyer's thesis and 
McDonald's cognate heuristic: Meyer,~1993; McDonald,~1982) 
involve matching the compound with meanings from the recent
context.  One way to do this within the \sll\ approach 
would be to incorporate the recent cotext as additional training data 
and give it more weight.
\item[Adjectival modifiers:] Adjectives exhibit similar syntactic ambiguities 
to those of noun compound modifiers.  Consider 
example~\ref{eg:conclusion_adj}, where \lingform{small} could specify the 
size of either the transaction or the fee.
\begin{examples}
\item small transaction fee   \label{eg:conclusion_adj}
\end{examples}
A model almost identical to that proposed in section~\ref{sec:cy_model} 
could be used for this task.  This would entail a conceptual representation of 
predicating adjectives, which might be based on the adjectival groupings 
found in WordNet, or on those extracted by Hatzivassiloglou~(1994).
\item[Interpreting nominalisations:] The experiments on noun compound 
paraphrasing concerned non-verbal nexus compounds.  One possible 
approach to nominalisations that is consistent with the meaning distributions 
theory is to use subject-verb-object statistics of the kind collected by 
Church~\etal~(1991a).  The model would be based on the joint distribution 
of predicates and their arguments, which would lead to preferences for the 
different semantic categories of nominalisations (see 
section~\ref{sec:cn_meaning}).
\end{description}


\cleardoublepage


\appendix
\cleardoublepage

\chapter{Half-binomial Result}
\label{appendix:halfbinomial}


In section~\ref{sec:dr_global}, the 
mathematics hinges on an expression for the
part of a binomial distribution that lies 
above the half-way point of the distribution.
This appendix contains some notes leading to a mathematical
result which I have been referring to as the 
{\em half-binomial result}.  This result provides 
a simpler expression for this probability.


We are interested in that part of a binomial distribution 
which lies above the half-way point of the distribution.
Let $p$ be the chance of success on each trial and $n$ be
the number of trials.  Since the distribution is discrete,
when $n$ is even some of the distribution lies exactly
on the half-way point (that is, \halfn).  In this case, 
we would like to weight the contribution from this part
by $\frac{1}{2}$.  

Thus, the probability we are interested in is:

\begin{equation}
\halfbin(p,n) = {\sum_{i=\halfnincup}^{n} {n \choose i} p^i (1-p)^{n-i}}
	+ 
	\frac{1}{2} \even{n} {n \choose \halfn} p^{\halfn} (1-p)^{\halfn}
\end{equation}

Here, and below, $\even{n}$ is $1$ when $n$ is even and 
$0$ otherwise.  $\odd{n}$ is the reverse.
By expansion, we can arrive at a recurrence as follows:

First consider the probability for $n-1$.

\begin{equation}
\halfbin(p,n-1) = {\sum_{i=\halfnup}^{n-1} {{n-1} \choose i} p^i (1-p)^{n-i-1}}
	+ \frac{1}{2} 
	\odd{n} {{n-1} \choose \halfndec} p^{\halfndec} (1-p)^{\halfndec}
\end{equation}

We want to find an expression for $\halfbin(p, n)$ in terms of
$\halfbin(p, n-1)$.  Proceed by splitting the first choice function 
in $\halfbin(p, n)$ into two halves using Pascal's relation.

\begin{eqnarray*}
\halfbin(p,n) & =  & {\sum_{i=\halfnincup}^{n} 
		\left( {{n-1} \choose {i-1}} + {{n-1} \choose {i}} \right) 
		p^i (1-p)^{n-i}}
	+
	\frac{1}{2} \even{n} {n \choose \halfn} p^{\halfn} (1-p)^{\halfn} \\
 & = & {\sum_{i=\halfnincup}^{n} {{n-1} \choose {i-1}} p^i (1-p)^{n-i}} \\
 & &	+ {\sum_{i=\halfnincup}^{n} {{n-1} \choose {i}} p^i (1-p)^{n-i}} \\
 & &	+
	\frac{1}{2} \even{n} {n \choose \halfn} p^{\halfn} (1-p)^{\halfn} \\
 & = & p {\sum_{j=\halfndecup}^{n-1} 
		{{n-1} \choose {j}} p^j (1-p)^{n-j-1}} \\
 & &	+ (1-p) {\sum_{i=\halfnincup}^{n} 
		{{n-1} \choose {i}} p^i (1-p)^{n-i-1}} \\
 & &	+
	\frac{1}{2} \even{n} {n \choose \halfn} p^{\halfn} (1-p)^{\halfn} \\
 & = & p \left( \odd{n} 
	{{n-1} \choose \halfndec} p^{\halfndec} (1-p)^{n-1-\halfndec} + 
	\sum_{j=\halfnup}^{n-1} {{n-1} \choose {j}} p^j (1-p)^{n-j-1} \right) \\
 & &	+ (1-p) \left( {\sum_{i=\halfnup}^{n-1} 
		{{n-1} \choose {i}} p^i (1-p)^{n-i-1}}  
		- \even{n} {{n-1} \choose \halfn} 
		p^{\halfn} (1-p)^{\halfn-1} \right) \\
 & &	+
	\frac{1}{2} \even{n} {n \choose \halfn} p^{\halfn} (1-p)^{\halfn} \\
 & = & p \left( \halfbin(p, n-1) + \odd{n} \frac{1}{2}
	{{n-1} \choose \halfndec} p^{\halfndec} (1-p)^{\halfndec} \right) \\
 & &	+ (1-p) \left( \begin{array}{rl} \halfbin(p, n-1)  
			& - \odd{n} \frac{1}{2}
		{{n-1} \choose \halfndec} p^{\halfndec} (1-p)^{\halfndec} \\
			& - \even{n} 
		{{n-1} \choose \halfn} p^{\halfn} (1-p)^{\halfn-1} 
		\end{array} \right) \\
 & &	+
	\frac{1}{2} \even{n} {n \choose \halfn} p^{\halfn} (1-p)^{\halfn} \\
 & = & \halfbin(p, n-1) \\
 & & 	+ \odd{n} \left( 
	\frac{p}{2} {{n-1} \choose \halfndec} p^{\halfndec} (1-p)^{\halfndec} 
	- \frac{1-p}{2} {{n-1} \choose \halfndec} 
	p^{\halfndec} (1-p)^{\halfndec} \right) \\
 & & 	+ \even{n} \left( 
	\frac{1}{2} {n \choose \halfn} p^{\halfn} (1-p)^{\halfn} 
	- (1-p) {{n-1} \choose \halfn} p^{\halfn} (1-p)^{\halfn-1} \right) 
	\label{eq:recurrence} 
\end{eqnarray*}

Since $\frac{1}{2}{n \choose \halfn} = {{n-1} \choose \halfn}$, the term
for even $n$ is zero.  Therefore, we are left with the simple recurrence:

\begin{equation}
\halfbin(p, n) = \halfbin(p, n-1) + \odd{n} (p-\frac{1}{2})
	{{n-1} \choose \halfndec} p^{\halfndec} (1-p)^{\halfndec} 
\end{equation}

Since each increase in $n$ only adds a new term, we can re-express
this as a sum, which is the {\em half-binomial result}.  It holds for
all $n \ge 0$.

\begin{equation}
\halfbin(p, n) = \frac{1}{2} + \sum_{i=0}^{\halfnup-1} (p-\frac{1}{2})
		{2i \choose i} p^{i} (1-p)^{i}
\end{equation}

Notice that this sum (unlike the expression we began with) 
does not contain terms that are dependent on $n$.  Only the
index of summation depends on $n$.  The corollary 
below follows from the fact that the term representing
even $n$ in the recurrence above is zero.

\begin{corollary}
For all even $n \ge 2$, $\halfbin(p, n) = \halfbin(p, n-1)$.
\end{corollary} 

Observing that terms of the sum are always positive when $p > 0.5$
leads to the following corollary.

\begin{corollary}
For all $n \ge 1$ and $p > \frac{1}{2}$, 
$\halfbin(p, n) \ge \halfbin(p, n-1)$.
\end{corollary}

\cleardoublepage

\chapter{Instructions to Human Judges}
\label{appendix:humaninstructions}

\section*{Instructions}

Welcome to my compound noun parsing experiment.                                                                                                             
The purpose is to establish how well people can parse compound nouns 
without using context.  For this, I have extracted 215 three-word compounds, 
which I will present to you for analysis.  

Three word compounds have two possible syntactic readings.  
In each example, you will be provided with both graphical and 
bracketed alternatives.  Here are two examples:

\begin{description}
\item[Left-Branching:] ((romance novel) cover)

Since we are not referring here to a \scare{novel cover} 
for \scare{romance}, but rather the \scare{cover} of a \scare{romance novel}, 
this example is left-branching.  
[accompanied by a box diagram showing romance novel linked to cover]

\item[Right-Branching:] (toy (coffee grinder))

Since we are not referring here to a \scare{grinder} for \scare{toy coffee}, 
but rather a \scare{coffee grinder} that is a \scare{toy}, 
this example is right-branching.
[accompanied by a box diagram showing coffee grinder linked to toy]

\end{description}

Your task is to choose which of these alternatives is 
most probably the correct one in the (unknown) context 
from which the compound was extracted.  
Since you do not know the context, you must make your best guess.  
Try to imagine possible meanings and contexts for each compound 
and choose the analysis you think is most likely.  
You will only be allowed to choose exactly one.

Choose your answer for each compound carefully.  
You will not be able to go back once you have clicked 
on the OK button.  Feel free to mark the check boxes 
to the left of the OK button if they are appropriate.  
You can also add any comments you like about each example 
in the comment box.  Click on the HELP button to review 
these instructions at any time.

Some points to note:                                                                                                                                          
\begin{itemize}
\item There aren't necessarily equal numbers of 
left- and right-branching compounds.
\item The compounds are presented in random order.
\item The compounds come from encyclopedia entries.
\end{itemize}

The entire task should take approximately 50--60 minutes.  
Please take your time.  Think for as long as you wish: 
we want you to answer as well as you can.  

\section*{Answer Dialogue}

After the instructions above had been presented to the
subject, a dialogue was displayed for each test noun compound.
The dialogue had the following elements:

\begin{itemize}
\item the test compound shown in large font.
\item two alternative analyses shown graphically.
\item two alternative analyses shown by bracketing.
\item radio buttons for left- and right-branching.
\item display with fraction of test completed.
\item HELP and OK buttons.
\item a free text comment box.
\item three check boxes, labelled:
	\begin{itemize}
	\item Contains a word you've never seen.
	\item Neither answer is appropriate.
	\item Both would fit equally well.
	\end{itemize}
\end{itemize}

\cleardoublepage

\chapter{Test Set for Noun Compound Parsing}
\label{appendix:cytest}

The following list of noun compounds forms the test set
used for the experiments on noun compound parsing.
It was produced from Grolier's using the method described
in section~\ref{sec:cy_method}.  Each was then assigned
one of the following analyses: left-branching~(L),
right-branching~(R), indeterminate~(I) and extraction
error~(E).  This answer is shown in the last 
column.

For comparison, the second column gives the prediction made
by the best statistical model when using tagged training
data (see section~\ref{sec:cy_comparisons} and in particular 
figure~\ref{fig:cy_tagged_tag_accuracy}).  This
model achieves 80.7\% accuracy.


\begin{center}

\begin{tabular}{|l|c|c|} \hline
Noun triple	& Best model & Correct Answer \\
\hline
minority business development	&	L	&	L	\\
satellite data systems	&	R	&	R	\\
disaster relief assistance	&	L	&	L	\\
county extension agents	&	R	&	R	\\
world food production	&	R	&	R	\\
granary storage baskets	&	L	&	R	\\
customs enforcement vehicles	&	L	&	L	\\
airport security improvements	&	L	&	L	\\
mountain summit areas	&	L	&	L	\\
law enforcement agencies	&	L	&	L	\\
college commencement poem	&	R	&	R	\\
health education institutions	&	L	&	L	\\
country music theme	&	L	&	L	\\
sea transportation hub	&	L	&	L	\\
principle organ systems	&	R	&	E	\\
quality movie animation	&	R	&	E	\\
army ant behavior	&	L	&	L	\\
fossil ant species	&	L	&	I	\\
missile defense systems	&	R	&	L	\\
world petroleum production	&	R	&	R	\\
representative percussion instruments	&	L	&	E	\\
Arab independence movements	&	L	&	L	\\
speech recognition system	&	L	&	L	\\
\hline
\end{tabular}

\begin{tabular}{|l|c|c|} \hline
Noun triple	& Best model & Correct Answer \\
\hline
production passenger vehicles	&	R	&	R	\\
revenue ton miles	&	L	&	R	\\
combustion chemistry technology	&	L	&	L	\\
science fiction novels	&	L	&	L	\\
missile guidance system	&	R	&	L	\\
sea bass species	&	L	&	L	\\
radiation energy conversion	&	L	&	L	\\
tobacco mosaic disease	&	R	&	I	\\
science fiction novels	&	L	&	L	\\
energy distribution properties	&	R	&	L	\\
college football player	&	L	&	I	\\
science fiction writer	&	L	&	L	\\
science fiction themes	&	L	&	L	\\
science fiction writer	&	L	&	L	\\
breeder technology development	&	L	&	L	\\
landmark majority opinions	&	R	&	R	\\
television news personality	&	R	&	I	\\
community college system	&	L	&	L	\\
town council members	&	L	&	L	\\
war crimes prosecutor	&	L	&	L	\\
health insurance laws	&	L	&	L	\\
science fiction satire	&	L	&	L	\\
death penalty statutes	&	L	&	L	\\
Calvinist peasant family	&	R	&	R	\\
exhibition ballroom dancers	&	L	&	R	\\
music hall performer	&	R	&	L	\\
volume alkali production	&	R	&	R	\\
decades child specialists	&	R	&	E	\\
child development specialists	&	L	&	L	\\
fact problem behavior	&	R	&	E	\\
skyscraper office buildings	&	R	&	R	\\
sea warfare doctrine	&	L	&	L	\\
reservoir evaporation losses	&	L	&	R	\\
communications satellite system	&	L	&	L	\\
data storage device	&	L	&	L	\\
error correction data	&	L	&	L	\\
automobile ignition systems	&	R	&	R	\\
computer ad data	&	R	&	E	\\
performance improvement method	&	L	&	L	\\
computer music studio	&	L	&	L	\\
privacy protection agency	&	R	&	L	\\
law enforcement interception	&	L	&	L	\\
computer education enthusiasts	&	L	&	L	\\
citizen conservation organizations	&	R	&	R	\\
development assistance efforts	&	L	&	L	\\
energy conservation law	&	L	&	L	\\
\hline
\end{tabular}

\begin{tabular}{|l|c|c|} \hline
Noun triple	& Best model & Correct Answer \\
\hline
citizens twenty-one years	&	L	&	E	\\
engine lubrication system	&	R	&	I	\\
Monday night football	&	L	&	L	\\
law enforcement agents	&	L	&	L	\\
law enforcement agencies	&	L	&	L	\\
law enforcement officials	&	L	&	L	\\
intelligence reconnaissance sources	&	R	&	R	\\
coalition war cabinet	&	R	&	R	\\
cell plasma membrane	&	R	&	I	\\
science fiction writer	&	L	&	L	\\
data management effort	&	L	&	L	\\
speech communication skills	&	R	&	I	\\
hair cell destruction	&	L	&	L	\\
bird pox viruses	&	L	&	I	\\
college extension personnel	&	R	&	R	\\
tobacco mosaic virus	&	R	&	I	\\
currency brokerage office	&	R	&	L	\\
countries education systems	&	R	&	E	\\
student personality traits	&	R	&	R	\\
hydrogen energy system	&	R	&	L	\\
fiber optics systems	&	L	&	L	\\
health enforcement agencies	&	R	&	L	\\
minority business enterprises	&	L	&	L	\\
law enforcement agency	&	L	&	L	\\
college basketball players	&	L	&	I	\\
law enforcement organizations	&	L	&	L	\\
law enforcement agencies	&	L	&	L	\\
studio office buildings	&	R	&	R	\\
speech transmission systems	&	L	&	L	\\
century manuscript illumination	&	L	&	E	\\
century manuscript illumination	&	L	&	E	\\
quality assurance department	&	L	&	L	\\
infantry infiltration tactics	&	R	&	R	\\
law enforcement officials	&	L	&	L	\\
law enforcement officials	&	L	&	L	\\
tobacco mosaic virus	&	R	&	I	\\
tobacco mosaic virus	&	R	&	I	\\
army ordnance depot	&	R	&	I	\\
highway transportation systems	&	R	&	I	\\
science fiction writer	&	L	&	L	\\
sperm cell production	&	L	&	L	\\
world grape production	&	R	&	R	\\
world food production	&	R	&	R	\\
Romans concert hall	&	R	&	E	\\
Sunday afternoon football	&	L	&	L	\\
country music singer	&	L	&	L	\\
\hline
\end{tabular}

\begin{tabular}{|l|c|c|} \hline
Noun triple	& Best model & Correct Answer \\
\hline
tenor sax players	&	L	&	L	\\
health maintenance organizations	&	R	&	L	\\
health maintenance organizations	&	R	&	L	\\
hospital payment system	&	L	&	L	\\
countries health insurance	&	L	&	E	\\
tenor saxophone player	&	L	&	L	\\
television drama series	&	L	&	R	\\
television suspense series	&	R	&	R	\\
lymph node enlargement	&	L	&	L	\\
law enforcement activities	&	L	&	L	\\
law enforcement standards	&	L	&	L	\\
cell plasma membrane	&	R	&	I	\\
repertory theater movement	&	L	&	L	\\
hospital dissection theaters	&	L	&	R	\\
government construction standards	&	R	&	R	\\
daylight hours hummingbirds	&	L	&	E	\\
kidney artery disease	&	R	&	L	\\
origins quota system	&	L	&	L	\\
valleys irrigation systems	&	R	&	E	\\
food storage facilities	&	L	&	L	\\
government ethics law	&	L	&	R	\\
period infantry armies	&	L	&	E	\\
century infantry tactics	&	R	&	E	\\
chicken pox infection	&	L	&	L	\\
information storage technology	&	L	&	L	\\
data storage systems	&	L	&	L	\\
debt repayment problems	&	L	&	L	\\
county borough corporations	&	L	&	R	\\
world petroleum crisis	&	L	&	R	\\
luxury furniture industry	&	L	&	L	\\
gridiron street pattern	&	R	&	R	\\
Buddhist hell scenes	&	R	&	R	\\
ethics committee investigation	&	L	&	L	\\
sea bass family	&	L	&	L	\\
computer industry entrepreneur	&	L	&	L	\\
day management responsibilities	&	L	&	R	\\
granite office tower	&	R	&	R	\\
satellite news distribution	&	L	&	R	\\
health maintenance organization	&	R	&	L	\\
college basketball facilities	&	L	&	I	\\
bladder outlet obstruction	&	L	&	L	\\
war crimes trials	&	L	&	L	\\
advice newspaper columns	&	R	&	R	\\
television mystery series	&	R	&	R	\\
missile defense weapons	&	R	&	L	\\
music hall comedian	&	R	&	L	\\
\hline
\end{tabular}

\begin{tabular}{|l|c|c|} \hline
Noun triple	& Best model & Correct Answer \\
\hline
workers compensation law	&	L	&	L	\\
canon law system	&	L	&	L	\\
student democracy movement	&	R	&	R	\\
world petroleum producers	&	L	&	R	\\
life insurance policy	&	L	&	L	\\
life insurance policy	&	L	&	L	\\
coalition war cabinet	&	R	&	R	\\
teacher education college	&	L	&	L	\\
college basketball commentator	&	L	&	L	\\
management information systems	&	L	&	R	\\
weapons delivery systems	&	L	&	L	\\
emergency medicine specialist	&	L	&	L	\\
health maintenance organizations	&	R	&	L	\\
tobacco mosaic virus	&	R	&	I	\\
pagan Arab tribes	&	R	&	I	\\
civilian undergraduate colleges	&	L	&	R	\\
law enforcement officials	&	L	&	L	\\
system drainage area	&	L	&	R	\\
quantum interference device	&	R	&	L	\\
monsoon regions rainfall	&	L	&	E	\\
fertility mystery cult	&	R	&	L	\\
city government activities	&	L	&	L	\\
century concert music	&	R	&	R	\\
music chorale melodies	&	L	&	R	\\
union representation elections	&	R	&	R	\\
government ephemeris offices	&	R	&	R	\\
war crimes tribunals	&	L	&	L	\\
sperm storage vessel	&	L	&	L	\\
nucleus evaporation products	&	R	&	R	\\
fission energy production	&	L	&	L	\\
weapon delivery systems	&	L	&	L	\\
weapons production facilities	&	R	&	L	\\
weapon delivery systems	&	L	&	L	\\
food energy calories	&	L	&	L	\\
assistant majority leader	&	R	&	R	\\
university opera workshops	&	R	&	R	\\
war crimes trials	&	L	&	L	\\
law enforcement resources	&	L	&	L	\\
energy storage element	&	L	&	L	\\
law enforcement agencies	&	L	&	L	\\
law enforcement agencies	&	L	&	L	\\
barn owl family	&	L	&	L	\\
computer memory units	&	L	&	I	\\
Buddhist temple precincts	&	L	&	R	\\
family cigar business	&	R	&	R	\\
world transportation artery	&	L	&	R	\\
\hline
\end{tabular}

\begin{tabular}{|l|c|c|} \hline
Noun triple	& Best model & Correct Answer \\
\hline
community barn raisings	&	R	&	R	\\
deputy music director	&	R	&	R	\\
twenty-one member nations	&	L	&	E	\\
war crimes indictments	&	L	&	L	\\
night warfare capabilities	&	L	&	L	\\
gasoline storage tanks	&	L	&	L	\\
cylinder phonograph system	&	L	&	R	\\
television news photography	&	L	&	L	\\
students laboratory instruction	&	R	&	E	\\
world umbrella organization	&	R	&	R	\\
years planetarium projectors	&	L	&	E	\\
venom delivery system	&	L	&	L	\\
polarizer prism system	&	R	&	R	\\
navigation guidance system	&	R	&	L	\\
law enforcement agencies	&	L	&	L	\\
science fiction writer	&	L	&	L	\\
engine combustion temperatures	&	L	&	R	\\
alpha particle source	&	L	&	L	\\
century population growth	&	L	&	E	\\
government poverty statistics	&	R	&	R	\\
combustion turbine generators	&	L	&	R	\\
combustion turbine unit	&	L	&	R	\\
years parent involvement	&	L	&	E	\\
landslide election victories	&	R	&	R	\\
government news agency	&	R	&	R	\\
computer output device	&	R	&	R	\\
computer hardware technology	&	L	&	L	\\
alpha particle bombardment	&	L	&	L	\\
child guidance movement	&	R	&	L	\\
granite valley temple	&	R	&	R	\\
family newspaper business	&	R	&	R	\\
computer radar systems	&	R	&	R	\\
government frequency allocations	&	R	&	R	\\
aperture synthesis systems	&	L	&	L	\\
world news roundup	&	L	&	L	\\
uranium disintegration series	&	R	&	R	\\
room temperature radon	&	L	&	E	\\
laser radar systems	&	L	&	L	\\
music industry designation	&	L	&	L	\\
army ammunition depot	&	R	&	I	\\
college football player	&	L	&	I	\\
imitation rococo interiors	&	L	&	L	\\
war college instructor	&	L	&	L	\\
decades city leaders	&	R	&	E	\\
chicken sarcoma virus	&	R	&	I	\\
deputy assistant secretary	&	L	&	I	\\
\hline
\end{tabular}

\begin{tabular}{|l|c|c|} \hline
Noun triple	& Best model & Correct Answer \\
\hline
peasant redemption payments	&	L	&	R	\\
city government elections	&	L	&	L	\\
army tank commander	&	R	&	I	\\
rider ropes cattle	&	R	&	E	\\
college basketball player	&	L	&	I	\\
college economics textbook	&	L	&	I	\\
luxury apartment buildings	&	L	&	L	\\
assistant division commander	&	R	&	R	\\
science curriculum development	&	L	&	L	\\
assistant majority leader	&	R	&	R	\\
city sewerage systems	&	L	&	I	\\
missile guidance systems	&	R	&	L	\\
household laundry products	&	R	&	R	\\
community welfare resources	&	L	&	R	\\
detection investigation committee	&	L	&	L	\\
centuries song accompaniments	&	L	&	E	\\
student achievement measurements	&	L	&	L	\\
Buddhist stupa mountain	&	R	&	I	\\
radar ocean surveillance	&	R	&	R	\\
communications satellite organization	&	L	&	L	\\
missile defense systems	&	R	&	L	\\
missile defense system	&	R	&	L	\\
government policy decisions	&	R	&	I	\\
war crimes tribunal	&	L	&	L	\\
college football player	&	L	&	I	\\
protein digestion products	&	L	&	L	\\
coalition civilian government	&	R	&	R	\\
world property ownership	&	L	&	E	\\
swine flu virus	&	L	&	L	\\
swine flu virus	&	L	&	L	\\
millennium Arab traders	&	L	&	E	\\
tenor saxophone player	&	L	&	L	\\
tenor saxophone player	&	L	&	L	\\
news bureau chiefs	&	L	&	L	\\
law enforcement agencies	&	L	&	L	\\
law enforcement officials	&	L	&	L	\\
civilian population losses	&	L	&	I	\\
precision navigation systems	&	L	&	I	\\
river valley communities	&	L	&	L	\\
college student governments	&	L	&	L	\\
development policy decisions	&	L	&	L	\\
computer graphics systems	&	L	&	I	\\
computer graphics system	&	L	&	I	\\
tobacco mosaic virus	&	R	&	I	\\
river January temperatures	&	L	&	E	\\
pagan fertility goddess	&	R	&	R	\\
\hline
\end{tabular}

\begin{tabular}{|l|c|c|} \hline
Noun triple	& Best model & Correct Answer \\
\hline
country bumpkin nephew	&	L	&	L	\\
country music revivals	&	L	&	L	\\
bile pigment metabolism	&	L	&	L	\\
law enforcement agencies	&	L	&	L	\\
world wool production	&	R	&	R	\\
vapor density methods	&	L	&	L	\\
science fiction writer	&	L	&	L	\\
security council action	&	L	&	L	\\
Amazon frontier region	&	R	&	R	\\
\hline
\end{tabular}

\end{center}

\cleardoublepage

\chapter{Test Set for Noun Compound Paraphrasing}
\label{appendix:cetest}

The following list of noun compounds forms the test set
used for the experiments on noun compound paraphrasing.
A random sample of 400 was selected from the 24,251 noun
pairs extracted from Grolier's using the method described
in section~\ref{sec:ce_method}.  These were then assigned
one of the following paraphrases: 
\lingform{of}~(O), \lingform{for}~(R), \lingform{in}~(I),
\lingform{on}~(N), \lingform{at}~(A), \lingform{from}~(F), 
\lingform{with}~(W), \lingform{about}~(T) and 
non-prepositional~(X).  This answer is shown in the last 
column.  The second column contains a code describing the
type for non-prepositional noun pairs: extraction error~(E),
verbal-nexus~(V) and copula~(B).

For comparison, the third column gives the prediction made
by the lexically parametrised statistical model using
maximum likelihood estimates (see section~\ref{sec:ce_results}
and in particular table~\ref{tb:ce_results_mle8}).  This
model achieves 40\% accuracy.

\vspace{1cm}

\begin{center}

\begin{tabular}{|l|l|c|c|} \hline
Noun pair	& Type	& Lexical model & Correct answer \\
\hline
fusion devices	&		&	R	&	W	\\
computation skills	&		&	R	&	A	\\
pigment accumulation	&	V-subj	&	O	&	X	\\
metallurgy industry	&		&	I	&	R	\\
pest species	&	B	&	F	&	X	\\
world food	&	E	&	I	&	X	\\
fossil fauna	&	B	&	F	&	X	\\
civilian population	&	B	&	O	&	X	\\
passengers hostage	&	E	&	O	&	X	\\
government agencies	&		&	N	&	O	\\
sea mammals	&		&	A	&	F	\\
Arab seafarers	&	B	&	O	&	X	\\
health problems	&		&	W	&	O	\\
deputy governor	&	B	&	O	&	X	\\
city legislature	&		&	I	&	O	\\
championship bout	&		&	I	&	R	\\
\hline
\end{tabular}

\begin{tabular}{|l|l|c|c|} \hline
Noun pair	& Type	& Lexical model & Correct answer \\
\hline
magic password	&	E	&	I	&	X	\\
carbon atoms	&	B-chem	&	W	&	X	\\
relations agency	&		&	R	&	R	\\
sea lanes	&		&	F	&	I	\\
oxygen atoms	&	B-chem	&	W	&	X	\\
child welfare	&		&	O	&	O	\\
concert music	&		&	R	&	A	\\
property owners	&	V-obj	&	W	&	X	\\
disease organisms	&		&	F	&	O	\\
laser technology	&		&	F	&	W	\\
newspaper subscriptions	&		&	R	&	R	\\
activity spectrum	&		&	O	&	O	\\
antibiotic regimen	&		&	O	&	O	\\
baccalaureate curriculum	&		&	O	&	R	\\
transportation system	&		&	R	&	R	\\
Arab origin	&	B	&	O	&	X	\\
Arab world	&		&	W	&	O	\\
concert appearances	&		&	I	&	I	\\
sea animals	&		&	A	&	O	\\
welfare agencies	&		&	N	&	R	\\
computer catalog	&		&	O	&	N	\\
hydrogen atoms	&	B	&	W	&	X	\\
submarine mountain	&	E	&	N	&	X	\\
hydrogen atoms	&	B	&	W	&	X	\\
alpha particle	&	B-chem	&	O	&	X	\\
television director	&	V-obj	&	R	&	X	\\
anatomy professor	&		&	O	&	O	\\
vehicle industry	&		&	I	&	R	\\
machinery operations	&		&	F	&	W	\\
warfare equipment	&		&	R	&	R	\\
country estate	&		&	I	&	I	\\
assistant secretary	&	B	&	O	&	X	\\
security pacts	&		&	O	&	O	\\
river valleys	&		&	N	&	W	\\
quadrant elevation	&		&	I	&	I	\\
banana industry	&		&	O	&	R	\\
jute products	&		&	O	&	O	\\
government patronage	&		&	O	&	F	\\
dairy barn	&		&	O	&	R	\\
battery technology	&		&	R	&	R	\\
football player	&	V-obj	&	N	&	X	\\
cattle rustler	&	V-obj	&	I	&	X	\\
trial lawyers	&		&	R	&	R	\\
drama critic	&	V-obj	&	R	&	X	\\
Arab conquest	&	V-subj	&	O	&	X	\\
warrior prince	&	B	&	O	&	X	\\
\hline
\end{tabular}

\begin{tabular}{|l|l|c|c|} \hline
Noun pair	& Type	& Lexical model & Correct answer \\
\hline
cancer production	&	V-obj	&	O	&	X	\\
protein molecules	&	B	&	F	&	X	\\
Sunday restrictions	&		&	N	&	N	\\
theater history	&		&	A	&	O	\\
life imprisonment	&		&	R	&	R	\\
family members	&		&	O	&	O	\\
storage batteries	&		&	R	&	R	\\
plutonium theft	&		&	O	&	O	\\
union leader	&		&	F	&	O	\\
priority areas	&		&	O	&	O	\\
Buddhist laity	&	B	&	F	&	X	\\
subsistence cultivation	&		&	O	&	R	\\
language family	&		&	F	&	O	\\
business investment	&		&	I	&	I	\\
business education	&		&	I	&	T	\\
business education	&		&	I	&	T	\\
climate pattern	&		&	N	&	O	\\
football player	&	V-obj	&	N	&	X	\\
lieutenant governors	&	B	&	O	&	X	\\
sector investment	&	V-subj	&	I	&	X	\\
fossil assemblage	&	V-obj	&	W	&	X	\\
typewriter mechanisms	&		&	R	&	R	\\
recreation area	&		&	R	&	R	\\
cattle industry	&		&	R	&	R	\\
cattle population	&		&	O	&	O	\\
ceramics products	&		&	F	&	O	\\
phonograph pickups	&		&	O	&	R	\\
monastery buildings	&		&	A	&	A	\\
apartment dwellers	&		&	W	&	I	\\
reaction mixture	&		&	O	&	R	\\
oxygen atoms	&	B-chem	&	W	&	X	\\
peasant rebellion	&	V-subj	&	O	&	X	\\
insect pests	&	B	&	N	&	X	\\
music director	&	V-obj	&	O	&	X	\\
city management	&	V-obj	&	O	&	X	\\
law systems	&		&	O	&	O	\\
business applications	&		&	O	&	I	\\
mountain valleys	&		&	N	&	I	\\
community education	&		&	I	&	T	\\
logic unit	&		&	N	&	R	\\
computer novices	&		&	I	&	W	\\
computer memory	&		&	R	&	R	\\
application areas	&		&	R	&	O	\\
information sources	&		&	O	&	O	\\
property law	&		&	W	&	T	\\
wilderness areas	&		&	F	&	O	\\
\hline
\end{tabular}

\begin{tabular}{|l|l|c|c|} \hline
Noun pair	& Type	& Lexical model & Correct answer \\
\hline
convenience foods	&		&	R	&	W	\\
business holdings	&		&	I	&	O	\\
corrosion resistance	&	V-obj	&	F	&	X	\\
world economies	&		&	I	&	O	\\
trio sonatas	&		&	R	&	R	\\
dairy cattle	&		&	F	&	R	\\
opera performances	&	V-obj	&	A	&	X	\\
fiction writer	&	V-obj	&	R	&	X	\\
mountain glaciers	&		&	F	&	N	\\
theater director	&	V-obj	&	A	&	X	\\
pigment granules	&		&	I	&	O	\\
road competitions	&		&	R	&	N	\\
temperature distribution	&	V-obj	&	A	&	X	\\
government buildings	&		&	N	&	R	\\
ballet genres	&		&	I	&	O	\\
prison poems	&		&	I	&	T	\\
vase paintings	&		&	N	&	N	\\
musk deer	&		&	O	&	W	\\
patron goddesses	&	B	&	W	&	X	\\
government intervention	&	V-subj	&	I	&	X	\\
cleavage division	&	B	&	O	&	X	\\
food resources	&	B	&	R	&	X	\\
cell membrane	&		&	O	&	O	\\
extinction theory	&		&	T	&	T	\\
bird droppings	&		&	O	&	F	\\
monkey pox	&		&	O	&	I	\\
hull maintenance	&	V-obj	&	O	&	X	\\
memory system	&		&	I	&	R	\\
pottery vessels	&		&	N	&	O	\\
population density	&		&	O	&	O	\\
business sector	&		&	O	&	O	\\
kidney disease	&		&	F	&	I	\\
Arab unity	&	V-subj	&	W	&	X	\\
family business	&		&	W	&	O	\\
decomposition reactions	&	B	&	N	&	X	\\
quantum theory	&		&	O	&	T	\\
storage capacity	&		&	R	&	R	\\
period classifications	&		&	O	&	I	\\
world soul	&		&	F	&	O	\\
reaction products	&	V-subj	&	F	&	X	\\
town halls	&		&	R	&	R	\\
life sciences	&		&	I	&	T	\\
plasma membrane	&		&	F	&	R	\\
food shortages	&		&	O	&	O	\\
photography movement	&		&	R	&	I	\\
terrorist activities	&		&	O	&	O	\\
\hline
\end{tabular}

\begin{tabular}{|l|l|c|c|} \hline
Noun pair	& Type	& Lexical model & Correct answer \\
\hline
mystery novels	&		&	N	&	T	\\
customs administrations	&	V-obj	&	W	&	X	\\
antenna rods	&	B	&	O	&	X	\\
shorthand device	&		&	I	&	R	\\
car odor	&		&	O	&	O	\\
country music	&		&	F	&	I	\\
food industry	&		&	R	&	R	\\
bile duct	&		&	I	&	R	\\
world wars	&		&	I	&	O	\\
rococo spirit	&		&	O	&	O	\\
trio sonata	&		&	R	&	R	\\
world community	&		&	I	&	O	\\
government war	&	E	&	W	&	X	\\
coalition cabinet	&		&	F	&	I	\\
music theory	&		&	O	&	T	\\
Jesuit origin	&		&	O	&	I	\\
hardware business	&		&	W	&	I	\\
coronation portal	&		&	A	&	T	\\
savanna areas	&		&	O	&	O	\\
frontier problems	&		&	N	&	N	\\
city dwellers	&		&	I	&	I	\\
family tradition	&		&	T	&	O	\\
gestation period	&		&	O	&	O	\\
city population	&		&	O	&	I	\\
magic beings	&	E	&	W	&	X	\\
recreation areas	&		&	R	&	R	\\
handicraft products	&	V-subj	&	R	&	X	\\
unit cell	&	B	&	W	&	X	\\
crime novelist	&		&	W	&	T	\\
sea monster	&		&	W	&	F	\\
treaty relationships	&		&	W	&	O	\\
business operations	&	V-obj	&	N	&	X	\\
lieutenant governor	&	B	&	O	&	X	\\
dominion status	&		&	O	&	O	\\
cancer cells	&		&	F	&	W	\\
child custody	&		&	N	&	O	\\
government intervention	&	V-subj	&	I	&	X	\\
hotel management	&	V-obj	&	A	&	X	\\
excavation skills	&		&	W	&	R	\\
life scientists	&		&	T	&	T	\\
impulse transmission	&	V-obj	&	O	&	X	\\
species determination	&	V-obj	&	T	&	X	\\
absorption hygrometers	&		&	O	&	W	\\
petroleum wealth	&		&	F	&	O	\\
transportation equipment	&		&	R	&	R	\\
family sagas	&		&	T	&	T	\\
\hline
\end{tabular}

\begin{tabular}{|l|l|c|c|} \hline
Noun pair	& Type	& Lexical model & Correct answer \\
\hline
mountain barrier	&	B	&	I	&	X	\\
elites resentment	&	E	&	O	&	X	\\
consonant systems	&		&	R	&	O	\\
language literature	&		&	T	&	I	\\
population density	&		&	O	&	O	\\
war captives	&		&	A	&	F	\\
worker satisfaction	&		&	F	&	O	\\
population explosion	&		&	O	&	O	\\
rationalist thinkers	&	B	&	O	&	X	\\
hardware technology	&		&	I	&	O	\\
insurance industry	&		&	R	&	R	\\
intelligence community	&		&	W	&	R	\\
catalog illustrations	&		&	R	&	I	\\
transmission system	&		&	R	&	R	\\
war crimes	&		&	A	&	I	\\
production facilities	&		&	R	&	R	\\
uplands temperatures	&	E	&	N	&	X	\\
theater orchestra	&		&	A	&	I	\\
violin concerto	&		&	R	&	R	\\
impeachment trial	&		&	R	&	R	\\
population density	&		&	O	&	O	\\
mountain king	&	E	&	F	&	X	\\
policy options	&		&	N	&	T	\\
meat products	&		&	F	&	W	\\
mountain country	&		&	F	&	W	\\
union security	&	V-obj	&	R	&	X	\\
business economics	&		&	F	&	R	\\
drainage basins	&		&	W	&	O	\\
coalition government	&		&	I	&	I	\\
luxury hotels	&		&	R	&	W	\\
population growth	&	V-subj	&	O	&	X	\\
heath family	&		&	O	&	O	\\
customs union	&		&	W	&	T	\\
kerosene lamps	&		&	O	&	W	\\
emergency detention	&		&	R	&	I	\\
faculty members	&		&	O	&	O	\\
lieutenant governor	&	B	&	O	&	X	\\
war secretary	&		&	O	&	R	\\
symphony orchestras	&		&	F	&	R	\\
poultry pests	&		&	I	&	R	\\
century Americans	&	E	&	R	&	X	\\
war god	&		&	O	&	O	\\
genre painter	&	V-obj	&	O	&	X	\\
guild members	&		&	O	&	O	\\
petroleum industry	&		&	F	&	R	\\
temperature variations	&	V-obj	&	A	&	X	\\
\hline
\end{tabular}

\begin{tabular}{|l|l|c|c|} \hline
Noun pair	& Type	& Lexical model & Correct answer \\
\hline
drainage patterns	&		&	N	&	O	\\
minority businesses	&		&	T	&	O	\\
cotton cultivation	&	V-obj	&	O	&	X	\\
majority leader	&		&	F	&	O	\\
money policy	&		&	O	&	T	\\
policy makers	&	V-obj	&	I	&	X	\\
ancestor spirits	&		&	O	&	O	\\
satellite system	&		&	O	&	W	\\
poultry products	&		&	R	&	F	\\
census population	&		&	F	&	N	\\
petroleum products	&		&	F	&	F	\\
opposition coalition	&		&	W	&	I	\\
government policy	&		&	O	&	O	\\
deputy director	&	B	&	O	&	X	\\
altitude variations	&	V-subj	&	A	&	X	\\
cattle town	&		&	N	&	R	\\
dairy products	&	V-subj	&	F	&	X	\\
fission products	&	V-subj	&	O	&	X	\\
weapons policy	&		&	N	&	N	\\
protein source	&		&	O	&	O	\\
ocean basins	&		&	I	&	O	\\
choice species	&		&	O	&	O	\\
backwoods protagonist	&		&	I	&	F	\\
vibration ratio	&		&	O	&	O	\\
university education	&		&	A	&	I	\\
car driver	&	V-obj	&	I	&	X	\\
antelope species	&		&	O	&	O	\\
carbon atom	&	B-chem	&	W	&	X	\\
valve systems	&		&	O	&	O	\\
opposition leaders	&	V-obj	&	I	&	X	\\
university teachers	&		&	A	&	I	\\
expansion turbine	&		&	R	&	W	\\
pagan teachings	&	E	&	O	&	X	\\
temple portico	&		&	N	&	O	\\
food preparation	&	V-obj	&	R	&	X	\\
education journals	&		&	I	&	T	\\
petroleum transportation	&	V-obj	&	N	&	X	\\
chemistry laboratories	&		&	R	&	R	\\
petroleum products	&		&	F	&	F	\\
Buddhist philosophy	&		&	F	&	O	\\
population expansion	&	V-subj	&	O	&	X	\\
university cabinets	&		&	F	&	O	\\
vortex atom	&	B	&	O	&	X	\\
cupboard doors	&		&	O	&	O	\\
separation negatives	&		&	O	&	F	\\
election laws	&		&	I	&	T	\\
\hline
\end{tabular}

\begin{tabular}{|l|l|c|c|} \hline
Noun pair	& Type	& Lexical model & Correct answer \\
\hline
anarchist conspirators	&	B	&	O	&	X	\\
coalition government	&		&	I	&	I	\\
laboratory quantities	&		&	I	&	R	\\
government action	&	V-subj	&	N	&	X	\\
construction quality	&		&	O	&	O	\\
parent education	&	V-obj	&	F	&	X	\\
television era	&		&	R	&	W	\\
letterpress composition	&	V-obj	&	O	&	X	\\
strength properties	&		&	O	&	O	\\
protein synthesis	&	V-obj	&	O	&	X	\\
frustration tolerance	&	V-obj	&	R	&	X	\\
incubation period	&		&	R	&	O	\\
January temperatures	&		&	I	&	I	\\
frequency output	&	V-obj	&	A	&	X	\\
aperture synthesis	&	V-obj	&	O	&	X	\\
ratings systems	&		&	O	&	O	\\
railway union	&		&	O	&	R	\\
education movement	&		&	R	&	R	\\
television newscaster	&		&	N	&	N	\\
warbler family	&		&	O	&	O	\\
equivalence principle	&		&	O	&	O	\\
rotation period	&		&	O	&	O	\\
world population	&		&	O	&	O	\\
household refrigeration	&		&	I	&	I	\\
business administration	&	V-obj	&	F	&	X	\\
office buildings	&		&	F	&	R	\\
affairs television	&	E	&	I	&	X	\\
boyar duma	&	E-foreign	&	O	&	X	\\
symphony orchestra	&		&	F	&	R	\\
country estate	&		&	I	&	I	\\
satellite system	&		&	O	&	O	\\
density measurements	&	V-obj	&	N	&	X	\\
sea lions	&		&	I	&	F	\\
Passover festival	&		&	O	&	R	\\
childhood sexuality	&		&	I	&	I	\\
television writer	&		&	N	&	R	\\
sea urchins	&		&	O	&	F	\\
horror tale	&		&	A	&	T	\\
shellfish crustaceans	&	B	&	O	&	X	\\
cargo carrier	&	V-obj	&	W	&	X	\\
shrub competition	&	V-subj	&	F	&	X	\\
hair follicles	&		&	I	&	R	\\
communications systems	&		&	R	&	R	\\
family connection	&		&	W	&	O	\\
soul music	&		&	R	&	T	\\
food products	&		&	F	&	F	\\
\hline
\end{tabular}

\begin{tabular}{|l|l|c|c|} \hline
Noun pair	& Type	& Lexical model & Correct answer \\
\hline
communications satellite	&		&	R	&	R	\\
management procedures	&		&	R	&	R	\\
baseball player	&	V-obj	&	I	&	X	\\
fiber optics	&		&	W	&	W	\\
construction industry	&		&	R	&	R	\\
county town	&		&	I	&	R	\\
estimation methods	&		&	I	&	R	\\
percentage composition	&		&	O	&	I	\\
altitude reconnaissance	&		&	F	&	A	\\
trolley cars	&	B	&	O	&	X	\\
student movement	&	V-subj	&	R	&	X	\\
suffrage committee	&		&	R	&	R	\\
suspension system	&		&	W	&	R	\\
health standards	&		&	R	&	O	\\
world championships	&		&	I	&	O	\\
gestation period	&		&	O	&	O	\\
arts museum	&		&	I	&	R	\\
tea room	&		&	R	&	R	\\
lab periods	&		&	R	&	I	\\
communications industries	&		&	R	&	R	\\
carrier system	&		&	N	&	W	\\
television production	&		&	N	&	R	\\
prohibition law	&		&	R	&	O	\\
tenor trombone	&	B	&	O	&	X	\\
pagan origins	&	E	&	O	&	X	\\
Sanskrit texts	&		&	F	&	I	\\
area basis	&		&	R	&	O	\\
arts colleges	&		&	I	&	R	\\
terrorist activities	&		&	O	&	O	\\
tumor production	&	V-obj	&	O	&	X	\\
industry revenues	&		&	F	&	I	\\
construction materials	&		&	R	&	R	\\
government officials	&		&	F	&	R	\\
cotton production	&	V-obj	&	R	&	X	\\
news events	&	B	&	I	&	X	\\
world war	&		&	I	&	O	\\
commonwealth status	&		&	I	&	I	\\
government intervention	&	V-subj	&	I	&	X	\\
food production	&	V-obj	&	O	&	X	\\
January temperature	&		&	I	&	I	\\
street scenes	&		&	F	&	N	\\
food industry	&		&	R	&	R	\\
eaves troughs	&		&	O	&	N	\\
lava fountains	&		&	O	&	O	\\
treatment systems	&		&	R	&	R	\\
puppet government	&	B	&	O	&	X	\\
\hline
\end{tabular}

\begin{tabular}{|l|l|c|c|} \hline
Noun pair	& Type	& Lexical model & Correct answer \\
\hline
frontier community	&		&	N	&	N	\\
temperature differences	&	V-subj	&	A	&	X	\\
frontier life	&		&	N	&	A	\\
marriage customs	&		&	O	&	T	\\
pertussis bacteria	&	B	&	O	&	X	\\
sophomore year	&	B	&	O	&	X	\\
crossroads village	&		&	O	&	A	\\
fiction representative	&	E	&	O	&	X	\\
puppet regimes	&	B	&	O	&	X	\\
settlement patterns	&		&	R	&	O	\\
luxury goods	&	B	&	R	&	X	\\
theater orchestra	&		&	A	&	R	\\
automobile factory	&		&	R	&	R	\\
television series	&		&	N	&	R	\\
room temperature	&		&	O	&	O	\\
laboratory applications	&		&	I	&	I	\\
\hline
\end{tabular}

\end{center}

\end{document}